%% file: manuscript.tex
\documentclass[pdflatex,sn-mathphys-num]{sn-jnl}

\usepackage{graphicx}
\usepackage{multirow}
\usepackage{amsmath,amssymb,amsfonts}
\usepackage{amsthm}
\usepackage{mathrsfs}
\usepackage[title]{appendix}
\usepackage{xcolor}
\usepackage{textcomp}
\usepackage{manyfoot}
\usepackage{booktabs}
\usepackage{algorithm}
\usepackage{algorithmicx}
\usepackage{algpseudocode}
\usepackage{listings}
\usepackage{lineno}
\usepackage{bibunits}

\usepackage{longtable}
\usepackage{pdflscape}
\usepackage{array}
\usepackage{ragged2e}
\usepackage{adjustbox}
\usepackage{tabularx}

\usepackage[T1]{fontenc}
\usepackage[utf8]{inputenc}

\usepackage[most]{tcolorbox}
\usepackage{enumitem}

\definecolor{promptgray}{RGB}{160,160,160}
\definecolor{promptbg}{RGB}{245,245,245}
\definecolor{promptborder}{RGB}{140,140,140}

\newtcolorbox{promptbox}[1]{
    enhanced,
    breakable,
    colback=promptbg,
    colframe=promptborder,
    coltitle=white,
    colbacktitle=promptgray,
    title=\textbf{#1},
    fonttitle=\bfseries,
    sharp corners,
    boxrule=0.6pt,
    left=8pt,
    right=8pt,
    top=8pt,
    bottom=8pt,
    titlerule=0pt,
    boxed title style={
        sharp corners,
        boxrule=0pt,
        colframe=promptgray,
        colback=promptgray,
        left=8pt,
        right=8pt,
        top=3pt,
        bottom=3pt
    }
}

\usepackage{tocloft}

\makeatletter
\def\ps@plain{%
    \let\@mkboth\@gobbletwo
    \let\@oddhead\@empty
    \let\@evenhead\@empty
    \def\@oddfoot{%
        \vbox to 18pt{%
            \vfill
            \reset@font
            \rmfamily
            \hfil\thepage\hfil
        }%
    }%
    \let\@evenfoot\@oddfoot
}
\makeatother

\usepackage{hyperref}

\theoremstyle{thmstyleone}%
\theoremstyle{thmstyletwo}%

\theoremstyle{thmstylethree}%

\begin{document}

\addtocontents{toc}{\protect\setcounter{tocdepth}{-1}}

\title[Article Title]{Teaching LLMs How ICU Physicians Approach Clinical Reasoning Through OMOP-Aligned Retrieval Improves Reasoning Across Clinical Domains}


\author[1,2]{\fnm{Miguel} \sur{Contreras}}
\author[1,2]{\fnm{Scott} \sur{Siegel}}
\author[1,2]{\fnm{Subhash} \sur{Nerella}}
\author[1,2]{\fnm{Jessica} \sur{Sena}}
\author[2,3]{\fnm{Jiaqing} \sur{Zhang}}
\author[1,2]{\fnm{Heng} \sur{Sun}}
\author[4]{\fnm{Hruday Tej} \sur{Akkaladevi}}
\author[5]{\fnm{Peiyu} \sur{Lu}}
\author[5]{\fnm{Jordan} \sur{Rosen}}

\author[6]{\fnm{Sumit} \sur{Kapoor}}
\author[1]{\fnm{Sasank} \sur{Desaraju}}
\author[7]{\fnm{Grace R.} \sur{Thompson}}
\author[8]{\fnm{Jacob} \sur{Purcell}}
\author[8]{\fnm{Michael} \sur{Petrauskis}}
\author[7]{\fnm{Philip KW.} \sur{Hong}}
\author[9]{\fnm{Meghan} \sur{Brennan}}
\author[10]{\fnm{Sarah} \sur{Chrabaszcz}}
\author[8]{\fnm{Tierra} \sur{Smith}}
\author[8]{\fnm{Ronnie} \sur{Ren}}
\author[7]{\fnm{Michel S.} \sur{Kabbash}}
\author[2,5]{\fnm{Ceyhun} \sur{Haziroglu}}
\author[2,5]{\fnm{Rushi} \sur{Patel}}
\author[8]{\fnm{Gabriel} \sur{Gomez}}
\author[5]{\fnm{Charlotte} \sur{Chaiklin}}
\author[8]{\fnm{Randy} \sur{Leung}}
\author[9]{\fnm{Kenneth N.} \sur{John}}
\author[7]{\fnm{Whitman} \sur{Wiggins}}
\author[8]{\fnm{Philip} \sur{Kayser}}

\author[11]{\fnm{Vincent} \sur{Bird}}
\author[12]{\fnm{Maria} \sur{Bruzzone}}
\author[7]{\fnm{Tyler J.} \sur{Loftus}}
\author[2,13]{\fnm{Azra} \sur{Bihorac}}
\author*[1,2]{\fnm{Parisa} \sur{Rashidi}}\email{parisa.rashidi@ufl.edu}

\affil[1]{\orgdiv{Department of Biomedical Engineering}, \orgname{University of Florida},
  \orgaddress{\city{Gainesville}, \state{FL}, \country{USA}}}

\affil[2]{\orgdiv{Intelligent Clinical Care Center (IC3)}, \orgname{University of Florida},
  \orgaddress{\city{Gainesville}, \state{FL}, \country{USA}}}

\affil[3]{\orgdiv{Department of Electrical and Computer Engineering}, \orgname{University of Florida},
  \orgaddress{\city{Gainesville}, \state{FL}, \country{USA}}}

\affil[4]{\orgname{Google},
  \orgaddress{\city{Seattle}, \state{WA}, \country{USA}}}

\affil[5]{\orgdiv{Department of Medicine}, \orgname{University of Florida},
  \orgaddress{\city{Gainesville}, \state{FL}, \country{USA}}}

\affil[6]{\orgdiv{Department of Critical Care Medicine}, \orgname{University of Pittsburgh},
  \orgaddress{\city{Pittsburgh}, \state{PA}, \country{USA}}}

\affil[7]{\orgdiv{Department of Surgery}, \orgname{University of Florida},
  \orgaddress{\city{Gainesville}, \state{FL}, \country{USA}}}

\affil[8]{\orgdiv{Department of Emergency Medicine}, \orgname{University of Florida},
  \orgaddress{\city{Gainesville}, \state{FL}, \country{USA}}}

\affil[9]{\orgdiv{Department of Anesthesiology}, \orgname{University of Florida},
  \orgaddress{\city{Gainesville}, \state{FL}, \country{USA}}}

\affil[10]{\orgdiv{Department of Orthopaedic Surgery and Sports Medicine}, \orgname{University of Florida},
  \orgaddress{\city{Gainesville}, \state{FL}, \country{USA}}}

\affil[11]{\orgdiv{Department of Urology}, \orgname{University of Florida},
  \orgaddress{\city{Gainesville}, \state{FL}, \country{USA}}}

\affil[12]{\orgdiv{Department of Neurology}, \orgname{University of Florida},
  \orgaddress{\city{Gainesville}, \state{FL}, \country{USA}}}

\affil[13]{\orgdiv{Division of Nephrology, Department of Medicine}, \orgname{University of Florida},
  \orgaddress{\city{Gainesville}, \state{FL}, \country{USA}}}

\abstract{Clinical decision-making relies on identifying relevant patient information to guide diagnosis and treatment, a challenge that is especially difficult in the data-dense and rapidly changing intensive care unit (ICU). Large language models (LLMs) could support this task. However, existing applications and datasets mostly emphasize surface-level retrieval or factual recall rather than the inductive and deductive reasoning clinicians practice to select and reason over decision-relevant evidence. We hypothesized that training LLMs on expert ICU reasoning could yield clinical reasoning skills that generalize beyond critical care. Here we introduce ICU-REACT, a reasoning dataset developed with 19 clinicians through a clinician-in-the-loop framework to teach LLMs to perform information retrieval and context-aware clinical reasoning in the ICU. Using ICU-REACT, we fine-tuned Clin-REACT models spanning 8B-70B parameters and three model families. Across five clinical reasoning benchmarks, Clin-REACT consistently outperformed its backbone models and open-source general-purpose and medical LLMs. Gains extended to different tasks including script concordance tests, and downstream diagnosis and treatment tasks. These findings suggest that expert reasoning supervision in critical care can improve broader clinical reasoning, although prospective evaluation is needed before real-world clinical use.}



\maketitle

\begin{bibunit}[sn-mathphys-num]

\section{Introduction}\label{introduction}

The intensive care unit (ICU) is a fast-paced and constantly evolving environment where clinicians must make timely decisions based on large amounts of patient data. These decisions often require quickly identifying and interpreting relevant information from electronic health record (EHR) data, including vital signs, laboratory results, medications, and imaging reports \cite{lijovic_leveraging_2025}. Navigating these systems remains complex and time-consuming, limiting clinicians' ability to efficiently extract and synthesize decision-relevant information \cite{murray_medknowts_2021}.

Large language models (LLMs) have shown potential to reduce the burden of navigating EHR data by enabling natural-language interaction with patient records \cite{ahsan_retrieving_2024, shi_ehragent_2024, li_scoping_2024}. However, existing applications remain largely focused on surface-level tasks such as retrieving information from clinical notes or extracting discrete variables. Clinical reasoning entails substantially more complex tasks. Clinicians must identify relevant data to then generate a probability-ranked set of hypotheses consistent with the available observations, perform diagnostic testing to evaluate those hypotheses, and implement patient-specific treatment plans. In this sense, identifying decision-relevant information is itself part of the reasoning process rather than a separate retrieval task. Therefore, we hypothesized that the ICU, with its dense information, clinical complexity, and uncertainty, could provide a high-yield setting for teaching inductive and deductive clinical reasoning. By learning to identify the most relevant evidence and to perform multi-layered clinical reasoning with incomplete information, models may develop clinical reasoning skills that transfer across clinical domains.

Realizing this potential requires training resources that supervise not only the final clinical answer, but also how relevant evidence is selected and used to reach it. However, existing datasets primarily target general medical question-answering \cite{jin_what_2020, hendrycks_measuring_2021}, isolated EHR retrieval tasks \cite{fleming_medalign_2024, wu_instruction_2024}, entity linking \cite{zhao_evaluating_2025}, consistency checking \cite{kwon_ehrcon_2024}, or individual reasoning dimensions such as temporal reasoning in longitudinal records \cite{cui_timer_2025}. These resources have advanced clinical language modeling, but they offer limited supervision of how clinicians identify relevant patient information, place it in context, and use it to support a clinical decision. A similar pattern has been seen at the model level. Earlier medical LLMs were trained primarily to acquire and reproduce medical knowledge \cite{chen_meditron-70b_2023}, with performance commonly assessed using traditional multiple-choice benchmarks \cite{jin_what_2020, pal_medmcqa_2022, zuo_medxpertqa_2025, hendrycks_measuring_2021}. More recent medical LLMs increasingly incorporate reasoning-focused training \cite{sellergren_medgemma_2026, chen_towards_2025, wang_baichuan-m1_2025, team_baichuan-m2_2025}, representing an important shift beyond factual recall. Yet their training and evaluation still rely heavily on the same conventional multiple-choice benchmarks, which provide only a limited view of whether they can identify the most relevant information and reason through realistic patient contexts.

More recent benchmarks have begun to address this gap by evaluating more complex forms of clinical reasoning. Some datasets have focused on script concordance tests that assess agreement with expert judgment under uncertainty \cite{mccoy_assessment_2025}, while others have evaluated broader clinical reasoning using structured cases \cite{qiu_quantifying_2025, chiu_simulating_2025}. Other benchmarks have extended this paradigm to EHR-grounded reasoning in the emergency department \cite{mehandru_er-reason_2025}, and to sequential diagnosis requiring models to iteratively gather information, update hypotheses, and select subsequent tests \cite{nori_sequential_2025}. Importantly, a recent dataset introduced long-context ICU evaluation tasks centered on patient assessment and action recommendations \cite{shen_realicu_2026}. Together, these efforts represent important progress toward more realistic evaluation of clinical reasoning tasks, but they are still designed primarily to assess model performance rather than teach models how to find, prioritize, and reason over clinically relevant information. This limitation highlights the need for clinician-curated datasets that can support both the training and evaluation of multi-stage clinical reasoning.

To address this gap, we introduce ICU-REACT (Intensive Care Unit Reasoning for Electronic Health Record-Anchored Decision Support Tasks), a clinician-supervised dataset designed to teach LLMs how to identify and reason over decision-relevant information in realistic ICU scenarios. ICU-REACT was developed by first creating a seed set through a clinician-in-the-loop framework involving 19 clinicians across multiple specialties. The clinician-curated seed dataset was subsequently augmented for model training using a self-instruct methodology \cite{wang_self-instruct_2023, zhang_alpacareinstruction-tuned_2025}. Each sample links an actionable clinical question to a patient context, a set of relevant EHR variables, and a rationale explaining why those variables are important to the decision. Clinical variables are mapped to the Observational Medical Outcomes Partnership (OMOP) Common Data Model, providing a standardized representation for information retrieval across heterogeneous EHR systems. We use ICU-REACT to develop Clin-REACT (Clinical Reasoning for Electronic Health Record-Anchored Decision Support Tasks), a family of LLMs fine-tuned to identify decision-relevant patient information and generate explicit, context-grounded reasoning connecting that evidence to clinical decisions.

We then test our broader hypothesis: whether reasoning supervision confined to the ICU can improve clinical reasoning beyond the domain in which it was learned. Clin-REACT models are evaluated on the held-out ICU-REACT test set and four independent external clinical reasoning benchmarks spanning critical care, emergency medicine, and general clinical reasoning: SCT-Bench \cite{mccoy_assessment_2025}, ER-Reason \cite{mehandru_er-reason_2025}, MedRBench \cite{qiu_quantifying_2025}, and VivaBench \cite{chiu_simulating_2025}. Across this evaluation suite, Clin-REACT models consistently outperformed their backbone models and performed strongly against open-source general-purpose and medical LLMs. Notably, these gains extended beyond ICU information retrieval and reasoning to downstream diagnostic and treatment tasks in other clinical settings. These results support our hypothesis that clinician supervision in a complex, high-acuity environment can teach models skills that transfer beyond the setting in which they were trained. Together, ICU-REACT and Clin-REACT provide a framework for training and evaluating clinical LLMs not only on retrieving patient information or generating answers, but on identifying which evidence matters and reasoning over it. Prospective evaluation with practicing clinicians remains necessary to determine clinical utility, safety, and readiness for real-world use.

\section{Results}\label{results}

\begin{figure}[htbp]
\centering
\includegraphics[width=0.95\linewidth]{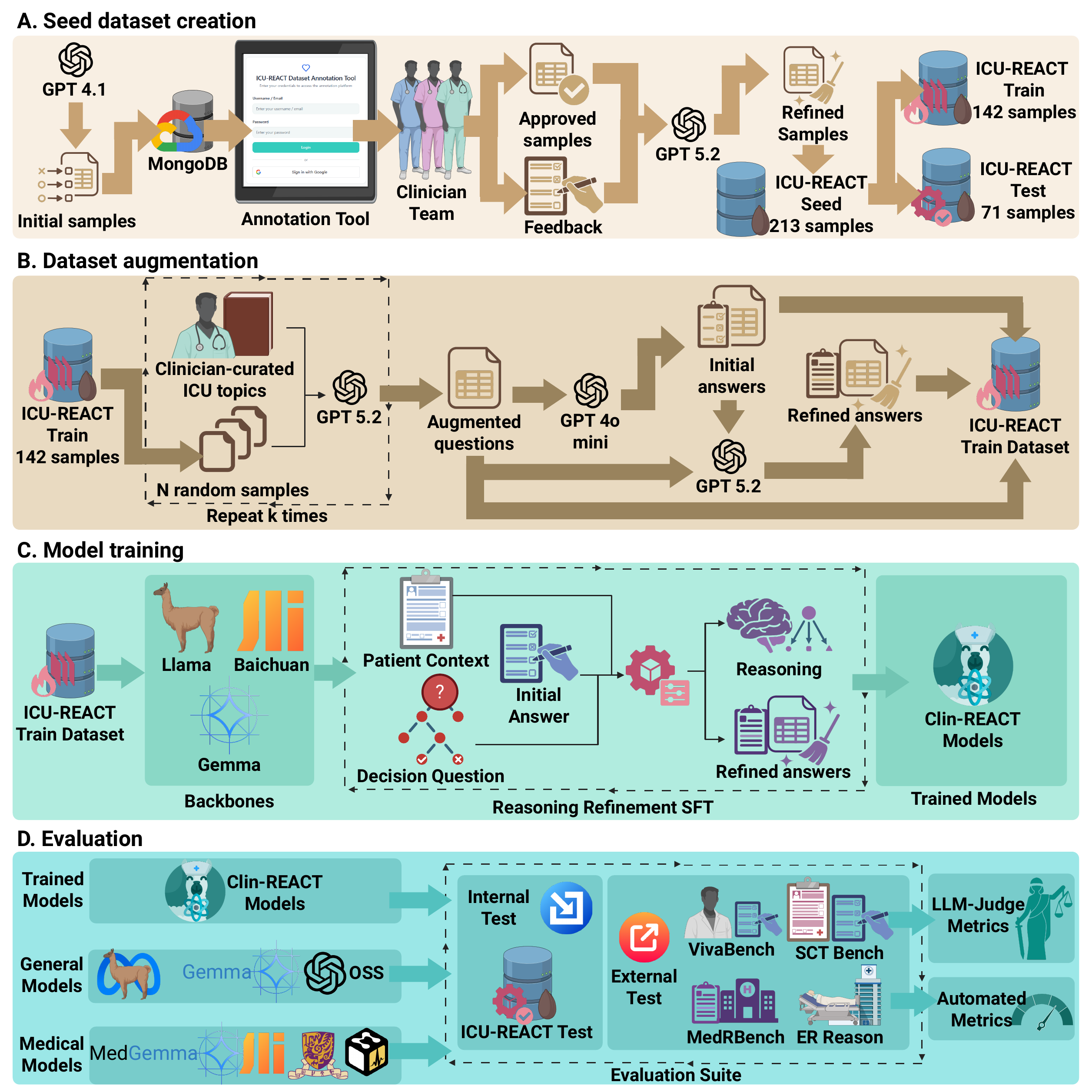}
\caption{Workflow overview of the ICU-REACT and Clin-REACT pipeline. (A) Seed dataset creation: GPT-4.1, selected for its state-of-the-art instruction-following performance at the time of initial generation, produces candidate samples that are stored in MongoDB and reviewed by clinicians through an online annotation tool. Approved samples are then refined with GPT-5.2, the state-of-the-art model available at the time of the experiments, using clinicians' written feedback to produce the ICU-REACT seed set ($n=213$). (B) Dataset augmentation: the seed set is divided into a training seed ($n=142$) and held-out test set ($n=71$). Clinician-curated ICU topics and $n$ randomly sampled training examples are used as few-shot context for GPT-5.2 to generate new questions. GPT-4o-mini is intentionally used to produce imperfect initial answers requiring refinement, after which GPT-5.2 generates improved answers using the question and initial response. This process is repeated for $k$ rounds until the target training-set size is reached. (C) Model training: Llama 3.1 8B Instruct, Baichuan M1 14B Instruct, Gemma 4 31B, and Llama 3.3 70B Instruct are fine-tuned through a single supervised fine-tuning stage to generate refined answers and corresponding reasoning from the patient context, decision question, and initial answer. ICU-REACT-Train-Small contained 10,000 samples and was used for the 8B model; ICU-REACT-Train-Medium contained 5,307 hard reasoning-refinement samples and was used for the 14B and 31B models; and ICU-REACT-Train-Large contained 27,973 reasoning-refinement samples and was used for the 70B model. Fine-tuning produced the Clin-REACT 8B, 14B, 31B, and 70B models. (D) Evaluation: Clin-REACT models are compared with open-source general-purpose and medical LLMs on the held-out ICU-REACT test set and four external benchmarks (SCT-Bench, ER-Reason, MedRBench, and VivaBench) using automated and LLM-judge metrics.}
\label{fig:workflow_overview}

\end{figure}

Figure~\ref{fig:workflow_overview} summarizes the overall workflow for developing ICU-REACT, training Clin-REACT, and evaluating clinical reasoning performance. We first created the ICU-REACT seed dataset using LLM-assisted generation, followed by structured review and refinement by 19 clinicians to ensure that the final samples reflected realistic ICU decision-making scenarios. We then augmented this seed set to create the training sets, fine-tuned Llama \cite{grattafiori_llama_2024}, Gemma \cite{team_gemma_2025}, and Baichuan \cite{wang_baichuan-m1_2025} backbones using low-rank adaptation (LoRA) supervised fine-tuning (SFT) to obtain Clin-REACT models, and evaluated performance against open-source general-purpose and medical LLMs on internal and external benchmarks.

\subsection{ICU-REACT captures multidimensional clinical reasoning, enabling reasoning-focused training}\label{results_dataset}

We created ICU-REACT, a decision-focused clinical reasoning and information retrieval dataset for the ICU, developed through a clinician-in-the-loop annotation framework for fine-tuning and benchmarking LLMs. Specifically, the seed set was curated through LLM-assisted generation and subsequently reviewed and refined by 19 clinicians across multiple specialties, resulting in 213 final samples, a size that is consistent with previous literature \cite{wang_self-instruct_2023, zhang_alpacareinstruction-tuned_2025}. The seed set was then split into training ($n=142$) and held-out test ($n=71$) sets. The training seed was augmented via a self-instruct pipeline into three variant-specific training sets: ICU-REACT-Train-Small (10,000 samples to train Clin-REACT 8B), Medium (5,307 hard reasoning-refinement samples to train Clin-REACT 14B and 31B), and Large (27,973 reasoning-refinement samples to train Clin-REACT 70B). The Small, Medium, and Large labels refer to the size of the Clin-REACT models they were designed to train, rather than the size of the training datasets. Each dataset was constructed to balance training-data scale and reasoning complexity with the capacity of the corresponding model architecture and size. Full composition and topic distributions for the training sets are provided in Supplementary Fig.~\ref{fig:train_datasets}.

For benchmarking, ICU-REACT-Test (the held-out test set) comprised 71 clinician-validated questions spanning nine critical-care topics, with greater representation of common ICU problems such as respiratory failure ($n = 16$), hemodynamic instability/shock ($n = 11$), and renal failure/electrolyte disorders ($n = 9$). Rather than testing isolated facts or individual EHR variables, questions were designed to require integration of multiple aspects of the patient record that clinicians routinely consider together when making decisions. A typical question required information from 4 clinical data categories (IQR 3-5), 7 more specific data sub-domains (IQR 5-10), and 18 individual patient variables (IQR 13–26.5). These commonly included vital signs, laboratory results, medications, and other physiologic measurements, reflecting the multidimensional information routinely synthesized during ICU care. Detailed test-set composition, topic distributions, and data-category breakdowns are provided in Supplementary Fig.~\ref{fig:test_dataset}.

\begin{figure*}[h]
\centering
\includegraphics[width=0.85\linewidth]{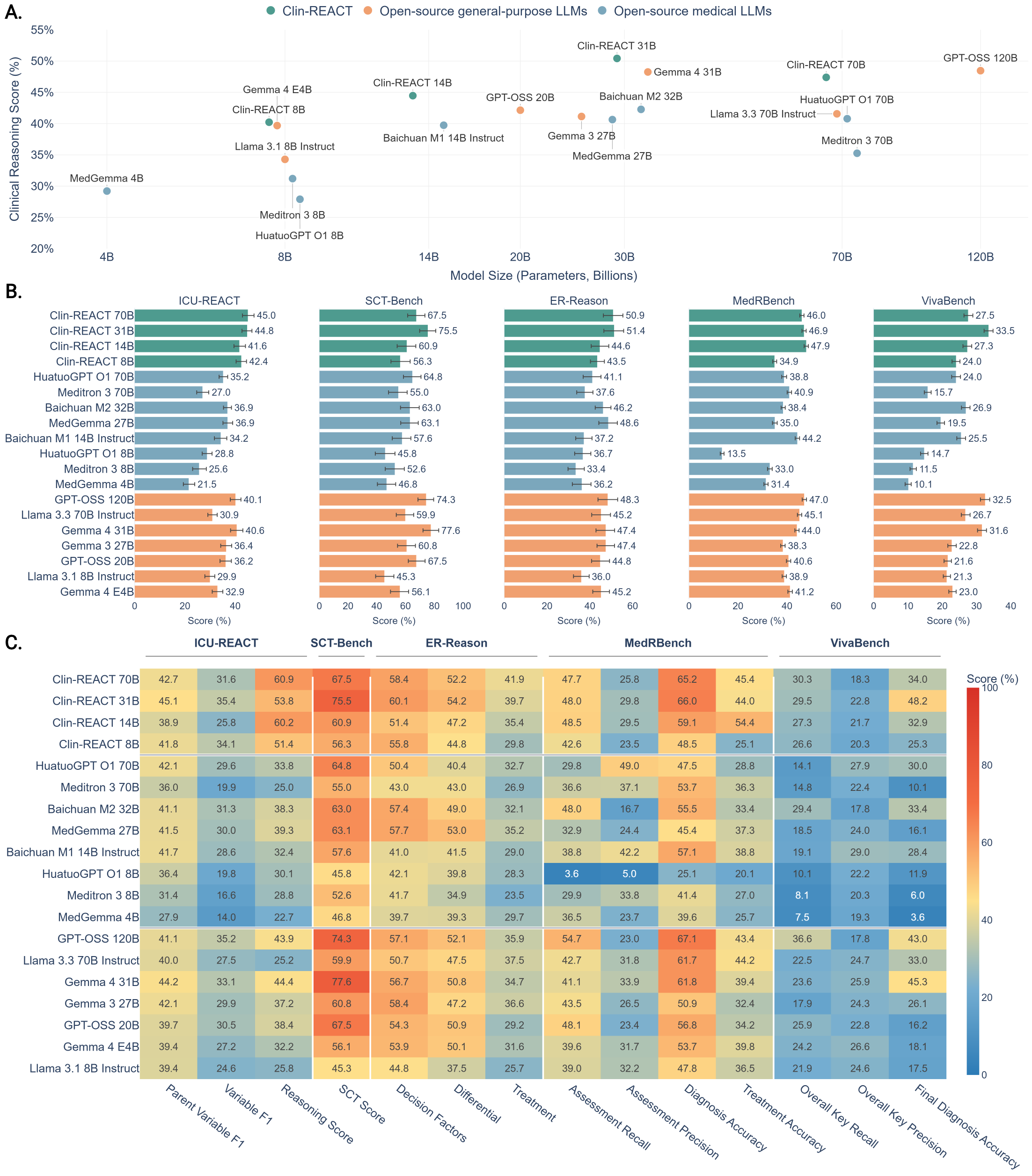}
\caption{Overall performance. (A) Relationship between model size and macro-average performance across all five clinical reasoning benchmarks evaluated (ICU-REACT, SCT-Bench, ER-Reason, MedRBench, and VivaBench). Models are colored by group: Clin-REACT models, open-source general-purpose LLMs, and open-source medical LLMs. (B) Benchmark-level performance of all models on ICU-REACT, SCT-Bench, ER-Reason, MedRBench, and VivaBench. Bars denote mean scores across each benchmark's metrics, and error bars represent 95\% confidence intervals. (C) Metric-level heatmap summarizing performance across the individual evaluation dimensions within each benchmark. Models are ordered by group and performance is displayed as percentage scores.}
\label{fig:overall_perform}

\end{figure*}

\subsection{Clin-REACT training transfers across diverse clinical reasoning benchmarks}\label{results_overall_perform}

Clin-REACT training improved performance beyond the in-domain ICU-REACT tasks, with Clin-REACT models generally performing better across four external clinical-reasoning benchmarks with different task formats and evaluation criteria (Fig.~\ref{fig:overall_perform}). Clin-REACT 31B achieved the highest macro score across all five benchmarks at 50.4 ($\pm$15.5 SD), compared with 48.5 ($\pm$15.8) for GPT-OSS 120B and 48.3 ($\pm$17.5) for Gemma 4 31B. As expected, the largest gains were seen on ICU-REACT, where the three best-performing models were Clin-REACT variants. Clin-REACT 70B achieved the highest score at 45.0 (95\% CI: 42.8-47.4), compared with 40.6 (38.1-43.0) for the strongest baseline.

Importantly, these gains carried over to benchmarks that differed from ICU-REACT in clinical setting, task structure, and scoring methodology. Clin-REACT models achieved the highest overall scores on three of the four external benchmarks, with the best-performing variants reaching 51.4 (46.5-55.8) on ER-Reason, 47.9 (47.0-48.7) on MedRBench, and 33.5 (32.2-34.8) on VivaBench. They also remained competitive on SCT-Bench: Clin-REACT 31B scored 75.5 (69.8-81.0), compared with 77.6 (72.1-83.1) for the top-performing model, while Clin-REACT 70B outperformed the strongest medical model (67.5 vs.\ 64.8). Together, these results suggest that the skills learned during Clin-REACT training extended beyond the source benchmark to a broader range of clinical-reasoning tasks.

Metric-level results showed a similar pattern. On ICU-REACT, Clin-REACT models showed their largest gains in information retrieval and reasoning, including the highest parent-variable F1 (45.1, 95\% CI: 42.0-48.3) and reasoning score (60.9, 58.2-63.4). On the external benchmarks, these gains appeared in different ways, including better identification of decision factors, differential reasoning, and treatment planning on ER-Reason; stronger assessment recommendation recall and diagnosis and treatment accuracy on MedRBench; and higher key-information recall and final-diagnosis accuracy on VivaBench. Thus, improvements in clinically grounded retrieval and reasoning transferred to tasks requiring different combinations of information selection, diagnosis, and treatment planning.

When Clin-REACT models did not achieve the highest score on a given metric, the differences often reflected precision-recall tradeoffs rather than a clear loss of capability. For example, on MedRBench, Clin-REACT 31B had lower assessment-recommendation recall than GPT-OSS 120B (48.0 vs.\ 54.7) but higher precision (29.8 vs.\ 23.0), while achieving similar diagnosis accuracy (66.0 vs.\ 67.1). A similar pattern appeared on VivaBench, where Clin-REACT 31B showed higher key-information precision but lower recall than GPT-OSS 120B. Overall, the consistency of these results across diverse external benchmarks suggests that Clin-REACT training yielded transferable clinical-reasoning skills rather than simply improving performance on ICU-REACT. Detailed model- and metric-level comparisons with confidence intervals and statistical testing are provided in Supplemental Section~\ref{all_baseline_comparisons}, with comparisons against four proprietary frontier LLMs reported in Supplemental Section~\ref{compare_frontier}.

\begin{figure*}[h]
\centering
\includegraphics[width=0.85\linewidth]{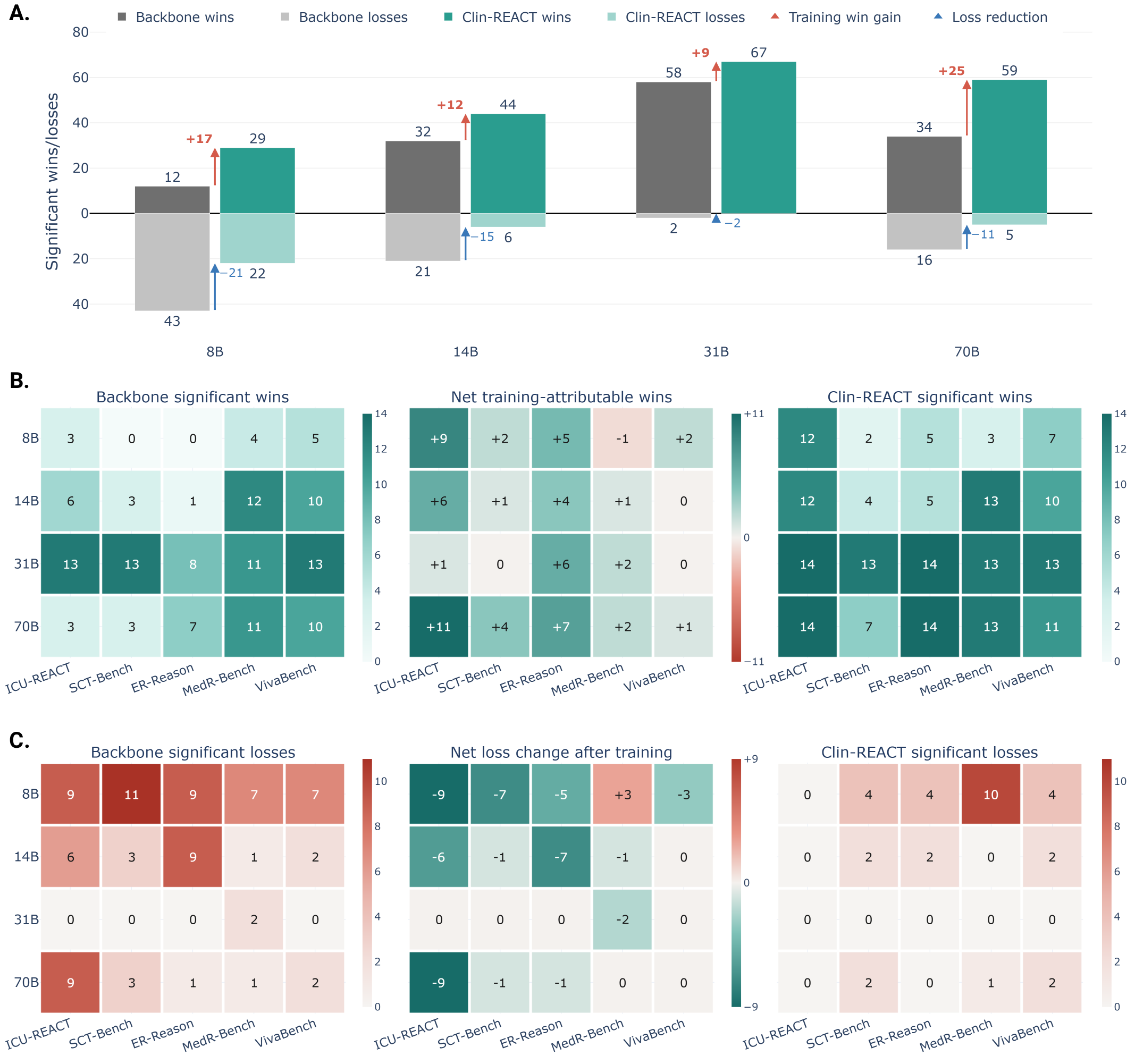}
\caption{Clin-REACT performance gains. Each Clin-REACT model is compared against its untrained backbone (8B, 14B, 31B, and 70B) across five benchmarks. A "win" is a pairwise comparison in which a model significantly outperforms a competing model, and a "loss" is one in which it is significantly outperformed; significance was determined using two-sided Wilcoxon signed-rank tests ($p < 0.05$). Comparing a backbone's win/loss counts against its corresponding fine-tuned Clin-REACT model isolates the effect of ICU-REACT training. (A) Aggregate counts of significant wins (bars above the axis) and losses (bars below the axis) summed across all benchmark comparisons, shown for each backbone (gray) and its corresponding Clin-REACT model (teal). The red arrow marks the increase in wins and the blue arrow the decrease in losses attributable to training. Across all four scales, Clin-REACT increases wins and reduces losses. (B) Significant wins broken down by benchmark: backbone wins (left), Clin-REACT wins (right), and the net difference between them (middle). In the middle heatmap, positive values indicate additional wins gained through training and negative values indicate wins lost. (C) Significant losses broken down by benchmark: backbone losses (left), Clin-REACT losses (right), and the net change (middle). In the middle heatmap, negative values indicate fewer losses after training (an improvement in performance) and positive values indicate more losses (a decrease in performance).}
\label{fig:perform_gains}

\end{figure*}

\subsection{Fine-tuning consistently improves performance over backbone models across clinical reasoning benchmarks}\label{results_baseline_compare}

Pairwise comparisons with 14 baseline models across five clinical-reasoning benchmarks showed that Clin-REACT training consistently improved the performance of each backbone model (Fig. \ref{fig:perform_gains}). Across model sizes, Clin-REACT variants achieved more significant wins and fewer significant losses than their corresponding backbones, with improvements extending beyond ICU-REACT to external reasoning tasks.

The gains were especially clear for the 8B and 70B models. Clin-REACT 8B increased from 12 significant wins with the Llama 3.1 8B Instruct backbone to 29, while reducing significant losses from 43 to 22. The largest improvement was on ICU-REACT (+9 wins), but gains were also seen on ER-Reason (+5), SCT-Bench (+2), and VivaBench (+2). Similarly, Clin-REACT 70B increased from 34 to 59 significant wins and reduced losses from 16 to 5 relative to Llama 3.3 70B Instruct. These improvements were seen across all five benchmarks, including substantial gains on ER-Reason (+7) and SCT-Bench (+4).

Clin-REACT training also strengthened the already competitive 14B and 31B backbones. Clin-REACT 14B increased significant wins from 32 to 44 and reduced losses from 21 to 6, with improvements extending beyond ICU-REACT to ER-Reason, SCT-Bench, and MedRBench. Clin-REACT 31B showed the strongest comparison profile, with 67 significant wins and no significant losses across 70 comparisons. Although its Gemma 4 31B backbone was already highly competitive, with 58 wins and only two losses, Clin-REACT training added nine wins and dropped both losses, with most of the additional gains occurring on ER-Reason and MedRBench.

Overall, these pairwise comparisons show that Clin-REACT training improved the relative performance of every backbone across a diverse set of clinical-reasoning tasks. The consistency of these improvements across model sizes and external benchmarks provides further evidence that the reasoning and information-retrieval skills learned during training transferred beyond the ICU setting. Heatmaps showing the magnitude of differences between each Clin-REACT variant and all baseline models, including gains over their corresponding backbones, are provided in Supplemental Section~\ref{all_baseline_comparisons}.

\begin{figure*}[h]
\centering
\includegraphics[width=0.95\linewidth]{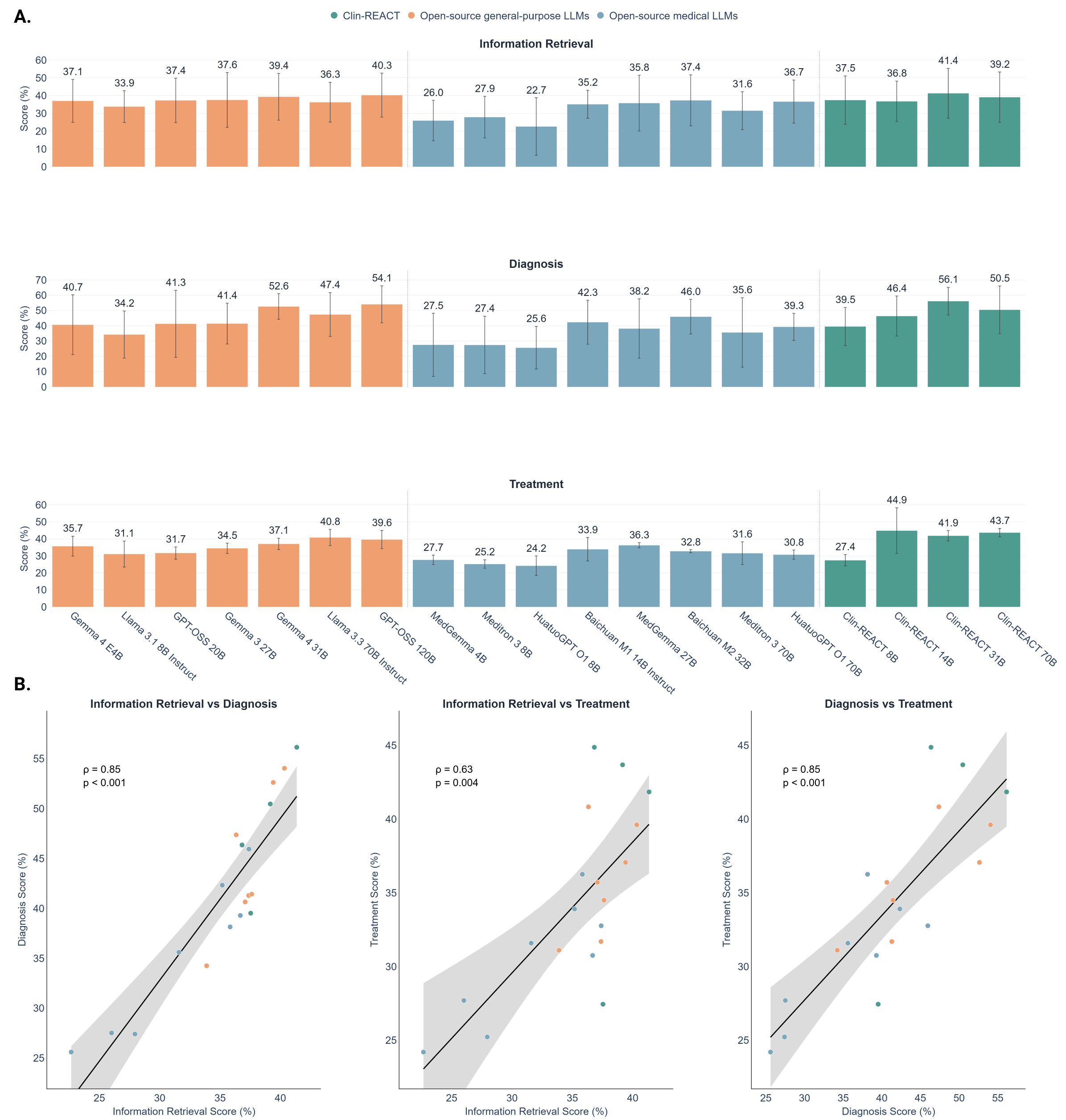}
\caption{Performance across different task types. (A) Task-level scores for information retrieval, diagnosis, and treatment across Clin-REACT models, open-source general-purpose LLMs, and open-source medical LLMs. Task scores were calculated by aggregating relevant benchmark metrics within each task domain. Bars denote mean performance across included metrics, and error bars denote the standard deviation across metrics. (B) Pairwise associations between model performance across information retrieval, diagnosis, and treatment categories. Points represent individual models, colored by model group, and fitted lines show the estimated linear relationships with shaded 95\% confidence intervals. Associations were evaluated using Spearman rank correlation tests, with correlation coefficients and corresponding p-values shown in each panel.}
\label{fig:perform_task}

\end{figure*}

\subsection{Performance gains extend from information retrieval reasoning to downstream diagnosis and treatment}\label{results_perform_task}

To assess whether Clin-REACT performance generalized across different aspects of clinical reasoning, benchmark-specific metrics were grouped into three task categories: information retrieval, diagnosis, and treatment (Fig.~\ref{fig:perform_task}A). Information retrieval included ICU-REACT parent variable F1 and variable F1 scores, ER-Reason decision factors accuracy, MedRBench assessment recommendation precision and recall, and VivaBench overall key-information precision and recall. Diagnosis included ER-Reason differential accuracy, MedRBench diagnosis accuracy, and VivaBench final diagnosis accuracy. Treatment included ER-Reason treatment planning accuracy and MedRBench treatment accuracy.

Clin-REACT models performed strongly across all three task categories, rather than showing gains only in information retrieval, the most emphasized task during training. Clin-REACT 31B achieved the highest overall scores for information retrieval and diagnosis, at 41.4 (SD 14.0) and 56.1 (SD 9.1), respectively, while Clin-REACT 14B achieved the highest treatment score at 44.9 (SD 13.4). Clin-REACT 70B was also competitive across all three categories. Together, these results show that Clin-REACT variants matched or outperformed the strongest general-purpose and medical baselines across information retrieval, diagnosis, and treatment, suggesting that the skills learned during training extended to multiple stages of clinical reasoning.

Comparisons with the corresponding backbone models showed that improvements on these tasks was attributable to Clin-REACT training. Clin-REACT 31B and 70B improved over Gemma 4 31B and Llama 3.3 70B Instruct, respectively, across all three categories. Clin-REACT 14B also improved in retrieval and diagnosis and showed a particularly large gain in treatment performance relative to Baichuan M1 14B Instruct (44.9 vs.\ 33.9). Clin-REACT 8B improved in retrieval and diagnosis compared with Llama 3.1 8B Instruct, although treatment performance had a slight decrease (27.4 vs.\ 31.1). Overall, these findings suggest that the benefits of Clin-REACT training generally extended beyond information retrieval to downstream diagnosis and treatment tasks.

Performance across the three task categories was also positively associated across models (Fig.~\ref{fig:perform_task}B). Information-retrieval and diagnosis scores were strongly correlated ($\rho=0.85$, $p<0.001$), as were diagnosis and treatment scores ($\rho=0.85$, $p<0.001$); information retrieval was moderately correlated with treatment performance ($\rho=0.63$, $p=0.004$). These associations indicate that models that were better at identifying clinically relevant information also tended to perform better on downstream diagnosis and treatment tasks, highlighting information selection as an important step of clinical reasoning.

\begin{figure*}[h]
\centering
\includegraphics[width=0.90\linewidth]{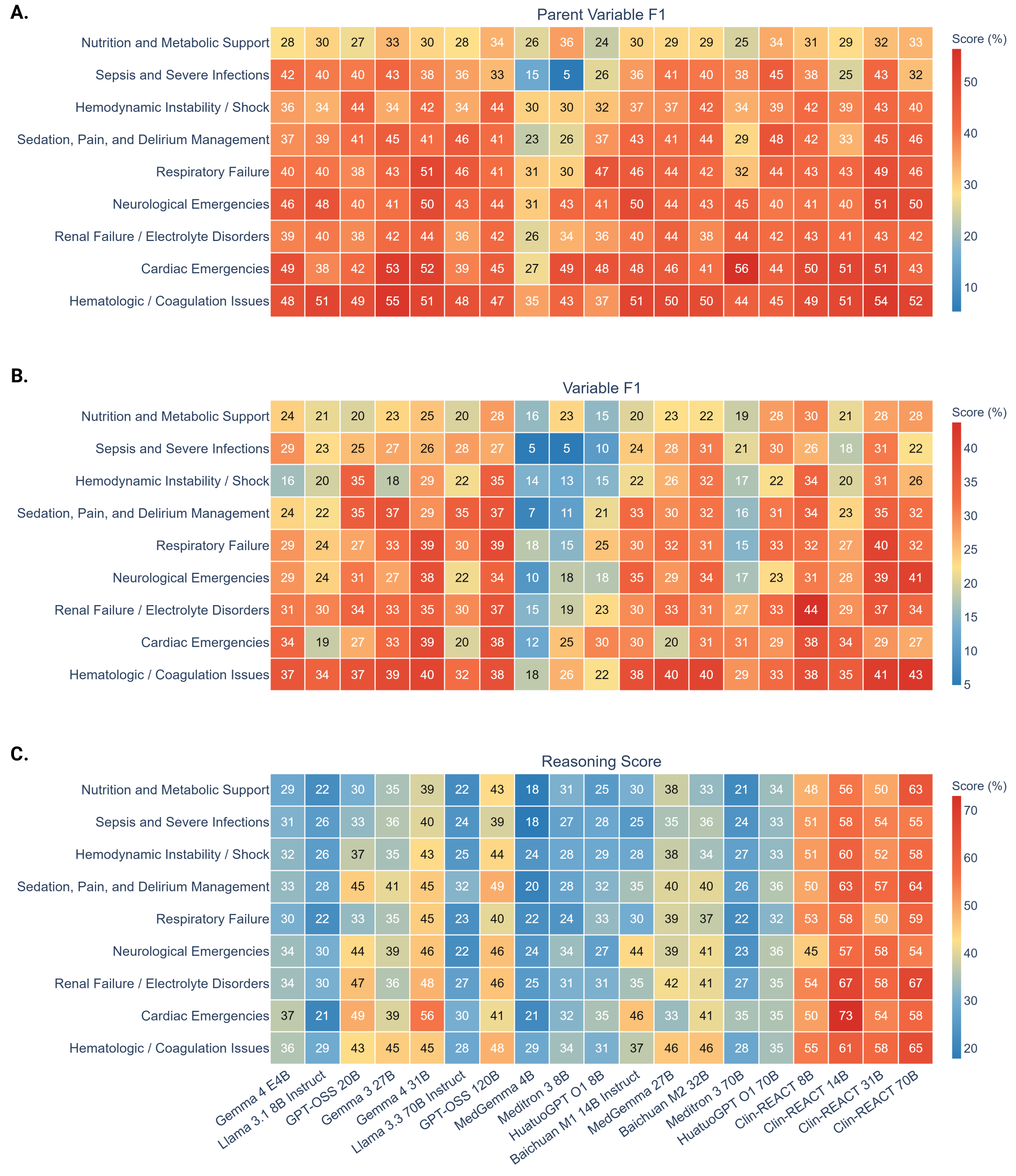}
\caption{Performance across topics in ICU-REACT test set. Heatmaps showing model performance across the nine topic areas in ICU-REACT test set for (A) parent variable F1, (B) variable F1, and (C) reasoning score. Rows denote ICU topics and columns denote evaluated models, ordered from left to right with open-source general-purpose LLMs followed by open-source medical LLMs and Clin-REACT models. Cell values represent percentage scores, with warmer colors indicating higher performance.}
\label{fig:perform_topics}

\end{figure*}

\subsection{Performance improvements are consistent across diverse clinical content}\label{results_perform_content}

Performance across the nine ICU topic categories showed that Clin-REACT improvements were spread across a broad range of clinical problems rather than driven by only a few conditions (Fig.~\ref{fig:perform_topics}). The clearest pattern was in reasoning quality, where a Clin-REACT variant achieved the highest score in every ICU topic. These gains were seen across major critical-care areas, including sepsis and severe infections, hemodynamic instability and shock, respiratory failure, renal and electrolyte disorders, cardiac emergencies, hematologic and coagulation disorders, and sedation, pain, and delirium management. For example, Clin-REACT achieved reasoning scores of 58 versus 40 for the strongest baseline in sepsis and severe infections, 60 versus 44 in hemodynamic instability and shock, 59 versus 45 in respiratory failure, and 67 versus 48 in renal failure and electrolyte disorders.

Information-retrieval performance varied more across topics but remained competitive with the strongest baselines. Clin-REACT models achieved the highest variable-level retrieval scores in several high-acuity areas, including respiratory failure and renal failure and electrolyte disorders, and matched or closely approached the strongest baselines in sepsis and severe infections and hemodynamic instability and shock. Parent-variable retrieval followed a similar pattern, with the best Clin-REACT variants generally within 1-2 points of the top baseline across these major ICU domains.

Overall, these results show that Clin-REACT training improved reasoning across a wide range of ICU presentations while maintaining strong retrieval of clinically relevant information. Similar gains were also seen across related clinical domains in the external benchmarks, further supporting the transfer of these capabilities beyond the ICU setting (Supplemental Section~\ref{cat_perform}).

\begin{figure*}[h]
\centering
\includegraphics[width=0.80\linewidth]{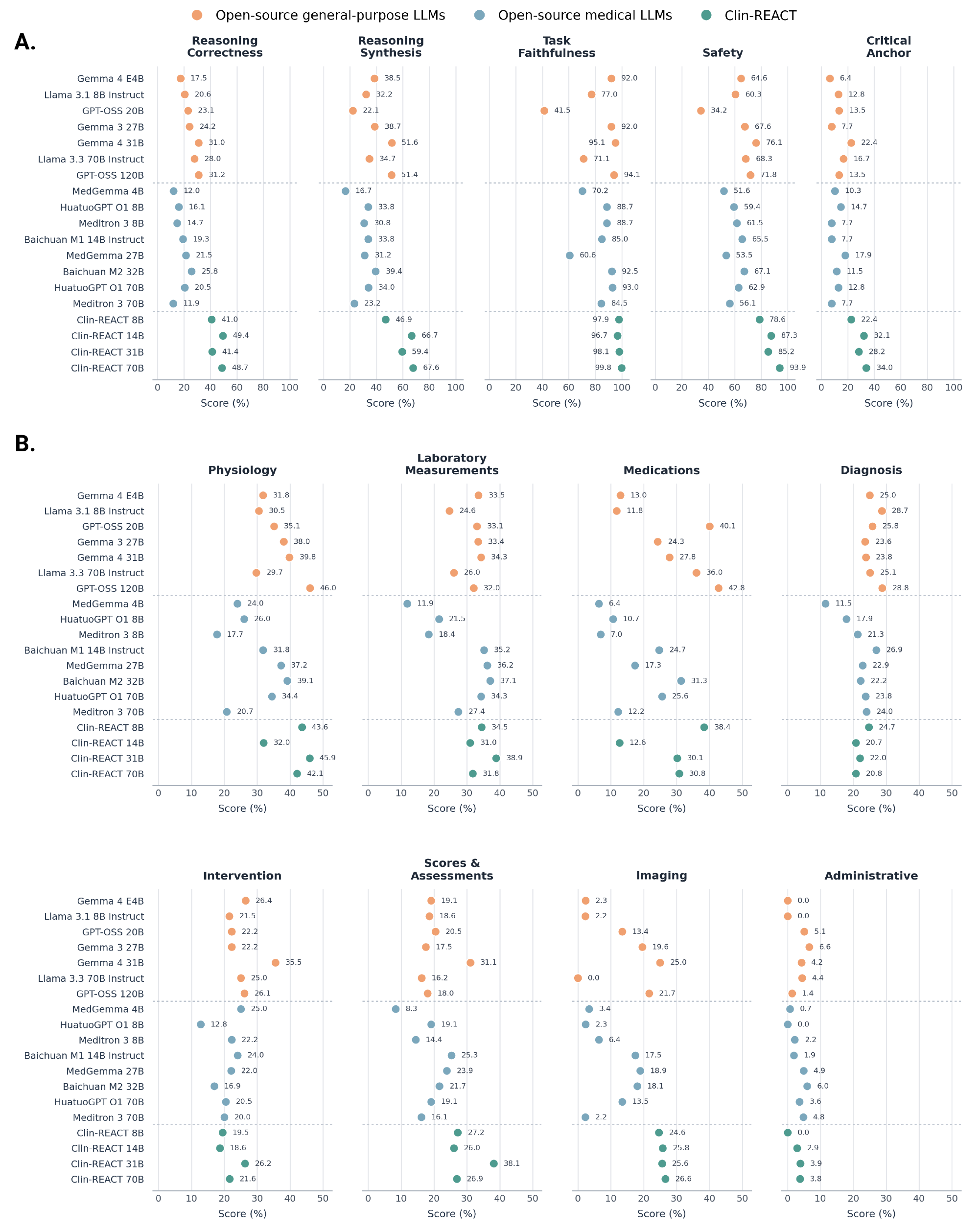}
\caption{Performance across rubric dimensions and clinical data categories. (A) Cleveland dot plots summarizing model performance across the five ICU-REACT rubric dimensions: reasoning correctness, reasoning synthesis, task faithfulness, safety, and critical anchor. (B) Cleveland dot plots showing variable-level F1 performance stratified by the clinical data categories represented in ICU-REACT questions, including physiology, laboratory measurements, medications, diagnoses, interventions, scores and assessments, imaging, and administrative information. Each dot represents the score for a given model on a specific dimension or category. Models are grouped by open-source general-purpose LLMs, open-source medical LLMs, and Clin-REACT models.}
\label{fig:perform_dim_cat}

\end{figure*}

\subsection{Training improves scores across evaluation dimensions}\label{results_perform_dimensions}

To better understand how Clin-REACT training changed the quality of model responses beyond overall benchmark scores, we evaluated five evaluation dimensions on ICU-REACT: reasoning correctness, reasoning synthesis, task faithfulness, safety, and critical-anchor identification (Fig.~\ref{fig:perform_dim_cat}A). Clin-REACT models consistently outperformed general-purpose and medical baselines in both reasoning correctness and synthesis, with correctness scores ranging from 41.0\% to 49.4\% and synthesis scores from 46.9\% to 67.6\%. These results suggest that Clin-REACT models were not only more likely to provide clinically appropriate explanations, but also better at bringing relevant evidence together into a coherent rationale.

The clearest advantages were seen in task faithfulness and safety. Clin-REACT models achieved task-faithfulness scores of 96.7\%-99.8\%, compared with 41.5\%-95.1\% for general-purpose models and 60.6\%-93.0\% for medical models. Safety followed a similar pattern, with Clin-REACT scores (78.6\%-93.9\%) exceeding both general-purpose (34.2\%-76.1\%) and medical (51.6\%-67.1\%) baselines.

Identifying critical anchors remained the most difficult dimension for all models, although Clin-REACT still achieved the strongest overall performance (22.4\%-34.0\%). Overall, these findings show that Clin-REACT training improved not only the correctness and coherence of clinical reasoning, but also how well models followed the intended task and avoided unsafe responses. At the same time, reliably identifying the most clinically important evidence remains an important area for improvement.

\subsection{Information retrieval scores improve across multiple clinically relevant data categories}\label{results_perform_data_categories}

Information-retrieval performance differed across clinical data types, but Clin-REACT models showed clear strengths in retrieving laboratory measurements, structured scores and assessments, and imaging findings (Fig.~\ref{fig:perform_dim_cat}B). A Clin-REACT variant achieved the highest F1 score among all evaluated models in each of these categories, reaching 38.9\% for laboratory measurements, 38.1\% for scores and assessments, and 26.6\% for imaging. Clin-REACT models also remained competitive for physiology and medications, nearly matching the strongest general-purpose model for physiology (45.9\% vs.\ 46.0\%) and outperforming the strongest medical baseline for medication retrieval.

Performance was more variable for diagnosis, intervention, and administrative information. Clin-REACT models remained competitive with medical baselines for intervention retrieval but had lower performance compared to the strongest general-purpose model, while diagnosis retrieval was modestly lower than the best general-purpose and medical baselines. Administrative information was challenging for all models and had the lowest retrieval scores overall. These results show that Clin-REACT training improved retrieval across several clinically important data types, while also highlighting diagnosis, intervention, and administrative information as areas for further improvement. Similar category-level gains were seen on external benchmarks, suggesting that these retrieval skills extended beyond the ICU-REACT test set (Supplemental Section~\ref{info_seek_cat}).

\begin{figure*}[h]
\centering
\includegraphics[width=1.0\linewidth]{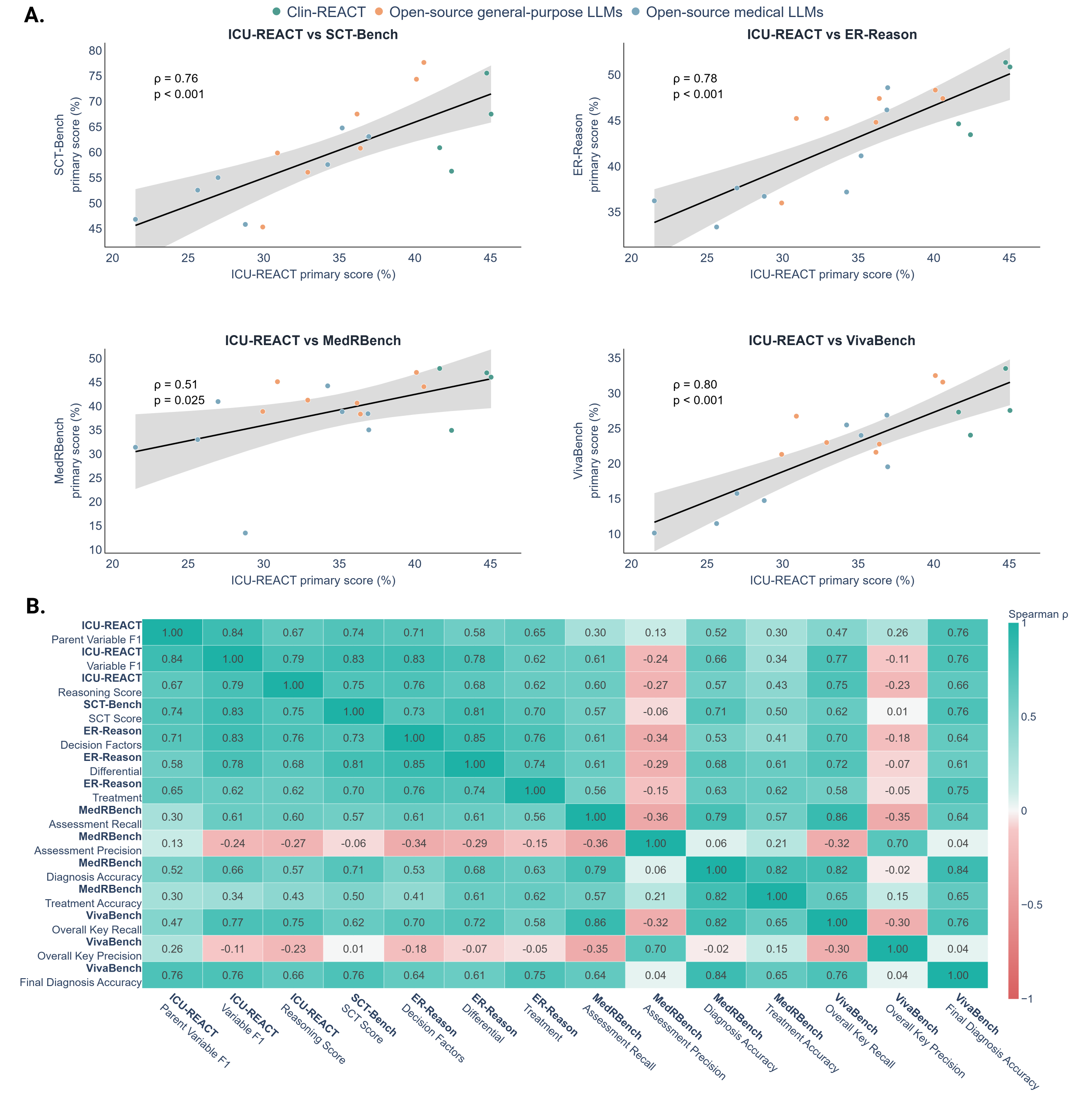}
\caption{Performance correlations between clinical reasoning benchmarks. (A) Associations between models’ ICU-REACT average scores and their average scores on SCT-Bench, ER-Reason, MedRBench, and VivaBench. Points represent individual models and are colored by model group, with solid lines indicating fitted linear trends with shaded 95\% confidence intervals. Associations were evaluated using Spearman rank correlation tests, with correlation coefficients and corresponding p-values shown in each panel. (B) Heatmap of pairwise Spearman rank correlations among individual metrics across the five clinical reasoning benchmarks evaluated. Cell values denote correlation coefficients, with positive and negative associations represented by teal and red shading, respectively.}
\label{fig:perform_corr}

\end{figure*}

\subsection{ICU-REACT performance aligns with external clinical reasoning benchmarks}\label{results_corr_bench}

To assess concordance between ICU-REACT and external measures of clinical reasoning, we calculated model-level Spearman rank correlations between ICU-REACT performance and scores on SCT-Bench, ER-Reason, MedRBench, and VivaBench (Fig.~\ref{fig:perform_corr}). ICU-REACT showed strong positive correlations with SCT-Bench (\(\rho=0.76\), ($p<0.001$)), ER-Reason (\(\rho=0.78\), ($p<0.001$)), and VivaBench (\(\rho=0.80\), ($p<0.001$)), and a more moderate but significant correlation with MedRBench (\(\rho=0.51\), ($p=0.025$)). Models that performed well on ICU-REACT also tended to perform well on external clinical-reasoning benchmarks, suggesting that ICU-REACT captures skills that are relevant to broader clinical reasoning.

Metric-level correlations showed a similar pattern across reasoning dimensions (Fig.~\ref{fig:perform_corr}B). ICU-REACT retrieval and reasoning metrics were strongly correlated to one another and were positively associated with external measures of decision-factor identification, differential diagnosis, treatment planning, and final diagnosis. The strongest cross-benchmark relationships were seen with SCT-Bench, ER-Reason decision-factor performance, and VivaBench final diagnosis accuracy, while diagnosis and treatment metrics were also closely aligned across MedRBench and VivaBench. In contrast, precision metrics showed weaker or occasionally inverse relationships with recall measures, reflecting differences in retrieval behavior and precision-recall tradeoffs across models. Overall, these correlations suggest that ICU-REACT measures reasoning abilities that overlap with established clinical-reasoning benchmarks.

\section{Discussion}\label{discussion}

In this study, we developed ICU-REACT, a decision-focused dataset for evaluating information retrieval and clinical reasoning in critical care, and Clin-REACT, a suite of fine-tuned LLMs optimized for information retrieval and reasoning in the ICU. The ICU-REACT dataset was designed to improve clinical reasoning of LLMs and address an important challenge in evaluating LLMs for critical care: assessing whether models can identify relevant patient information under different decision scenarios and use it to generate clinically coherent, task-aligned, and safe reasoning in complex ICU scenarios \cite{hager_evaluation_2024, shi_large_2025}. Our results demonstrate that reasoning-refinement training on a domain-constrained dataset can yield broad, transferable improvements in clinically relevant reasoning and information-selection behavior that extend well beyond the ICU setting, while surpassing strong open-source general-purpose and medical LLMs.

A central finding of this study is that Clin-REACT models improved clinical reasoning performance not only on in-domain ICU tasks but also across external benchmarks that were not part of training. Although ICU-REACT is constrained to critical-care scenarios, Clin-REACT models achieved leading or highly competitive performance on ER-Reason, a related critical-care benchmark, as well as on more general clinical reasoning benchmarks such as SCT-Bench, MedRBench, and VivaBench. For instance, Clin-REACT 31B achieved the highest overall macro score across all benchmarks (50.4 ± 15.5 SD) and the best performance on ER-Reason (51.4) and VivaBench (33.5), while remaining competitive on SCT-Bench and MedRBench (Fig.~\ref{fig:overall_perform}). Importantly, these improvements were not limited to tasks that closely resembled ICU-REACT or to a single type of evaluation. The external benchmarks tested different aspects of clinical reasoning, including decision-making under uncertainty, diagnosis and treatment selection, information seeking, and concept-level reasoning, using scoring approaches ranging from agreement with expert response distributions to retrieval and accuracy deterministic metrics. The fact that Clin-REACT improved across these diverse tasks and evaluation methods makes it less likely that the gains were driven only by memorization, dataset-specific response patterns, or alignment with an LLM evaluator. Instead, the results suggest that the reasoning skills learned from ICU-REACT transferred beyond the training setting, improving clinically relevant reasoning and information selection across different clinical contexts and task formats.

Clin-REACT training produced consistent improvements over each model's respective backbone across model families and scales. Relative to their backbones, Clin-REACT variants gained net statistically significant wins on our pairwise benchmark comparisons at every scale, ranging from 9 net wins for Clin-REACT 31B over its already strong Gemma 4 31B backbone to 25 net wins for Clin-REACT 70B over Llama 3.3 70B Instruct (Fig.~\ref{fig:perform_gains}). Notably, Clin-REACT 31B achieved 67 significant wins with no significant losses across 70 comparisons, and removed both of the losses observed for its backbone. The gains on SCT-Bench provide a particularly clear example of this transfer. Rather than simply selecting a diagnosis or semantically matching an expected response, SCT-Bench asks models to quantify how new clinical information changes the likelihood of a diagnostic hypothesis under uncertainty. Despite this task formulation not being explicitly represented in ICU-REACT training, Clin-REACT improved over its respective backbone by +11.0\% for the 8B model ($p < 0.01$), +3.3\% for the 14B model ($p > 0.05$), and +7.6\% for the 70B model ($p < 0.01$), showing better agreement with clinician response distributions and providing evidence that the learned reasoning behaviors generalized beyond the structure of the training task itself. These improvements were seen across different model families (Llama, Gemma, and Baichuan) indicating that the benefits of ICU-REACT training were not specific to a single architecture. The magnitude of the gains, however, varied across models. Stronger backbones, such as Gemma 4 31B, showed smaller improvements, given they already performed well before fine-tuning. On the other hand, models with more room to improve showed larger gains. Importantly, our data contamination analyses found no evidence of substantial overlap or benchmark leakage between ICU-REACT training data and the external benchmarks that could readily explain these results (Supplemental Section~\ref{data_contamination}), supporting the interpretation that the improvements reflect transferable learning rather than memorization.

Notably, these gains were achieved through lightweight low-rank adaptation (LoRA) fine-tuning rather than extensive retraining. Clin-REACT models were therefore highly parameter-efficient, with smaller variants often matching or outperforming much larger general-purpose and medical LLMs. For example, Clin-REACT 14B achieved higher overall clinical reasoning performance than MedGemma 27B and Baichuan M2 32B, while Clin-REACT 8B exceeded larger medical models such as Baichuan M1 14B Instruct and Meditron 3 70B. This pattern also held at larger model sizes. Clin-REACT 31B achieved the highest overall performance, outperforming HuatuoGPT O1 70B and GPT-OSS 120B despite having roughly one quarter as many parameters as the latter. Clin-REACT 70B also clearly outperformed HuatuoGPT O1 70B and exceeded GPT-OSS 120B on ICU-REACT and ER-Reason. The fact that relatively lightweight fine-tuning could produce models that matched or surpassed much larger and more heavily trained systems suggests that the gains came from the quality of ICU-REACT and the reasoning-refinement training approach, rather than model size or training scale alone.

Beyond information retrieval, Clin-REACT models also improved on downstream diagnosis and treatment tasks. In our task-level analysis, Clin-REACT models achieved the strongest or highly competitive performance across all three categories. Clin-REACT 31B had the highest scores for information retrieval (41.4) and diagnosis (56.1), while Clin-REACT 14B achieved the highest treatment score (44.9) (Fig.~\ref{fig:perform_task}A). Relative to their backbones, Clin-REACT models improved information retrieval and diagnostic performance at all four scales. This pattern is reinforced by the moderate-to-strong correlations we observed among task categories (information retrieval and diagnosis, $\rho=0.85$; diagnosis and treatment, $\rho=0.85$; information retrieval and treatment, $\rho=0.63$) (Fig.~\ref{fig:perform_task}B). One possible explanation for these gains is that better identification and prioritization of clinically relevant information gives models stronger evidence for subsequent diagnostic and treatment decisions. This may help explain why Clin-REACT training improved not only information retrieval, but also diagnostic performance on external benchmarks such as MedRBench and VivaBench. Models that more effectively recognize which findings are relevant may also be better positioned to reach the correct diagnosis or treatment decision. Although these associations do not establish that improved retrieval directly causes better downstream performance, they suggest that the benefits of Clin-REACT training extend across multiple stages of clinical decision-making rather than being limited to information extraction alone.

The reasoning-refinement approach seemed to play an important role in these results. Among the training tasks we tested, asking models to improve a flawed initial response had the best results, outperforming alternatives such as variable selection and context or question generation (see ablations in Supplemental Section~\ref{ablations}). One possible explanation is that models learn more from correcting imperfect reasoning than from simply imitating correct answers, since the refinement process exposes them to failure modes and the steps needed to fix them \cite{alazraki_no_2025, an_can_2024}. Our t-SNE analyses support this interpretation, showing greater diversity among refined reasoning texts than initial reasoning texts (Supplementary Fig.~\ref{fig:train_datasets}), suggesting that refinement provides a richer training signal. More broadly, these findings support a framework in which clinician-curated reasoning serves as the foundation, LLM-based augmentation amplifies that signal, and fine-tuning teaches models reasoning and information-selection skills that transfer across clinical tasks.

Our analyses also revealed meaningful limitations of existing LLMs on ICU clinical reasoning that ICU-REACT was able to surface. Across ICU topics and rubric dimensions, general-purpose and medical models frequently struggled, whereas Clin-REACT models showed consistent improvement. A Clin-REACT variant achieved the highest reasoning score across all nine ICU topic categories, with pronounced advantages in critical domains such as sepsis, hemodynamic instability/shock, respiratory failure, and renal failure (Fig.~\ref{fig:perform_topics}). The evaluation-dimension analysis was especially informative in areas where many models still struggle. For safety, general-purpose models scored 34.2-76.1 and medical models 51.6-67.1, showing that even strong baselines sometimes produced recommendations or overlooked considerations that could raise safety concerns. In contrast, Clin-REACT models performed better (78.6-93.9) (Fig.~\ref{fig:perform_dim_cat}A). Critical-anchor identification was even more challenging. This metric measured whether models recognized the most important clinical evidence supporting a decision. General-purpose models scored only 6.4-22.4 and medical models 7.7-17.9, suggesting that many models failed to center their reasoning on the most relevant findings. Clin-REACT achieved the strongest performance (22.4-34.0), but scores remained relatively low across all models, highlighting reliable identification of the most consequential clinical evidence as an important remaining challenge.

Clin-REACT models similarly improved on reasoning correctness, reasoning synthesis, and task faithfulness, and demonstrated more comprehensive information retrieval across clinical data categories, achieving the strongest F1 in three of eight categories, including laboratory measurements, scores and assessments, and imaging, and remaining competitive in other categories (Fig.~\ref{fig:perform_dim_cat}B). Notably, general-purpose and medical models tended to rely more heavily on routinely documented data, such as laboratory results, physiology, and medications, while more often missing clinically interpretive information such as scores/assessments, and imaging. Clin-REACT training helped shift attention toward these data types. Clin-REACT 31B achieved the highest F1 for scores and assessments (38.1), while Clin-REACT 70B performed best for imaging (26.6). This suggests that training with ICU-REACT encouraged models to consider a broader and more clinically complete set of evidence, rather than defaulting to the information that is most readily available.

A notable observation is that open-source medical LLMs generally underperformed on these clinical reasoning benchmarks, often having lower performance than general-purpose models of comparable or smaller size. This finding is consistent with recent results comparing general-purpose and specialized clinical models \cite{vishwanath_general-purpose_2026}. Because many medical LLMs are optimized on multiple-choice question-answering benchmarks, this pattern raises the possibility that such training incentivizes pattern recognition and answer memorization rather than the generative, multi-step reasoning required for realistic clinical decision-making \cite{thapa_reasoning_2026, kim_questioning_2025}. Our analysis comparing performance across medical multiple-choice and clinical-reasoning benchmarks further supports this interpretation, showing that stronger multiple-choice performance did not consistently correspond to stronger performance on clinically grounded reasoning tasks (see Supplemental Section~\ref{mc_cr_compare}). If so, strong performance on multiple-choice medical benchmarks may not fully reflect a model’s clinical reasoning ability. Open-ended frameworks such as ICU-REACT, which require models to identify relevant information and reason through it, may provide a more realistic assessment of models intended for clinical use.

Finally, our correlation analyses support ICU-REACT's validity as a clinical reasoning benchmark. Model-level ICU-REACT performance correlated moderately to strongly with SCT-Bench ($\rho=0.76$), ER-Reason ($\rho=0.78$), and VivaBench ($\rho=0.80$), and more moderately but still significantly with MedRBench ($\rho=0.51$) (Fig.~\ref{fig:perform_corr}A). These relationships were also seen at the metric level, with ICU-REACT metrics being closely correlated with decision-factor identification, diagnosis, and recall metrics across the external benchmarks. More broadly, the correlation patterns across metrics and datasets showed that models were ranked similarly across different evaluations (Fig.~\ref{fig:perform_corr}B). Together, these findings suggest that ICU-REACT captures a meaningful signal of clinical reasoning ability that is consistent with established benchmarks, while also testing ICU-specific skills that other benchmarks do not directly assess. The weaker or occasionally negative correlations among precision metrics also highlight an important precision-recall trade-off in clinical reasoning, which a decision-focused benchmark such as ICU-REACT may help reveal.

Our study has several limitations. First, our evaluation relied on automated metrics and LLM-as-judge evaluations rather than direct expert review of Clin-REACT outputs. Although LLM-based evaluation can scale across many models and has shown reasonable agreement with human raters, it cannot fully replace clinician judgment of response quality, safety, and clinical appropriateness. A related concern is that GPT-5.2 was used both to generate training dataset and to evaluate selected endpoints, including ICU-REACT reasoning. This teacher-judge overlap could favor responses that resemble the model’s preferred reasoning style. However, this limitation does not apply equally across all of our evaluations. Clin-REACT also improved on independently scored outcomes, including agreement with human expert response distributions on SCT-Bench and deterministic measures of diagnosis, treatment, and information retrieval on MedRBench, VivaBench, and ICU-REACT. In addition, ICU-REACT reasoning was scored using rubrics derived from clinician ground-truth responses. The consistency of the results across these different evaluation approaches makes evaluator preference alone unlikely to explain the overall gains, although prospective evaluation by practicing clinicians will be essential.

Second, the ICU-REACT test set is relatively small, with 71 questions. Although this is similar in scale to related benchmarks such as ER-Reason and covers nine common critical-care topics, it cannot represent the full range of cases and edge conditions encountered in ICU practice. ICU-REACT should therefore be viewed as a measure of reasoning across representative ICU scenarios rather than a comprehensive assessment of critical-care reasoning. Third, Clin-REACT models were trained using LoRA fine-tuning alone. This lightweight approach was effective and allowed smaller models to compete with or outperform much larger models. However, we did not explore more advanced approaches such as reinforcement learning from verifiable rewards or preference-based optimization methods such as Group Relative Policy Optimization (GRPO). These strategies may provide additional improvements in reasoning quality. Fourth, although the seed examples used to build ICU-REACT were reviewed by clinicians, the augmented training corpus itself was not clinically validated. The synthetic data may therefore contain factual errors, stylistic patterns, or biases inherited from the teacher model. This limits how strongly the augmented corpus can be described as clinically validated. Nevertheless, the consistent gains across independently developed external benchmarks and different evaluation methods suggest that the training data captured useful and transferable reasoning signals rather than simply reinforcing patterns specific to the synthetic corpus.

Finally, ICU-REACT has limitations related to annotation and clinical representation. Although the dataset was reviewed by clinicians from several specialties, the annotator pool did not include some sub specialties that are highly relevant to ICU care, including neurology, nephrology, and cardiology. This may limit the depth of validation in those areas. In addition, nearly all annotators came from a single institution, with only one clinician from an external site. As a result, some aspects of the dataset may reflect institution-specific practices or conventions, which could limit generalization across health systems.

These limitations point to several directions for future work. One important next step is to integrate Clin-REACT directly with EHR systems using our OMOP-aligned taxonomy of clinical variables, allowing models to retrieve and reason over structured patient data in real time. More advanced training strategies, including reinforcement learning approaches such as GRPO, may further improve reasoning beyond supervised fine-tuning. Prospective evaluation with practicing clinicians will also be critical for assessing safety, usefulness, and performance in realistic clinical settings. Finally, developing ICU-REACT datasets tailored to specific ICU environments, such as medical, surgical, cardiac, or neurological intensive care, could help train models that better reflect the patient populations, workflows, and decision-making needs of each setting.

\section{Methods}\label{methods}

\subsection{Dataset}\label{methods_dataset}

We developed ICU-REACT, a clinician-supervised dataset designed to teach LLMs to perform context-aware clinical reasoning and information retrieval. The dataset creation had two main stages: seed dataset creation and dataset augmentation.

\begin{figure*}[h]
\centering
\includegraphics[width=1\linewidth]{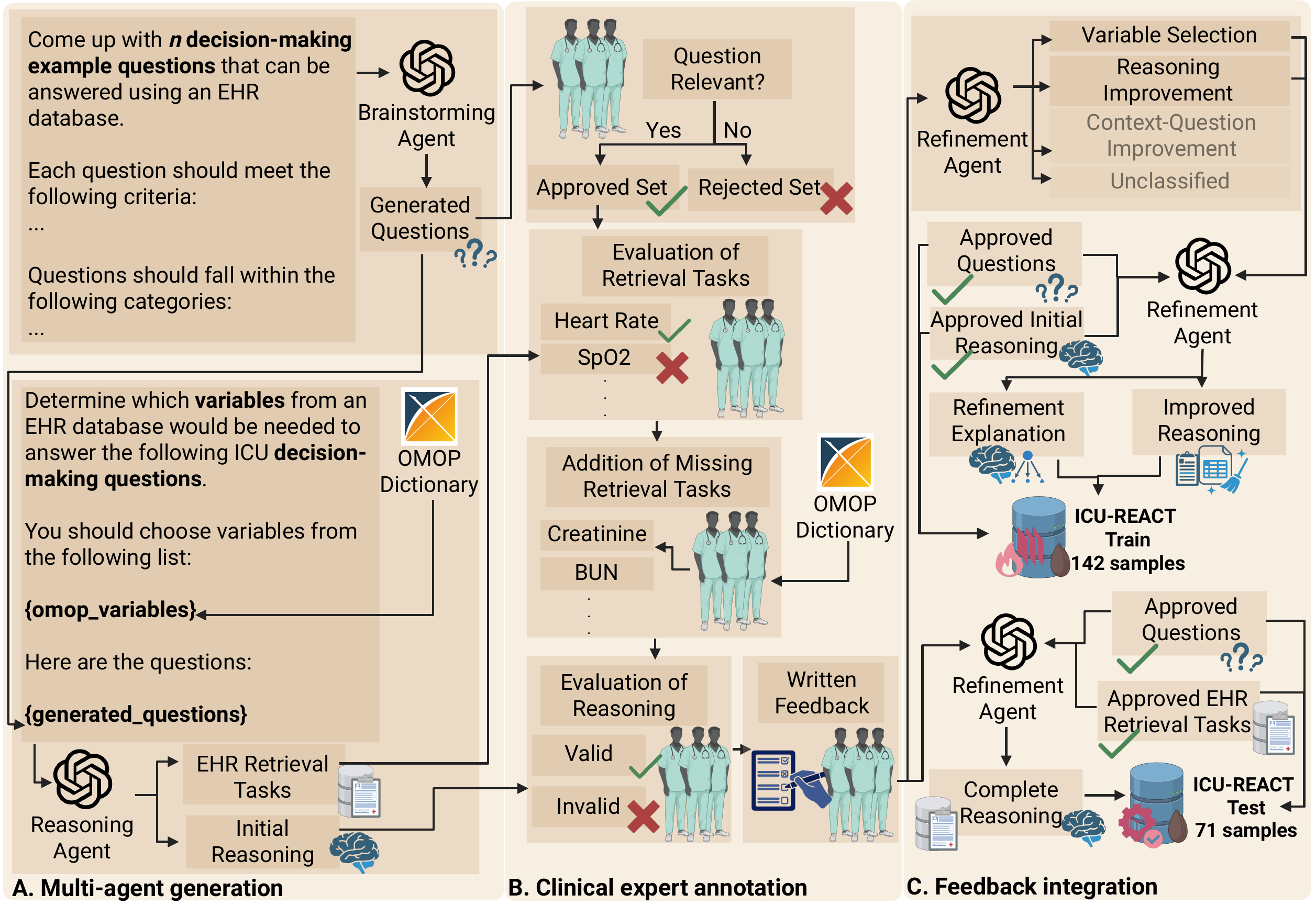}
\caption{Seed dataset creation overview. (A) A brainstorming agent generates candidate ICU decision-making questions that can be answered with data from an EHR database according to predefined clinical criteria and topic categories. A reasoning agent then identifies the EHR retrieval tasks and initial clinical reasoning needed to answer each question, with variables standardized using the OMOP dictionary. (B) Clinical experts review generated questions for relevance, evaluate the completeness and appropriateness of proposed retrieval tasks, add missing variables when necessary, and assess the validity of the initial reasoning while providing written feedback. (C) Written clinician feedback is incorporated by a refinement agent through distinct pipelines for the training and test seed sets. For ICU-REACT-Train, feedback is first categorized as variable-selection, reasoning improvement, context–question improvement, or unclassified. Feedback concerning variable selection and reasoning improvement is then combined with the original patient context–question pair and initial reasoning to generate both an explanation of required revisions and an improved reasoning response. The resulting training samples contain the context, question, initial reasoning, refinement explanation, and improved reasoning. For ICU-REACT-Test, approved questions, retrieval tasks, and written clinician feedback are directly provided to the refinement agent to generate complete reasoning responses incorporating all required EHR variables.}
\label{fig:seed_creation}

\end{figure*}

\subsubsection{Seed dataset creation}\label{methods_seed}

We first create the ICU-REACT seed set through LLM-driven generation and clinician expert curation. Specifically, we first employ a multi-agent system responsible for generating diverse, clinically relevant decision-making questions and associated EHR retrieval tasks using controlled prompts and domain ontologies. These initial generations were stored in an online MongoDB and subsequently refined through a structured annotation process conducted by 19 clinicians, including physicians, residents, and medical trainees. Clinicians conducted the review through our online annotation tool to ensure the clinical validity and relevance of the generated data to real-world ICU workflows. All EHR data elements were mapped to a manually crafted subset of standardized concepts from the Observational Medical Outcomes Partnership (OMOP) Common Data Model to facilitate interoperability and reproducibility across EHR systems. The overall process is summarized on Fig.~\ref{fig:seed_creation}.

\paragraph{Initial generation}\label{methods_initgen}

The first stage of dataset creation leveraged a multi-agent generation framework consisting of two GPT-4.1-based agents: a Brainstorming Agent and a Reasoning Agent. The GPT-4.1 model was selected for its state-of-the-art instruction-following performance at the time of initial generation.

The Brainstorming Agent was prompted to generate $n$ example decision-making questions that could be answered using structured or unstructured EHR data (Fig.~\ref{fig:seed_creation}A). Each question was required to meet specific criteria: it must represent a clinically actionable decision (for example, initiating or adjusting therapy, ordering diagnostics, or assessing patient trajectory), it must be answerable using available EHR data, and it must fall within nine ICU decision domains: Nutrition and Metabolic Support, Sepsis and Severe Infections, Hemodynamic Instability / Shock, Sedation, Pain, and Delirium Management, Respiratory Failure, Neurological Emergencies, Renal Failure / Electrolyte Disorders, Cardiac Emergencies, Hematologic / Coagulation Issues. 

For each generated question, the Reasoning Agent determined which EHR variables would be needed to answer it (Fig.~\ref{fig:seed_creation}A). Variable selection was constrained to a manually crafted subset of standardized terms within the OMOP dictionary (see Supplemental Section \ref{omop_var}), ensuring consistency and reproducibility across datasets. The Reasoning Agent produced a structured list of EHR retrieval tasks (for example, heart rate, creatinine, or SpO$_2$) along with a concise reasoning paragraph explaining why those variables were required to address the clinical question. These outputs (decision-making questions, variable retrieval targets, and reasoning explanations) formed the preliminary set of EHR retrieval tasks for expert review.

\paragraph{Clinician annotation}\label{methods_annotation}

The second stage involved clinical expert annotation and validation conducted by a team of 19 clinicians across critical care, general surgery, emergency medicine, internal medicine, anesthesiology, and orthopedics. This process ensured the relevance, completeness, and reasoning validity of all generated samples (Fig.~\ref{fig:seed_creation}B).

To facilitate efficient data review and collection, we developed a custom web-based annotation platform (see Supplemental Section~\ref{annotation_tool}) that served as the primary interface between clinicians and the data engineering team. The tool provided an interactive environment for reviewing each sample, including the hypothetical patient context, decision-making question, retrieved variables, and model-generated reasoning. Clinicians had the option to approve, reject, or modify each sample, as well as provide written feedback explaining their decisions. All annotations were automatically stored in a centralized MongoDB database on Google Cloud, ensuring version control and traceability.

Clinicians independently reviewed each generated question within the platform to determine its clinical relevance and clarity. To minimize bias, we ensured that each sample was reviewed by two clinicians. Questions that were classified as irrelevant by at least one clinician were excluded, while approved questions proceeded to evaluation of retrieval accuracy. Annotators verified whether the variables identified by the model were relevant for answering each question. Variables that were considered relevant by at least one clinician were kept, while variables that were considered irrelevant by both clinicians were removed. Clinicians then had the option to add missing variables from our manually crafted OMOP dictionary subset to maintain consistent terminology (Fig.~\ref{fig:seed_creation}B).

Following this, clinicians assessed the reasoning explanations provided by the model. Each rationale was evaluated for logical coherence, clinical accuracy, and relevance to the decision-making task. Samples were marked as valid if the reasoning demonstrated appropriate clinical logic supported by the corresponding EHR variables, or invalid if the reasoning was incomplete or clinically inconsistent (Fig.~\ref{fig:seed_creation}B). Approved samples were then split into two sets: an ICU-REACT train seed set for LLM-driven augmentation for model training and an ICU-REACT test set for evaluation of models. This separation ensured that augmented data would not overlap with test cases, mitigating data leakage and helping preserve the integrity of subsequent model assessments. All validation results and written reviewer notes were collected through the annotation platform for subsequent feedback integration for both sets.

\paragraph{Feedback integration}\label{methods_feedback}

To refine approved samples using written clinician feedback, we implemented a feedback integration pipeline leveraging GPT-5.2 as a refinement agent. The GPT-5.2 model was selected for its state-of-the-art status at the time of the experiments. Specifically, we used two different routes for feedback integration on both ICU-REACT seed sets. For the ICU-REACT test set, written feedback was directly inputted into the refinement agent along with the approved questions and EHR retrieval tasks to generate a complete reasoning response to each question including all EHR variables (Fig.~\ref{fig:seed_creation}C).

For refinement of the ICU-REACT train seed set, clinician comments were first classified by the refinement agent into three feedback types: variable selection feedback, reasoning improvement feedback, and context-question improvement feedback, with remaining comments tagged as 'unclassified' (Fig.~\ref{fig:seed_creation}C). Variable selection feedback captured issues related to missing, irrelevant, or incorrectly mapped EHR variables; reasoning improvement feedback identified gaps in clinical logic, completeness, or interpretability; and context-question improvement feedback described concerns about the clarity or alignment of the patient context and decision-making question.

We focused on feedback related to variable selection and reasoning improvement, as these directly addressed the model outputs used for clinical reasoning supervision. Each refinement sample was constructed from the original context-question pair and the initial reasoning generated by the GPT-4.1-based Reasoning Agent described in Section~\ref{methods_initgen}. This information, together with the clinician feedback, was provided to the refinement agent, which was prompted to produce two outputs: an explanation of how the initial reasoning should be improved and a corresponding improved reasoning response, both aligned with the clinician feedback (Fig.~\ref{fig:seed_creation}C).

The train seed was then consolidated as samples containing the patient context, decision-making question, initial reasoning, explanation of the needed reasoning improvements, and improved reasoning (Fig.~\ref{fig:seed_creation}C). This structure preserved the original model rationale while adding clinician-guided supervision that explicitly teaches how clinical reasoning should be revised, providing the curated foundation for subsequent data augmentation.

\subsubsection{Dataset augmentation}\label{methods_augmentation}

\begin{figure*}[h]
\centering
\includegraphics[width=0.95\linewidth]{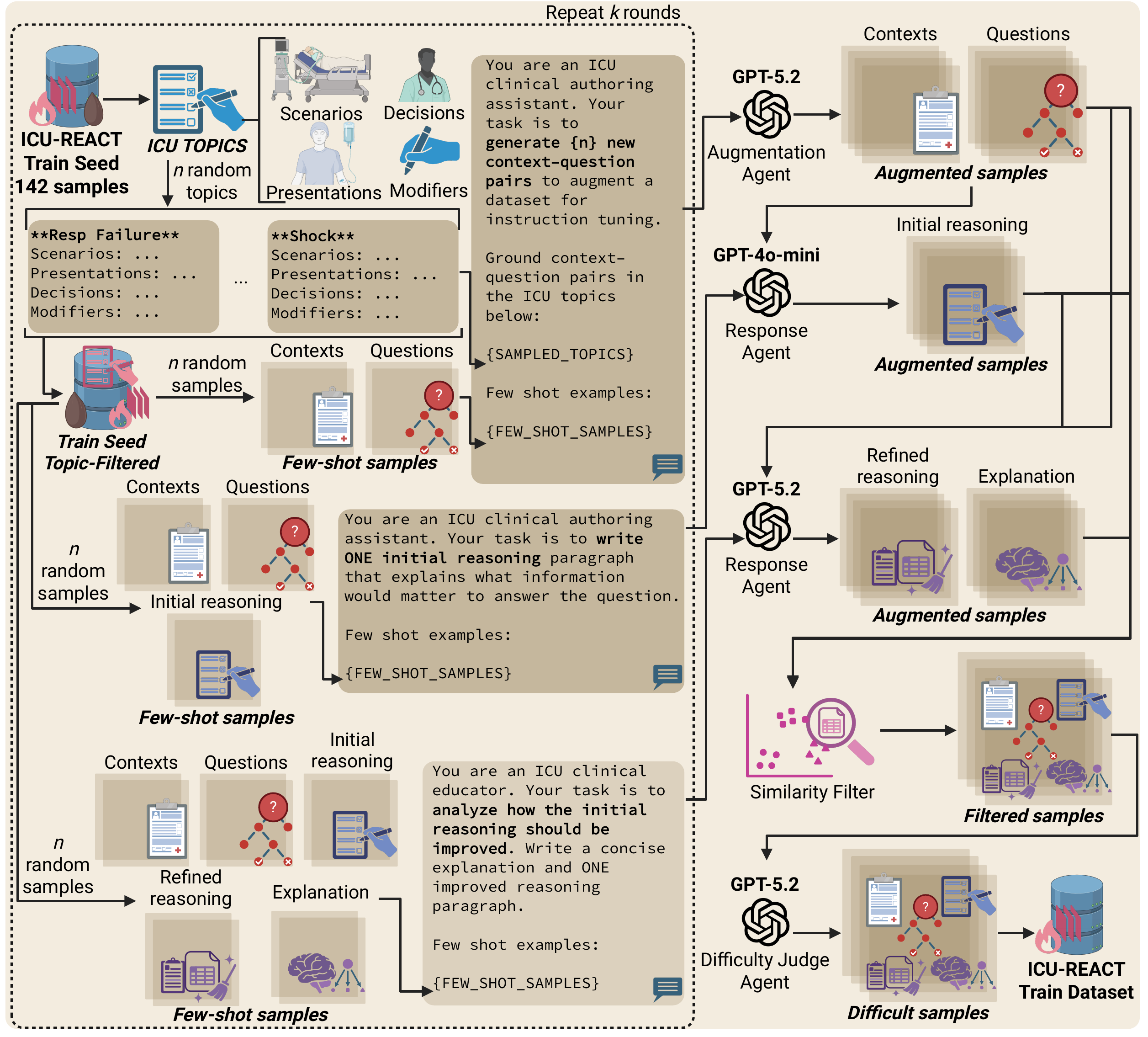}
\caption{Dataset augmentation overview. Beginning with the clinician-validated ICU-REACT-Train seed set, the pipeline iteratively sampled ICU topics together with topic-specific scenarios, patient presentations, clinical modifiers, and decision types to guide generation of diverse context–question pairs. At each generation stage, few-shot examples were randomly sampled from the train seed dataset and incorporated into the corresponding prompts to preserve the expert-aligned structure, terminology, and reasoning schema of the seed examples. A GPT-5.2 augmentation agent generated new patient contexts and ICU decision-making questions, after which a GPT-4o-mini agent produced initial reasoning responses. A GPT-5.2 refinement agent then generated a refinement explanation and an improved reasoning response from the initial reasoning, using additional seed-derived few-shot examples to maintain consistency with clinician-validated reasoning patterns. To reduce redundancy, augmented samples were vectorized, grouped through nearest-neighbor search, and compared within groups using ROUGE-L; samples with ROUGE-L scores of at least 0.7 were removed. Finally, a GPT-5.2 difficulty-judging agent categorized retained samples as easy, medium, or hard, and samples were filtered according to difficulty level to construct training sets tailored to each model architecture and scale, producing the final ICU-REACT-Train datasets.}
\label{fig:data_augmentation}

\end{figure*}

Following the construction of the ICU-REACT train seed dataset, we implemented an automated augmentation pipeline to expand the dataset's size to allow for model training (Fig.~\ref{fig:data_augmentation}).

The objective of the augmentation process was to increase the diversity of clinical reasoning examples while maintaining the structure, terminology, and expert-aligned reasoning schema established in the original seed dataset. To achieve this, a GPT-5.2 augmentation agent was used to generate the new context-question pairs through carefully controlled prompting. Specifically, a clinician-guided ICU topic list was used to iteratively prompt the augmentation agent to produce diverse patient contexts and decision-making questions, by sampling \textit{n} topics at each augmentation iteration. To further improve diversity of the dataset, we used ChatGPT to create lists of patient scenarios, presentations, modifiers and decisions within each ICU topic, sampling \textit{n} items of each category at each augmentation iteration. The full list of topics with their corresponding scenarios, presentations, modifiers and decisions can be found on Supplemental Section \ref{icu_topics}. Following this, a GPT-4o-mini agent was used to generate initial reasonings for each augmented pair, and subsequently, a GPT-5.2 agent was used to generate refined reasonings along with explanations for the refinement based on the initial reasonings, mirroring the schema and structure of the clinician-validated train seed dataset. At each generation stage, we randomly sampled few-shot examples from the ICU-REACT train seed dataset and incorporated them into the corresponding prompts to preserve the expert-aligned structure, terminology, and reasoning schema.

To reduce redundancy among augmented samples, we applied a similarity filter by vectorizing samples and grouping them using nearest-neighbor search. Samples within each group were then compared pairwise using ROUGE-L, and samples with a ROUGE-L score of 0.7 or higher were removed. Finally, we further filtered samples by difficulty level. Specifically, a GPT-5.2 difficulty judge agent was used to classify samples in one of three difficulty categories: easy, medium, and hard. Subsequently, samples were filtered according to difficulty level to construct training sets tailored to each model architecture and scale, producing the final ICU-REACT Train Datasets.

\begin{figure*}[h]
\centering
\includegraphics[width=0.90\linewidth]{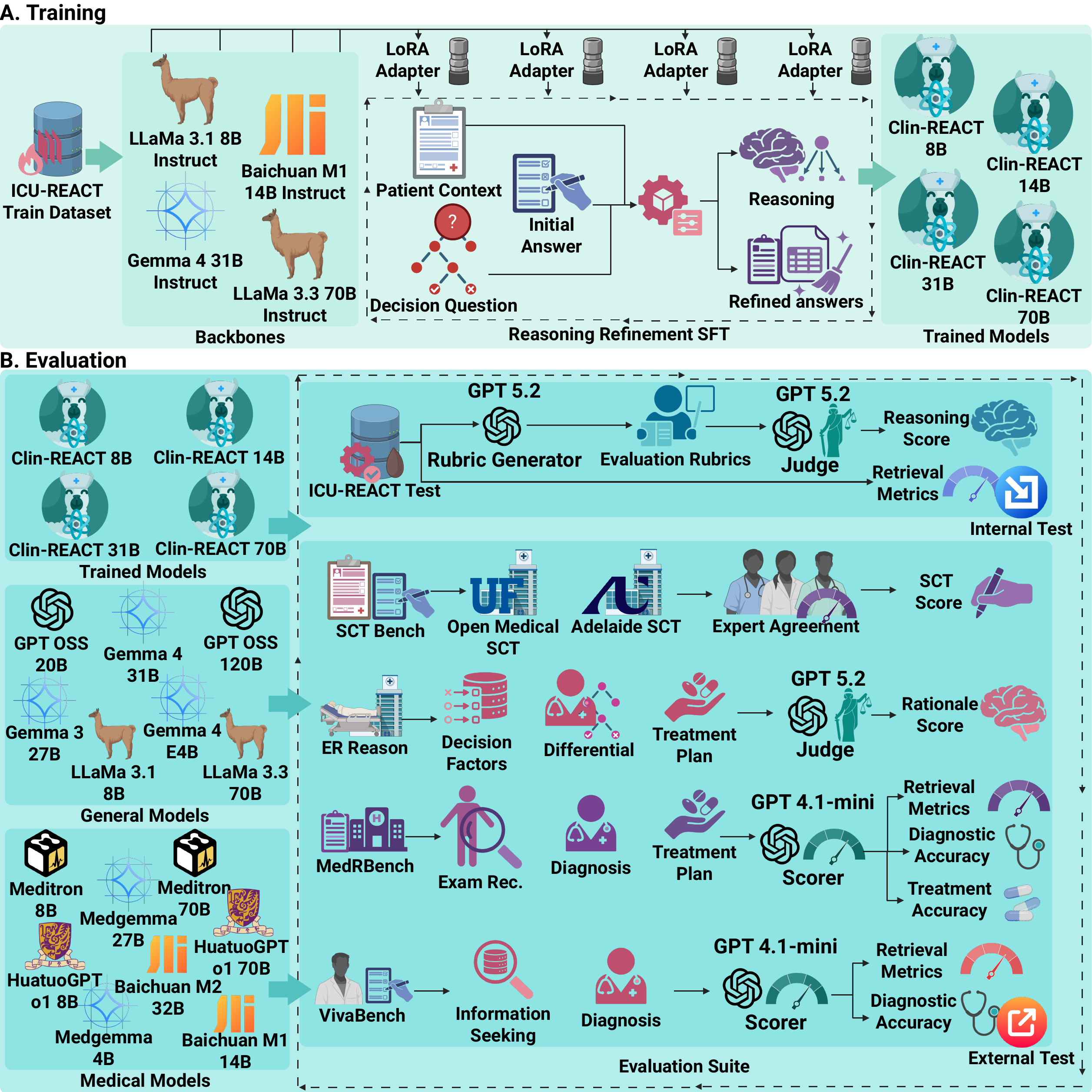}
\caption{Model Training and Evaluation. (A) Supervised fine-tuning of four open-source backbone models using the ICU-REACT-Train datasets. Low-rank adaptation (LoRA) adapters were trained through reasoning-refinement supervision, with the patient context, decision-making question, and initial response provided as model inputs, and the refinement explanation and improved reasoning response used as target outputs. This process yielded Clin-REACT 8B, 14B, 31B, and 70B models. (B) Evaluation of Clin-REACT models alongside open-source general-purpose and medical LLM baselines. In-domain performance was assessed on the held-out ICU-REACT-Test set using automated variable-retrieval precision, recall, and F1 metrics, together with GPT-5.2 rubric-based reasoning scores. External evaluation included SCT-Bench, which measured agreement with expert response distributions under clinical uncertainty; ER-Reason, which evaluated precision and recall of clinically relevant concepts in acute-care rationales; MedRBench, which assessed examination recommendation, diagnostic, and treatment performance; and VivaBench, which evaluated information-seeking and final diagnostic accuracy in multi-turn clinical cases.}
\label{fig:model_train_eval}

\end{figure*}

\subsection{Model training}\label{methods_model_train}

We developed Clin-REACT, a set of fine-tuned LLMs for decision-focused clinical reasoning and information retrieval in the ICU. Clin-REACT models were initialized from Llama, Gemma, and Baichuan base models and trained on the ICU-REACT Train Datasets using supervised fine-tuning (SFT) (Fig.~\ref{fig:model_train_eval}A).

\subsubsection{Base models}\label{methods_base_models}

We trained Clin-REACT variants from four base models: Llama 3.1 8B, Llama 3.3 70B, Gemma 4 31B, and Baichuan M1 14B. These models span different architectures and parameter counts, allowing to assess the robustness of ICU-REACT as a training dataset for improving decision-focused clinical reasoning beyond a single model family and size. Training multiple Clin-REACT variants also provided a set of models with different size-performance tradeoffs, supporting evaluation across both more efficient and higher-capacity deployment settings.

\subsubsection{Training strategy}\label{methods_train_strat}

The models were trained with an assistant-only causal language modeling objective through parameter-efficient fine-tuning (PEFT). We used low-rank adaptation (LoRA) to fine-tune each base model by freezing the original model weights and learning small trainable low-rank update matrices for selected linear layers~\cite{hu_lora_2021}. For a pretrained weight matrix $W_0 \in \mathbb{R}^{d \times k}$, LoRA parameterized the adapted weight as:

\[
W = W_0 + \Delta W = W_0 + \frac{\alpha}{r}BA,
\]

where $A \in \mathbb{R}^{r \times k}$ and $B \in \mathbb{R}^{d \times r}$ are trainable low-rank matrices, $r \ll \min(d,k)$ is the LoRA rank, and $\alpha$ is a scaling factor that controlled the magnitude of the learned update. During training, the pretrained weights $W_0$ remained fixed, and only the LoRA parameters $A$ and $B$ were optimized, substantially reducing the number of trainable parameters while adapting the model to ICU-REACT reasoning refinement.

Each training example consisted of a patient context $c$, a decision-making question $q$, an initial reasoning response $r^{(0)}$, and a target assistant response $a$. The target response $a$ corresponded to the reasoning refinement and contained two components: an explanation of how the initial reasoning should be improved and the final improved reasoning. The models were optimized only on the assistant response tokens, while the input tokens from the context, question, and initial reasoning were used as conditioning information:

\[
\mathcal{L}_{\text{refine}}(\theta)
= - \sum_{i=1}^{N} \sum_{t=1}^{|a_i|}
\log P_\theta\!\left(a_{i,t} \mid a_{i,<t}, c_i, q_i, r_i^{(0)}\right)
\]

where $N$ is the number of training examples, $c_i$ denotes the patient context for example $i$, $q_i$ denotes the corresponding decision-making question, and $r_i^{(0)}$ denotes the initial reasoning generated before refinement. The target assistant response $a_i = (a_{i,1}, \ldots, a_{i,|a_i|})$ is the reasoning refinement sequence, consisting of both the explanation of the needed refinement and the improved reasoning. The term $a_{i,t}$ denotes the $t$-th target token, $a_{i,<t}$ denotes all preceding assistant-response tokens, and $P_\theta$ is the conditional token probability assigned by the model with parameters $\theta$. Because the loss was computed only over the assistant response tokens $a_i$, the model learned to use the patient context, decision-making question, and initial reasoning as input while being supervised to generate the refined reasoning output.

\subsection{Model evaluation}\label{methods_model_eval}

We evaluated Clin-REACT models and compared their performance against open-source general-purpose and medical LLMs on internal and external benchmarks to measure their performance on clinical reasoning (Fig.~\ref{fig:model_train_eval}B).

\subsubsection{Benchmarks}\label{methods_benchmarks}

We evaluated Clin-REACT on five clinical reasoning benchmarks that probe ICU-specific reasoning, general clinical reasoning, and acute care decision-making:
\begin{itemize}
    \item \textbf{ICU-REACT Test (in-domain) (n = 71).} A held-out test set drawn from the ICU-REACT distribution, used to measure in-domain generalization for both reasoning generation and structured variable retrieval. This benchmark directly reflects the target use case of ICU decision support. For evaluation, we measured automated retrieval metrics (i.e., precision, recall and F1 scores) for variable domains/categories and individual variables, as well as LLM-judge-based reasoning score using LLM-generated evaluation rubrics. Specifically, we used GPT-5.2 to first generate evaluation rubrics for each sample on the test dataset following similar methodologies from previous literature~\cite{arora_healthbench_2025}. We then used a GPT-5.2-based judge which scored model responses strictly following the evaluation rubrics. The final reasoning score was calculated from adding all points given on each rubric criteria and normalizing the score based on the maximum possible number of criteria points.
    \item \textbf{SCT-Bench (n = 174).} \cite{mccoy_assessment_2025} A script concordance test (SCT)-style benchmark assessing clinical reasoning under uncertainty. Specifically, we used the public set of this benchmark which contains the Open Medical SCT from the University of Florida and the Adelaide SCT from Adelaide University. Performance was computed using the SCT scoring procedure that rewards alignment with expert response distributions, emphasizing calibration rather than single-label correctness.
    \item \textbf{ER-Reason (n = 72).} \cite{mehandru_er-reason_2025} An emergency medicine reasoning benchmark evaluating acute care decision-making and prioritization in time-sensitive settings, providing an out-of-domain but clinically adjacent stress test. We focused on the rationale module of this dataset, which contains physician-written explanations for decision factors, differential diagnosis, and treatment planning across 72 emergency room cases. To evaluate model outputs, we used a GPT-5.2-based LLM-judge approach to assess the quality of each generated rationale by quantifying recall and precision of relevant medical concepts present in the original physician-written rationales for each category.
    \item \textbf{MedRBench (n = 1,338).} \cite{qiu_quantifying_2025} A reasoning-focused benchmark built from publicly-available structured real-world clinical case reports, evaluating models across examination recommendation, diagnostic decision-making, and treatment planning, with explicit assessment of final clinical outputs. Prior to evaluation, we excluded samples that triggered Azure content-policy filters during processing with Azure-hosted GPT helper or judge models, and used the resulting filtered set consistently across all evaluated models to ensure an identical evaluation cohort. Performance of all models was measured following the same methodology as in the original paper~\cite{qiu_quantifying_2025}, using GPT-4.1-mini instead of GPT-4o as the helper and judge model due to budget constraints. We focused on examination recommendation recall and precision, diagnosis accuracy and treatment accuracy.
    \item \textbf{VivaBench (n = 934).} \cite{chiu_simulating_2025} A multi-turn benchmark that simulates viva voce-style clinical examinations built using clinical vignettes from publicly available repositories, requiring models to iteratively gather relevant history, physical examination findings, and diagnostic investigations before synthesizing a final diagnosis. Prior to evaluation, we excluded samples that triggered Azure content-policy filters during processing with Azure-hosted GPT helper or judge models, and used the resulting filtered set consistently across all evaluated models to ensure an identical evaluation cohort. Performance of all models was measured following the same methodology as in the original paper~\cite{chiu_simulating_2025}, using GPT-4.1-mini instead of GPT-4.1 as the mapper and judge model due to budget constraints. We focused on information seeking recall and precision, and diagnosis accuracy. 
\end{itemize}

Together, these benchmarks quantified (i) in-domain performance on ICU decision-making and variable retrieval (ICU-REACT Test), (ii) robustness to uncertainty and partial information (SCT-Bench and VivaBench), and (iii) generalization to broader clinical reasoning and acute-care workflows (ER-Reason and MedRBench).

\subsubsection{Baselines}\label{methods_baselines}

We compared Clin-REACT model variants against both open-source general-purpose LLMs and medical-domain LLMs. The general-purpose comparison models included GPT-OSS models (GPT-OSS 20B and GPT-OSS 120B)~\cite{openai_gpt-oss-120b_2025}, Gemma models (Gemma 3 27B and Gemma 4 31B)~\cite{team_gemma_2025}, and Llama models (Llama 3.1 8B Instruct and Llama 3.3 70B Instruct)~\cite{grattafiori_llama_2024}. The medical-domain comparison models included Meditron models (Meditron 3 8B and Meditron 3 70B)~\cite{sallinen_llama-3-meditron_2025}, MedGemma models (MedGemma 4B and MedGemma 27B)~\cite{sellergren_medgemma_2026}, HuatuoGPT-o1 models (HuatuoGPT-o1 8B and HuatuoGPT-o1 70B)~\cite{chen_towards_2025}, and Baichuan models (Baichuan M1 14B Instruct and Baichuan M2 32B)~\cite{wang_baichuan-m1_2025, team_baichuan-m2_2025}. This comparison set was designed to contextualize Clin-REACT performance against models spanning different parameter scales, model families, and degrees of medical specialization.

\subsubsection{Statistical Analyses}\label{methods_stat_analyses}

We performed statistical analyses to compare Clin-REACT model variants against zero-shot baseline models across the five clinical reasoning benchmarks. Because all models were evaluated on the same benchmark instances within each dataset, model comparisons were treated as paired analyses. All benchmark metrics operated on the $[0,100]$ range, with higher scores indicating better performance.

A non-parametric bootstrap procedure with 1000 iterations was employed
to estimate the mean and 95\% confidence intervals of the evaluation metrics across each benchmark. In each iteration, a resampled dataset equal in size to the test set of the benchmark being evaluated was generated via random sampling with replacement.

For pairwise model comparisons, statistical significance was assessed using the two-sided Wilcoxon signed-rank test. For a given comparison, we calculated the paired difference in score between the two models across benchmark instances and tested whether the median paired difference differed from zero.

\section{Data availability}

The ICU-REACT dataset, including the seed training set, the ICU-REACT-Train-Small, Medium, and Large augmented training sets, and the held-out ICU-REACT-Test benchmark with its sample-specific evaluation rubrics, is publicly available through Hugging Face at \url{https://huggingface.co/datasets/macontreras98/ICU-REACT}. The external clinical-reasoning benchmarks evaluated in this study are available from their respective original sources. SCT-Bench is publicly available at \url{https://github.com/SCT-Bench/sctpublic}; MedRBench is publicly available at \url{https://github.com/MAGIC-AI4Med/MedRBench}; and VivaBench is publicly available through Hugging Face at \url{https://huggingface.co/datasets/chychiu/VivaBench}. ER-Reason is available through PhysioNet at \url{https://physionet.org/content/er-reason/1.0.0/}. Access to ER-Reason requires PhysioNet credentialing, completion of the Collaborative Institutional Training Initiative (CITI Program) ``Data or Specimens Only Research'' training module, agreement to the applicable data use agreement, and approval of a project-specific access request by the data contributors.

\section{Code availability}

The code for dataset processing and augmentation, model training and inference, benchmark evaluation, and all analyses reported in this study is freely available at \url{https://github.com/iheallab/icureact}. The trained Clin-REACT model checkpoints are publicly available through Hugging Face: Clin-REACT 8B at \url{https://huggingface.co/macontreras98/Llama-Clin-REACT-8B}, Clin-REACT 14B at \url{https://huggingface.co/macontreras98/Clin-REACT-14B}, Clin-REACT 31B at \url{https://huggingface.co/macontreras98/Clin-REACT-31B}, and Clin-REACT 70B at \url{https://huggingface.co/macontreras98/Llama-Clin-REACT-70B}. The model checkpoints are distributed subject to the respective licenses and terms of their underlying base models.

\section{Acknowledgments}

A.B and P.R. were supported by NIH/NINDS R01 NS120924 and NIH/NIBIB R01 EB029699.


\putbib[sn-bibliography]

\end{bibunit}

\input{supplement}

\end{document}

%% file: supplement.tex
\clearpage


\pagenumbering{arabic}
\renewcommand{\thepage}{\arabic{page}}
\setcounter{page}{1}

\begin{center}
    {\LARGE\bfseries Supplemental Material}
\end{center}

\vspace{-0.5em}

\appendix

\renewcommand{\thesection}{S\arabic{section}}
\setcounter{section}{0}

\renewcommand{\thesubsection}{\thesection.\arabic{subsection}}
\setcounter{subsection}{0}

\renewcommand{\thefigure}{S\arabic{figure}}
\setcounter{figure}{0}

\renewcommand{\thetable}{S\arabic{table}}
\setcounter{table}{0}

\renewcommand*{\theHsection}{supp.\arabic{section}}
\renewcommand*{\theHsubsection}
    {supp.\arabic{section}.\arabic{subsection}}
\renewcommand*{\theHfigure}{supp.\arabic{figure}}
\renewcommand*{\theHtable}{supp.\arabic{table}}


\addtocontents{toc}{\protect\setcounter{tocdepth}{2}}

\setcounter{tocdepth}{2}
\tableofcontents

\clearpage

\begin{bibunit}[sn-mathphys-num]


\section{Annotation workflow details}\label{annotation_tool}

\begin{figure}[htbp]
\centering
\includegraphics[width=0.95\linewidth]{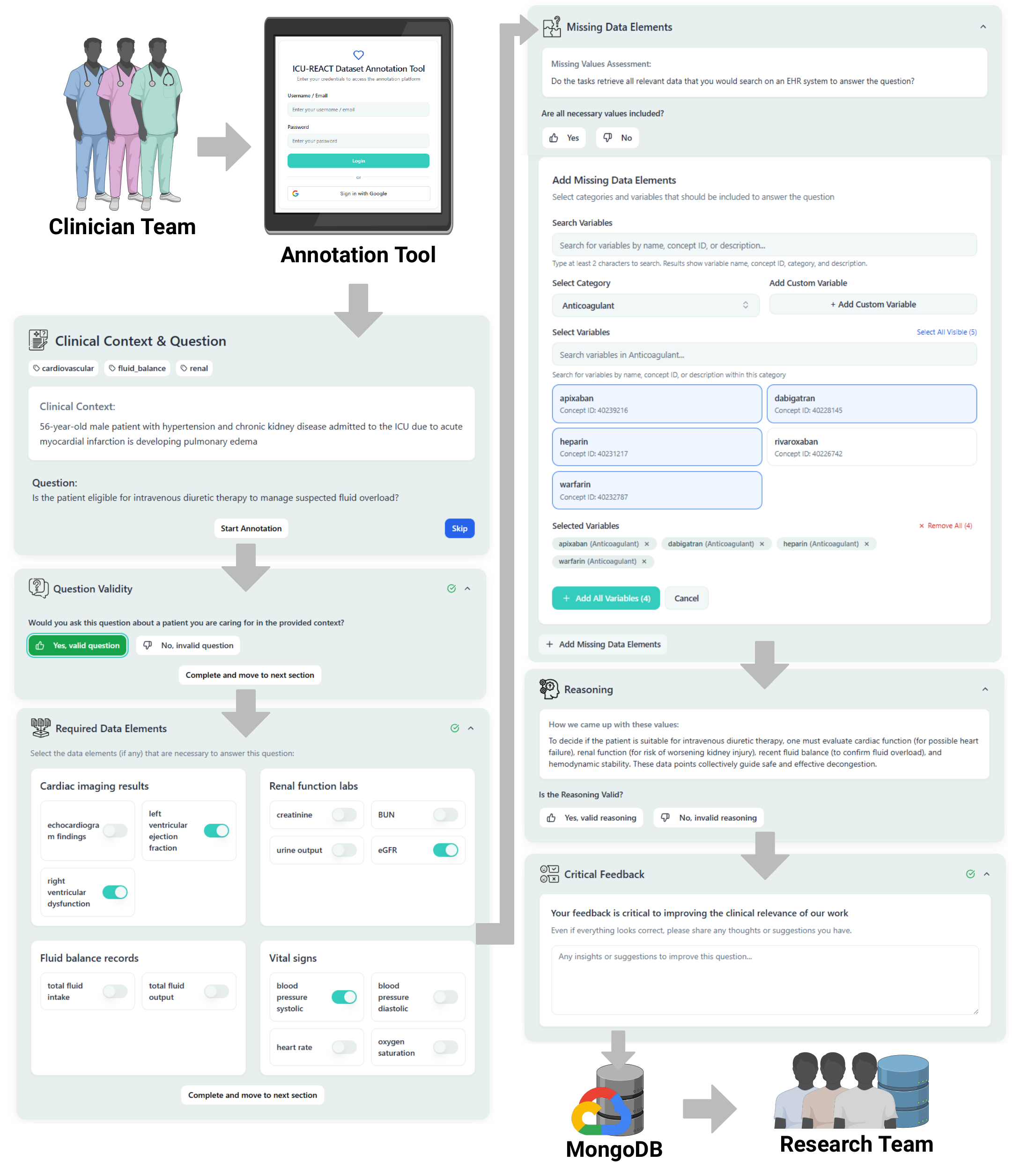}
\caption{Clinician annotation workflow and interface for the ICU-REACT seed dataset. The panels show screen captures of the web-based annotation tool for a single example being annotated, with arrows indicating the order of the workflow. A clinician team logs in and is presented with the Clinical Context and Question, tagged by relevant clinical domains, and chooses to annotate or skip the item. The annotation then proceeds through sequential sections: (i) Question Validity, confirming whether the question is one a clinician would ask for such a patient; (ii) Required Data Elements, toggling the variables needed to answer the question, organized by category; (iii) Missing Data Elements, indicating whether the retrieved data are complete and, if not, adding further variables by name or concept ID or as custom entries; (iv) Reasoning, judging whether the provided rationale is valid; and (v) Critical Feedback, an open-text field for additional suggestions. Completed annotations are written to a MongoDB store and passed to the research team for use in the construction of the ICU-REACT seed dataset. Note: the example shown is for illustration of the interface only and is not a real nor validated annotated sample.}
\label{fig:annotation_tool}
\end{figure}

To ensure the clinical validity of ICU-REACT, candidate items were reviewed by a team of clinicians through a web-based annotation tool (Fig.~\ref{fig:annotation_tool}). After authenticating, an annotator was shown the clinical context and question for a single item, along with its clinical-domain tags, and could either begin the annotation or skip items outside their expertise. The review then proceeded through five sequential stages. In Question Validity, the annotator judged whether the question was one they would plausibly ask when caring for a patient in the given context. If the annotator considered the question to be not relevant, then annotation for the item was terminated and would proceed to the next item. In Required Data Elements, they selected the variables needed to answer the question from category-organized toggles (e.g., cardiac imaging results, renal function labs, fluid balance records, and vital signs). In Missing Data Elements, they assessed whether the item's retrieved data captured everything they would query in an EHR and, when necessary, added further variables by searching the OMOP-based variable taxonomy by name or concept ID, or by entering custom variables. In Reasoning, they judged whether the accompanying rationale linking the selected data to the clinical decision was sound. Finally, in Critical Feedback, they could leave free-text suggestions to improve the item even when all prior checks passed. Such written feedback could be as detailed or concise as the annotator wished. Completed annotations were persisted to a MongoDB database and returned to the research team, who aggregated the clinician judgments to filter, correct, and finalize items for the ICU-REACT seed dataset.

\begin{figure*}[h]
\centering
\includegraphics[width=0.90\linewidth]{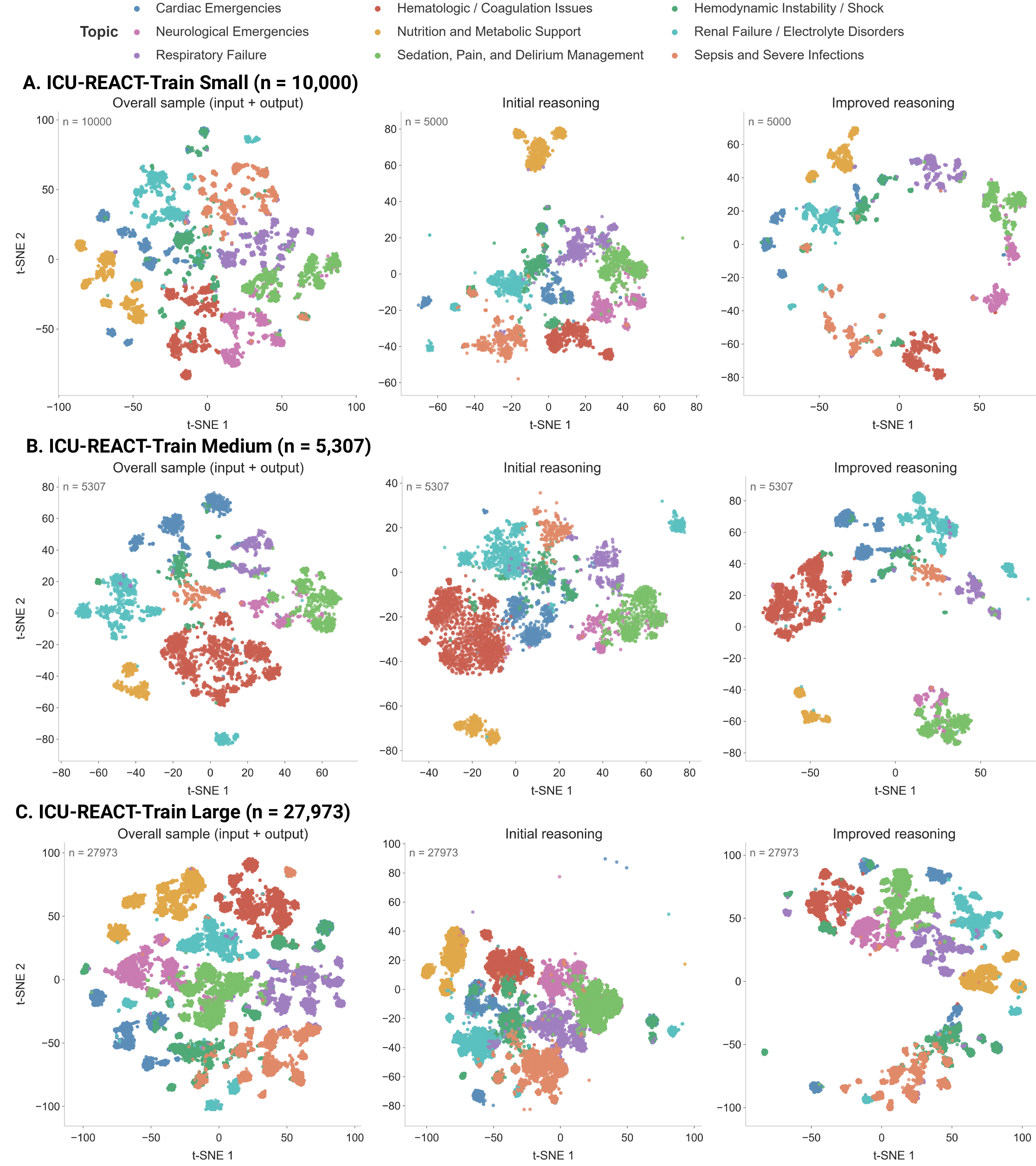}
\caption{Topic distribution of the ICU-REACT-Train datasets. t-distributed stochastic neighbor embedding (t-SNE) projections of the ICU-REACT-Train Small (A), Medium (B), and Large (C) datasets, with samples colored according to the primary ICU topic represented in each question. Dataset names (Small, Medium, Large) refer to the parameter scale of the target model each set was used to train. For each dataset scale, the left panel depicts the overall distribution of input-output training examples, whereas the middle and right panels separately visualize examples corresponding to the initial-reasoning and improved-reasoning stages.}
\label{fig:train_datasets}

\end{figure*}

\begin{figure*}[h]
\centering
\includegraphics[width=0.95\linewidth]{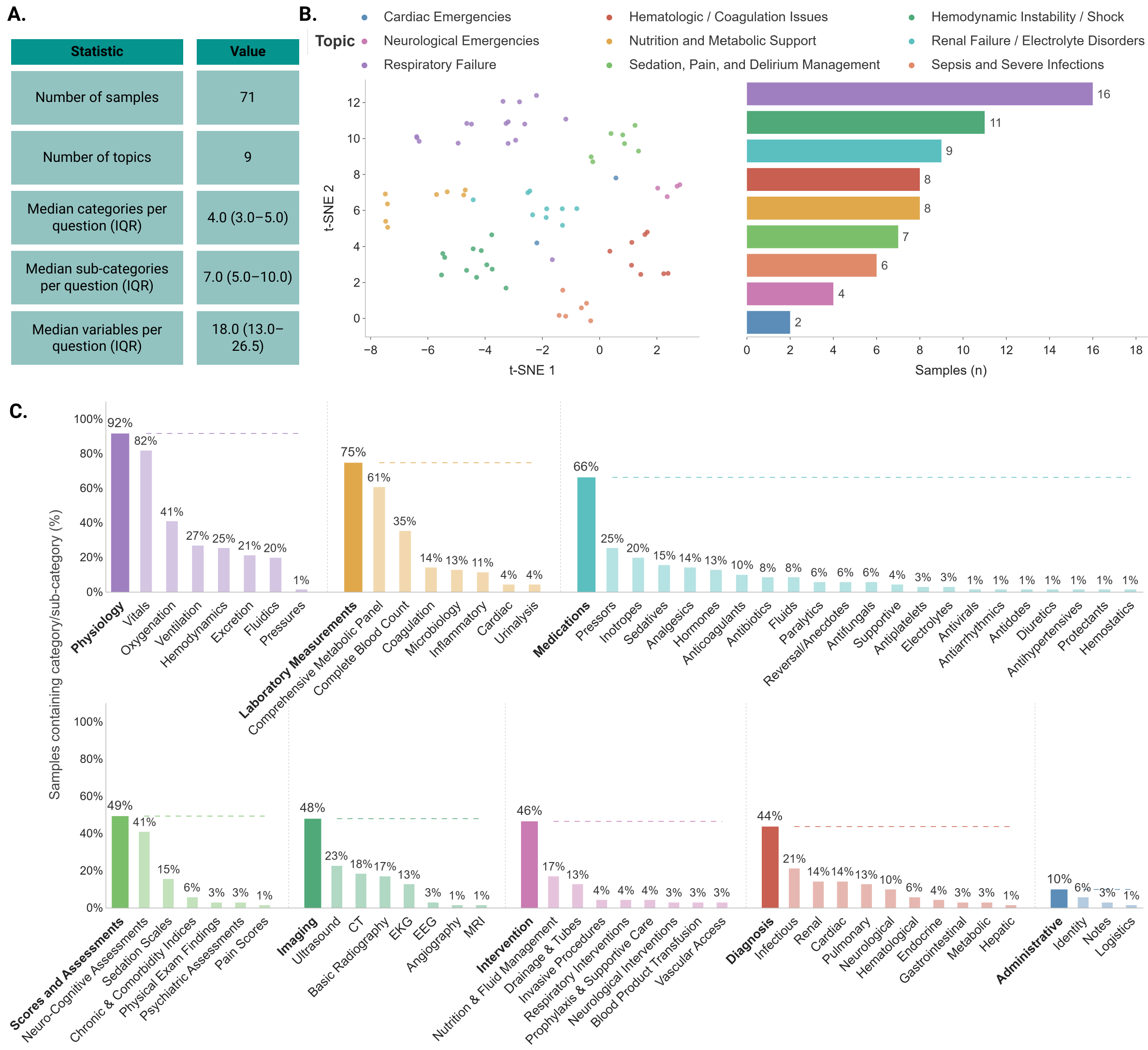}
\caption{ICU-REACT-Test dataset description. (A) Summary of dataset composition, including the number of samples and clinical topics, and the median number of categories, sub-categories, and variables represented per question. (B) t-distributed stochastic neighbor embedding (t-SNE) visualization of questions according to their clinical content, colored by the nine ICU topic areas, alongside the number of samples on each topic. (C) Proportion of test-set questions containing each information category, grouped into physiologic variables, laboratory measurements, medications, scores and assessments, imaging, interventions, diagnoses, and administrative information.}
\label{fig:test_dataset}

\end{figure*}

\begin{figure}[htbp]
\centering
\includegraphics[width=0.95\linewidth]{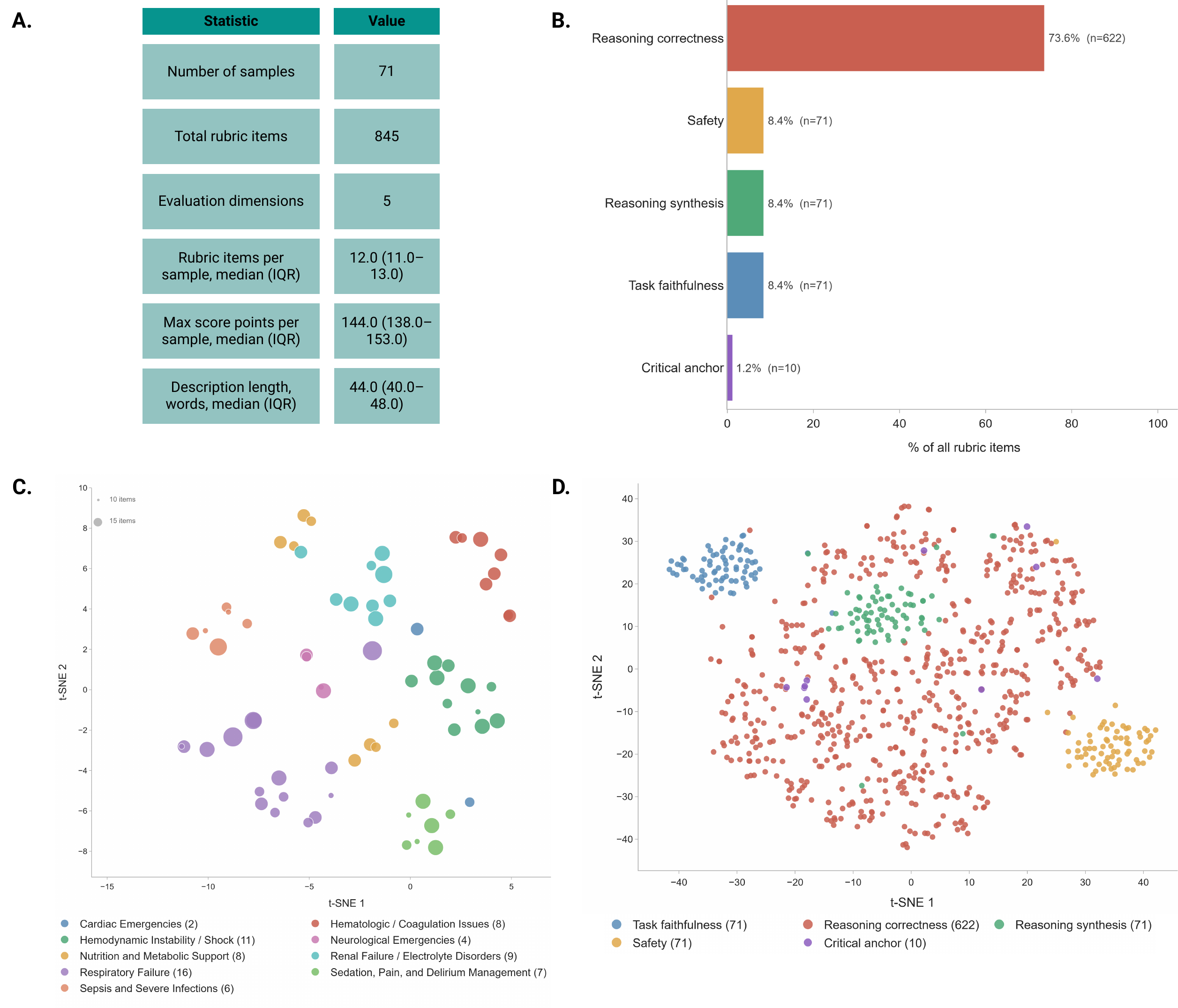}
\caption{Rubric statistics and item distribution for the LLM-Judge evaluation on the ICU-REACT test set. (A) Summary statistics of the rubric set, comprising 71 samples and 845 rubric items across 5 evaluation dimensions, together with the per-sample median (IQR) of rubric items, maximum score points, and rubric-description length in words. (B) Distribution of rubric items across the five evaluation dimensions (Reasoning correctness, Safety, Reasoning synthesis, Task faithfulness, and Critical anchor), expressed as a percentage of all 845 rubric items with item counts in parentheses. (C) Two-dimensional t-distributed stochastic neighbor embedding (t-SNE) projection at the sample level, in which each marker is one of the 71 samples, sized by its number of rubric items and colored by its ICU-REACT clinical topic; legend counts give the number of samples per topic. (D) Two-dimensional t-SNE projection at the rubric-item level, in which each point is one of the 845 rubric items, colored by its evaluation dimension; legend counts give the number of items per dimension.}
\label{fig:rubrics_stats}
\end{figure}

\section{ICU-REACT dataset details}\label{dataset_details}

\subsection{ICU-REACT-Train}
ICU-REACT-Train-Small (Clin-REACT 8B) splits its 10{,}000 examples evenly into 5{,}000 reasoning-refinement and 5{,}000 variable-selection examples, pairing the reasoning task with explicit information-retrieval supervision at the smallest scale, whereas ICU-REACT-Train-Medium (5{,}307 examples; Clin-REACT 14B and 31B) and ICU-REACT-Train-Large (27{,}973 examples; Clin-REACT 70B) consist solely of reasoning-refinement examples. Each such example is a two-stage target, in which an initial reasoning trace is revised into an improved trace that the model learns to produce. Figure~\ref{fig:train_datasets} shows t-SNE projections of the three sets colored by primary ICU topic: for each scale, the left panel depicts the full set of input--output examples, while the middle and right panels separate the initial- and improved-reasoning stages, illustrating the overall sample distribution and its shift between stages.

\subsection{ICU-REACT-Test}
The full topic distribution of ICU-REACT-Test (Fig.~\ref{fig:test_dataset}B) is respiratory failure ($n=16$), hemodynamic instability/shock ($n=11$), renal failure/electrolyte disorders ($n=9$), nutrition and metabolic support ($n=8$), hematologic/coagulation issues ($n=8$), sedation, pain, and delirium management ($n=7$), sepsis and severe infections ($n=6$), neurological emergencies ($n=4$), and cardiac emergencies ($n=2$), with a two-dimensional t-SNE projection of the question embeddings showing coherent clustering by topic. Ordered by prevalence, the top-level information categories are physiology (92\%), laboratory measurements (75\%), medications (66\%), scores and assessments (49\%), imaging (48\%), intervention (46\%), diagnosis (44\%), and administrative (10\%) (Fig.~\ref{fig:test_dataset}C); within each, prevalence concentrated in a few dominant sub-categories, including vitals (82\%) within physiology, the comprehensive metabolic panel (61\%) within laboratory measurements, and pressors (25\%) within medications.

\subsection{Evaluation rubrics}
Model outputs on ICU-REACT-Test were scored by an LLM-Judge against item-level rubrics (Fig.~\ref{fig:rubrics_stats}). The rubric set covers all 71 test samples with 845 items across five evaluation dimensions, at a median of 12.0 items per sample (IQR 11.0--13.0), a median maximum of 144.0 score points per sample (IQR 138.0--153.0), and a median description length of 44.0 words (IQR 40.0--48.0) (Fig.~\ref{fig:rubrics_stats}A). Items are dominated by reasoning correctness (73.6\%, $n=622$), followed by safety, reasoning synthesis, and task faithfulness (each 8.4\%, $n=71$), and critical anchor (1.2\%, $n=10$) (Fig.~\ref{fig:rubrics_stats}B). Two-dimensional t-SNE projections color the items by their source sample's clinical topic (Fig.~\ref{fig:rubrics_stats}C) and by evaluation dimension (Fig.~\ref{fig:rubrics_stats}D), with both topics and dimensions forming visibly separated clusters.

\section{Training hyperparameters}\label{train_params}

All Clin-REACT variants were fine-tuned with low-rank adaptation (LoRA), leaving the backbone weights frozen and updating only the injected adapters. We used identical adapter settings across all four scales ($r=16$, $\alpha=32$, dropout $0.05$); in preliminary experiments, varying the rank, scaling factor, and dropout produced no appreciable change in downstream performance, so a single configuration was retained across scales to keep adapter capacity constant. Optimization used a learning rate of $1\times10^{-5}$ with a weight decay of $0.01$ and a linear warmup over the first $100$ steps. All models were trained with a per-device batch size of $4$ and $4$ gradient accumulation steps, giving an effective batch size of $16$. Clin-REACT 8B was trained for $5$ epochs and the remaining variants for $3$. Each training corpus was split $90{:}10$ into training and validation partitions, with the validation partition used only for checkpoint selection. Table~\ref{tab:clinreact_hyperparameters} reports the full configuration for each variant.

\begin{table}[htbp]
\centering
\footnotesize
\caption{Training hyperparameters for each Clin-REACT variant.}
\label{tab:clinreact_hyperparameters}
\begin{tabular}{lcccc}
\toprule
Hyperparameter & Clin-REACT 8B & Clin-REACT 14B & Clin-REACT 31B & Clin-REACT 70B \\
\midrule
LoRA rank ($r$) & 16 & 16 & 16 & 16 \\
LoRA alpha ($\alpha$) & 32 & 32 & 32 & 32 \\
LoRA dropout & 0.05 & 0.05 & 0.05 & 0.05 \\
Learning rate & $1\times 10^{-5}$ & $1\times 10^{-5}$ & $1\times 10^{-5}$ & $1\times 10^{-5}$ \\
Weight decay & 0.01 & 0.01 & 0.01 & 0.01 \\
Warmup steps & 100 & 100 & 100 & 100 \\
Batch size (per device) & 4 & 4 & 4 & 4 \\
Gradient accumulation steps & 4 & 4 & 4 & 4 \\
Effective batch size & 16 & 16 & 16 & 16 \\
Epochs & 5 & 3 & 3 & 3 \\
Training samples (total) & 10,000 & 5,307 & 5,307 & 27,973 \\
Training samples (train split) & 9,000 & 4,776 & 4,776 & 25,176 \\
Train:validation split (\%) & 90:10 & 90:10 & 90:10 & 90:10 \\
\bottomrule
\end{tabular}
\end{table}

\begin{figure}[htbp]
\centering
\includegraphics[width=0.95\linewidth]{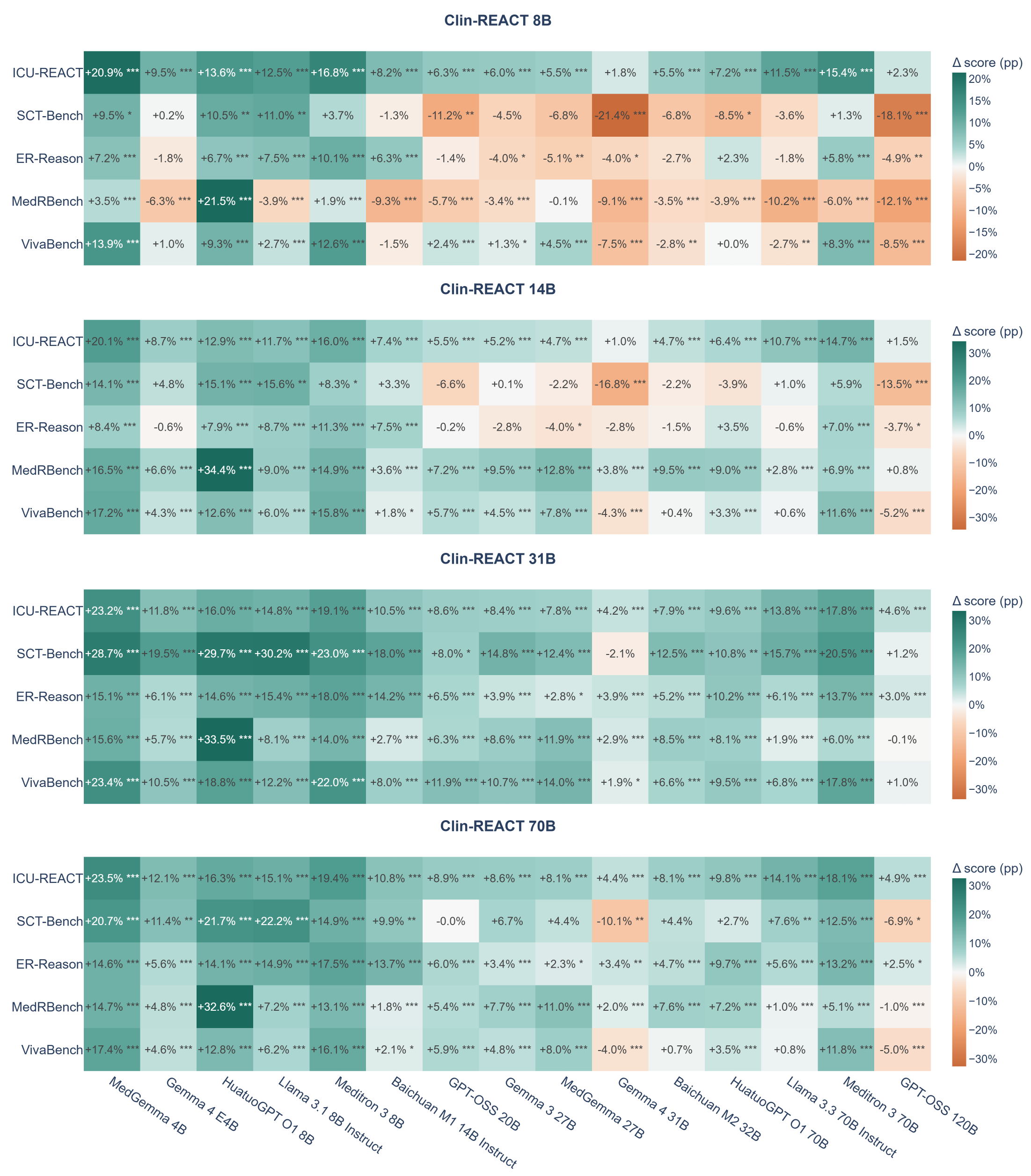}
\caption{Per-benchmark score differences between each Clin-REACT model and every baseline model across five clinical-reasoning benchmarks. Panels correspond to the four Clin-REACT models (top to bottom: Clin-REACT 8B, 14B, 31B, and 70B). Each cell reports the difference in score ($\Delta$, percentage points) between the Clin-REACT model and a baseline model on one benchmark (rows), computed as the Clin-REACT model minus the baseline using each benchmark's average-across-metrics score; green indicates the Clin-REACT model scored higher, orange indicates the baseline scored higher, and color intensity represents the magnitude of the difference. Columns are the 15 open-source baseline models (medical and general purpose), ordered by parameter count; rows are the five clinical-reasoning benchmarks (ICU-REACT, SCT-Bench, ER-Reason, MedRBench, and VivaBench). Numeric labels give each cell's $\Delta$, and asterisks denote the statistical significance of the paired difference (two-sided Wilcoxon signed-rank test; *, $p < 0.05$; **, $p < 0.01$; ***, $p < 0.001$). Each panel uses an independent color scale.}
\label{fig:heatmaps_comparisons}
\end{figure}

\begin{figure}[htbp]
\centering
\includegraphics[width=0.95\linewidth]{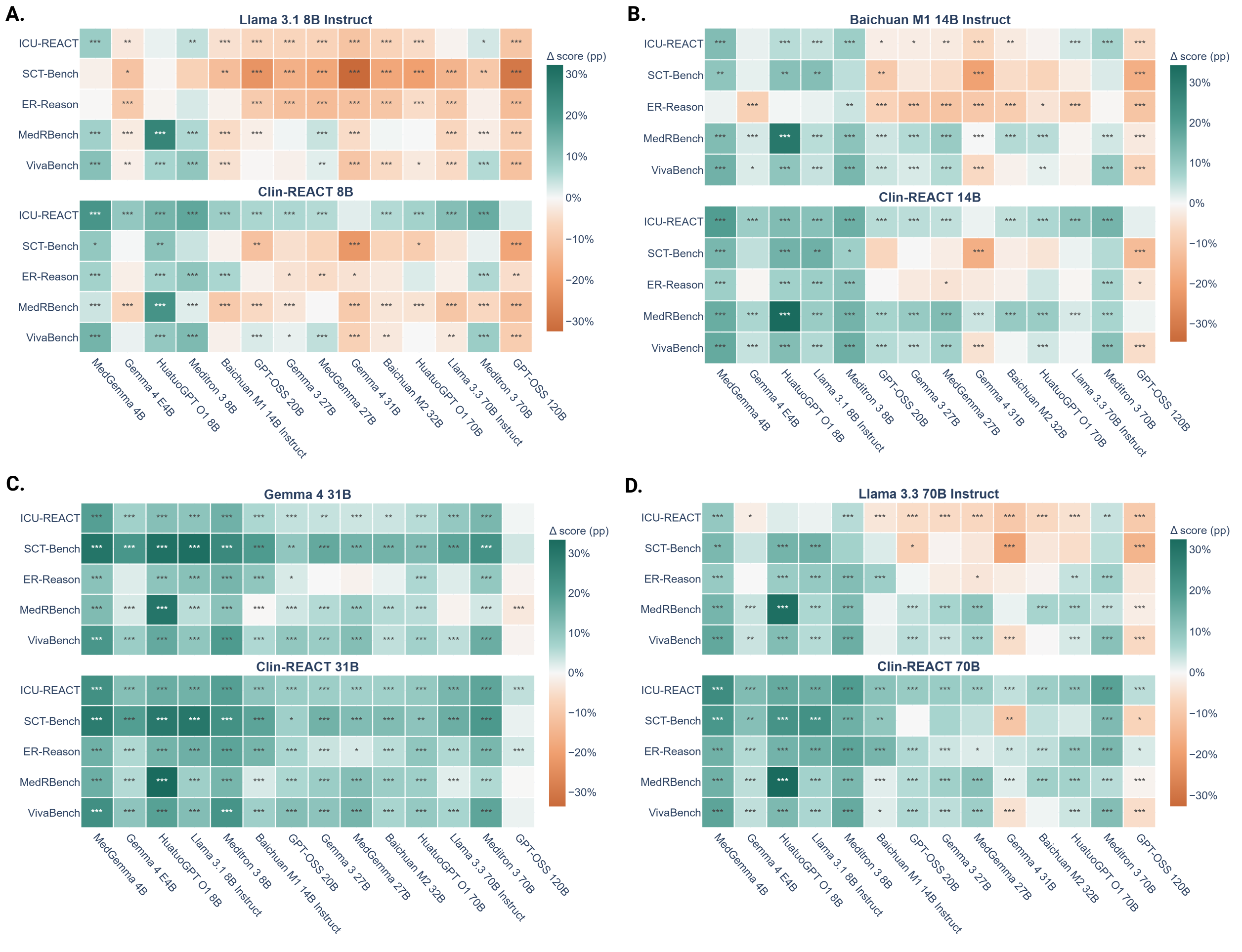}
\caption{Per-benchmark score differences of each Clin-REACT model and its backbone against baseline models across five clinical-reasoning benchmarks. Each panel pairs a backbone (top heatmap) with its fine-tuned Clin-REACT counterpart (bottom heatmap) at a single scale: (A) Llama 3.1 8B Instruct and Clin-REACT 8B, (B) Baichuan M1 14B Instruct and Clin-REACT 14B, (C) Gemma 4 31B and Clin-REACT 31B, and (D) Llama 3.3 70B Instruct and Clin-REACT 70B. Each cell reports the difference in score ($\Delta$, percentage points) between the featured model (backbone or Clin-REACT) and a baseline model on one benchmark (rows), computed as the featured model minus the baseline using each benchmark's average-across-metrics score; green indicates the featured model scored higher, orange indicates the baseline scored higher, and color intensity represents the magnitude of the difference. Columns are the baseline models (all evaluated baselines except the panel's backbone), ordered by parameter count; rows are the five clinical-reasoning benchmarks (ICU-REACT, SCT-Bench, ER-Reason, MedRBench, and VivaBench), and the top and bottom heatmaps of each panel share the same columns and rows. Asterisks denote the statistical significance of the paired difference (two-sided Wilcoxon signed-rank test; *, $p < 0.05$; **, $p < 0.01$; ***, $p < 0.001$). Each panel uses an independent color scale.}
\label{fig:heatmaps_backbones}
\end{figure}

\begin{sidewaystable*}[p]
\centering
\normalsize
\caption{Performance of Clin-REACT models against baselines across clinical reasoning benchmarks.}
\label{tab:all_model_results}
\setlength{\tabcolsep}{1.8pt}
\renewcommand{\arraystretch}{1.20}
\begin{adjustbox}{max width=\linewidth}
\begin{tabular}{l@{\hspace{0.8pt}}cccccccccccccc}
\toprule
\multirow{2}{*}{\textbf{Model}} & \multicolumn{3}{c}{\textbf{ICU-REACT}} & \multicolumn{1}{c}{\textbf{SCT-Bench}} & \multicolumn{3}{c}{\textbf{ER-Reason}} & \multicolumn{4}{c}{\textbf{MedR-Bench}} & \multicolumn{3}{c}{\textbf{VivaBench}} \\
\cmidrule(lr){2-4} \cmidrule(lr){5-5} \cmidrule(lr){6-8} \cmidrule(lr){9-12} \cmidrule(lr){13-15}
 & \textbf{Parent F1} & \textbf{Variable F1} & \textbf{Reasoning} & \textbf{SCT} & \textbf{Decision} & \textbf{Differential} & \textbf{Treatment} & \textbf{Assess. Rec.} & \textbf{Assess. Prec.} & \textbf{Diagnosis} & \textbf{Treatment} & \textbf{Overall Recall} & \textbf{Overall Precision} & \textbf{Final Dx} \\
\midrule
Clin-REACT 70B & \shortstack{42.7 \\ (39.4--46.3)} & \shortstack{31.6 \\ (28.9--34.3)} & \textbf{\shortstack{60.9 \\ (58.2--63.4)}}\textsuperscript{***} & \shortstack{67.5 \\ (61.6--73.0)} & \shortstack{58.4 \\ (52.9--63.5)} & \shortstack{52.2 \\ (46.2--57.9)} & \textbf{\shortstack{41.9 \\ (36.3--47.3)}} & \shortstack{47.7 \\ (46.1--49.6)} & \shortstack{25.8 \\ (24.7--27.1)} & \shortstack{65.2 \\ (62.1--68.3)} & \shortstack{45.4 \\ (41.1--49.8)} & \underline{\shortstack{30.3 \\ (29.1--31.4)}} & \shortstack{18.3 \\ (17.6--19.0)} & \shortstack{34.0 \\ (31.2--37.0)} \\
Clin-REACT 31B & \textbf{\shortstack{45.1 \\ (42.0--48.3)}} & \textbf{\shortstack{35.4 \\ (32.9--37.9)}} & \shortstack{53.8 \\ (51.7--55.8)} & \underline{\shortstack{75.5 \\ (69.8--80.6)}} & \textbf{\shortstack{60.1 \\ (54.2--65.7)}}\textsuperscript{*} & \textbf{\shortstack{54.2 \\ (48.3--60.7)}} & \shortstack{39.7 \\ (34.1--45.3)} & \shortstack{48.0 \\ (46.2--49.7)} & \underline{\shortstack{29.8 \\ (28.4--31.1)}} & \underline{\shortstack{66.0 \\ (62.7--69.0)}} & \shortstack{44.0 \\ (39.8--48.3)} & \shortstack{29.5 \\ (28.4--30.6)} & \underline{\shortstack{22.8 \\ (21.9--23.7)}} & \textbf{\shortstack{48.2 \\ (45.4--51.5)}} \\
Clin-REACT 14B & \shortstack{38.9 \\ (35.5--42.6)} & \shortstack{25.8 \\ (23.2--28.3)} & \shortstack{60.2 \\ (57.8--62.8)} & \shortstack{60.9 \\ (54.7--66.8)} & \shortstack{51.4 \\ (45.5--57.3)} & \shortstack{47.2 \\ (42.1--52.7)} & \shortstack{35.4 \\ (30.1--40.9)} & \underline{\shortstack{48.5 \\ (46.9--50.3)}} & \shortstack{29.5 \\ (28.2--30.6)} & \shortstack{59.1 \\ (55.7--62.2)} & \textbf{\shortstack{54.4 \\ (50.0--58.5)}}\textsuperscript{***} & \shortstack{27.3 \\ (26.3--28.4)} & \shortstack{21.7 \\ (20.8--22.7)} & \shortstack{32.9 \\ (29.7--35.9)} \\
Clin-REACT 8B & \shortstack{41.8 \\ (38.7--44.7)} & \shortstack{34.1 \\ (31.5--36.8)} & \shortstack{51.4 \\ (49.1--53.6)} & \shortstack{56.3 \\ (49.7--62.2)} & \shortstack{55.8 \\ (50.9--60.6)} & \shortstack{44.8 \\ (39.9--49.7)} & \shortstack{29.8 \\ (25.4--34.5)} & \shortstack{42.6 \\ (41.0--44.3)} & \shortstack{23.5 \\ (22.3--24.8)} & \shortstack{48.5 \\ (45.1--52.0)} & \shortstack{25.1 \\ (21.6--29.3)} & \shortstack{26.6 \\ (25.4--27.6)} & \shortstack{20.3 \\ (19.5--21.1)} & \shortstack{25.3 \\ (22.3--28.2)} \\
\midrule
HuatuoGPT O1 70B & \shortstack{42.1 \\ (39.3--45.1)} & \shortstack{29.6 \\ (27.2--32.2)} & \shortstack{33.8 \\ (32.3--35.1)} & \shortstack{64.8 \\ (58.6--71.2)} & \shortstack{50.4 \\ (45.0--55.7)} & \shortstack{40.4 \\ (34.3--46.5)} & \shortstack{32.7 \\ (27.2--38.2)} & \shortstack{29.8 \\ (28.2--31.4)} & \textbf{\shortstack{49.0 \\ (47.0--51.3)}}\textsuperscript{***} & \shortstack{47.5 \\ (44.4--50.8)} & \shortstack{28.8 \\ (24.9--33.0)} & \shortstack{14.1 \\ (13.3--14.9)} & \shortstack{27.9 \\ (26.3--29.3)} & \shortstack{30.0 \\ (27.1--33.1)} \\
Meditron 3 70B & \shortstack{36.0 \\ (31.7--39.8)} & \shortstack{19.9 \\ (16.7--22.9)} & \shortstack{25.0 \\ (23.5--26.5)} & \shortstack{55.0 \\ (48.8--61.9)} & \shortstack{43.0 \\ (37.6--48.2)} & \shortstack{43.0 \\ (37.2--49.4)} & \shortstack{26.9 \\ (21.5--32.3)} & \shortstack{36.6 \\ (34.9--38.3)} & \shortstack{37.1 \\ (35.5--38.7)} & \shortstack{53.7 \\ (50.5--57.0)} & \shortstack{36.3 \\ (32.4--41.1)} & \shortstack{14.8 \\ (13.9--15.8)} & \shortstack{22.4 \\ (21.0--23.7)} & \shortstack{10.1 \\ (8.1--11.9)} \\
Baichuan M2 32B & \shortstack{41.1 \\ (38.6--43.9)} & \shortstack{31.3 \\ (29.1--33.4)} & \shortstack{38.3 \\ (36.6--40.0)} & \shortstack{63.0 \\ (56.4--68.8)} & \shortstack{57.4 \\ (52.2--62.8)} & \shortstack{49.0 \\ (43.3--54.5)} & \shortstack{32.1 \\ (27.2--37.1)} & \shortstack{48.0 \\ (46.4--49.7)} & \shortstack{16.7 \\ (15.9--17.6)} & \shortstack{55.5 \\ (52.0--58.9)} & \shortstack{33.4 \\ (29.0--38.2)} & \shortstack{29.4 \\ (28.3--30.6)} & \shortstack{17.8 \\ (17.1--18.6)} & \shortstack{33.4 \\ (30.4--36.6)} \\
MedGemma 27B & \shortstack{41.5 \\ (38.6--44.4)} & \shortstack{30.0 \\ (27.5--32.4)} & \shortstack{39.3 \\ (37.3--41.3)} & \shortstack{63.1 \\ (56.8--68.9)} & \shortstack{57.7 \\ (51.6--63.4)} & \underline{\shortstack{53.0 \\ (46.8--59.7)}} & \shortstack{35.2 \\ (30.1--40.5)} & \shortstack{32.9 \\ (30.9--34.8)} & \shortstack{24.4 \\ (22.8--26.2)} & \shortstack{45.4 \\ (42.1--48.6)} & \shortstack{37.3 \\ (33.0--41.9)} & \shortstack{18.5 \\ (17.6--19.4)} & \shortstack{24.0 \\ (22.7--25.3)} & \shortstack{16.1 \\ (13.6--18.4)} \\
Baichuan M1 14B Instruct & \shortstack{41.7 \\ (38.4--45.2)} & \shortstack{28.6 \\ (25.4--31.7)} & \shortstack{32.4 \\ (30.3--34.5)} & \shortstack{57.6 \\ (51.7--63.7)} & \shortstack{41.0 \\ (35.4--46.9)} & \shortstack{41.5 \\ (36.0--47.1)} & \shortstack{29.0 \\ (24.5--33.9)} & \shortstack{38.8 \\ (37.1--40.4)} & \shortstack{42.2 \\ (40.4--44.0)} & \shortstack{57.1 \\ (53.6--60.2)} & \shortstack{38.8 \\ (34.6--43.4)} & \shortstack{19.1 \\ (18.2--20.0)} & \textbf{\shortstack{29.0 \\ (27.6--30.4)}}\textsuperscript{***} & \shortstack{28.4 \\ (25.7--31.4)} \\
HuatuoGPT O1 8B & \shortstack{36.4 \\ (32.8--39.8)} & \shortstack{19.8 \\ (17.3--22.2)} & \shortstack{30.1 \\ (28.4--31.9)} & \shortstack{45.8 \\ (39.5--52.2)} & \shortstack{42.1 \\ (36.6--47.8)} & \shortstack{39.8 \\ (34.8--44.5)} & \shortstack{28.3 \\ (22.9--34.0)} & \shortstack{3.6 \\ (2.8--4.4)} & \shortstack{5.0 \\ (4.0--6.1)} & \shortstack{25.1 \\ (22.4--27.7)} & \shortstack{20.1 \\ (16.6--23.9)} & \shortstack{10.1 \\ (9.4--10.8)} & \shortstack{22.2 \\ (20.5--23.7)} & \shortstack{11.9 \\ (10.1--14.0)} \\
Meditron 3 8B & \shortstack{31.4 \\ (26.9--35.8)} & \shortstack{16.6 \\ (13.5--19.6)} & \shortstack{28.8 \\ (27.0--30.5)} & \shortstack{52.6 \\ (46.0--59.2)} & \shortstack{41.7 \\ (36.7--46.6)} & \shortstack{34.9 \\ (29.2--40.9)} & \shortstack{23.5 \\ (18.3--29.0)} & \shortstack{29.9 \\ (28.2--31.4)} & \shortstack{33.8 \\ (32.1--35.8)} & \shortstack{41.4 \\ (38.0--44.4)} & \shortstack{27.0 \\ (23.0--30.7)} & \shortstack{8.1 \\ (7.3--8.8)} & \shortstack{20.3 \\ (18.6--22.1)} & \shortstack{6.0 \\ (4.6--7.6)} \\
MedGemma 4B & \shortstack{27.9 \\ (24.3--31.5)} & \shortstack{14.0 \\ (11.5--16.6)} & \shortstack{22.7 \\ (21.0--24.7)} & \shortstack{46.8 \\ (40.5--53.2)} & \shortstack{39.7 \\ (32.7--46.7)} & \shortstack{39.3 \\ (34.2--44.7)} & \shortstack{29.7 \\ (24.5--35.1)} & \shortstack{36.5 \\ (34.8--38.4)} & \shortstack{23.7 \\ (22.5--25.0)} & \shortstack{39.6 \\ (36.4--43.1)} & \shortstack{25.7 \\ (22.0--29.7)} & \shortstack{7.5 \\ (6.9--8.1)} & \shortstack{19.3 \\ (17.4--20.9)} & \shortstack{3.6 \\ (2.5--4.9)} \\
\midrule
GPT-OSS 120B & \shortstack{41.1 \\ (38.3--44.1)} & \underline{\shortstack{35.2 \\ (32.4--37.9)}} & \shortstack{43.9 \\ (41.7--46.0)} & \shortstack{74.3 \\ (68.8--79.5)} & \shortstack{57.1 \\ (51.0--63.5)} & \shortstack{52.1 \\ (46.3--57.7)} & \shortstack{35.9 \\ (30.7--41.3)} & \textbf{\shortstack{54.7 \\ (52.8--56.4)}}\textsuperscript{***} & \shortstack{23.0 \\ (21.9--24.1)} & \textbf{\shortstack{67.1 \\ (63.5--70.2)}} & \shortstack{43.4 \\ (39.0--47.9)} & \textbf{\shortstack{36.6 \\ (35.3--37.8)}}\textsuperscript{***} & \shortstack{17.8 \\ (17.1--18.6)} & \shortstack{43.0 \\ (39.8--46.0)} \\
Llama 3.3 70B Instruct & \shortstack{40.0 \\ (36.9--43.2)} & \shortstack{27.5 \\ (25.0--29.8)} & \shortstack{25.2 \\ (23.8--26.6)} & \shortstack{59.9 \\ (53.4--65.6)} & \shortstack{50.7 \\ (44.7--56.5)} & \shortstack{47.5 \\ (41.2--54.3)} & \underline{\shortstack{37.5 \\ (31.9--43.2)}} & \shortstack{42.7 \\ (40.9--44.5)} & \shortstack{31.8 \\ (30.5--33.2)} & \shortstack{61.7 \\ (58.1--65.0)} & \underline{\shortstack{44.2 \\ (39.6--48.8)}} & \shortstack{22.5 \\ (21.5--23.5)} & \shortstack{24.7 \\ (23.5--25.8)} & \shortstack{33.0 \\ (30.0--36.0)} \\
Gemma 4 31B & \underline{\shortstack{44.2 \\ (41.1--47.7)}} & \shortstack{33.1 \\ (30.3--35.9)} & \underline{\shortstack{44.4 \\ (42.2--46.7)}} & \textbf{\shortstack{77.6 \\ (72.6--82.6)}} & \shortstack{56.7 \\ (51.0--62.4)} & \shortstack{50.8 \\ (45.1--56.4)} & \shortstack{34.7 \\ (29.0--40.2)} & \shortstack{41.1 \\ (39.4--42.8)} & \shortstack{33.9 \\ (32.4--35.4)} & \shortstack{61.8 \\ (58.4--65.0)} & \shortstack{39.4 \\ (35.3--44.0)} & \shortstack{23.6 \\ (22.7--24.5)} & \shortstack{25.9 \\ (24.9--27.0)} & \underline{\shortstack{45.3 \\ (42.0--48.2)}} \\
Gemma 3 27B & \shortstack{42.1 \\ (39.1--45.3)} & \shortstack{29.9 \\ (27.1--32.5)} & \shortstack{37.2 \\ (35.2--39.0)} & \shortstack{60.8 \\ (54.6--67.1)} & \underline{\shortstack{58.4 \\ (53.0--63.6)}} & \shortstack{47.2 \\ (41.5--53.4)} & \shortstack{36.6 \\ (31.3--41.8)} & \shortstack{43.5 \\ (41.8--45.2)} & \shortstack{26.5 \\ (25.3--27.9)} & \shortstack{50.9 \\ (47.5--54.0)} & \shortstack{32.4 \\ (28.2--37.1)} & \shortstack{17.9 \\ (17.0--18.7)} & \shortstack{24.3 \\ (23.0--25.6)} & \shortstack{26.1 \\ (23.4--29.2)} \\
GPT-OSS 20B & \shortstack{39.7 \\ (36.5--42.9)} & \shortstack{30.5 \\ (28.1--33.2)} & \shortstack{38.4 \\ (35.4--40.8)} & \shortstack{67.5 \\ (61.9--73.4)} & \shortstack{54.3 \\ (48.9--60.0)} & \shortstack{50.9 \\ (45.2--56.5)} & \shortstack{29.2 \\ (24.7--33.6)} & \shortstack{48.1 \\ (46.1--49.9)} & \shortstack{23.4 \\ (22.3--24.7)} & \shortstack{56.8 \\ (53.6--60.2)} & \shortstack{34.2 \\ (29.9--38.8)} & \shortstack{25.9 \\ (24.8--26.9)} & \shortstack{22.8 \\ (21.6--23.9)} & \shortstack{16.2 \\ (13.8--18.6)} \\
Gemma 4 E4B & \shortstack{39.4 \\ (36.3--42.5)} & \shortstack{27.2 \\ (24.3--30.0)} & \shortstack{32.2 \\ (30.6--33.8)} & \shortstack{56.1 \\ (49.5--62.3)} & \shortstack{53.9 \\ (48.2--59.8)} & \shortstack{50.1 \\ (44.7--55.5)} & \shortstack{31.6 \\ (26.7--36.9)} & \shortstack{39.6 \\ (37.8--41.4)} & \shortstack{31.7 \\ (30.1--33.4)} & \shortstack{53.7 \\ (50.2--57.2)} & \shortstack{39.8 \\ (35.5--44.4)} & \shortstack{24.2 \\ (23.3--25.3)} & \shortstack{26.6 \\ (25.4--27.7)} & \shortstack{18.1 \\ (15.7--20.6)} \\
Llama 3.1 8B Instruct & \shortstack{39.4 \\ (36.2--42.7)} & \shortstack{24.6 \\ (22.2--27.0)} & \shortstack{25.8 \\ (24.3--27.3)} & \shortstack{45.3 \\ (38.4--51.9)} & \shortstack{44.8 \\ (39.5--50.2)} & \shortstack{37.5 \\ (32.6--42.7)} & \shortstack{25.7 \\ (20.7--30.7)} & \shortstack{39.0 \\ (37.3--40.7)} & \shortstack{32.2 \\ (30.8--33.6)} & \shortstack{47.8 \\ (44.5--50.9)} & \shortstack{36.5 \\ (32.4--41.1)} & \shortstack{21.9 \\ (21.0--22.9)} & \shortstack{24.6 \\ (23.4--25.7)} & \shortstack{17.5 \\ (15.1--19.9)} \\
\bottomrule
\end{tabular}
\end{adjustbox}
\vspace{2mm}
\begin{minipage}{0.98\linewidth}
\tiny 
Values are mean percentages with 95\% confidence intervals in parentheses across a 1,000-iteration bootstrap with replacement. For each metric, the highest-scoring Clin-REACT model and highest-scoring non-Clin-REACT model are selected. The higher of those two scores is shown in bold, and the lower score is underlined. Significance symbols indicate the paired Wilcoxon signed-rank p-value comparing those two models: \textsuperscript{*}$p<0.05$, \textsuperscript{**}$p<0.01$, and \textsuperscript{***}$p<0.001$. Abbreviations: Parent F1, parent-variable F1 score; Variable F1, individual-variable F1 score; Reasoning, reasoning score; SCT, SCT score; Decision, decision factors; Assess. Rec., assessment recall; Assess. Prec., assessment precision; Final Dx, final diagnosis accuracy. Included model groups: Clin-REACT, Open-source medical LLMs, Open-source general-purpose LLMs.
\end{minipage}
\end{sidewaystable*}

\section{Comparisons against all baselines}\label{all_baseline_comparisons}

A detailed comparison of each Clin-REACT model against all baseline models can be seen in Fig.~\ref{fig:heatmaps_comparisons}. When compared with its backbone architecture (Llama 3.1 8B Instruct), Clin-REACT 8B achieved statistically significant gains of +12.5\% ($p < 0.001$), +11.0\% ($p < 0.01$), +7.5\% ($p < 0.001$) and +2.7\% ($p < 0.001$) on ICU-REACT, SCT-Bench, ER-Reason, and VivaBench, respectively, with slight degradation on MedRBench (-3.9\%, $p < 0.001$). On the other hand, Clin-REACT 14B improved across all benchmarks with respect to its backbone (Baichuan M1 14B Instruct), with mostly significant gains of +7.4\% ($p < 0.001$), +3.3\% ($p > 0.05$), +7.5\% ($p < 0.001$), +3.6\% ($p < 0.001$) and +1.8\% ($p < 0.05$) on ICU-REACT, SCT-Bench, ER-Reason, MedRBench, and VivaBench, respectively.

Similarly, Clin-REACT 31B outperformed its backbone (Gemma 4 31B) on four of five benchmarks, with statistically significant gains of +4.2\% on ICU-REACT ($p < 0.001$), +3.9\% on ER-Reason ($p < 0.001$), +2.9\% on MedRBench ($p < 0.001$), and +1.9\% on VivaBench ($p < 0.05$). Although performance decreased slightly on SCT-Bench (-2.1\%), this difference was not statistically significant ($p > 0.05$). Clin-REACT 70B demonstrated the largest improvements over its backbone (Llama 3.3 70B Instruct), achieving gains of +14.1\% ($p < 0.001$), +7.6\% ($p < 0.01$), +5.6\% ($p < 0.001$), and +1.0\% ($p < 0.001$) on ICU-REACT, SCT-Bench, ER-Reason, and MedRBench, respectively, with a small, although non-significant, improvement of +0.8\% on VivaBench ($p < 0.05$).

Among similarly sized models, Clin-REACT 8B demonstrated substantial advantages over other small baseline models. Compared with MedGemma 4B, HuatuoGPT O1 8B, and Meditron 3 8B, Clin-REACT 8B showed significant advantages across all five benchmarks, including advantages of +20.9\%, +13.6\%, and +16.8\% on ICU-REACT, respectively (all $p < 0.001$). Its strongest relative gains were observed against HuatuoGPT O1 8B on MedRBench (+21.5\%, $p < 0.001$) and against Meditron 3 8B on ICU-REACT (+16.8\%, $p < 0.001$), ER-Reason (+10.1\%, $p < 0.001$), and VivaBench (+12.6\%, $p < 0.001$). Clin-REACT 8B also outperformed Gemma 4 E4B on ICU-REACT (+9.5\%, $p < 0.001$) and SCT-Bench (+0.2\%, $p > 0.05$), although its performance was more mixed on ER-Reason and MedRBench.

Clin-REACT 14B similarly outperformed most comparably sized models, including the 8B baselines and GPT-OSS 20B. Relative to HuatuoGPT O1 8B, it achieved gains of +12.9\%, +15.1\%, +7.9\%, +34.4\%, and +12.6\% across ICU-REACT, SCT-Bench, ER-Reason, MedRBench, and VivaBench, respectively (all $p < 0.001$). It also consistently exceeded Meditron 3 8B, with improvements ranging from +8.3\% on SCT-Bench to +16.0\% on ICU-REACT (all $p < 0.001$). Compared with GPT-OSS 20B, Clin-REACT 14B achieved significant gains on ICU-REACT (+5.5\%, $p < 0.001$), MedRBench (+7.2\%, $p < 0.001$), and VivaBench (+5.7\%, $p < 0.001$), while differences on SCT-Bench and ER-Reason were not statistically significant.

Within the 20--32B parameter range, Clin-REACT 31B showed broad and consistent advantages over GPT-OSS 20B, Gemma 3 27B, MedGemma 27B, Gemma 4 31B, and Baichuan M2 32B. In particular, Clin-REACT 31B outperformed GPT-OSS 20B by +8.6\% on ICU-REACT, +8.0\% on SCT-Bench, +6.5\% on ER-Reason, +6.3\% on MedRBench, and +11.9\% on VivaBench (all $p < 0.001$). It also achieved significant gains over Gemma 3 27B of +8.4\%, +14.8\%, +3.9\%, +8.6\%, and +10.7\% across the five benchmarks, respectively. Compared with Gemma 4 31B, Clin-REACT 31B improved performance on ICU-REACT (+4.2\%, $p < 0.001$), ER-Reason (+3.9\%, $p < 0.001$), MedRBench (+2.9\%, $p < 0.001$), and VivaBench (+1.9\%, $p < 0.05$), with a non-significant decrease on SCT-Bench (-2.1\%).

Among models in the 70B parameter range, Clin-REACT 70B consistently outperformed HuatuoGPT O1 70B, Llama 3.3 70B Instruct, and Meditron 3 70B. Compared with HuatuoGPT O1 70B, it achieved gains of +9.8\% on ICU-REACT, +2.7\% on SCT-Bench, +9.7\% on ER-Reason, +7.2\% on MedRBench, and +3.5\% on VivaBench (all $p < 0.001$). Relative to Meditron 3 70B, Clin-REACT 70B showed particularly large improvements on ICU-REACT (+18.1\%, $p < 0.001$), SCT-Bench (+12.5\%, $p < 0.001$), ER-Reason (+13.2\%, $p < 0.001$), and VivaBench (+11.8\%, $p < 0.001$). These findings indicate that Clin-REACT fine-tuning conferred performance improvements not only over smaller models, but also over comparably sized general-purpose and medical LLMs. A visualization of the improvement of Clin-REACT models over their backbones when compared to baseline models can be seen in Fig.~\ref{fig:heatmaps_backbones}. Detailed results across all metrics can be found on Table~\ref{tab:all_model_results}.

\section{Ablations}\label{ablations}

We conducted two complementary ablation studies to characterize how Clin-REACT's performance depends on (i) the composition of self-supervised training tasks used to construct the ICU-REACT train datasets, and (ii) the scale of the training sets. Both studies were run on Llama-3.1-8B-Instruct and Llama-3.3-70B-Instruct backbones under identical pre-processing, split, and evaluation protocols, with all models evaluated on the held-out ICU-REACT test set.

\subsection{Training tasks}

We tested generation of the ICU-REACT train datasets through five self-supervised augmentation tasks, each contributing a distinct form of clinical supervision. Each task was framed from the perspective of an ICU clinician and operated over patient context and decision-making question pairs:
\begin{itemize}

\item \textbf{Variable Selection Reasoning.} Given a patient context and a decision-making question, produced one coherent paragraph identifying which clinical variables are most relevant to the decision and explaining why, drawing on our curated ICU variable and taxonomy framework organized by category and sub-category. Rather than emitting a final answer, the task taught models to select and justify the decision-informing data that a clinician would review, spanning severity, trajectory, contraindications, response to therapy, and safety monitoring.

\item \textbf{Reasoning Refinement.} Given a patient context, decision question, and an initial reasoning draft, produced a concise explanation of how the reasoning should be improved followed by a single revised reasoning paragraph. This critique-and-improve supervision targeted what is missing, overemphasized, or clinically misprioritized in the draft, sharpening intermediate reasoning quality and clinical coherence.

\item \textbf{Question Generation.} Given only a patient context, produced one specific ICU decision-making question that would guide immediate clinical action (e.g., regarding fluids, vasopressors, ventilation, antibiotics, sedation, or diagnostics), teaching models to identify the salient, actionable decision implied by a case.

\item \textbf{Context Generation.} Given only a decision-making question, produced one realistic patient context in which that question would be clinically applicable, complete with enough ICU-relevant detail (organ support, physiology, trajectory, comorbidities, acute problem) to justify why the question would be asked. This would potentially improve models' grounding in plausible clinical scenarios.

\item \textbf{Context/Question Refinement.} Given an under specified initial context--question pair, produced a shared explanation of why it is too under specified for ICU decision-making, followed by all clinically distinct scenario-specific formulations, each with a scenario label, clinical rationale, refined question, and refined context. This task acted as quality control, disambiguating vague inputs into coherent, decision-ready scenarios.

\end{itemize}

\begin{figure*}[h]
\centering
\includegraphics[width=0.95\linewidth]{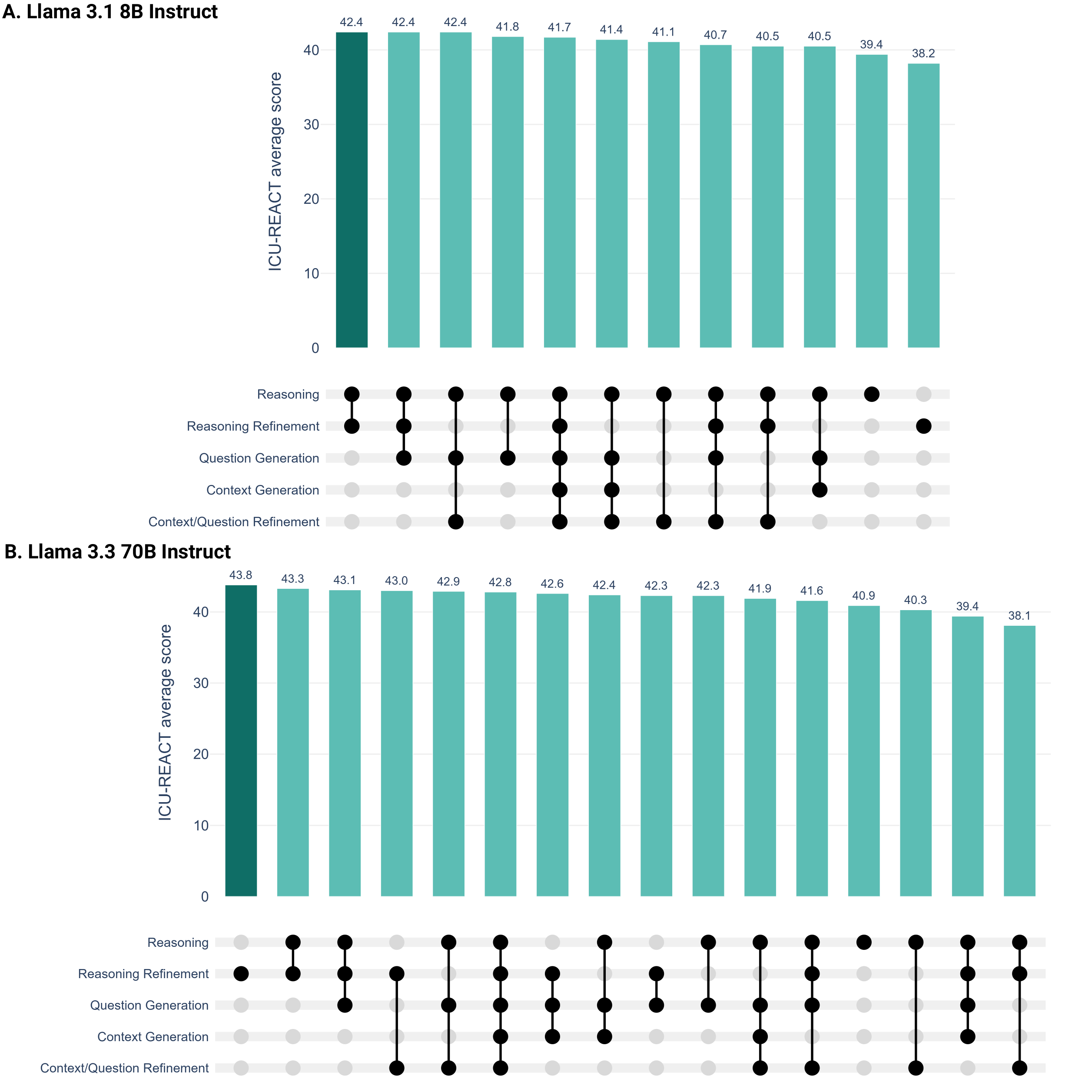}
\caption{Training task ablation. ICU-REACT average score for models trained on each combination of the five self-supervised augmentation tasks (Variable Selection Reasoning, Reasoning Refinement, Question Generation, Context Generation, and Context/Question Refinement) for (A) Llama-3.1-8B-Instruct and (B) Llama-3.3-70B-Instruct. Filled dots in the lower matrix denote the tasks included in a given configuration, connected vertically when combined; bars are sorted in descending order of average score, and the best-performing configuration in each panel is highlighted in dark teal.}
\label{fig:task_ablation}

\end{figure*}

\begin{figure*}[h]
\centering
\includegraphics[width=0.95\linewidth]{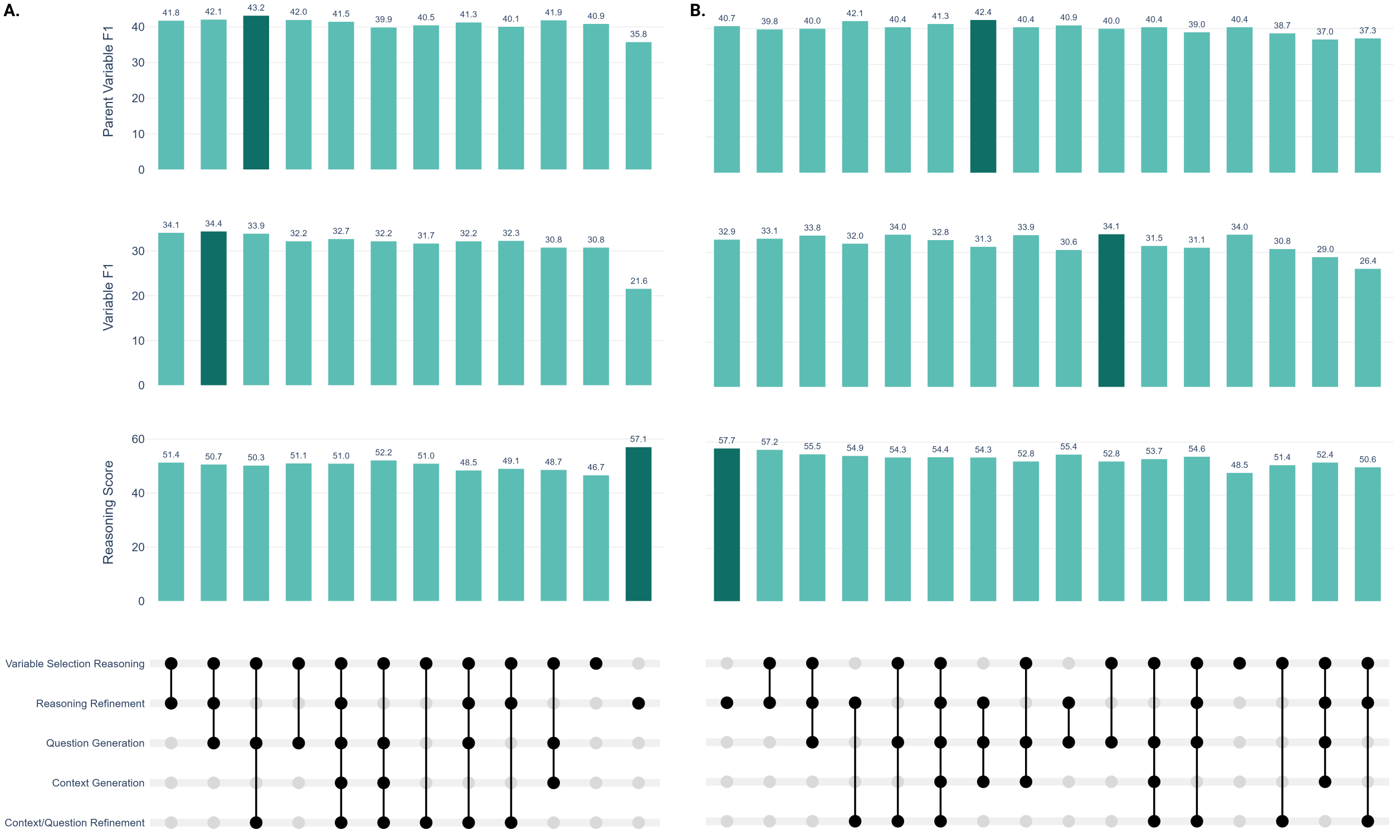}
\caption{Training task ablation decomposed by evaluation metric. ICU-REACT performance decomposed into parent variable F1 (top row), variable F1 (middle row), and reasoning score (bottom row) across training-task combinations for (A) Llama-3.1-8B-Instruct and (B) Llama-3.3-70B-Instruct. Filled dots in the lower matrix denote the tasks included in each configuration; within each metric row, the best-performing configuration is highlighted in dark teal.}
\label{fig:task_metric_ablation}

\end{figure*}

\subsection{Task ablation}

To isolate the contribution of each augmentation task, we trained both backbones on different task combinations, keeping the dataset size fixed, and evaluated each configuration on the ICU-REACT test set (Fig.~\ref{fig:task_ablation}). For Llama-3.1-8B-Instruct, the best configuration combined Variable Selection Reasoning and Reasoning Refinement, achieving an average score of 42.4 (Fig.~\ref{fig:task_ablation}A). The top three configurations each retained the Variable Selection Reasoning task, while the top two also retained the Reasoning Refinement task. Performance bottomed out at 39.4 and 38.2 when the model was trained on only Variable Selection Reasoning and Reasoning Refinement, respectively, suggesting that multi-task training benefited the 8B model the most. A slightly different pattern was observed for Llama-3.3-70B-Instruct, where the best configuration reached 43.8 by only training on the Reasoning Refinement task (Fig.~\ref{fig:task_ablation}B). Moreover, the top four configurations all included the Reasoning Refinement task. Performance generally declined with the addition of other tasks to the Reasoning Refinement task, suggesting that the 70B model benefited the most from a single task training on a critique task.

We decomposed performance into each ICU-REACT metric (parent variable F1, variable F1, and reasoning score) to assess whether the optimal task composition holds across evaluation dimensions (Fig.~\ref{fig:task_metric_ablation}). Across both backbones, Reasoning Refinement alone maximized the reasoning score (57.1 for 8B, 57.7 for 70B), but its effect on variable identification was scale-dependent. For Llama-3.1-8B-Instruct, training on Reasoning Refinement in isolation degraded both parent variable F1 (35.8) and variable F1 (21.6) to their lowest values, and pairing it with Variable Selection Reasoning was required to recover strong F1 performance (parent variable F1 41.8, variable F1 34.1) while retaining a still-robust reasoning score of 51.4 (Fig.~\ref{fig:task_metric_ablation}A). For Llama-3.3-70B-Instruct, by contrast, Reasoning Refinement alone maintained robust variable identification (parent variable F1 40.7, variable F1 32.9) and achieved the highest reasoning score (57.7), remaining competitive with the best configurations which used Variable Selection Reasoning (parent variable F1 42.4, variable F1 34.1) (Fig.~\ref{fig:task_metric_ablation}B). Thus, critique-based supervision consistently improved reasoning quality, but only the larger model absorbed this benefit without a corresponding trade-off in variable identification.

Taken together, these results indicate that the optimal training-task composition is backbone-dependent and metric-dependent, yet the Reasoning Refinement task emerged as the consistent throughline across both axes. Across both backbones, Reasoning Refinement drove the reasoning score, reflecting the central role of critique-based supervision in improving clinical reasoning quality. At 70B scale, this benefit came at no cost to variable identification, as Reasoning Refinement alone preserved robust parent variable and variable F1. At 8B scale, Reasoning Refinement in isolation degraded F1 performance, and pairing it with Variable Selection Reasoning was necessary to recover strong variable identification while maintaining a robust reasoning score. In both cases, the Reasoning Refinement task was consistently present in the top-performing configurations, underscoring the central role of critique-based supervision in driving downstream ICU-REACT performance across model scales.

\begin{figure*}[h]
\centering
\includegraphics[width=0.95\linewidth]{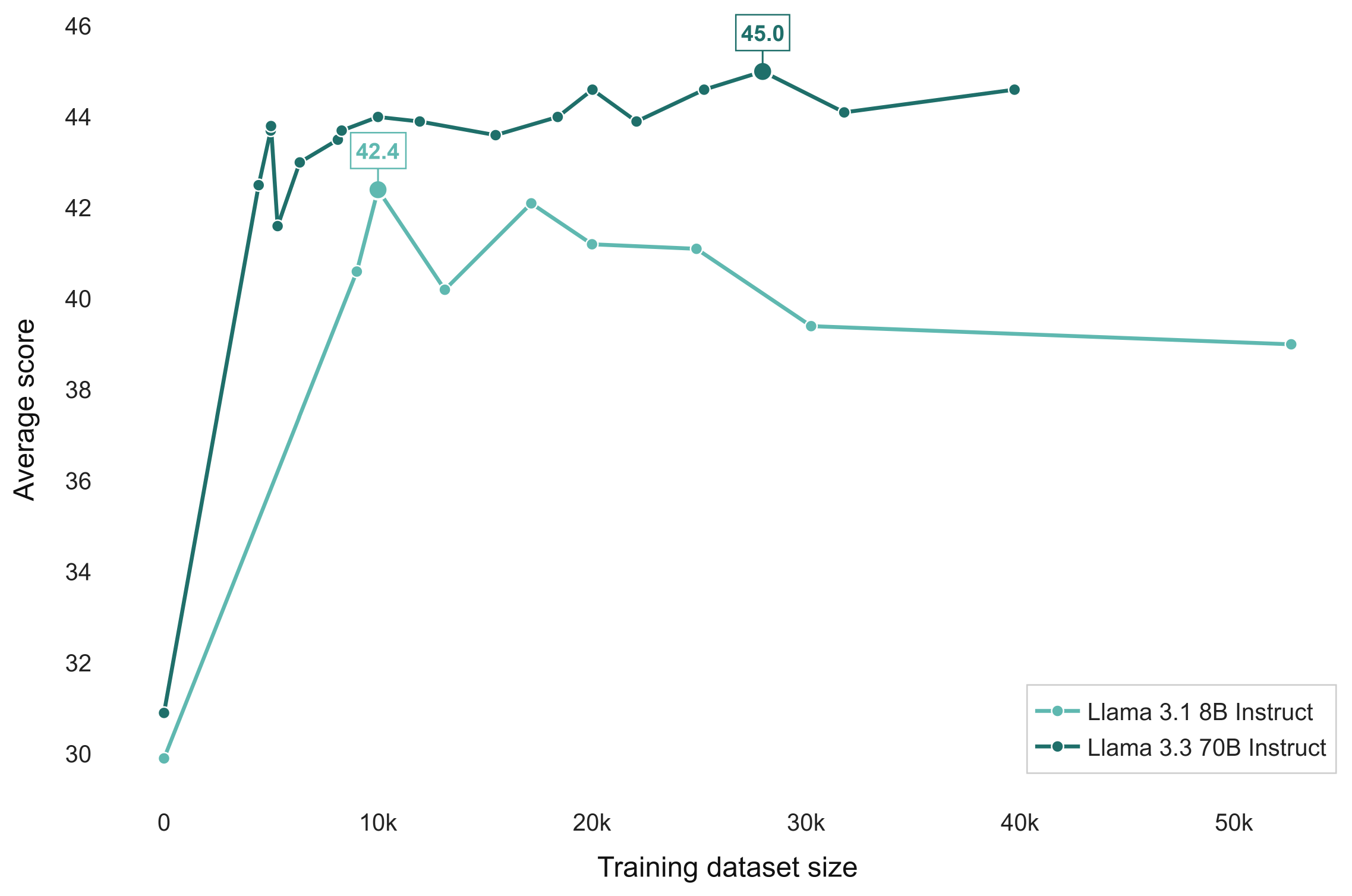}
\caption{Training dataset size ablation. Average ICU-REACT test score as a function of training dataset size for Llama-3.1-8B-Instruct (light teal) and Llama-3.3-70B-Instruct (dark teal). Each point is a model fine-tuned on the indicated number of examples; the leftmost point of each series (size 0) denotes the corresponding zero-shot instruct baseline. Highest performance for each model are highlighted with boxed scores.}
\label{fig:size_ablation}

\end{figure*}

\subsection{Dataset size ablation}

We characterized scaling behavior by varying the number of training examples while holding the task composition fixed (Reasoning Refinement and Variable Selection Reasoning for Llama 3.1 8B Instruct and Reasoning Refinement for Llama 3.3 70B Instruct) (Figure~\ref{fig:size_ablation}). Both backbones exhibited steep early gains followed by rapid saturation. Starting from their zero-shot baselines (29.9 for the 8B model and 30.9 for the 70B model), both models improved sharply within the first several thousand examples. Llama-3.3-70B-Instruct rose above 43 by roughly 8,000 examples and continued to improve gradually thereafter, peaking at 45.0 near 28,000 examples before plateauing. Llama-3.1-8B-Instruct peaked earlier, reaching 42.4 at around 10,000 examples, after which performance was unstable and declined gradually with additional data, settling below 40 at the largest training sizes. This divergence suggests that the larger backbone continues to benefit from additional training data well beyond the point at which the smaller model saturates, and that the 8B model is prone to mild degradation when trained on substantially more data than its effective capacity supports. Taken together, the two ablations indicate that a compact, reasoning-focused training mixture on the order of 10k--30k examples captures most of the attainable performance, with the optimal scale increasing with backbone size.

\begin{figure*}[htbp]
\centering
\includegraphics[width=0.95\linewidth]{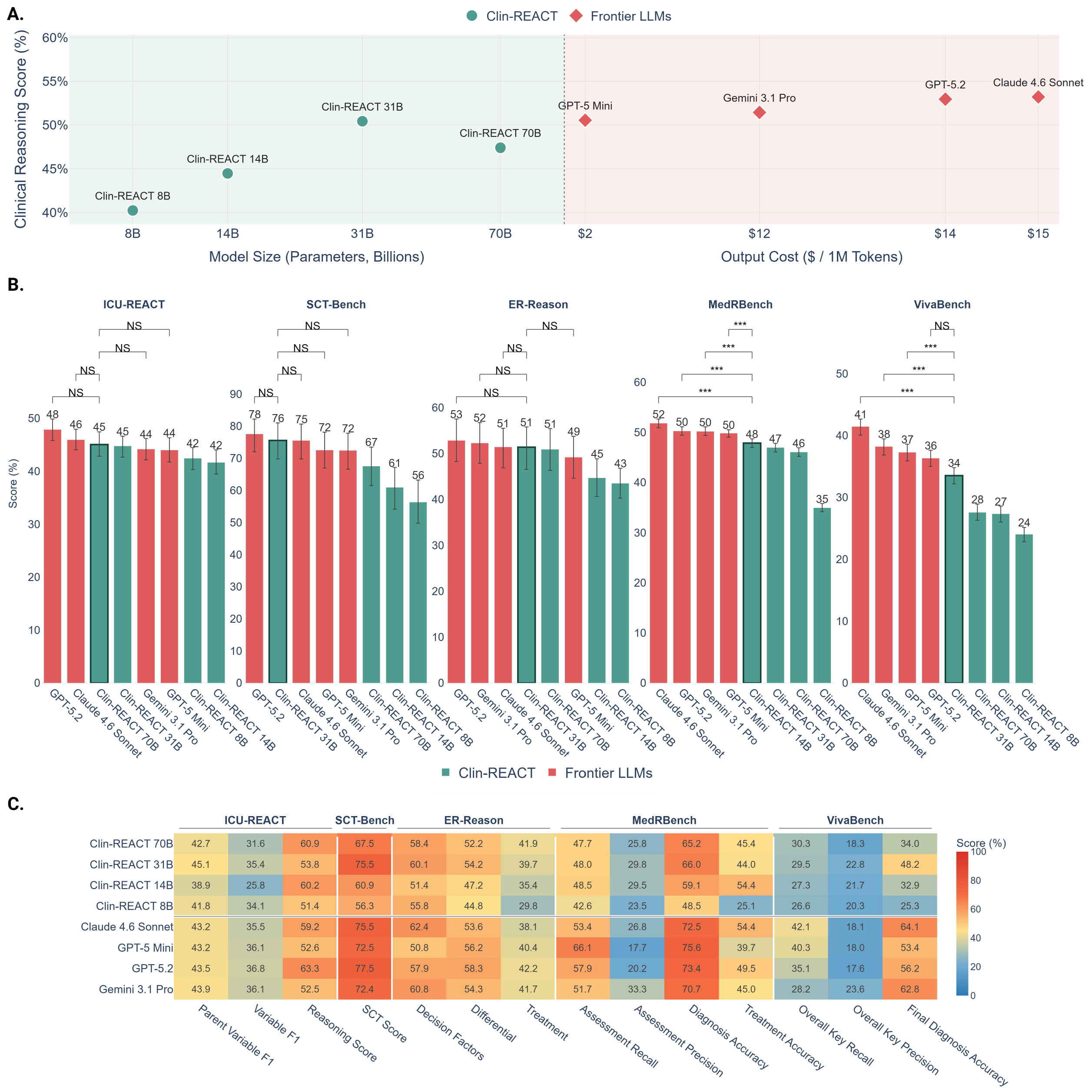}
\caption{Comparison of Clin-REACT models against proprietary frontier
LLMs on clinical reasoning. (A) Clinical-reasoning score across five benchmarks for both model classes, plotted against parameter count (left region, in billions) for Clin-REACT models (circles, $n = 4$) and against output cost (right region, \$ per 1M tokens) for frontier LLMs (diamonds, $n = 4$), with a vertical divider separating the two scales. Each clinical-reasoning score is the mean across five clinical-reasoning benchmarks (ICU-REACT, SCT-Bench, ER-Reason, MedRBench, and VivaBench). (B) Per-benchmark comparison of the four Clin-REACT models
(teal) against the four frontier LLMs (red) across the five clinical-reasoning
benchmarks. Bars are ordered by score within each panel, and the highest-scoring
Clin-REACT model per benchmark is outlined. Error bars denote 95\% CIs; brackets indicate pairwise comparisons between the best Clin-REACT model on each benchmark and frontier LLMs (two-sided Wilcoxon signed-rank test; NS, not significant; \textbf{***},
$p < 0.001$). (C) Heatmap of component sub-metrics underlying each aggregate
benchmark score in (B), grouped by benchmark: ICU-REACT (Parent Variable F1,
Variable F1, Reasoning Score), SCT-Bench (SCT Score), ER-Reason (Decision
Factors, Differential, Treatment), MedRBench (Assessment Recall, Assessment
Precision, Diagnosis Accuracy, Treatment Accuracy), and VivaBench (Overall Key
Recall, Overall Key Precision, Final Diagnosis Accuracy). Cell color encodes
score (0--100\%), with warmer colors indicating higher scores.}
\label{fig:frontier_comparison}

\end{figure*}

\begin{sidewaystable*}[p]
\centering
\small
\caption{Performance of Clin-REACT models against proprietary frontier models across clinical reasoning benchmarks.}
\label{tab:frontier_model_results}
\setlength{\tabcolsep}{1.8pt}
\renewcommand{\arraystretch}{1.20}
\begin{adjustbox}{max width=\linewidth}
\begin{tabular}{l@{\hspace{0.8pt}}cccccccccccccc}
\toprule
\multirow{2}{*}{\textbf{Model}} & \multicolumn{3}{c}{\textbf{ICU-REACT}} & \multicolumn{1}{c}{\textbf{SCT-Bench}} & \multicolumn{3}{c}{\textbf{ER-Reason}} & \multicolumn{4}{c}{\textbf{MedR-Bench}} & \multicolumn{3}{c}{\textbf{VivaBench}} \\
\cmidrule(lr){2-4} \cmidrule(lr){5-5} \cmidrule(lr){6-8} \cmidrule(lr){9-12} \cmidrule(lr){13-15}
 & \textbf{Parent F1} & \textbf{Variable F1} & \textbf{Reasoning} & \textbf{SCT} & \textbf{Decision} & \textbf{Differential} & \textbf{Treatment} & \textbf{Assess. Rec.} & \textbf{Assess. Prec.} & \textbf{Diagnosis} & \textbf{Treatment} & \textbf{Overall Recall} & \textbf{Overall Precision} & \textbf{Final Dx} \\
\midrule
Clin-REACT 70B & \shortstack{42.7 \\ (39.4--46.3)} & \shortstack{31.6 \\ (28.9--34.3)} & \underline{\shortstack{60.9 \\ (58.2--63.4)}} & \shortstack{67.5 \\ (61.6--73.0)} & \shortstack{58.4 \\ (52.9--63.5)} & \shortstack{52.2 \\ (46.2--57.9)} & \underline{\shortstack{41.9 \\ (36.3--47.3)}} & \shortstack{47.7 \\ (46.1--49.6)} & \shortstack{25.8 \\ (24.7--27.1)} & \shortstack{65.2 \\ (62.1--68.3)} & \shortstack{45.4 \\ (41.1--49.8)} & \underline{\shortstack{30.3 \\ (29.1--31.4)}} & \shortstack{18.3 \\ (17.6--19.0)} & \shortstack{34.0 \\ (31.2--37.0)} \\
Clin-REACT 31B & \textbf{\shortstack{45.1 \\ (42.0--48.3)}} & \underline{\shortstack{35.4 \\ (32.9--37.9)}} & \shortstack{53.8 \\ (51.7--55.8)} & \underline{\shortstack{75.5 \\ (69.8--80.6)}} & \underline{\shortstack{60.1 \\ (54.2--65.7)}} & \underline{\shortstack{54.2 \\ (48.3--60.7)}} & \shortstack{39.7 \\ (34.1--45.3)} & \shortstack{48.0 \\ (46.2--49.7)} & \underline{\shortstack{29.8 \\ (28.4--31.1)}} & \underline{\shortstack{66.0 \\ (62.7--69.0)}} & \shortstack{44.0 \\ (39.8--48.3)} & \shortstack{29.5 \\ (28.4--30.6)} & \underline{\shortstack{22.8 \\ (21.9--23.7)}} & \underline{\shortstack{48.2 \\ (45.4--51.5)}} \\
Clin-REACT 14B & \shortstack{38.9 \\ (35.5--42.6)} & \shortstack{25.8 \\ (23.2--28.3)} & \shortstack{60.2 \\ (57.8--62.8)} & \shortstack{60.9 \\ (54.7--66.8)} & \shortstack{51.4 \\ (45.5--57.3)} & \shortstack{47.2 \\ (42.1--52.7)} & \shortstack{35.4 \\ (30.1--40.9)} & \underline{\shortstack{48.5 \\ (46.9--50.3)}} & \shortstack{29.5 \\ (28.2--30.6)} & \shortstack{59.1 \\ (55.7--62.2)} & \underline{\shortstack{54.4 \\ (50.0--58.5)}} & \shortstack{27.3 \\ (26.3--28.4)} & \shortstack{21.7 \\ (20.8--22.7)} & \shortstack{32.9 \\ (29.7--35.9)} \\
Clin-REACT 8B & \shortstack{41.8 \\ (38.7--44.7)} & \shortstack{34.1 \\ (31.5--36.8)} & \shortstack{51.4 \\ (49.1--53.6)} & \shortstack{56.3 \\ (49.7--62.2)} & \shortstack{55.8 \\ (50.9--60.6)} & \shortstack{44.8 \\ (39.9--49.7)} & \shortstack{29.8 \\ (25.4--34.5)} & \shortstack{42.6 \\ (41.0--44.3)} & \shortstack{23.5 \\ (22.3--24.8)} & \shortstack{48.5 \\ (45.1--52.0)} & \shortstack{25.1 \\ (21.6--29.3)} & \shortstack{26.6 \\ (25.4--27.6)} & \shortstack{20.3 \\ (19.5--21.1)} & \shortstack{25.3 \\ (22.3--28.2)} \\
\midrule
Claude 4.6 Sonnet & \shortstack{43.2 \\ (40.4--46.2)} & \shortstack{35.5 \\ (33.4--37.6)} & \shortstack{59.2 \\ (56.8--61.5)} & \shortstack{75.5 \\ (70.1--80.8)} & \textbf{\shortstack{62.4 \\ (57.1--67.6)}} & \shortstack{53.6 \\ (48.0--59.4)} & \shortstack{38.1 \\ (32.3--43.7)} & \shortstack{53.4 \\ (51.6--55.2)} & \shortstack{26.8 \\ (25.5--28.2)} & \shortstack{72.5 \\ (69.5--75.6)} & \textbf{\shortstack{54.4 \\ (50.0--59.1)}} & \textbf{\shortstack{42.1 \\ (40.9--43.4)}}\textsuperscript{***} & \shortstack{18.1 \\ (17.4--18.7)} & \textbf{\shortstack{64.1 \\ (61.0--67.1)}}\textsuperscript{***} \\
GPT-5 Mini & \shortstack{43.2 \\ (40.0--46.6)} & \shortstack{36.1 \\ (33.6--38.8)} & \shortstack{52.6 \\ (50.4--55.0)} & \shortstack{72.5 \\ (66.9--77.7)} & \shortstack{50.8 \\ (44.1--57.0)} & \shortstack{56.2 \\ (50.8--61.9)} & \shortstack{40.4 \\ (34.7--45.9)} & \textbf{\shortstack{66.1 \\ (64.4--67.7)}}\textsuperscript{***} & \shortstack{17.7 \\ (17.0--18.6)} & \textbf{\shortstack{75.6 \\ (72.8--78.5)}}\textsuperscript{***} & \shortstack{39.7 \\ (34.9--44.5)} & \shortstack{40.3 \\ (39.1--41.6)} & \shortstack{18.0 \\ (17.3--18.7)} & \shortstack{53.4 \\ (50.5--56.5)} \\
GPT-5.2 & \shortstack{43.5 \\ (40.3--46.7)} & \textbf{\shortstack{36.8 \\ (34.2--39.2)}} & \textbf{\shortstack{63.3 \\ (61.0--65.5)}} & \textbf{\shortstack{77.5 \\ (72.4--82.4)}} & \shortstack{57.9 \\ (50.9--63.9)} & \textbf{\shortstack{58.3 \\ (52.9--63.8)}}\textsuperscript{*} & \textbf{\shortstack{42.2 \\ (36.8--47.2)}} & \shortstack{57.9 \\ (55.9--59.7)} & \shortstack{20.2 \\ (19.3--21.1)} & \shortstack{73.4 \\ (70.3--76.2)} & \shortstack{49.5 \\ (44.8--54.3)} & \shortstack{35.1 \\ (33.9--36.3)} & \shortstack{17.6 \\ (16.9--18.3)} & \shortstack{56.2 \\ (53.2--59.2)} \\
Gemini 3.1 Pro & \underline{\shortstack{43.9 \\ (40.8--47.2)}} & \shortstack{36.1 \\ (33.6--38.5)} & \shortstack{52.5 \\ (50.3--54.6)} & \shortstack{72.4 \\ (66.7--77.6)} & \shortstack{60.8 \\ (54.9--66.1)} & \shortstack{54.3 \\ (48.7--59.9)} & \shortstack{41.7 \\ (35.5--46.8)} & \shortstack{51.7 \\ (49.9--53.5)} & \textbf{\shortstack{33.3 \\ (31.7--34.8)}}\textsuperscript{***} & \shortstack{70.7 \\ (67.5--73.6)} & \shortstack{45.0 \\ (40.5--49.2)} & \shortstack{28.2 \\ (27.1--29.3)} & \textbf{\shortstack{23.6 \\ (22.6--24.6)}} & \shortstack{62.8 \\ (59.6--66.0)} \\
\bottomrule
\end{tabular}
\end{adjustbox}
\vspace{2mm}
\begin{minipage}{0.98\linewidth}
\scriptsize 
Values are mean percentages with 95\% confidence intervals in parentheses across a 1,000-iteration bootstrap with replacement. For each metric, the highest-scoring Clin-REACT model and highest-scoring frontier model are selected. The higher of those two scores is shown in bold, and the lower score is underlined. Significance symbols indicate the paired Wilcoxon signed-rank p-value comparing those two models: \textsuperscript{*}$p<0.05$, \textsuperscript{**}$p<0.01$, and \textsuperscript{***}$p<0.001$. Abbreviations: Parent F1, parent-variable F1 score; Variable F1, individual-variable F1 score; Reasoning, reasoning score; SCT, SCT score; Decision, decision factors; Assess. Rec., assessment recall; Assess. Prec., assessment precision; Final Dx, final diagnosis accuracy. Included model groups: Clin-REACT, Frontier LLMs.
\end{minipage}
\end{sidewaystable*}

\section{Comparison with Frontier Models}\label{compare_frontier}

We benchmarked the four Clin-REACT models against four proprietary frontier LLMs across five clinical-reasoning benchmarks: GPT-5.2, GPT-5 Mini, Claude 4.6 Sonnet, and Gemini 3.1 Pro (Fig.~\ref{fig:frontier_comparison}; Table~\ref{tab:frontier_model_results}). Averaged across benchmarks, Clin-REACT 31B achieved a macro clinical-reasoning score of 50.4\% (SD, 15.5), closely matching GPT-5 Mini at 50.5\% (SD, 13.3) and approaching Gemini 3.1 Pro at 51.4\% (SD, 12.9), GPT-5.2 at 52.9\% (SD, 15.1), and Claude 4.6 Sonnet at 53.2\% (SD, 13.2). This performance was achieved despite Clin-REACT 31B being an open-weight model roughly an order of magnitude smaller and incurring no per-token inference cost, whereas frontier output prices ranged from \$2 to \$15 per 1M tokens (Fig.~\ref{fig:frontier_comparison}A).

On three of the five benchmarks (ICU-REACT, SCT-Bench, and ER-Reason) the best Clin-REACT model was statistically indistinguishable from all frontier models (Fig.~\ref{fig:frontier_comparison}B). On ICU-REACT, Clin-REACT 70B achieved an aggregate score of 45.0\% (95\% CI, 42.8--47.4), compared with 47.9\% (45.8--49.9) for GPT-5.2, 45.9\% (44.1--47.9) for Claude 4.6 Sonnet, 44.2\% (42.1--46.2) for Gemini 3.1 Pro, and 44.0\% (41.7--46.3) for GPT-5 Mini. At the component level, Clin-REACT 31B achieved the highest parent-variable F1 of any model (45.1\% versus 43.9\% for Gemini 3.1 Pro), and its remaining sub-metrics did not differ significantly from the best frontier scores (Table~\ref{tab:frontier_model_results}). On SCT-Bench, Clin-REACT 31B scored 75.5\% (69.8--81.0), equal to Claude 4.6 Sonnet at 75.5\% (69.8--80.7) and close to GPT-5.2 at 77.5\% (72.0--82.2). On ER-Reason, Clin-REACT 31B scored 51.4\% (46.5--55.8), compared with 52.8\% (48.2--57.5) for GPT-5.2, 52.2\% (47.8--56.7) for Gemini 3.1 Pro, and 51.4\% (46.9--55.5) for Claude 4.6 Sonnet. Only the ER-Reason differential-diagnosis sub-metric reached significance (GPT-5.2, 58.3\%, versus Clin-REACT 31B, 54.2\%; $p<0.05$) (Table~\ref{tab:frontier_model_results}). Thus, on focused reasoning and information-retrieval tasks, the Clin-REACT models were competitive with systems many times their size and cost.

Frontier models retained a clearer advantage on MedRBench and VivaBench. On MedRBench, the highest-performing Clin-REACT model, Clin-REACT 14B, achieved 47.9\% (47.0--48.7), compared with 51.8\% (50.9--52.6) for Claude 4.6 Sonnet, 50.2\% (49.4--51.1) for GPT-5.2, 50.2\% (49.4--51.1) for Gemini 3.1 Pro, and 49.8\% (49.0--50.5) for GPT-5 Mini; all four frontier models significantly exceeded the best Clin-REACT model ($p<0.001$). On VivaBench, Clin-REACT 31B achieved 33.5\% (32.2--34.8), compared with 41.4\% (40.1--42.7) for Claude 4.6 Sonnet, 38.2\% (36.8--39.4) for Gemini 3.1 Pro, 37.3\% (35.9--38.6) for GPT-5 Mini, and 36.3\% (35.0--37.6) for GPT-5.2. Three of the four frontier models significantly exceeded the best Clin-REACT model on VivaBench, whereas the difference from GPT-5.2 was not significant.

At the component level, frontier models showed their largest advantages on several MedRBench and VivaBench metrics. On MedRBench, GPT-5 Mini achieved higher assessment recall than Clin-REACT 14B (66.1\% versus 48.5\%) and higher diagnosis accuracy than Clin-REACT 31B (75.6\% versus 66.0\%). However, Clin-REACT 14B matched Claude 4.6 Sonnet on treatment accuracy (54.4\% for both models). On VivaBench, Claude 4.6 Sonnet achieved higher overall key recall than Clin-REACT 70B (42.1\% versus 30.3\%) and higher final-diagnosis accuracy than Clin-REACT 31B (64.1\% versus 48.2\%). In contrast, overall key precision was similar between Gemini 3.1 Pro and Clin-REACT 31B (23.6\% versus 22.8\%), with the difference not reaching statistical significance. Taken together, these results indicate that scale- and cost-efficient open models can be trained to cloasely approach or even match proprietary frontier models on targeted clinical-reasoning tasks.

\section{Performance stratified by clinical content}\label{cat_perform}

\begin{figure}[htbp]
\centering
\includegraphics[width=0.90\linewidth]{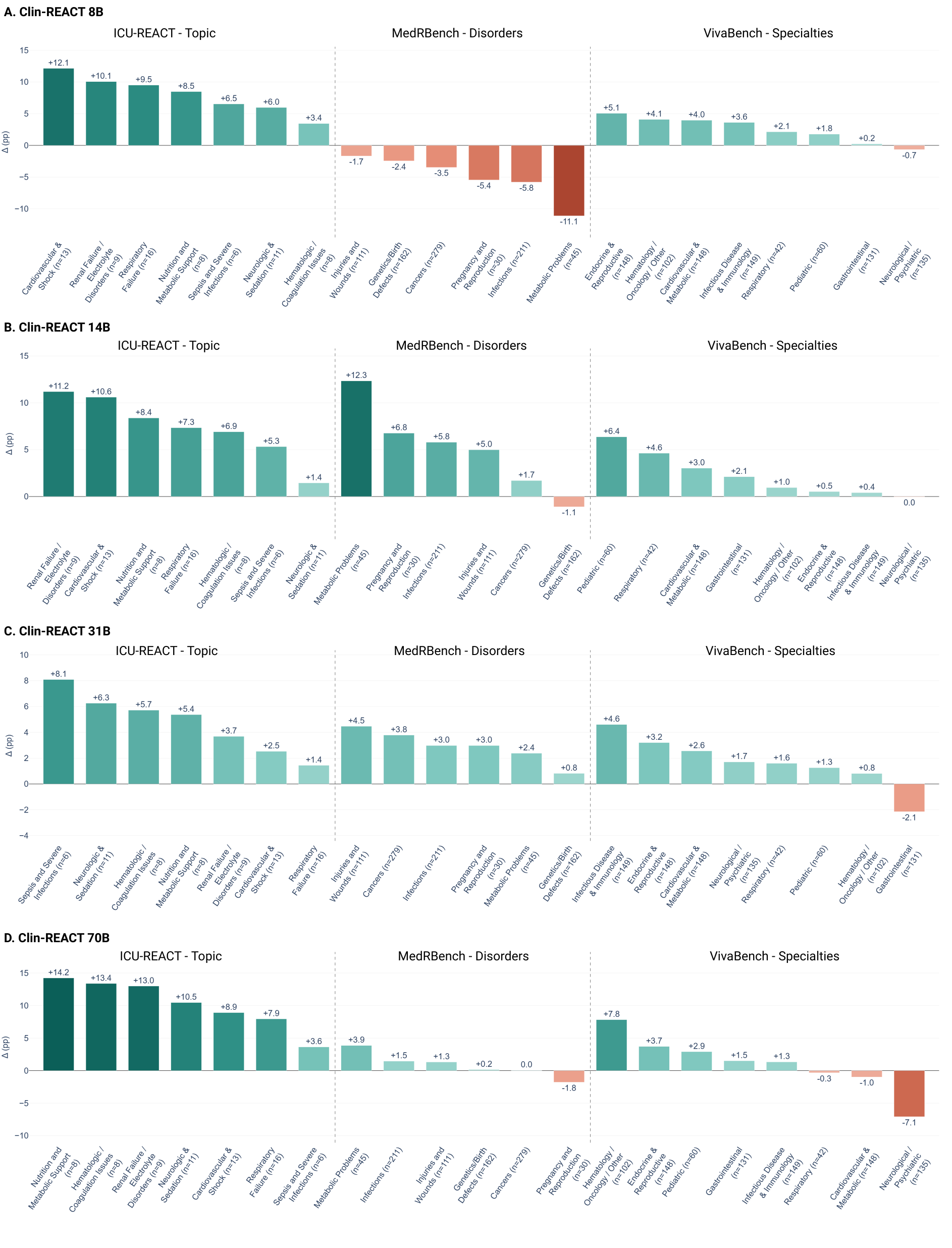}
\caption{Per-clinical content training gains of Clin-REACT models over backbones across clinical-reasoning benchmarks. Gains are shown for (A) Clin-REACT 8B (backbone: Llama 3.1 8B Instruct), (B) Clin-REACT 14B (backbone: Baichuan M1 14B Instruct), (C) Clin-REACT 31B (backbone: Gemma 4 31B), and (D) Clin-REACT 70B (backbone: Llama 3.3 70B Instruct). Each bar reports the change in score ($\Delta$, percentage points) for a single clinical category, computed as the Clin-REACT model minus its corresponding backbone using each benchmark's average-across-metrics score; bar color encodes the sign and magnitude of the change, with teal representing improvement over the backbone, red representing regression, and darker shades denoting larger magnitudes. Within each benchmark section, bars are sorted in descending order of $\Delta$, so category order varies across panels; numeric labels show each category's $\Delta$ and axis labels show the number of test samples ($n$). The three sections per panel, separated by dashed vertical lines, correspond to ICU-REACT topics (7 categories), MedRBench disorder groups (6 categories), and VivaBench specialty groups (8 categories).}
\label{fig:category_gains}
\end{figure}

Clin-REACT models demonstrated gains over their corresponding backbones across clinical categories, with several related content areas showing parallel improvements across benchmarks (Fig.~\ref{fig:category_gains}). For example, Clin-REACT 14B improved substantially in ICU-REACT Nutrition and Metabolic Support (+8.4 percentage points [pp]) and in the related MedRBench Metabolic Problems category (+12.3 pp), while also gaining on VivaBench Cardiovascular and Metabolic conditions (+3.0 pp). Similarly, Clin-REACT 31B improved on ICU-REACT Sepsis and Severe Infections (+8.1 pp), MedRBench Infections (+3.0 pp), and VivaBench Infectious Disease and Immunology (+4.6 pp). Hematologic content also showed cross-benchmark gains for several variants, including Clin-REACT 70B on ICU-REACT Hematologic/Coagulation Issues (+13.4 pp) and VivaBench Hematology/Oncology/Other (+7.8 pp). These patterns suggest that training gains in some ICU-REACT topics transferred to clinically related categories in external benchmarks, although the correspondence was not uniform across model sizes or datasets. For example, Clin-REACT 8B improved across all ICU-REACT topics but regressed across all MedRBench disorder groups, while Clin-REACT 70B showed strong ICU-REACT gains but declined in selected VivaBench specialties, most notably Neurological/Psychiatric conditions ($-7.1$ pp).

\begin{figure}[htbp]
\centering
\includegraphics[width=0.95\linewidth]{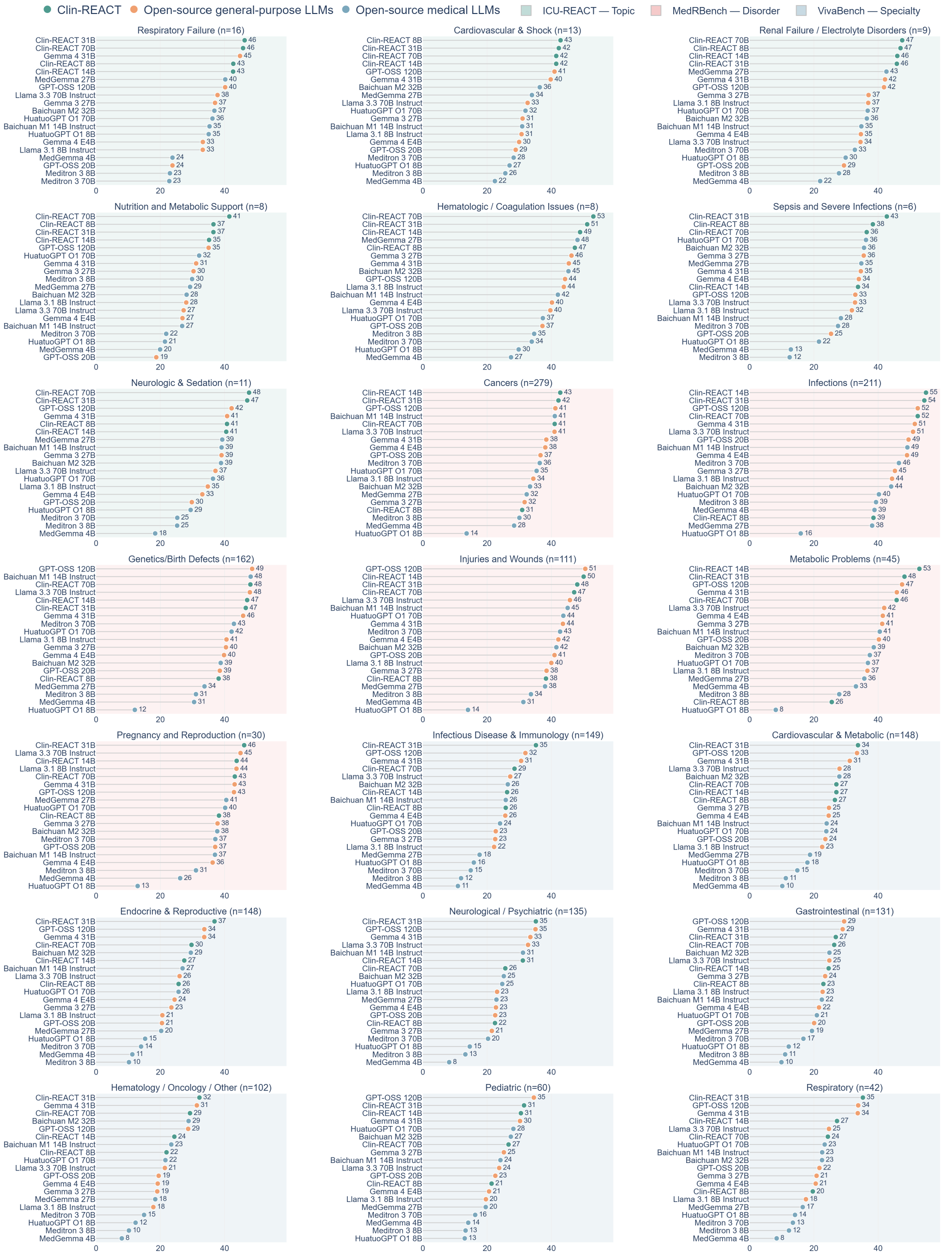}
\caption{Per-clinical content comparison of Clin-REACT and baseline models across clinical-reasoning benchmarks. Each panel ranks all evaluated models by score within a single clinical category, drawn from one of the three benchmarks that carry per-sample categorical labels: ICU-REACT topics (green panels, 7 categories), MedRBench disorder groups (pink panels, 6 categories), and VivaBench specialty groups (blue panels, 8 categories). Panel titles give the category name and its number of test samples ($n$). Within each panel, points are ordered by score (highest at top, ranked independently per panel) and colored by model class: Clin-REACT (teal, $n = 4$), open-source general purpose LLMs (orange, $n = 9$), and open-source medical LLMs (blue, $n = 6$); numeric labels give each model's score within the category. Scores are calculated as the average score across metrics within each benchmark.}
\label{fig:category_ranks}
\end{figure}

\begin{figure}[htbp]
\centering
\includegraphics[width=0.95\linewidth]{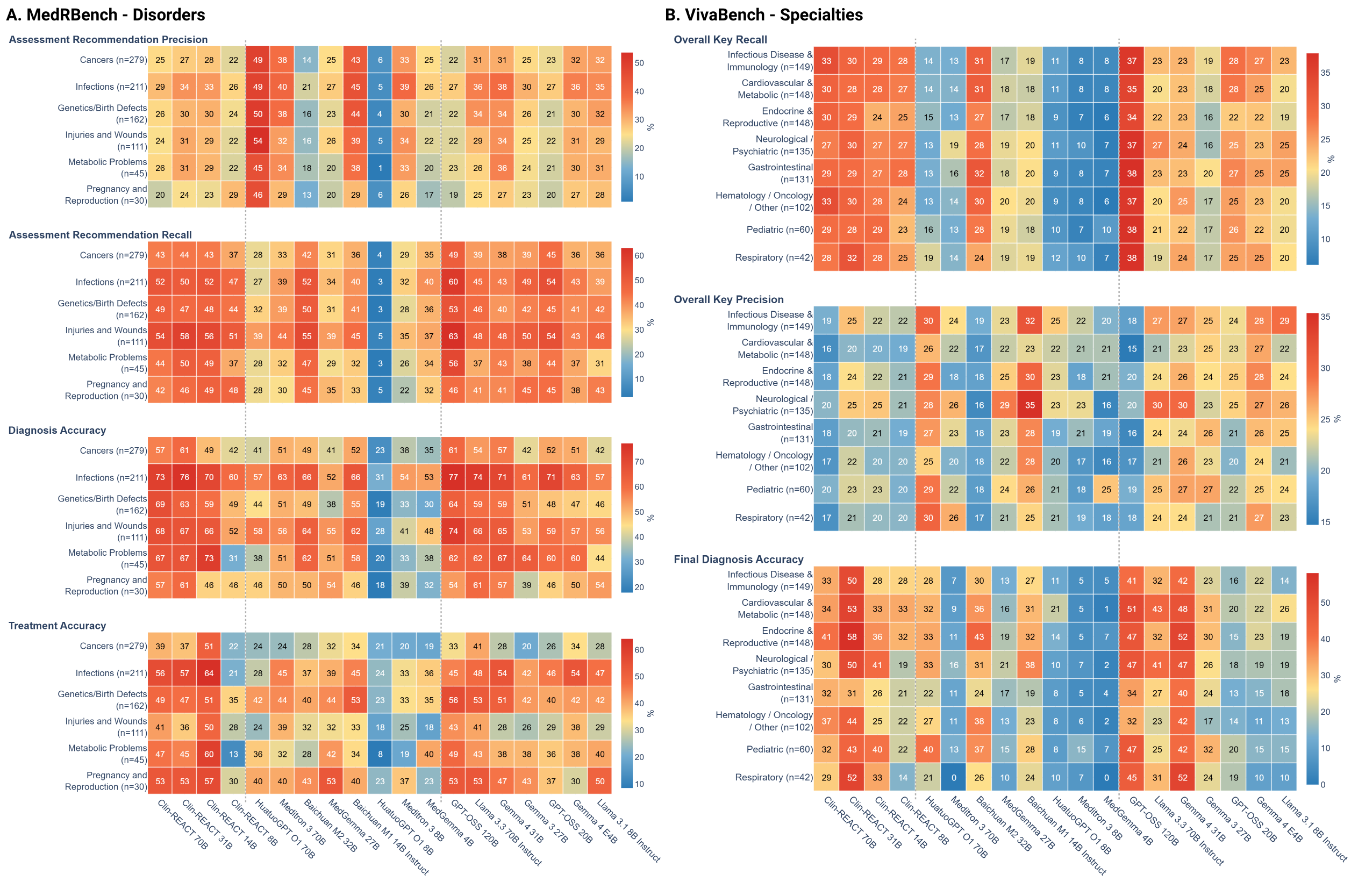}
\caption{Per-clinical content, per-metric heatmap of Clin-REACT and baseline models' performance on MedRBench and VivaBench. Each cell reports a model's score (\%) within a single clinical category (rows) for one evaluation metric (heatmap blocks). Columns comprise all evaluated models, grouped by class and separated by dashed vertical lines: Clin-REACT (left, $n = 4$), open-source medical LLMs (center, $n = 8$), and open-source general purpose LLMs (right, $n = 7$). Row labels give the category name and its number of test samples ($n$); numeric labels give each cell's score. Warmer colors indicate higher scores, with each metric block using an independent color scale such shading is comparable within a block. (A) MedRBench disorder groups (6 categories) across four metrics: Assessment Recommendation Precision, Assessment Recommendation Recall, Diagnosis Accuracy, and Treatment Accuracy. (B) VivaBench specialty groups (8 categories) across three metrics: Overall Key Recall, Overall Key Precision, and Final Diagnosis Accuracy.}
\label{fig:categories_performance}
\end{figure}

Compared with the evaluated open-source baseline models, Clin-REACT models ranked strongly on most clinical categories across benchmarks (Fig.~\ref{fig:category_ranks}). A Clin-REACT variant achieved the highest score in each of the seven ICU-REACT topic categories, and multiple Clin-REACT variants frequently occupied the leading positions within the same category. Clin-REACT models also achieved the highest category-level scores in four of the six MedRBench disorder groups, including Cancers, Infections, Metabolic Problems, and Pregnancy and Reproduction. On VivaBench, Clin-REACT models led or tied for the highest score in most specialty groups, including Infectious Disease and Immunology, Cardiovascular and Metabolic, Endocrine and Reproductive, Neurological/Psychiatric, Hematology/Oncology/Other, and Respiratory conditions. Performance was comparatively weaker in the Pediatric and Gastrointestinal specialties, where general-purpose baseline models ranked highest. Detailed metric-level differences are provided in Fig.~\ref{fig:categories_performance}.

\section{Information seeking performance by data category}\label{info_seek_cat}

\begin{figure}[htbp]
\centering
\includegraphics[width=0.95\linewidth]{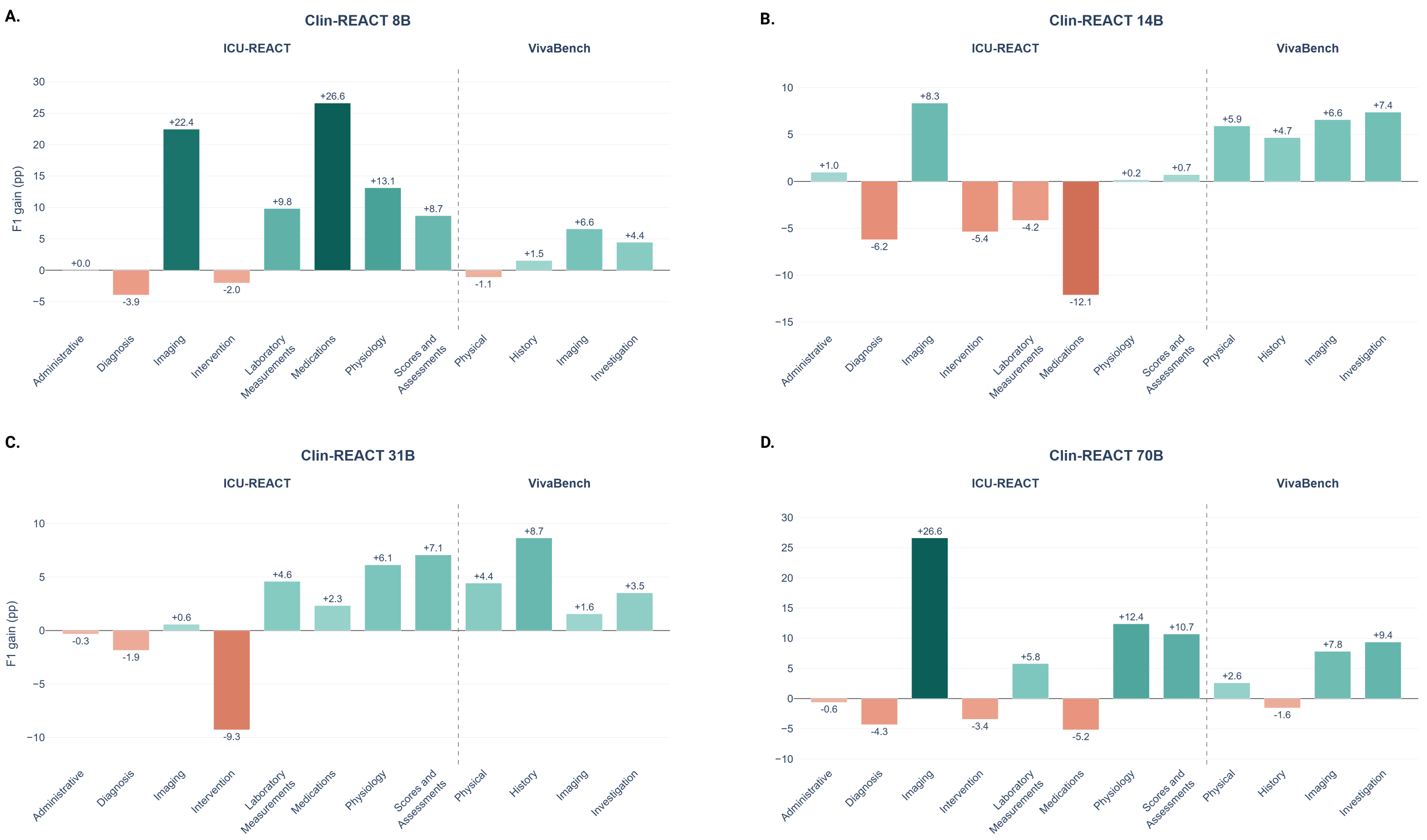}
\caption{Per-category information-retrieval gains of Clin-REACT models over backbones on clinical-reasoning benchmarks. Gains are shown for (A) Clin-REACT 8B (backbone: Llama 3.1 8B Instruct), (B) Clin-REACT 14B (backbone: Baichuan M1 14B Instruct), (C) Clin-REACT 31B (backbone: Gemma 4 31B), and (D) Clin-REACT 70B (backbone: Llama 3.3 70B Instruct) across ICU-REACT and VivaBench. Each bar reports the change in information-retrieval F1 ($\Delta$, percentage points) for a single content category, computed as the Clin-REACT model's F1 minus its corresponding backbone's F1; bar color encodes the sign and magnitude of the change, with teal representing improvement over the backbone, red representing regression, and darker shades denoting larger magnitudes. Categories appear in a fixed order that is consistent across panels, and numeric labels give each category's $\Delta$. The two sections per panel, separated by a dashed vertical line, correspond to ICU-REACT variable categories (8 categories) and VivaBench finding categories (4 categories).}
\label{fig:info_retrieve_gains}
\end{figure}

Clin-REACT training produced category-specific improvements in information retrieval, with the most consistent cross-benchmark transfer observed for imaging-related information (Fig.~\ref{fig:info_retrieve_gains}). All four Clin-REACT variants improved imaging F1 on both ICU-REACT and VivaBench. The largest gains were achieved by Clin-REACT 70B and Clin-REACT 8B on ICU-REACT imaging (+26.6 and +22.4 percentage points [pp], respectively), accompanied by gains of +7.8 and +6.6 pp on VivaBench imaging. Clin-REACT 14B similarly improved imaging retrieval on ICU-REACT (+8.3 pp) and VivaBench (+6.6 pp), while Clin-REACT 31B showed smaller but positive gains on both benchmarks (+0.6 and +1.6 pp). This consistent pattern suggests that the retrieval improvements learned from ICU-focused training transferred particularly well to imaging information in an external clinical-reasoning benchmark.

\begin{figure}[htbp]
\centering
\includegraphics[width=0.95\linewidth]{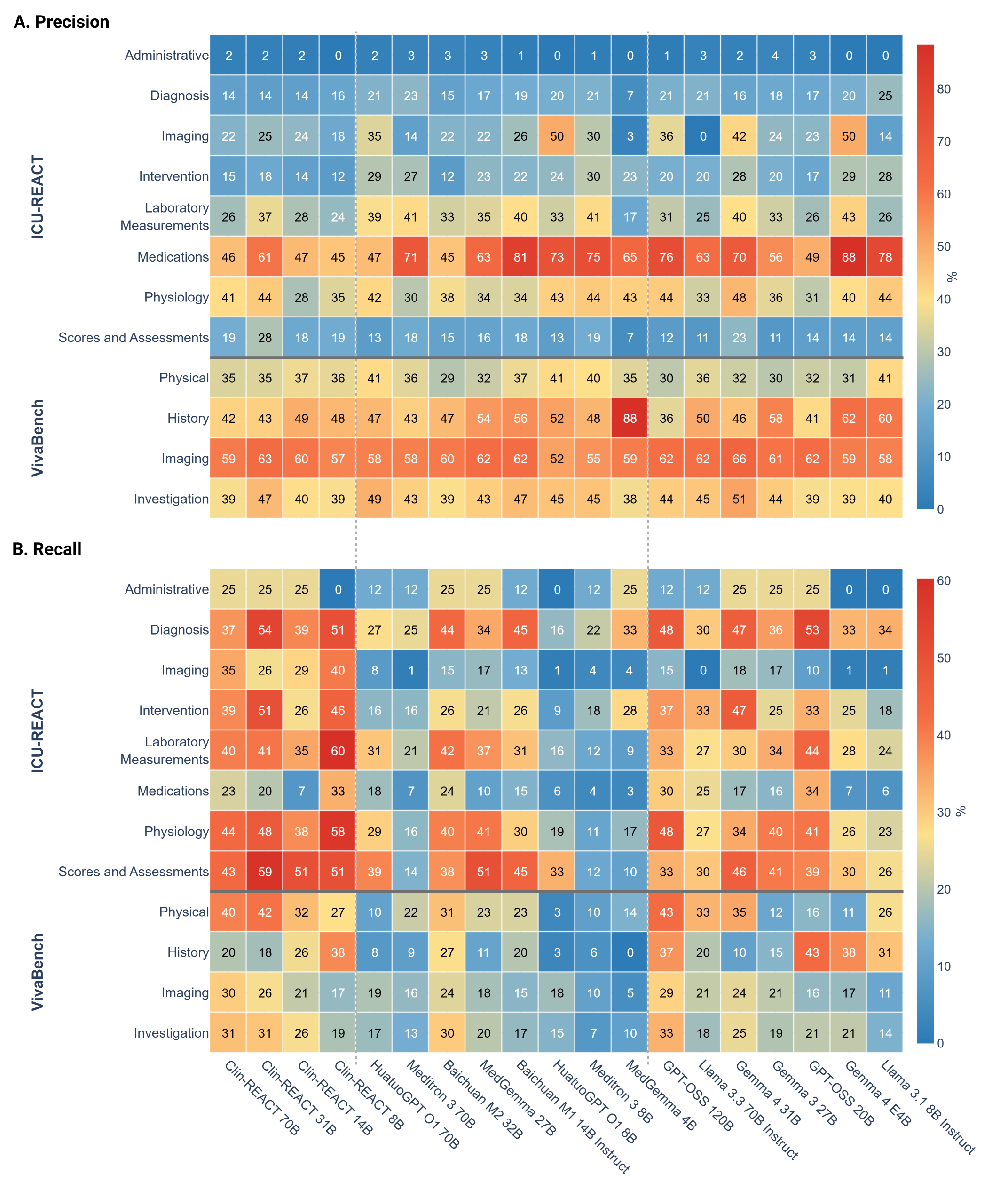}
\caption{Per-category heatmaps of information-retrieval of Clin-REACT and baseline models on ICU-REACT and VivaBench. Information-retrieval is shown as (A) precision and (B) recall. Each cell reports a model's value (\%) for a single content category (rows). Columns comprise all evaluated models, grouped by class and separated by dashed vertical lines: Clin-REACT (left, $n = 4$), open-source medical LLMs (center, $n = 8$), and open-source general purpose LLMs (right, $n = 7$). Rows are grouped and separated by a horizontal line into ICU-REACT variable categories (8 categories) and VivaBench finding categories (4 categories); numeric labels give each cell's value. Warmer colors indicate higher values, with each panel using an independent color scale such that shading is comparable within a panel.}
\label{fig:info_retrieve_heatmap}
\end{figure}

Other gains were more dependent on model size and data category. Clin-REACT 8B showed substantial ICU-REACT improvements for medications (+26.6 pp), physiology (+13.1 pp), laboratory measurements (+9.8 pp), and scores and assessments (+8.7 pp), while also improving VivaBench history (+1.5 pp) and investigation retrieval (+4.4 pp). Clin-REACT 31B and 70B likewise improved ICU-REACT physiology and scores and assessments, with Clin-REACT 31B showing its largest VivaBench gain for history (+8.7 pp) and Clin-REACT 70B for investigation (+9.4 pp). Clin-REACT 14B exhibited a different pattern, with weaker or negative gains in several ICU-REACT categories but consistent improvements across all four VivaBench finding categories, reaching +7.4 pp for investigation. Thus, gains transferred across benchmarks for several related information types, but the magnitude and direction of transfer varied across model sizes. The corresponding category-level precision and recall profiles indicate that these differences reflected distinct retrieval tradeoffs across model families and data types; detailed values are provided in Fig.~\ref{fig:info_retrieve_heatmap}.

\section{Multiple choice medical benchmarks vs clinical reasoning}\label{mc_cr_compare}

To assess whether Clin-REACT training preserved general medical knowledge, we compared each Clin-REACT model with its corresponding backbone across four medical multiple-choice benchmarks: MMLU-Med, MedMCQA, MedQA, and MedXpertQA. Clin-REACT training generally preserved performance on these benchmarks, particularly for the larger model variants (Table~\ref{tab:mcq-backbone-comparison}). Relative to their corresponding backbones, Clin-REACT 14B, 31B, and 70B showed only modest reductions in average multiple-choice accuracy of 1.2, 1.4, and 1.2 percentage points, respectively. These changes were not uniformly negative across individual benchmarks: Clin-REACT 31B improved on MMLU-Med (+0.7 points), Clin-REACT 70B improved on MedMCQA (+0.2) and MedXpertQA (+3.8), and Clin-REACT 14B improved on MedXpertQA (+0.7). Clin-REACT 8B showed a larger average reduction of 8.7 points, driven primarily by MedQA ($-19.8$ points), indicating that preservation of multiple-choice performance was less consistent at the smallest model scale.


\newcommand{\dn}[1]{{\color{red!70!black}$\blacktriangledown$\,(\,$-$#1\,)}}
\newcommand{\up}[1]{{\color{green!45!black}$\blacktriangle$\,(\,$+$#1\,)}}

\begin{table}[t]
\centering
\footnotesize
\setlength{\tabcolsep}{5pt}
\caption{Performance of Clin-REACT models against their
backbones on general medical multiple-choice benchmarks.}
\label{tab:mcq-backbone-comparison}
\begin{tabular}{lccccc}
\toprule
Model & Average & MMLU-Med & MedMCQA & MedQA & MedXpertQA \\
\midrule
Llama 3.1 8B Instruct & 43.3 & 64.0 & 47.5 & 47.1 & 14.8 \\
Clin-REACT 8B & 34.6 \dn{8.7} & 55.2 \dn{8.8} & 42.0 \dn{5.5} & 27.3 \dn{19.8} & 13.8 \dn{1.0} \\
\midrule
Baichuan M1 14B Instruct & 56.5 & 81.4 & 62.5 & 61.2 & 21.0 \\
Clin-REACT 14B & 55.3 \dn{1.2} & 78.8 \dn{2.6} & 61.3 \dn{1.2} & 59.5 \dn{1.7} & 21.7 \up{0.7} \\
\midrule
Gemma-4-31B-It & 80.2 & 94.2 & 78.5 & 92.4 & 55.8 \\
Clin-REACT 31B & 78.8 \dn{1.4} & 94.9 \up{0.7} & 78.4 \dn{0.1} & 89.8 \dn{2.6} & 52.2 \dn{3.6} \\
\midrule
Llama 3.3 70B Instruct & 57.5 & 73.6 & 57.0 & 77.8 & 21.5 \\
Clin-REACT 70B & 56.3 \dn{1.2} & 69.0 \dn{4.6} & 57.2 \up{0.2} & 73.6 \dn{4.2} & 25.3 \up{3.8} \\
\bottomrule
\end{tabular}
\begin{tablenotes}[flushleft]
\footnotesize
\item All values are accuracy (\%). Each block pairs an instruction-tuned
backbone (top) with its corresponding \mbox{Clin-REACT model} (bottom); arrows
indicate the change relative to that backbone.
\end{tablenotes}
\end{table}

\begin{figure*}[h]
\centering
\includegraphics[width=0.95\linewidth]{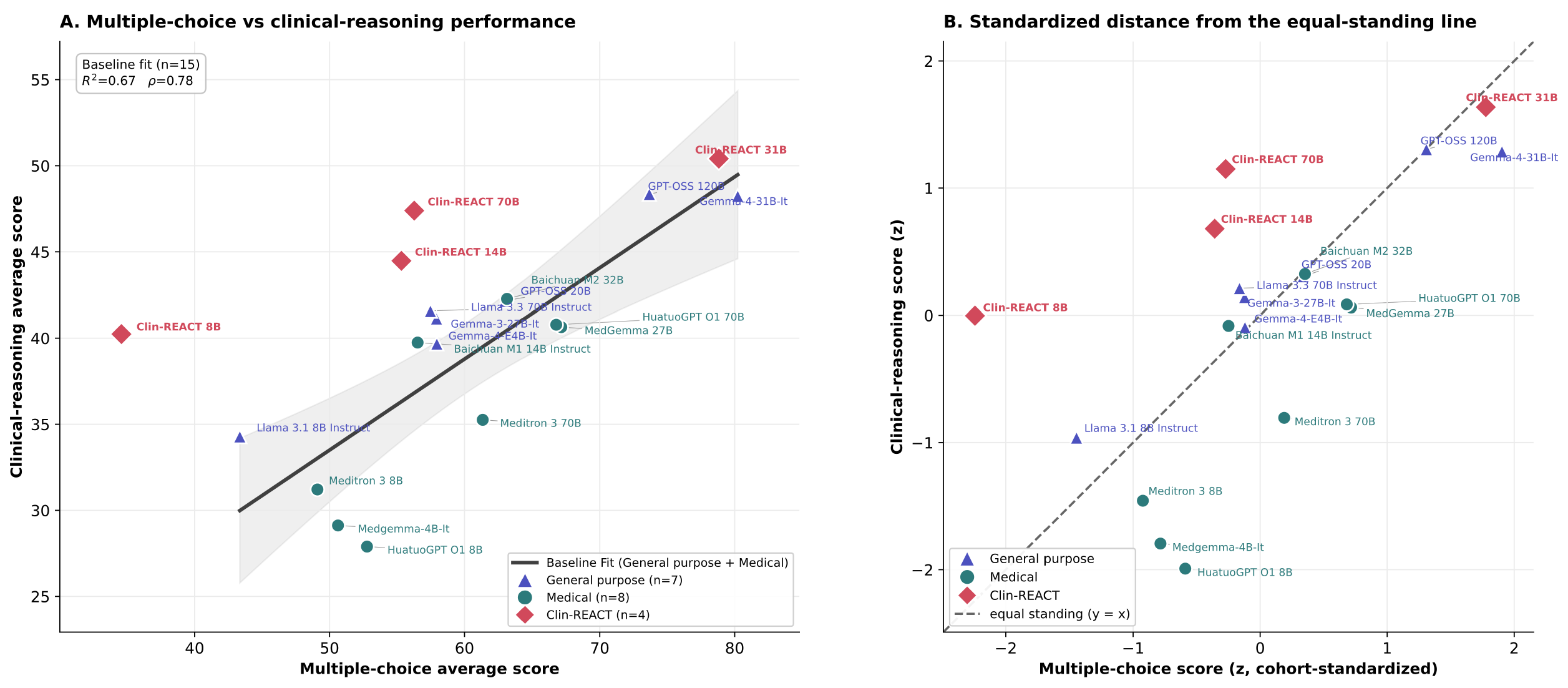}
\caption{Relationship between medical multiple-choice and clinical reasoning
benchmarks performance. (A) Clinical-reasoning average score versus
multiple-choice average score for all evaluated models. Each benchmark score is
first computed by averaging its constituent sub-metrics, and each axis is then
the mean across benchmarks (four multiple-choice benchmarks: MMLU-Med, MedMCQA,
MedQA, MedXpertQA; five clinical-reasoning benchmarks: ER-Reason, ICU-REACT,
MedRBench, SCT-Bench, VivaBench). Marker shape denotes model class:
general-purpose (triangles, $n = 7$), open-source medical (circles, $n = 8$),
and Clin-REACT (diamonds, $n = 4$). The solid line is an ordinary least-squares
fit to the baseline models only (general-purpose and medical, $n = 15$; shaded
band, 95\% confidence interval); Clin-REACT models are overlaid but excluded
from the fit so that the line describes the multiple-choice-to-reasoning
relationship among existing models. Goodness of fit for the baseline
relationship is inset ($R^{2} = 0.67$, Spearman $\rho = 0.78$).
(B) The same models after standardizing each axis to $z$-scores across
the full cohort ($n = 19$). The dashed diagonal ($y = x$) marks equal relative
standing on the two benchmark families: models above the line rank higher on
clinical reasoning than on multiple choice, and models below the line rank
lower.}
\label{fig:mc_cr_avg}

\end{figure*}

\begin{figure*}[h]
\centering
\includegraphics[width=0.95\linewidth]{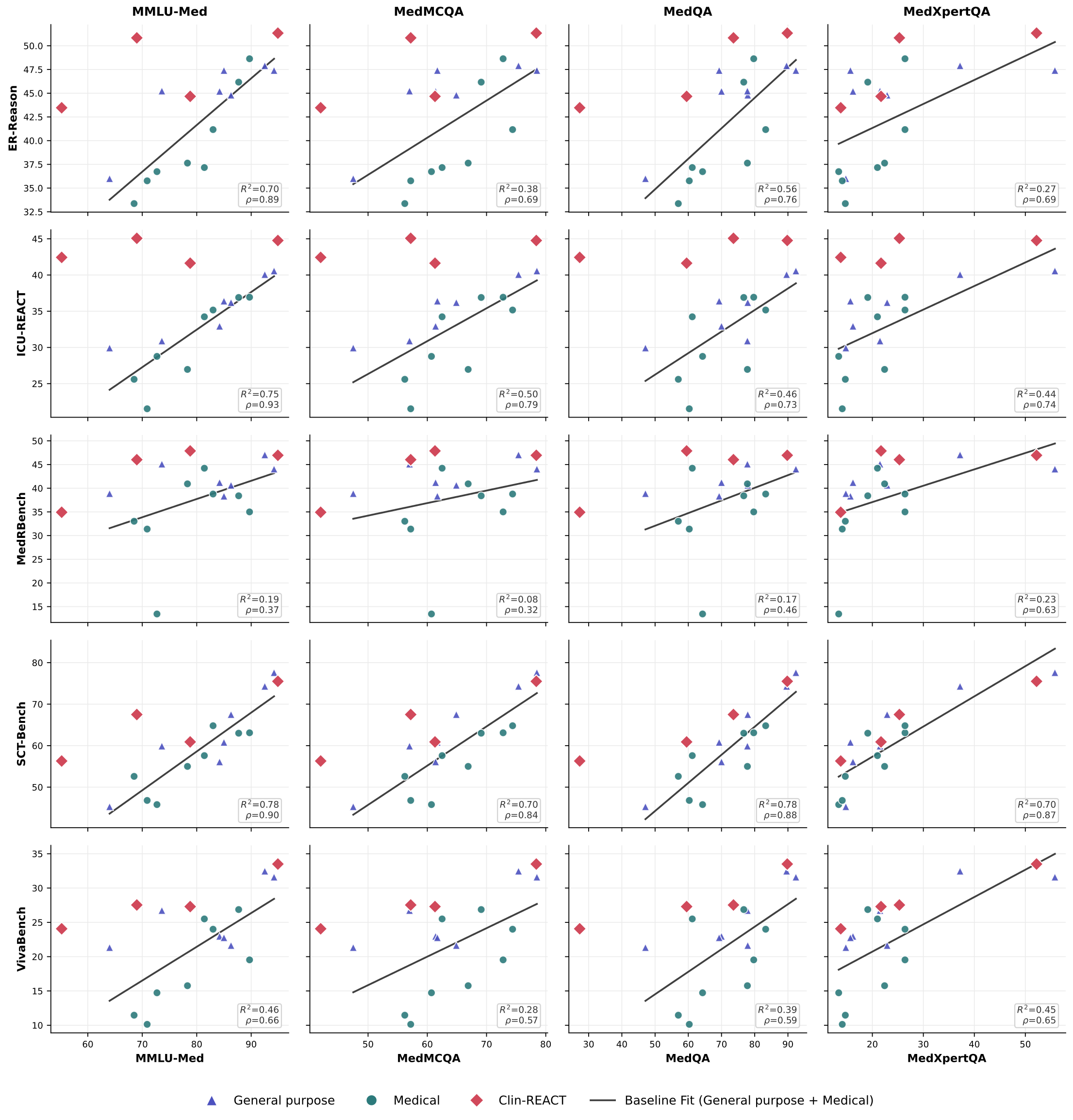}
\caption{Benchmark-level relationships between multiple-choice and
clinical-reasoning performance. Pairwise scatter plots for all 20 combinations
of the five clinical-reasoning benchmarks (rows) and four multiple-choice
benchmarks (columns). Each point is one model: general-purpose (triangles),
open-source medical (circles), and Clin-REACT (diamonds). Clinical-reasoning
benchmark scores are the average of the benchmark's sub-metrics. Within each
panel, the solid line is an ordinary least-squares fit to the baseline models
only (general-purpose and medical, $n = 15$), with the corresponding
coefficient of determination ($R^{2}$) and Spearman rank correlation ($\rho$)
inset; Clin-REACT models are overlaid but excluded from the fit and from the
reported statistics. Axes are shared within each row and column.}
\label{fig:mc_cr_scat}

\end{figure*}

To examine whether performance on medical multiple-choice benchmarks reflects broader clinical-reasoning ability, we conducted a correlation analysis between each model's average score across four multiple-choice benchmarks (MMLU-Med, MedMCQA, MedQA, and MedXpertQA) and its average score across five clinical-reasoning benchmarks (ICU-REACT, SCT-Bench, ER-Reason, MedRBench, and VivaBench) (Fig.~\ref{fig:mc_cr_avg}). Among the baseline general-purpose and medical models, multiple-choice and clinical-reasoning performance were positively correlated ($R^{2}=0.67$; Spearman $\rho=0.78$), but substantial departures from this overall trend indicated that strong multiple-choice performance did not consistently translate into strong clinical reasoning. For example, Meditron 3 70B, HuatuoGPT O1 70B, and MedGemma 27B achieved higher average multiple-choice scores than Llama 3.3 70B Instruct (61.4\%, 66.8\%, and 67.1\% versus 57.5\%, respectively), yet all three performed worse on the clinical-reasoning benchmarks (35.3\%, 40.8\%, and 40.6\% versus 41.6\%). Among the smaller models, the same pattern was evident when comparing Llama 3.1 8B Instruct with medical models of a similar scale. HuatuoGPT O1 8B, MedGemma 4B, and Meditron 3 8B all achieved higher average multiple-choice scores than Llama 3.1 8B Instruct (52.8\%, 50.6\%, and 49.1\% versus 43.4\%, respectively), yet each obtained a lower average clinical-reasoning score (27.9\%, 29.1\%, and 31.2\% versus 34.3\%). These examples show that models with stronger performance on knowledge-focused multiple-choice tasks may still underperform models with lower multiple-choice scores when evaluated on benchmarks requiring information retrieval, evidence integration, and clinical decision-making.

Clin-REACT models further illustrated this distinction. Clin-REACT 14B and Clin-REACT 70B achieved clinical-reasoning scores of 44.5\% and 47.4\%, respectively, despite multiple-choice scores of 55.3\% and 56.3\%; both therefore outperformed several models with substantially higher multiple-choice scores, including MedGemma 27B and HuatuoGPT O1 70B. The contrast was strongest for Clin-REACT 8B, which achieved the lowest multiple-choice average in the cohort (34.6\%) but a clinical-reasoning score of 40.2\%, exceeding Llama 3.1 8B Instruct (34.3\%), Meditron 3 70B (35.3\%), and several medical models with considerably higher multiple-choice performance. After cohort standardization, Clin-REACT 8B, 14B, and 70B consequently ranked substantially higher in clinical reasoning than in multiple choice and appeared above the equal-standing line, while Clin-REACT 31B remained strong on both benchmark families (Fig.~\ref{fig:mc_cr_avg}B). Together, these findings suggest that medical multiple-choice benchmarks capture an important component of medical knowledge but do not consistently represent the broader reasoning capacity required for clinically grounded tasks. Benchmark-level comparisons further supported the observed relationships (Fig.~\ref{fig:mc_cr_scat}).

\section{Data contamination analysis}\label{data_contamination}

\begin{figure}[htbp]
\centering
\includegraphics[width=0.95\linewidth]{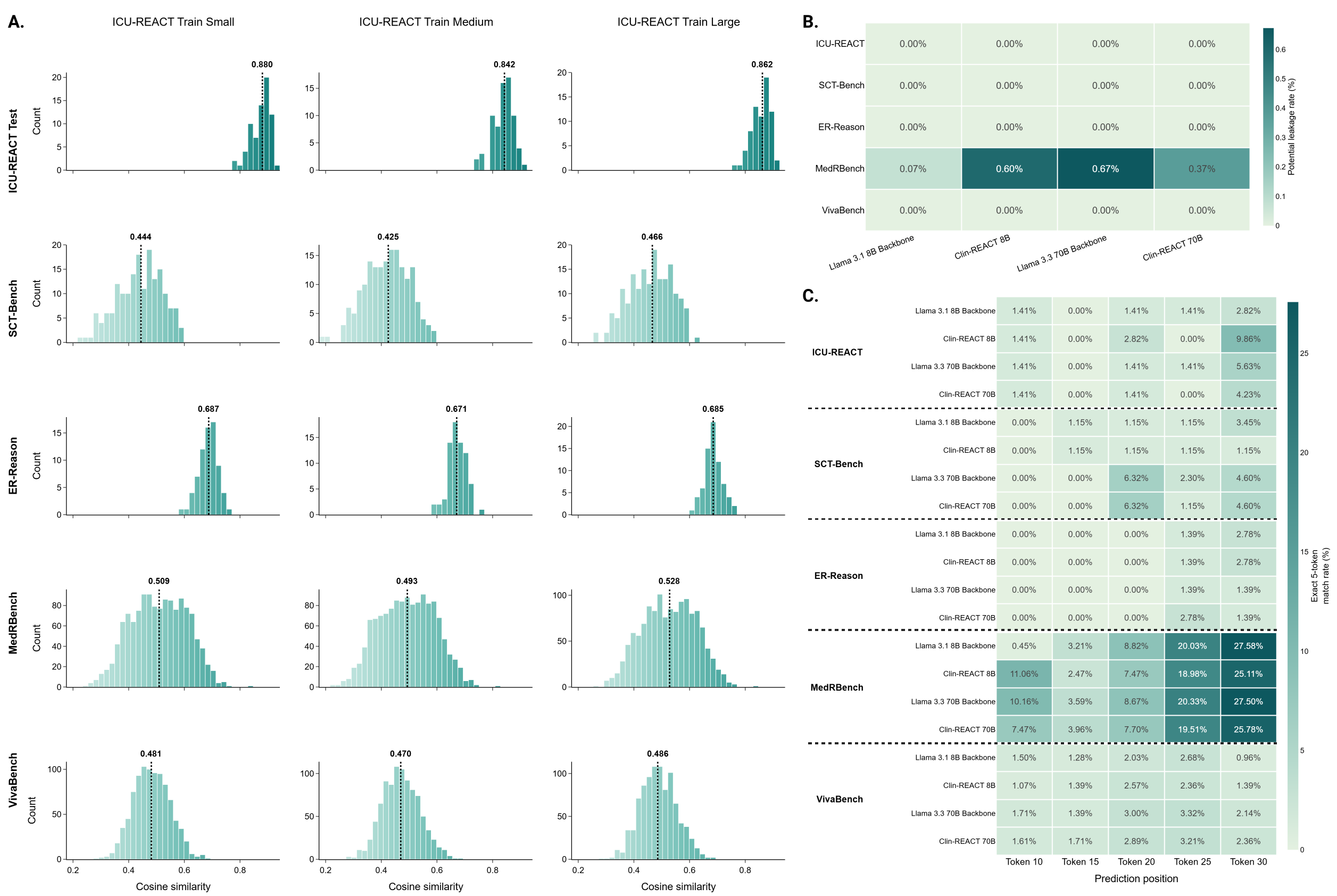}
\caption{Data-contamination analysis between ICU-REACT training sets and clinical reasoning evaluation benchmarks. (A) Distributions of nearest-neighbor cosine similarity between samples from five clinical reasoning benchmarks (rows: ICU-REACT Test, SCT-Bench, ER-Reason, MedRBench, and VivaBench) and the ICU-REACT Train Small, Medium, and Large datasets (columns). Bars show the number of evaluation samples within each cosine-similarity interval, with color intensity increasing with similarity; vertical dotted lines indicate the median similarity for each comparison, with the median value reported above each distribution. (B) Potential leakage rates for Clin-REACT 8B and 70B and their corresponding Llama 3.1 8B Instruct and Llama 3.3 70B Instruct backbones across the five clinical reasoning benchmarks. Each cell reports the percentage of samples satisfying the potential-leakage criterion, with darker teal indicating a higher rate. (C) Exact 5-token match rates at five prediction positions (tokens 10, 15, 20, 25, and 30) for the same four models across each benchmark. Rows are grouped by benchmark and separated by dashed horizontal lines; each cell reports the percentage of samples with an exact 5-token match at the corresponding prediction position, with darker teal indicating a higher match rate.}
\label{fig:data_contamination}
\end{figure}

To assess whether benchmark performance could be influenced by overlap between the Clin-REACT training data and evaluation sets, we conducted complementary semantic-similarity and completion-based contamination analyses (Fig.~\ref{fig:data_contamination}). First, for each sample in ICU-REACT Test and the four external benchmarks (SCT-Bench, ER-Reason, MedRBench, and VivaBench), we identified its nearest neighbor within each of the ICU-REACT Small, Medium, and Large training sets using cosine similarity of the input-output text embeddings. As expected, ICU-REACT Test showed the highest similarity to the ICU-REACT training sets (median cosine similarity, 0.842--0.880), whereas similarity was substantially lower for the external benchmarks: 0.425--0.466 for SCT-Bench, 0.671--0.687 for ER-Reason, 0.493--0.528 for MedRBench, and 0.470--0.486 for VivaBench (Fig.~\ref{fig:data_contamination}A). We subsequently performed a stricter completion-based analysis using Clin-REACT 8B and 70B and their respective Llama 3.1 8B Instruct and Llama 3.3 70B Instruct backbones. For each reference answer, the model was conditioned on prefixes truncated at token positions 10, 15, 20, 25, and 30 of the input, and the subsequent five generated tokens were compared with the corresponding five reference tokens. We defined a sample as exhibiting potential leakage when exact 5-token matches occurred at three or more of the five tested positions \cite{xu_benchmarking_2024, wu_bridge_2026}, thereby requiring repeated continuation-level agreement rather than an isolated phrase match.

Under this criterion, no potential leakage was detected for ICU-REACT, SCT-Bench, ER-Reason, or VivaBench for either Clin-REACT model or either backbone (Fig.~\ref{fig:data_contamination}B). The only non-zero rates occurred on MedRBench and remained low: 0.07\% for the Llama 3.1 8B backbone, 0.60\% for Clin-REACT 8B, 0.67\% for the Llama 3.3 70B backbone, and 0.37\% for Clin-REACT 70B. Importantly, the direction of change after fine-tuning was not consistent across scales, with potential leakage increasing relative to the 8B backbone but decreasing relative to the 70B backbone, providing no evidence of a systematic increase attributable to Clin-REACT training. Position-specific exact-match rates were higher than the stricter leakage rates, particularly for MedRBench at later prediction positions, where token-30 match rates ranged from 25.1\% to 27.6\% across models (Fig.~\ref{fig:data_contamination}C). However, these local matches were also prominent in the pretrained backbones and rarely persisted across enough positions to satisfy the potential-leakage criterion, suggesting that they more likely reflect reproducible or formulaic clinical language than sustained memorization of benchmark answers. Together, the semantic and completion-based analyses provide little evidence that the observed Clin-REACT performance gains were driven by systematic contamination of the evaluated benchmarks.

\section{ICU Variable and Taxonomy Framework}\label{omop_var}

We developed a structured ICU variable and taxonomy framework algined with OMOP concepts to support the building of the ICU-REACT dataset and future model deployment compatible with EHR systems. This framework was designed to reflect clinically meaningful domains consistent with established ICU severity scoring systems and outcome prediction models. Established ICU severity scores (APACHE, SAPS, SOFA) demonstrate that demographic, diagnostic, physiologic, laboratory, intervention, and admission variables are valuable determinants of patient outcomes~\cite{breslow_severity_2012, pellathy_intensive_2021}.

We categorized the organization of variables into eight domains: (1) Diagnosis, (2) Physiology, (3) Laboratory Values, (4) Imaging, (5) Medications, (6) Interventions, (7) Severity Scores and Clinical Assessments, and (8) Administrative Variables. The selection of these domains was guided by medical trainees and informed by prior work defining common data elements and data dictionaries for critical care research~\cite{murphy_development_2025, johnson_mimic-iii_2016}.

Within each domain, most variables were identified through the common practice of expert elicitation~\cite{reese_patient_2018}, from attending physicians, residents, and medical students with clinical training at the University of Florida College of Medicine. Prior work has defined critical care data elements through expert consensus approaches~\cite{murphy_development_2025, reese_patient_2018}, while our approach incorporates vignette-based expert elicitation to capture clinician reasoning in context-specific ICU scenarios. Participants were presented with standardized ICU clinical vignettes representing common critical care scenarios. For each vignette, participants selected the variables they considered clinically relevant for patient assessment, monitoring, and management. Responses were collected across participants, and variables endorsed through this process were added to the eight predefined domains.

Variables were consolidated to reduce redundancy and grouped into clinically coherent sub-domains; for instance, the Diagnosis domain was organized into organ system-based groupings (Cardiac, Pulmonary, Neurological, Infectious, Renal, Metabolic, Endocrine, Gastrointestinal, Hepatic, and Hematological) to reflect common etiologies encountered in ICU practice. This expert-driven data collection method aims to ensure that the final framework reflects real-world clinical reasoning and maintains structured organization for future applications. The full list of variables organized by clinical domain and sub-domains can be found on Table~\ref{tab:icu-variable-taxonomy}.

\begingroup
\small
\setlength{\tabcolsep}{4pt}
\renewcommand{\arraystretch}{1.15}
\setlength{\emergencystretch}{1em}
\begin{longtable}{>{\raggedright\arraybackslash}p{\dimexpr 0.220\linewidth-2\tabcolsep\relax}>{\raggedright\arraybackslash}p{\dimexpr 0.220\linewidth-2\tabcolsep\relax}>{\raggedright\arraybackslash}p{\dimexpr 0.560\linewidth-2\tabcolsep\relax}}
\caption{ICU variable taxonomy organized by clinical domain and sub-domain.}
\label{tab:icu-variable-taxonomy} \\
\toprule
\textbf{Domain} & \textbf{Subcategory} & \textbf{Variables} \\
\midrule
\endfirsthead
\multicolumn{3}{c}{\tablename\ \thetable{} -- continued from previous page} \\
\toprule
\textbf{Domain} & \textbf{Subcategory} & \textbf{Variables} \\
\midrule
\endhead
\midrule
\multicolumn{3}{r}{Continued on next page} \\
\endfoot
\bottomrule
\endlastfoot
Diagnosis & Cardiac & acute coronary syndrome; atrial fibrillation; bradycardia; cardiac arrest; cardiogenic shock; complete heart block; endocarditis; heart failure; myocardial infarction; pericardial effusion; supraventricular tachycardia; ventricular fibrillation; ventricular tachycardia; Aortic dissection; Cardiomyopathy; Myocarditis; Valvular heart disease; DVT; Pulmonary Embolism \\
Diagnosis & Pulmonary & respiratory acidosis; respiratory alkalosis; ventilator-\allowbreak{}associated pneumonia; ARDS; COPD exacerbation; aspiration pneumonitis; asthma exacerbation; hemothorax; pleural effusion; pneumonia; pneumothorax; pulmonary contusion; pulmonary edema; pulmonary embolism; pulmonary hypertension; respiratory failure; air embolism; fat embolism syndrome; Atelectasis; Pulmonary fibrosis; Bronchiectasis; Lung cancer \\
Diagnosis & Neurological & hemorrhage; encephalitis; myxedema coma; Guillain-\allowbreak{}Barr\'{e} syndrome; brain death; cerebral edema; encephalopathy; epidural hematoma; intracranial hemorrhage; ischemic stroke; myasthenia gravis; status epilepticus; subarachnoid hemorrhage; subdural hematoma; traumatic brain injury; Spinal cord injury; Brain tumor; Hydrocephalus; ICU-\allowbreak{}acquired weakness; Delirium (Hypoactive) \\
Diagnosis & Infectious & bacteremia; catheter-\allowbreak{}related bloodstream infection; fungemia; necrotizing fasciitis; anaphylaxis; sepsis; septic shock; COVID-\allowbreak{}19; Influenza; Cellulitis; Abscess; Ventilator-\allowbreak{}associated pneumonia; CLABSI; CAUTI \\
Diagnosis & Renal & urinary tract infection; adrenal insufficiency; acute kidney injury; acute tubular necrosis; chronic kidney disease; nephrotic syndrome; rhabdomyolysis; urinary retention; Nephrolithiasis; Glomerulonephritis; Hydronephrosis; Pyelonephritis; Hepatorenal Syndrome \\
Diagnosis & Metabolic & hyperkalemia; hypernatremia; hypokalemia; hyponatremia; metabolic acidosis; metabolic alkalosis; hyperthermia; hypothermia; malignant hyperthermia; multi-\allowbreak{}organ dysfunction syndrome; systemic inflammatory response syndrome; Hypomagnesemia; Hypermagnesemia; Hypophosphatemia; Hyperphosphatemia; Heat stroke \\
Diagnosis & Endocrine & diabetic ketoacidosis; hyperglycemia; hyperosmolar hyperglycemic state; hypoglycemia; thyroid storm; Hypothyroidism; Hyperthyroidism; Cushing's syndrome \\
Diagnosis & Gastrointestinal & Clostridioides difficile colitis; GI bleed; acute pancreatitis; bowel ischemia; bowel obstruction; cholecystitis; ileus; lower GI bleed; upper GI bleed; meningitis; hemorrhagic stroke; Gastritis; Peptic ulcer disease; Diverticulitis; Volvulus \\
Diagnosis & Hepatic & acute liver failure; ascites; esophageal varices; liver failure; Cirrhosis; Portal hypertension; Steatohepatitis; Hepatitis \\
Diagnosis & Hematological & anemia; deep vein thrombosis; disseminated intravascular coagulation; heparin-\allowbreak{}induced thrombocytopenia; thrombocytopenia; thrombotic thrombocytopenic purpura; Leukemia; Lymphoma; Sickle cell disease; Hemophilia; Hypovolemic shock \\
\addlinespace[3pt]
Physiology & Excretion & Urine output; Hourly urine output; Chest tube output; Drain output; Nasogastric tube output; Stool output; Bile output; Jackson-\allowbreak{}Pratt drain output; Hemovac drain output \\
Physiology & Fluidics & Fluid intake; Fluid balance; Daily net fluid balance; Cumulative fluid balance; Insensible losses; Bolus volume; Maintenance rate; Cumulative Balance \\
Physiology & Hemodynamics & Central venous pressure; Cardiac output; Cardiac index; Stroke volume; Stroke volume index; Ejection fraction; Systemic vascular resistance; Systemic vascular resistance index; Pulmonary vascular resistance; Pulmonary vascular resistance index; Pulmonary artery pressure; Pulmonary capillary wedge pressure; Right atrial pressure; Shock index; Stroke volume variation; Pulse pressure variation; Cardiac power output; Fluid responsiveness; Passive leg raise result; IVC distensibility index; Hemodynamic instability; Pulse Pressure \\
Physiology & Oxygenation & PaO2/\allowbreak{}FiO2 ratio; Oxygenation index; Alveolar-\allowbreak{}arterial oxygen gradient; Dead space fraction; Ventilatory ratio; Mixed venous oxygen saturation; Central venous oxygen saturation; Oxygen delivery; Oxygen consumption; Oxygen extraction ratio; End-\allowbreak{}tidal CO2; Venous-\allowbreak{}to-\allowbreak{}arterial CO2 gap; Work of breathing; Spontaneous Breathing Trial (SBT); Rapid shallow breathing index; Diaphragm excursion; pH; PaO2; PaCO2; Bicarbonate; Base excess; Lactate; Anion gap; Mechanical Power; Stress Index; Mean Airway Pressure; Spontaneous Effort; Cough Strength \\
Physiology & Pressures & Intra-\allowbreak{}abdominal pressure; Bladder pressure \\
Physiology & Ventilation & Mode of ventilation; Respiratory rate (set); Mandatory rate; Spontaneous rate; PEEP; FiO2; Pressure support; Volume control; Inspiratory time; Expiratory time; I:E ratio; Mean airway pressure; Lung volumes; Tidal volume; Minute ventilation; Expiratory minute volume; Peak inspiratory pressure; Plateau pressure; Driving pressure; Auto-\allowbreak{}PEEP; Mechanical power; Respiratory compliance; Respiratory compliance (static); Respiratory compliance (dynamic); Respiratory system resistance; Respiratory system elastance; Transpulmonary pressure; Airway resistance; Trigger sensitivity \\
Physiology & Vitals & Mean arterial pressure; Heart rate; Blood pressure systolic; Blood pressure diastolic; Pulse pressure; Respiratory rate; Oxygen saturation; Core temperature; Weight \\
\addlinespace[3pt]
Imaging & Angiography & Cerebral Angiography; Coronary Angiography; GI bleeding extravasation; Peripheral Angiography \\
Imaging & Basic Radiography & Abdominal X-\allowbreak{}ray (KUB); Atelectasis (X-\allowbreak{}ray); Cardiomegaly; Chest X-\allowbreak{}ray; Infiltrates; Line position (X-\allowbreak{}ray); Mediastinal widening; Pelvic X-\allowbreak{}ray; Pleural effusion (X-\allowbreak{}ray); Pneumothorax (X-\allowbreak{}ray); Portable Chest X-\allowbreak{}ray; Pulmonary edema (X-\allowbreak{}ray); Tube position (X-\allowbreak{}ray) \\
Imaging & CT & Abscess (CT); Aortic dissection (CT); Bowel ischemia (CT); Bowel obstruction (CT); Brain edema; CT Abdomen/\allowbreak{}Pelvis; CT Chest; CT Head; CT Spine; CTA Chest (PE protocol); CTA Head/\allowbreak{}Neck; Free air (CT); Intracranial hemorrhage; Midline shift; Pulmonary embolism (CT) \\
Imaging & EEG & EEG findings \\
Imaging & EKG & 12-\allowbreak{}lead EKG; EKG Rhythm; Ischemic EKG changes; EKG Findings; ST-\allowbreak{}segment changes \\
Imaging & Fluoroscopy & Barium Swallow; Diaphragmatic Fluoroscopy (Sniff Test) \\
Imaging & MRI & Acute stroke (MRI/\allowbreak{}DWI); Brain tumor (MRI); Cardiac MRI; MRI Brain; MRI Spine \\
Imaging & Scintigraphy & Bone Scan; V/\allowbreak{}Q Scan \\
Imaging & Ultrasound & B-\allowbreak{}lines (Ultrasound); Bedside Echo; Bedside IVC assessment; FAST Exam; Lung Ultrasound; Lung sliding; Abdominal Ultrasound; Bladder Ultrasound; Echocardiogram (TEE); Echocardiogram (TTE); Pericardial effusion findings; Renal Ultrasound; Valvular function findings; Vascular Ultrasound (DVT) \\
\addlinespace[3pt]
Medications & Analgesics & Acetaminophen; Fentanyl; Hydromorphone; Ketorolac; Methadone; Morphine; Remifentanil \\
Medications & Antiarrhythmics & Adenosine; Amiodarone; Digoxin; Diltiazem; Esmolol; Lidocaine; Metoprolol \\
Medications & Antibiotics & Azithromycin; Aztreonam; Cefepime; Ceftriaxone; Daptomycin; Gentamicin; Levofloxacin; Linezolid; Meropenem; Metronidazole; Piperacillin-\allowbreak{}Tazobactam; Vancomycin \\
Medications & Anticoagulants & Apixaban; Argatroban; Bivalirudin; Enoxaparin; Heparin; Rivaroxaban; Warfarin \\
Medications & Antidotes & Andexanet Alfa; Flumazenil; Idarucizumab; Naloxone; Protamine; Sugammadex; Vitamin K \\
Medications & Antifungals & Amphotericin B; Fluconazole; Micafungin; Voriconazole \\
Medications & Antihypertensives & Clevidipine; Enalaprilat; Hydralazine; Labetalol; Nicardipine; Nitroglycerin; Nitroprusside \\
Medications & Antiplatelets & Aspirin; Cangrelor; Clopidogrel; Ticagrelor \\
Medications & Antivirals & Acyclovir; Ganciclovir; Oseltamivir \\
Medications & Diuretics & Bumetanide; Furosemide; Metolazone; Spironolactone \\
Medications & Electrolytes & Calcium Chloride; Calcium Gluconate; Magnesium Sulfate; Potassium Chloride; Sodium Bicarbonate; Sodium Phosphate \\
Medications & Fluids & Albumin 25\%; Albumin 5\%; Lactated Ringers; Normal Saline \\
Medications & Hemostatics & Alteplase; Kcentra; Tranexamic Acid \\
Medications & Hormones & Desmopressin; Dexamethasone; Glucagon; Hydrocortisone; Insulin; Levothyroxine; Methylprednisolone; Octreotide \\
Medications & Inotropes & Dobutamine; Dopamine; Isoproterenol; Milrinone \\
Medications & Paralytics & Cisatracurium; Rocuronium; Succinylcholine; Vecuronium \\
Medications & Pressors & Angiotensin II; Ephedrine; Epinephrine; Norepinephrine; Phenylephrine; Vasopressin \\
Medications & Protectants & Famotidine; Metoclopramide; Ondansetron; Pantoprazole \\
Medications & Reversal/Anecdotes & Sugammadex; Vitamin K \\
Medications & Sedatives & Dexmedetomidine; Etomidate; Ketamine; Lorazepam; Midazolam; Propofol \\
Medications & Supportive & Alteplase; Prothrombin Complex Concentrate; Tranexamic Acid \\
\addlinespace[3pt]
Administrative & Ethics & Advance directives; DNR/\allowbreak{}DNI status; Family meeting notes; Hospice referral \\
Administrative & Habits & Alcohol use; Drug use history; Smoking status \\
Administrative & Identity & Age; Date of birth; Ethnicity; Gender; Height; Language; Medical record number; Name; Race \\
Administrative & Logistics & Admission date; Admission diagnosis; Admission source; Admission time; Admitting service; Discharge date; Hospital length of stay; ICU admission date; ICU length of stay \\
Administrative & Notes & Nutrition notes; Occupational therapy notes; Palliative care notes; Physical therapy notes; Respiratory therapy notes; Social work notes; Speech therapy notes \\
Administrative & Social & Employment status; Housing status; Insurance type; Next of kin \\
Administrative & Status & Allergies; BMI; Organ donor status; Pregnancy status; Vaccination status \\
\addlinespace[3pt]
Intervention & Advanced Life Support & ECMO; IABP; VAD; continuous renal replacement therapy; dialysis; hemodialysis; implantable cardioverter-\allowbreak{}defibrillator; peritoneal dialysis; permanent pacemaker; temporary pacemaker; therapeutic hypothermia \\
Intervention & Blood Product Transfusion & cryoprecipitate transfusion; plasma transfusion; platelet transfusion; red blood cell transfusion \\
Intervention & Drainage \& Tubes & Foley catheter; Hemovac drain; Jackson-\allowbreak{}Pratt drain; biliary drain; chest tube placement; nasogastric tube; orogastric tube; percutaneous endoscopic gastrostomy; rectal tube; suprapubic catheter; wound VAC \\
Intervention & Invasive Procedures & cardiac catheterization; cardioversion; colonoscopy; debridement; defibrillation; endoscopy; lumbar puncture; paracentesis; percutaneous coronary intervention; pericardiocentesis; skin graft; thoracentesis \\
Intervention & Neurological Interventions & external ventricular drain; intracranial pressure monitor; lumbar drain; train-\allowbreak{}of-\allowbreak{}four monitoring \\
Intervention & Nutrition \& Fluid Management & IV fluids administered; enteral feeding rate; enteral nutrition; fluid resuscitation; parenteral nutrition \\
Intervention & Procedures & tracheostomy care \\
Intervention & Prophylaxis \& Supportive Care & DVT; active dialysis orders; bladder scan; current oxygen delivery device; dvt prophylaxis; neuromuscular blockade; neuromuscular electrical stimulation; pressure ulcer staging; sedation vacation; sequential compression device; spontaneous awakening trial; stress ulcer prophylaxis; tracheostomy care; venous thromboembolism prophylaxis \\
Intervention & Respiratory Interventions & bilevel positive airway pressure; bronchoscopy; continuous positive airway pressure; extubation; high-\allowbreak{}flow nasal cannula; intubation; mechanical insufflation-\allowbreak{}exsufflation; oxygen via tracheostomy collar; prone positioning; recruitment maneuvers; reintubation; tracheostomy \\
Intervention & Vascular Access & PICC line; arterial line; central line; epidural catheter; intraosseous line; midline catheter; peripheral IV; pulmonary artery catheter \\
\addlinespace[3pt]
Laboratory Measurements & Cardiac & BNP; CK; CK-\allowbreak{}MB; NT-\allowbreak{}proBNP; Troponin I; Troponin T \\
Laboratory Measurements & Coagulation & Activated clotting time (ACT); Anti-\allowbreak{}Xa level; Antithrombin III; D-\allowbreak{}dimer; Fibrinogen; INR; PT; PTT; Protein C activity; Protein S activity; ROTEM; TEG - MA; TEG - R time; Thrombin time \\
Laboratory Measurements & Complete Blood Count & Basophils (absolute); Eosinophils (absolute); Hematocrit; Hemoglobin; Immature granulocytes (Bands); Lymphocytes (absolute); MCH; MCHC; MCV; Monocytes (absolute); Neutrophils (absolute); Nucleated RBCs; Peripheral smear; Platelet count; RBC count; RDW; Reticulocyte count; WBC count \\
Laboratory Measurements & Comprehensive Metabolic Panel & ALP; ALT; AST; Albumin; Anion gap; BUN; Bicarbonate; Calcium (total); Chloride; Creatinine; Direct Bilirubin; Glucose; Indirect Bilirubin; Ionized Calcium; Magnesium; Phosphorus; Potassium; Sodium; Total Bilirubin; Total Protein; Uric acid \\
Laboratory Measurements & Endocrine & ACTH; Cortisol (random); Free T3; Free T4; HbA1c; Insulin level; PTH; TSH \\
Laboratory Measurements & Inflammatory & CRP; ESR; Ferritin; Haptoglobin; Interleukin-\allowbreak{}6 (IL-\allowbreak{}6); LDH \\
Laboratory Measurements & Microbiology & Blood culture; C. diff PCR; CSF culture; Fungal culture; Gram stain; MRSA Screen; Procalcitonin; Sputum culture; Urine culture; Viral PCR (Respiratory Panel); Wound culture \\
Laboratory Measurements & Nutritional & Cystatin C; Folate; Prealbumin; Transferrin; Vitamin B12; Vitamin D \\
Laboratory Measurements & Toxicology & Acetaminophen level; Amikacin level; Carbamazepine level; Cyclosporine level; Digoxin level; Ethanol level; Gentamicin level; Lithium level; Phenytoin level; Salicylate level; Tacrolimus level; Tobramycin level; Urine drug screen; Valproic acid level; Vancomycin trough \\
Laboratory Measurements & Urinalysis & Urine bilirubin; Urine blood; Urine glucose; Urine ketones; Urine leukocyte esterase; Urine microscopic exam; Urine nitrite; Urine osmolality; Urine pH; Urine protein; Urine specific gravity \\
\addlinespace[3pt]
Scores and Assessments & Acute severity of illness scores & APACHE II score; LODS; OASIS; SAPS II; SOFA score; qSOFA \\
Scores and Assessments & Chronic \& Comorbidity Indices & Charlson Comorbidity Index; Clinical Frailty Scale; Elixhauser Comorbidity Index \\
Scores and Assessments & Neuro-Cognitive Assessments & FOUR Score; Glasgow Coma Scale; Hunt and Hess Scale; Mental status; Muscle strength; NIH Stroke Scale; Pupil response; Seizures; modified Rankin Scale \\
Scores and Assessments & Nutrition Scores & MUST; NUTRIC score \\
Scores and Assessments & Organ Function Scores & Balthazar score; Child-\allowbreak{}Pugh score; MELD score \\
Scores and Assessments & Pain Scores & BPS; CPOT; NVPS; Pain score \\
Scores and Assessments & Physical Exam Findings & Abdominal distension; Bowel sounds; Capillary refill; Peripheral edema; Skin turgor \\
Scores and Assessments & Psychiatric Assessments & CAM-\allowbreak{}ICU; Delirium \\
Scores and Assessments & Safety Risk Assessments & Braden Scale; CURB-\allowbreak{}65; MEWS; Morse Fall Scale; NEWS; Wells score \\
Scores and Assessments & Sedation Scales & RASS; SAS; Sedation level \\
\end{longtable}
\endgroup

\section{ICU Topics Dimensions}\label{icu_topics}

Prompting a language model repeatedly with only a topic label (e.g.\
``septic shock'') yields cases that vary in wording but converge on a
small number of prototypical presentations, which limits the diversity
gained from augmentation. To avoid this, we decomposed each ICU topic
into four axes along which real cases differ: the underlying
clinical scenario (the etiology or precipitating event), the
presentation (the bedside findings or trigger prompting a
decision), the management decision under consideration, and
contextual modifiers (comorbidities, competing risks, resource
constraints, and goals-of-care limitations that alter what is
appropriate). These axes are largely independent of one another: the
same presentation can arise from different etiologies, and the same
decision can be appropriate or contraindicated depending on the
modifier. Sampling them separately therefore produces variation that is
clinically meaningful rather than merely lexical, and lets a modest
number of enumerated items cover a much larger space of distinct cases.

We selected nine clinician-informed topics spanning the decision contexts that dominate
adult ICU care, and for each topic used ChatGPT to enumerate candidate
items along the four axes, seeded with the topic definition and
constrained to decisions that arise in routine intensive care. The
generated lists were then reviewed and edited to remove redundant or
implausible entries and to keep items at a comparable level of
granularity. The final set comprises 315 items: 96 clinical
scenarios, 84 presentations, 87 management decisions, and 48 contextual
modifiers, distributed across topics as shown in
Table~\ref{tab:icu-topic-dimensions}. Drawing a single item from each
axis would yield 50{,}017 distinct scenario specifications across the nine
topics, which allowed for diverse generation of training samples. 

\begingroup
\small
\setlength{\tabcolsep}{4pt}
\renewcommand{\arraystretch}{1.15}
\setlength{\emergencystretch}{1em}
\begin{longtable}{>{\raggedright\arraybackslash}p{\dimexpr 0.240\linewidth-2\tabcolsep\relax}>{\raggedright\arraybackslash}p{\dimexpr 0.200\linewidth-2\tabcolsep\relax}>{\raggedright\arraybackslash}p{\dimexpr 0.560\linewidth-2\tabcolsep\relax}}
\caption{ICU clinical topics and dimensions used to generate ICU-REACT training sets}
\label{tab:icu-topic-dimensions} \\
\toprule
\textbf{Topic} & \textbf{Dimension} & \textbf{Items} \\
\midrule
\endfirsthead
\multicolumn{3}{c}{\tablename\ \thetable{} -- continued from previous page} \\
\toprule
\textbf{Topic} & \textbf{Dimension} & \textbf{Items} \\
\midrule
\endhead
\midrule
\multicolumn{3}{r}{Continued on next page} \\
\endfoot
\bottomrule
\endlastfoot
Hemodynamic Instability / Shock & Clinical scenarios & Septic shock from pneumonia; Septic shock from intra-\allowbreak{}abdominal source; Postoperative hypotension after major surgery; Acute upper GI hemorrhage with volume loss; Traumatic hemorrhagic shock; Cardiogenic shock after acute MI; Acute decompensated heart failure with low output; Acute right-\allowbreak{}ventricular failure from massive PE; Cardiac tamponade with obstructive physiology; Anaphylactic distributive shock; Neurogenic/\allowbreak{}spinal shock after trauma; Adrenal/\allowbreak{}relative cortisol insufficiency in shock; Tension pneumothorax causing obstructive shock; Dialysis/\allowbreak{}CRRT-\allowbreak{}associated hypotension in ICU \\
Hemodynamic Instability / Shock & Presentations & Sustained low MAP despite initial fluids; Rising lactate and signs of poor perfusion; Cool, mottled extremities and delayed cap refill; Warm, vasodilatory shock with wide pulse pressure; Oliguria or declining urine output; Acute change in mental status from hypoperfusion; Tachycardia with narrow pulse pressure; Escalating vasopressor requirement over hours; Hypotension with new arrhythmia; Need for closer hemodynamic monitoring; Hypotension in setting of limited fluid tolerance (ARDS/\allowbreak{}CHF) \\
Hemodynamic Instability / Shock & Management decisions & Initiate norepinephrine as first-\allowbreak{}line vasopressor; Add a second vasopressor such as vasopressin; Start inotropic support for low cardiac output; Administer another fluid bolus vs hold further fluids; Transfuse packed RBCs for hemorrhagic component; Obtain bedside echocardiography to define etiology; Place arterial and/\allowbreak{}or central line for monitoring; Initiate stress-\allowbreak{}dose corticosteroids for refractory shock; Arrange source-\allowbreak{}control procedure for shock driver; Escalate to mechanical/\allowbreak{}advanced circulatory support; Begin conservative fluid strategy after stabilization \\
Hemodynamic Instability / Shock & Contextual modifiers & Patient has significant cardiopulmonary comorbidities; High risk of fluid overload/\allowbreak{}ARDS so resuscitate cautiously; Limited vascular access or line-\allowbreak{}related infection concern; Post-\allowbreak{}cardiac surgery physiology alters pressor choice; Immunocompromised host with atypical infection; Goals-\allowbreak{}of-\allowbreak{}care or DNR/\allowbreak{}DNI status limits escalation \\
\addlinespace[3pt]
Respiratory Failure & Clinical scenarios & ARDS secondary to severe pneumonia; Community-\allowbreak{}acquired pneumonia with hypoxemic failure; COPD exacerbation with hypercapnic respiratory failure; Obesity hypoventilation/\allowbreak{}OSA causing ventilatory failure; Postoperative atelectasis with impaired gas exchange; Aspiration event in intubated patient; Ventilator-\allowbreak{}associated pneumonia in ICU; Neuromuscular weakness (GBS/\allowbreak{}MG) causing hypoventilation; Cardiogenic pulmonary edema with respiratory compromise; Prolonged mechanical ventilation needing weaning; Post-\allowbreak{}extubation stridor or upper-\allowbreak{}airway compromise; Trauma/\allowbreak{}chest wall injury with poor ventilation; Severe asthma in ICU not responding to usual therapy \\
Respiratory Failure & Presentations & Worsening oxygen requirement despite current support; Increased work of breathing and tachypnea; Rising CO2 with somnolence or confusion; Ventilator dyssynchrony or high peak pressures; Difficulty clearing secretions or poor cough; Failure of HFNC/\allowbreak{}NIV trial; Hemodynamic instability related to high PEEP; Ready for spontaneous breathing trial; Failed prior extubation attempt; Recurrent desaturation with minimal activity; Concern for aspiration or VAP on current ventilator settings \\
Respiratory Failure & Management decisions & Escalate to non-\allowbreak{}invasive ventilation (e.g. BiPAP); Proceed with endotracheal intubation and mechanical ventilation; Adjust ventilator settings for lung-\allowbreak{}protective strategy; Increase or titrate PEEP/\allowbreak{}FiO2 to maintain oxygenation; Initiate prone positioning or recruitment maneuvers; Start weaning protocol/\allowbreak{}SBT and assess extubation readiness; Switch to HFNC from NIV for better tolerance; Request bronchoscopy/\allowbreak{}suctioning for secretion management; Consult for tracheostomy due to prolonged ventilation; Add sedation/\allowbreak{}analgesia to improve ventilator synchrony; Treat underlying cardiac/\allowbreak{}pulmonary edema to improve oxygenation \\
Respiratory Failure & Contextual modifiers & Patient is hemodynamically fragile and cannot tolerate high PEEP; High aspiration risk/\allowbreak{}enteral feeding ongoing; Difficult airway or prior failed intubation; Immunosuppressed patient at risk for atypical infections; Obesity/\allowbreak{}body habitus complicates ventilation strategy; Infection control/\allowbreak{}isolation requirements limit procedures \\
\addlinespace[3pt]
Sepsis and Severe Infections & Clinical scenarios & Pneumonia with sepsis requiring ICU admission; Complicated intra-\allowbreak{}abdominal infection/\allowbreak{}postoperative peritonitis; Central line-\allowbreak{}associated bloodstream infection; Urinary tract infection progressing to urosepsis; Skin/\allowbreak{}soft-\allowbreak{}tissue infection or necrotizing fasciitis; Infected indwelling device or prosthesis; Febrile neutropenia with suspected bacterial source; Ventilator-\allowbreak{}associated pneumonia in intubated patient; Biliary tract infection/\allowbreak{}cholangitis; Catheter-\allowbreak{}associated peritonitis in dialysis patient; Endocarditis suspected in ICU patient; Sepsis of unclear source requiring broad coverage \\
Sepsis and Severe Infections & Presentations & Fever or hypothermia with leukocytosis; Hypotension or rising vasopressor needs; New or worsening organ dysfunction (renal, hepatic, neuro); Persistent or recurrent bacteremia on cultures; Local signs of infection at catheter or wound site; Poor response to initial antibiotics; Elevated inflammatory markers or rising lactate; Respiratory decompensation due to infection; Need for source control identified on imaging; Immunocompromised host with atypical presentation \\
Sepsis and Severe Infections & Management decisions & Initiate broad-\allowbreak{}spectrum empiric antibiotics; De-\allowbreak{}escalate antibiotics based on culture and sensitivity; Remove or exchange an infected central line or device; Arrange surgical or percutaneous source control; Start or escalate vasopressors for septic shock; Add stress-\allowbreak{}dose steroids for refractory septic shock; Extend or shorten duration of antibiotic therapy; Switch to antifungal or antiviral coverage when indicated; Isolate patient and apply infection control measures; Initiate early enteral nutrition in septic patient; Consult infectious diseases for complex resistant organisms \\
Sepsis and Severe Infections & Contextual modifiers & Renal or hepatic dysfunction limiting antibiotic choice; Recent hospitalization or MDR organism risk; Pregnancy/\allowbreak{}postpartum infection considerations; Neutropenia/\allowbreak{}oncology patient with low inflammatory response; Source not yet identified; diagnostics still pending; Coagulopathy/\allowbreak{}bleeding risk complicating source-\allowbreak{}control procedure \\
\addlinespace[3pt]
Neurological Emergencies & Clinical scenarios & Traumatic brain injury after fall or MVC; Spontaneous intracranial hemorrhage; Acute ischemic stroke requiring ICU monitoring; Status epilepticus or recurrent seizures; Post-\allowbreak{}ictal coma not awakening as expected; ICU delirium in mechanically ventilated patient; Sedation-\allowbreak{}related depressed mental status; Suspected increased ICP from mass effect or edema; Neurosurgical postoperative patient with neuro changes; Meningitis/\allowbreak{}encephalitis with altered mental status \\
Neurological Emergencies & Presentations & Sudden drop in GCS or unresponsiveness; Unequal or sluggish pupils; Ongoing or recurrent seizure activity; New agitation or hyperactive delirium; Failure to awaken after sedation wean; Headache and vomiting suggesting increased ICP; Hypertension/\allowbreak{}bradycardia pattern concerning for herniation; Speech or focal motor deficits; Fever with meningismus in altered patient \\
Neurological Emergencies & Management decisions & Secure airway/\allowbreak{}intubate to protect from aspiration; Obtain emergent neuroimaging (CT/\allowbreak{}MRI/\allowbreak{}CTA); Initiate ICP-\allowbreak{}lowering measures (head-\allowbreak{}up, mannitol, HTS); Start or escalate antiepileptic medication; Adjust or lighten sedation to assess neuro status; Initiate delirium screening and non-\allowbreak{}pharm management; Control blood pressure within neuro-\allowbreak{}targeted range; Consult neurosurgery/\allowbreak{}neurology urgently; Place invasive monitoring (ICP/\allowbreak{}EVD) when indicated \\
Neurological Emergencies & Contextual modifiers & Anticoagulated/\allowbreak{}antiplatelet patient with bleeding risk; Cervical spine precautions or trauma constraints; Concurrent sepsis/\allowbreak{}hemodynamic instability; Renal/\allowbreak{}hepatic impairment limiting sedative choice; Difficult family/\allowbreak{}goals-\allowbreak{}of-\allowbreak{}care situation around prognosis \\
\addlinespace[3pt]
Renal Failure / Electrolyte Disorders & Clinical scenarios & Sepsis-\allowbreak{}associated AKI in ICU; Postoperative AKI after major surgery; Contrast-\allowbreak{}induced nephropathy; Cardiorenal syndrome with volume overload; Acute tubular necrosis from hypotension; Drug-\allowbreak{}induced nephrotoxicity; Obstructive uropathy recognized in ICU patient; Tumor lysis/\allowbreak{}metabolic ICU patient; Isolated severe hyperkalemia; Hypo/\allowbreak{}hypernatremia related to ICU therapies; CRRT-\allowbreak{}dependent patient with circuit or prescription issues \\
Renal Failure / Electrolyte Disorders & Presentations & Oliguria or anuria over several hours; Rising creatinine/\allowbreak{}BUN compared to baseline; Pulmonary/\allowbreak{}peripheral edema from fluid overload; Metabolic acidosis not improving; Dangerous electrolyte derangement (qualitative); Hemodynamic instability limiting diuresis; Need to start nephrotoxic/\allowbreak{}renally-\allowbreak{}cleared drug; CRRT filter clotting or inadequate clearance; Uremic symptoms or concerns in ICU patient \\
Renal Failure / Electrolyte Disorders & Management decisions & Initiate CRRT or intermittent hemodialysis; Temporize and treat hyperkalemia/\allowbreak{}electrolyte abnormality; Restrict fluids and use diuretics for volume management; Adjust or hold nephrotoxic/\allowbreak{}renally-\allowbreak{}cleared medications; Change CRRT prescription (dose, fluid removal, anticoagulation); Investigate and relieve obstruction; Consult nephrology for AKI not improving; Optimize hemodynamics to improve renal perfusion; Plan earlier RRT due to multi-\allowbreak{}organ failure; Correct sodium or acid-\allowbreak{}base disturbance cautiously \\
Renal Failure / Electrolyte Disorders & Contextual modifiers & Hemodynamic fragility limiting fluid removal; Liver disease or coagulopathy affecting dialysis access; Pediatric/\allowbreak{}small adult size affecting modality choice; Active sepsis or catheter infection risk; Goals of care limit initiation of chronic dialysis \\
\addlinespace[3pt]
Cardiac Emergencies & Clinical scenarios & New-\allowbreak{}onset atrial fibrillation with rapid ventricular response; Sustained ventricular tachycardia or VF arrest in ICU; Acute coronary syndrome in hemodynamically unstable patient; Post-\allowbreak{}cardiac arrest (ROSC) patient in ICU; Symptomatic bradycardia or high-\allowbreak{}grade AV block; Pericardial tamponade physiology; Post--cardiac surgery patient with arrhythmia; Decompensated heart failure with pulmonary edema; Takotsubo/\allowbreak{}stress cardiomyopathy presentation; Electrolyte-\allowbreak{}triggered arrhythmia in ICU \\
Cardiac Emergencies & Presentations & Arrhythmia with hypotension or chest discomfort; Recurrent or shockable rhythm post-\allowbreak{}ROSC; Persistent ST changes or ischemic symptoms; Low cardiac output with cool extremities; Syncope or presyncope in monitored patient; Rising filling pressures/\allowbreak{}JVD in ICU; Worsening dyspnea/\allowbreak{}pulmonary edema on monitor; Need for rate control before procedure; Bradycardia not responding to atropine \\
Cardiac Emergencies & Management decisions & Perform synchronized cardioversion or defibrillation; Choose rate vs rhythm control for AF; Start or adjust anticoagulation for arrhythmia; Optimize hemodynamics post-\allowbreak{}ROSC (fluids/\allowbreak{}pressors); Place temporary pacemaker or consult EP; Treat underlying ischemia/\allowbreak{}ACS and call cardiology; Diurese and afterload reduce in decompensated HF; Pericardiocentesis/\allowbreak{}pericardial drain for tamponade; Correct electrolytes/\allowbreak{}precipitating factors \\
Cardiac Emergencies & Contextual modifiers & Recent surgery/\allowbreak{}bleeding risk affecting anticoagulation; Chronic arrhythmia with different baseline HR goals; Renal dysfunction affecting medication choices; Do-\allowbreak{}not-\allowbreak{}shock or limited-\allowbreak{}resuscitation status; Limited catheterization/\allowbreak{}surgical resources overnight \\
\addlinespace[3pt]
Sedation, Pain, and Delirium Management & Clinical scenarios & Mechanically ventilated patient on continuous sedation; Trauma/\allowbreak{}TBI patient with agitation; Alcohol or benzodiazepine withdrawal in ICU; Postoperative patient with significant pain needs; Septic patient developing ICU delirium; Long-\allowbreak{}stay ICU patient with sleep disruption; Patient failing ventilator weaning due to oversedation; Elderly patient with hypoactive delirium; Patient with history of psychiatric disease on home meds \\
Sedation, Pain, and Delirium Management & Presentations & Agitation and RASS above target; Inadequate analgesia despite current regimen; Fluctuating confusion or CAM-\allowbreak{}ICU positive; Failure to awaken after sedation interruption; Autonomic hyperactivity suggesting withdrawal; Nighttime agitation/\allowbreak{}sundowning pattern; Ventilator dyssynchrony due to under/\allowbreak{}over-\allowbreak{}sedation; Family concern about over-\allowbreak{}sedation; Need for neuro exam but sedation too deep \\
Sedation, Pain, and Delirium Management & Management decisions & Adjust sedative/\allowbreak{}analgesic doses or agents; Initiate or repeat delirium screening and non-\allowbreak{}pharm measures; Treat alcohol/\allowbreak{}benzo withdrawal with appropriate protocol; Lighten sedation to facilitate weaning/\allowbreak{}mobilization; Add antipsychotic or alpha-\allowbreak{}2 agonist for agitation; Schedule pain medications instead of PRN only; Institute daily sedation interruption; Consult psych/\allowbreak{}pain service for complex regimen; Modify ventilator or sedation to improve synchrony \\
Sedation, Pain, and Delirium Management & Contextual modifiers & Renal/\allowbreak{}hepatic impairment limiting drug choice; Older adult/\allowbreak{}fall risk with sedative sensitivity; History of substance use disorder/\allowbreak{}tolerance; Concomitant sepsis/\allowbreak{}organ failure changing metabolism; Goals-\allowbreak{}of-\allowbreak{}care emphasizing comfort over wakefulness \\
\addlinespace[3pt]
Nutrition and Metabolic Support & Clinical scenarios & Intubated patient NPO requiring early enteral nutrition; Postoperative abdominal surgery with poor gastric tolerance; Severe pancreatitis with high aspiration risk; Septic/\allowbreak{}catabolic patient needing high protein/\allowbreak{}calorie support; ICU patient with uncontrolled hyperglycemia; Patient unable to maintain enteral access (dislodged tube); Prolonged ICU stay with inadequate oral intake; Burn/\allowbreak{}trauma patient with elevated metabolic demands; Patient with refeeding risk/\allowbreak{}malnutrition on admission \\
Nutrition and Metabolic Support & Presentations & High gastric residuals or vomiting with enteral feeds; Prolonged NPO status due to procedures; Concern for aspiration on current feeding route; Wide glucose variability or persistent hyperglycemia; Electrolyte shifts after starting nutrition; Inability to meet calorie goals with current regimen; Need to minimize fluid volume in feeds; Feeding interruptions due to imaging/\allowbreak{}procedures \\
Nutrition and Metabolic Support & Management decisions & Start or advance enteral nutrition; Switch to post-\allowbreak{}pyloric or jejunal feeding route; Initiate parenteral nutrition due to intolerance of enteral feeds; Adjust insulin regimen or initiate infusion for glycemic control; Modify caloric/\allowbreak{}protein goals based on catabolic state; Implement refeeding monitoring and electrolyte replacement; Use trophic/\allowbreak{}low-\allowbreak{}volume feeds when full feeds not tolerated; Coordinate feeding schedule around procedures; Consult nutrition/\allowbreak{}metabolic support team \\
Nutrition and Metabolic Support & Contextual modifiers & Renal/\allowbreak{}hepatic dysfunction affecting formula choice; Need for strict fluid restriction; Obesity/\allowbreak{}underweight requiring adjusted calculations; Immunocompromised or oncology patient; High aspiration risk/\allowbreak{}ventilated patient positioning \\
\addlinespace[3pt]
Hematologic / Coagulation Issues & Clinical scenarios & Active upper or lower GI bleeding; Postoperative surgical site bleeding; Coagulopathy from liver disease or sepsis; ICU patient on warfarin/\allowbreak{}DOAC with bleeding; Thrombocytopenia or suspected HIT; ICU patient at high risk for VTE due to immobility; DIC in setting of sepsis/\allowbreak{}trauma; Planned invasive procedure requiring correction \\
Hematologic / Coagulation Issues & Presentations & Ongoing oozing from line/\allowbreak{}wound sites; Drop in hemoglobin/\allowbreak{}hematocrit over short interval; Abnormal coagulation studies needing correction; Bleeding while on therapeutic anticoagulation; Need to start VTE prophylaxis but bleeding risk present; New thrombotic event while platelets are low; Multiorgan failure with DIC picture; Upcoming procedure with high bleeding risk \\
Hematologic / Coagulation Issues & Management decisions & Reverse or hold anticoagulation/\allowbreak{}antiplatelet therapy; Transfuse PRBCs/\allowbreak{}platelets/\allowbreak{}plasma/\allowbreak{}cryo as indicated; Start or withhold pharmacologic VTE prophylaxis; Evaluate and treat suspected HIT; Correct coagulopathy before procedures; Balance bleeding vs clotting in DIC/\allowbreak{}sepsis; Consult hematology for complex coagulopathy; Restart anticoagulation when bleeding controlled \\
Hematologic / Coagulation Issues & Contextual modifiers & Renal/\allowbreak{}hepatic impairment affecting drug clearance; Recent surgery with strict surgeon preferences; Limited blood product availability; High thrombotic risk (mechanical valve, recent PE); Palliative/\allowbreak{}comfort-\allowbreak{}focused goals of care \\
\end{longtable}
\endgroup

\clearpage
\section{Prompts}\label{prompts}

\subsection{Seed generation prompts}\label{seed_prompts}

\begin{promptbox}{Prompt 1}
\footnotesize
\textbf{Prompt for initial generation of ICU-REACT decision-making questions and retrieval tasks.}

\vspace{0.5em}

You are a medical expert. You have access to an electronic health record (EHR) database.

Generate \verb|{n_questions}| decision-making example questions that can be answered using an EHR database.

\vspace{1em}

\textbf{Each question should contain each of the following elements:}

\begin{enumerate}[leftmargin=1.5em]

    \item \textbf{Question that can be answered using an EHR database.} The question should meet the following requirements:

    \begin{itemize}[leftmargin=1.5em]
        \item Should be in the context of an Intensive Care Unit (ICU) setting.
        \item Should be actionable and discrete, focusing on decision-making that physicians would consider.
        \item Should reference one or more tables that would be found in an EHR database.
        \item Should be limited to individual patients.
        \item Should be applicable to most ICU unit types.
        \item Should cover different organ systems and topics relevant to the ICU setting.
        \item Can have multiple retrieval tasks from multiple tables.
        \item Should be unique and not highly similar to the examples provided.
        \item Should be formulated without restating the context of the question, but rather as a direct inquiry.
        \item Should be concise and focused, avoiding unnecessary complexity or jargon.
        \item Should be a single question, not a compound question or multiple questions in one.
        \item Should not ask about value trends or changes over time, as the focus is on discrete, actionable inquiries.
        \item Should not ask about current status of a patient, but rather about specific actions or decisions that can be made based on the available data.
    \end{itemize}

    \item \textbf{Context in which the question is asked.} The context should follow the template below:

    \vspace{0.5em}

    \begin{quote}
    {\footnotesize
    \texttt{[age][gender] patient with [past\_medical\_history] admitted to the ICU due to [surgery\_or\_current\_complication] is developing [critical\_condition]}
    }
    \end{quote}

    \item \textbf{Category of the question.} This should be a list of tags for the question, such as cardiovascular, respiratory, renal, or other relevant ICU categories.

    \item \textbf{ICU topic.} This should specify the relevant ICU topic that the question belongs to.

    \item \textbf{Retrieval tasks that need to be performed to answer the question.} Each task should include:

    \begin{itemize}[leftmargin=1.5em]
        \item Category of data that needs to be retrieved.
        \item List of variables to retrieve for the corresponding task. Do not include ``dose'', ``route'', ``frequency'', ``timestamp'', ``start time'', or similar fields in the variables list.
    \end{itemize}

    \item \textbf{Reasoning behind the retrieval tasks needed to answer the question.} This should be a step-by-step explanation of why each variable is important in the context of the question.

\end{enumerate}

\vspace{1em}

\textbf{Examples of retrieval tasks:}

\vspace{0.5em}

\begin{itemize}[leftmargin=1.5em]
    \item \textbf{Task:} Patient demographics \\
    \textbf{Variables:} [age, gender, race, ethnicity]

    \item \textbf{Task:} Hematology test results \\
    \textbf{Variables:} [WBC count, RBC count, Hemoglobin, Hematocrit, Platelet count]

    \item \textbf{Task:} Arterial blood gas (ABG) results \\
    \textbf{Variables:} [pH, pCO2, pO2, HCO3, SaO2, lactate]

    \item \textbf{Task:} Vital signs \\
    \textbf{Variables:} [heart rate, respiratory rate, blood pressure systolic, blood pressure diastolic, temperature, oxygen saturation]

    \item \textbf{Task:} Ventilator settings \\
    \textbf{Variables:} [FiO2, PEEP, tidal volume, respiratory rate set, plateau pressure, mode of ventilation]

    \item \textbf{Task:} Renal function labs \\
    \textbf{Variables:} [creatinine, BUN, urine output, eGFR]

    \item \textbf{Task:} Sepsis-related labs \\
    \textbf{Variables:} [procalcitonin, CRP, lactate, WBC count, temperature, blood cultures]

    \item \textbf{Task:} Vasopressor administration \\
    \textbf{Variables:} [norepinephrine, epinephrine, vasopressin, dobutamine]

    \item \textbf{Task:} Sedation and analgesia medications \\
    \textbf{Variables:} [propofol, midazolam, fentanyl, dexmedetomidine]

    \item \textbf{Task:} ICU admission details \\
    \textbf{Variables:} [ICU admission time, admitting service, ICU type, primary diagnosis]
\end{itemize}

\vspace{1em}

\textbf{Output format:}

Generate a JSON object with the generated examples following this template:

\vspace{0.5em}

\begin{lstlisting}
{
  "examples": [
    {
      "question": "User question",
      "context": "Patient context for the question",
      "category": "List of tags for the question",
      "icu_topic": "Relevant ICU topic the question belongs to",
      "retrieval_tasks": [
        {
          "task": "Task to retrieve information from the database",
          "variables": "List of variables to retrieve for corresponding task"
        }
      ],
      "reasoning": "Reasoning behind the question and corresponding retrieval tasks."
    }
  ]
}
\end{lstlisting}

\vspace{1em}

\textbf{Few-shot examples:}

Here are some examples of questions in JSON format:

\vspace{0.5em}

\begin{verbatim}
{examples}
\end{verbatim}

\end{promptbox}

\begin{promptbox}{Prompt 2}
\footnotesize
\textbf{Prompt for mapping ICU-REACT clinical questions to high-level ICU topics.}

\vspace{0.5em}

You are an ICU attending physician. Your task is to categorize clinical questions and their patient context, if provided, into high-level ICU topics.

\vspace{1em}

\textbf{ICU topic list. Use these labels exactly:}

\vspace{0.5em}

\begin{enumerate}[leftmargin=1.5em]

    \item \textbf{Hemodynamic Instability / Shock}

    \textbf{Types:} Septic, hypovolemic, cardiogenic, obstructive, such as pulmonary embolism or tamponade.

    \textbf{Management:}
    \begin{itemize}[leftmargin=1.5em]
        \item Rapid fluid resuscitation and monitoring of response.
        \item Use of vasopressors, with norepinephrine being the most common.
        \item Inotropes for impaired cardiac output.
        \item Monitoring lactate, central venous pressure, and echocardiography.
    \end{itemize}

    \item \textbf{Respiratory Failure}

    \textbf{Types:} Hypoxemic, such as acute respiratory distress syndrome or pneumonia; and hypercapnic, such as chronic obstructive pulmonary disease or neuromuscular weakness.

    \textbf{Management:}
    \begin{itemize}[leftmargin=1.5em]
        \item Oxygen supplementation, high-flow nasal cannula, or non-invasive ventilation.
        \item Intubation and mechanical ventilation.
        \item Adjusting ventilator settings, including FiO\textsubscript{2}, PEEP, and tidal volume.
        \item Weaning and extubation planning.
    \end{itemize}

    \item \textbf{Sepsis and Severe Infections}

    \textbf{Management:}
    \begin{itemize}[leftmargin=1.5em]
        \item Early recognition using qSOFA or SOFA.
        \item Blood cultures and broad-spectrum antibiotics.
        \item Source control, such as line removal or abscess drainage.
        \item Hemodynamic support with fluids and vasopressors.
    \end{itemize}

    \item \textbf{Neurological Emergencies}

    \textbf{Scenarios:} Stroke, intracranial hemorrhage, traumatic brain injury, seizures, delirium, and coma.

    \textbf{Management:}
    \begin{itemize}[leftmargin=1.5em]
        \item Neurological checks, including GCS, RASS, and pupillary response.
        \item Intracranial pressure monitoring and osmotic therapy, such as mannitol or hypertonic saline.
        \item Airway protection in reduced consciousness.
        \item Sedation and delirium management.
    \end{itemize}

    \item \textbf{Renal Failure / Electrolyte Disorders}

    \textbf{Management:}
    \begin{itemize}[leftmargin=1.5em]
        \item Monitor urine output and creatinine.
        \item Correct electrolyte imbalances, such as hyperkalemia and hyponatremia.
        \item Initiate renal replacement therapy, including CRRT or dialysis.
        \item Optimize fluid balance.
    \end{itemize}

    \item \textbf{Cardiac Emergencies}

    \textbf{Scenarios:} Arrhythmias, myocardial infarction, and post-cardiac arrest care.

    \textbf{Management:}
    \begin{itemize}[leftmargin=1.5em]
        \item ACLS protocols for arrest.
        \item Antiarrhythmics, such as amiodarone or lidocaine.
        \item Defibrillation or pacing.
        \item Hemodynamic optimization after return of spontaneous circulation.
    \end{itemize}

    \item \textbf{Sedation, Pain, and Delirium Management}

    \textbf{Management:}
    \begin{itemize}[leftmargin=1.5em]
        \item Balance sedation to ensure comfort while preserving wakefulness.
        \item Sedatives, including propofol, dexmedetomidine, and midazolam, and analgesics, including opioids.
        \item Delirium screening using CAM-ICU.
        \item Early mobilization and sleep hygiene.
    \end{itemize}

    \item \textbf{Nutrition and Metabolic Support}

    \textbf{Management:}
    \begin{itemize}[leftmargin=1.5em]
        \item Enteral nutrition, such as nasogastric or post-pyloric feeding.
        \item Parenteral nutrition if enteral nutrition is not feasible.
        \item Glycemic control while avoiding hypoglycemia and hyperglycemia.
    \end{itemize}

    \item \textbf{Hematologic / Coagulation Issues}

    \textbf{Scenarios:} Bleeding, thrombosis, and anticoagulation management.

    \textbf{Management:}
    \begin{itemize}[leftmargin=1.5em]
        \item Transfusion support, including packed red blood cells, platelets, and plasma.
        \item Deep vein thrombosis prophylaxis, such as heparin or compression devices.
        \item Reversal of anticoagulants in bleeding.
    \end{itemize}

\end{enumerate}

\vspace{1em}

\textbf{Rules:}

\begin{itemize}[leftmargin=1.5em]
    \item Always choose exactly one primary topic and return it under the key \texttt{topic}.
    \item Optionally choose any number of additional, clearly relevant topics and return them under the key \texttt{secondary\_topics}.
    \item Both \texttt{topic} and every entry in \texttt{secondary\_topics} must be copied from the topic titles above verbatim. Case and punctuation may differ, but the text should match.
    \item If the question is very general but clearly fits one domain, pick the best single topic.
    \item If multiple domains are clearly involved, pick the most central one as \texttt{topic} and add the others to \texttt{secondary\_topics}.
\end{itemize}

\vspace{1em}

\textbf{Return strict JSON only, with this schema:}

\vspace{0.5em}

\begin{lstlisting}
{
  "topic": "<one of the 9 topic titles>",
  "secondary_topics": ["<other topic title>", "..."]
}
\end{lstlisting}

\vspace{1em}

\textbf{Patient context, which may be empty:}

\vspace{0.5em}

\begin{lstlisting}
{context}
\end{lstlisting}

\vspace{1em}

\textbf{Clinical question:}

\vspace{0.5em}

\begin{lstlisting}
{question}
\end{lstlisting}

\end{promptbox}

\begin{promptbox}{Prompt 3}
\footnotesize
\textbf{Prompt for classifying clinician feedback on ICU-REACT samples.}

\vspace{0.5em}

You are an expert ICU clinician-educator and annotation quality assurance analyst.

You will be shown one ICU-REACT sample and one clinician feedback comment about that sample.

Your task is to classify the clinician feedback into exactly one of the following categories:

\vspace{0.5em}

\begin{enumerate}[leftmargin=1.5em]

    \item \textbf{\texttt{question\_context\_improvement}}

    Use this when the feedback is mainly about improving the patient context, the framing of the question, missing clinical state details, time anchoring, trajectory, care supports, realism, consistency, or making the decision fork clearer.

    \item \textbf{\texttt{variable\_selection}}

    Use this when the feedback is mainly about which variables should or should not be retrieved, missing key discriminating variables, irrelevant variables, over-selection, under-selection, or the need for different data elements.

    \item \textbf{\texttt{reasoning\_improvement}}

    Use this when the feedback is mainly about how the rationale or reasoning should be written, what logic should be explained, how variables should be connected to the decision, missing interpretation, weak justification, or reasoning structure.

    \item \textbf{\texttt{unclassified}}

    Use this when the feedback is not clearly about any of the above categories, is too vague, or does not contain actionable critique.

\end{enumerate}

\vspace{1em}

Return only valid JSON with exactly these keys:

\vspace{0.5em}

\begin{lstlisting}
{
  "label": "question_context_improvement" | "variable_selection" | "reasoning_improvement" | "unclassified",
  "confidence": 0.0,
  "rationale": "brief explanation"
}
\end{lstlisting}

\vspace{1em}

\textbf{Rules:}

\begin{itemize}[leftmargin=1.5em]
    \item Choose exactly one label.
    \item Do not output markdown.
    \item Confidence must be between 0 and 1.
    \item Base your decision primarily on the clinician feedback, using the sample only as context.
\end{itemize}

\vspace{1em}

\textbf{Classify the following clinician feedback.}

\vspace{0.5em}

\begin{lstlisting}
sample_id: {sample_id}
reviewer_label: {reviewer_label}
question: {question}
context: {context}
selected_variables: {selected_variables}
model_reasoning_text: {model_reasoning_text}
clinician_feedback:
{clinician_feedback}

Return only the JSON object.
\end{lstlisting}

\end{promptbox}

\begin{promptbox}{Prompt 4}
\footnotesize
\textbf{Prompt for refining reasoning on ICU-REACT train seed samples using clinician feedback.}

\vspace{0.5em}

You are an expert ICU clinician-educator improving ICU-REACT reasoning samples.

You will be given:

\begin{itemize}[leftmargin=1.5em]
    \item the current clinical question,
    \item the current patient context,
    \item the current reasoning text,
    \item clinician feedback comments that were already classified as \texttt{REASONING IMPROVEMENT} feedback.
\end{itemize}

Your task is to improve only the reasoning.

\vspace{1em}

\textbf{Core objective:}

Turn incomplete, unfocused, weakly justified, or clinically imprecise reasoning into stronger clinician-style reasoning that better uses the available context, question, and clinician feedback.

\vspace{1em}

\textbf{Important principles:}

\begin{enumerate}[leftmargin=1.5em]
    \item Preserve the original clinical intent unless the clinician feedback clearly indicates the reasoning is framed incorrectly.
    \item Do not modify the question.
    \item Do not modify the context.
    \item Improve the reasoning so it better explains:
    \begin{itemize}[leftmargin=1.5em]
        \item why the available clinical information matters for the decision or interpretation,
        \item how it relates to the specific clinical question,
        \item what trajectories, contraindications, alternative explanations, or missing discriminators should matter,
        \item and how a chart-reviewing clinician would think through the decision.
    \end{itemize}
    \item The reasoning should read like one coherent clinician-style explanatory paragraph, not bullets or JSON-like fragments.
    \item Do not invent unsupported facts, lab values, vitals, events, treatments, imaging findings, or outcomes that are not present in the sample or directly justified by the context or question.
    \item Do not simply restate the context; explain its clinical relevance.
    \item If the clinician feedback clearly implies more than one clinically distinct interpretation of how the reasoning should be strengthened, you may create multiple refinement paths.
    \item Prefer a small number of high-quality refinement paths over many weak ones.
\end{enumerate}

\vspace{1em}

\textbf{When to create multiple refinement paths:}

\begin{itemize}[leftmargin=1.5em]
    \item Create multiple paths only if the clinician feedback clearly supports distinct reasoning framings that would materially change the explanation.
    \item For example, if the feedback indicates the reasoning should differ depending on whether the main concern is severity assessment, treatment response, contraindication screening, or distinguishing competing etiologies, separate these into distinct reasoning paths only if they truly represent different coherent reasoning directions.
    \item If the feedback supports only one refinement, return exactly one path.
\end{itemize}

\vspace{1em}

\textbf{Critical rule:}

\begin{itemize}[leftmargin=1.5em]
    \item Never create multiple paths that are only superficial rewrites of the same reasoning.
    \item Each refinement path must represent a meaningfully distinct clinician-style reasoning emphasis or framing.
    \item If the feedback does not clearly support multiple refinements, return a single path.
\end{itemize}

\vspace{1em}

\textbf{What makes a strong reasoning refinement:}

\begin{itemize}[leftmargin=1.5em]
    \item The reasoning explicitly links the context and question to the clinical decision.
    \item The reasoning explains why the important details matter clinically.
    \item The reasoning reflects prioritization, trajectory, safety, alternatives, and decision relevance.
    \item The reasoning is concise but substantive.
    \item The reasoning clearly reflects the clinician feedback.
    \item The reasoning sounds like real ICU chart-review reasoning rather than generic commentary.
\end{itemize}

\vspace{1em}

\textbf{Current question:}

\vspace{0.5em}

\begin{lstlisting}
{question}
\end{lstlisting}

\vspace{1em}

\textbf{Current context:}

\vspace{0.5em}

\begin{lstlisting}
{context}
\end{lstlisting}

\vspace{1em}

\textbf{Current reasoning:}

\vspace{0.5em}

\begin{lstlisting}
{reasoning}
\end{lstlisting}

\vspace{1em}

\textbf{Clinician feedback:}

\vspace{0.5em}

\begin{lstlisting}
{feedback_text}
\end{lstlisting}

\vspace{1em}

\textbf{Return strict JSON only with exactly this schema:}

\vspace{0.5em}

\begin{lstlisting}
{
  "explanation": "concise clinician-style explanation of what is missing or weak in the current reasoning, written as a clean synthesis of the clinician feedback",
  "variant_strategy": "single_variant" | "multi_variant",
  "refinement_paths": [
    {
      "path_id": 1,
      "scenario_label": "short label for this reasoning refinement path",
      "clinical_rationale": "brief clinician-style rationale for why this is a distinct reasoning direction or emphasis",
      "improved_reasoning": "improved clinician-style reasoning paragraph",
      "specific_feedback_addressed": ["specific issue 1", "specific issue 2"]
    }
  ],
  "audit": {
    "main_issues_addressed": ["issue 1", "issue 2"],
    "preserved_original_intent": true,
    "used_feedback_count": 0
  }
}
\end{lstlisting}

\vspace{1em}

\textbf{Rules:}

\begin{itemize}[leftmargin=1.5em]
    \item Output valid JSON only.
    \item Do not wrap in markdown fences.
    \item \texttt{explanation} should be concise, specific, and faithful to the clinician feedback.
    \item \texttt{explanation} should read like brief clinician reasoning, not robotic meta-commentary.
    \item \texttt{explanation} should state why the current reasoning is insufficient for the decision, not just that it is ``too vague''.
    \item Do not include information in \texttt{explanation} that was not present in the clinician feedback.
    \item \texttt{refinement\_paths} must contain at least one item.
    \item Each \texttt{clinical\_rationale} and \texttt{improved\_reasoning} must be a non-empty string.
    \item \texttt{improved\_reasoning} must remain aligned with the provided question and context.
    \item \texttt{improved\_reasoning} must be one coherent paragraph in natural language.
    \item \texttt{improved\_reasoning} must not invent unsupported patient details.
    \item If feedback is weak or partial, make only minimal justified edits and return a single path.
    \item Usually return one to four refinement paths maximum.
\end{itemize}

\end{promptbox}

\begin{promptbox}{Prompt 5}
\footnotesize
\textbf{Prompt for selecting relevant variable-taxonomy sections using clinician feedback on ICU-REACT test set.}

\vspace{0.5em}

You are an ICU clinician and annotation reviewer. Your job in this step is only to decide which parts of the variable taxonomy are needed to:

\begin{enumerate}[leftmargin=1.5em]
    \item cover the original approved variables, and
    \item extract any additional variables implied or explicitly requested in the expert feedback.
\end{enumerate}

You will be given:

\begin{itemize}[leftmargin=1.5em]
    \item original variables that were already approved,
    \item expert feedback describing what clinicians want added or covered,
    \item a taxonomy index containing categories and their subcategories, but no variables yet.
\end{itemize}

\vspace{1em}

\textbf{Task:}

\begin{itemize}[leftmargin=1.5em]
    \item Select only the categories and subcategories that are relevant to:
    \begin{enumerate}[leftmargin=1.5em]
        \item mapping the original variables into the taxonomy, and
        \item capturing clinician-requested variables from expert feedback.
    \end{enumerate}
    \item Be conservative: if a variable or request could plausibly belong to a section, include it.
    \item Do not select irrelevant sections.
\end{itemize}

\vspace{1em}

\textbf{Original variables approved by at least one annotator:}

\vspace{0.5em}

\begin{lstlisting}
{variables_str}
\end{lstlisting}

\vspace{1em}

\textbf{Expert feedback:}

\vspace{0.5em}

\begin{lstlisting}
{feedback_text}
\end{lstlisting}

\vspace{1em}

\textbf{Taxonomy index. Choose from here:}

\vspace{0.5em}

\begin{lstlisting}
{taxonomy_index}
\end{lstlisting}

\vspace{1em}

\textbf{Return only the following JSON, with no text outside JSON:}

\vspace{0.5em}

\begin{lstlisting}
{
  "selected_sections": [
    {
      "category": "CategoryName",
      "subcategories": ["SubcatA", "SubcatB"]
    }
  ]
}
\end{lstlisting}

\vspace{1em}

\textbf{Constraints:}

\begin{itemize}[leftmargin=1.5em]
    \item Category and subcategory names must match the taxonomy index exactly.
    \item Only include subcategories that belong to the specified category.
    \item If nothing applies, which should be rare, return:
\end{itemize}

\vspace{0.5em}

\begin{lstlisting}
{
  "selected_sections": []
}
\end{lstlisting}

\end{promptbox}

\begin{promptbox}{Prompt 6}
\footnotesize
\textbf{Prompt for selecting refined variables from chosen taxonomy sections on ICU-REACT test set.}

\vspace{0.5em}

You are an ICU clinician and annotation reviewer. Your job in this step is to map the original approved variables to the taxonomy variables provided, when needed, and to select any additional variables requested by expert feedback, constrained to variables by section.

You will be given:

\begin{itemize}[leftmargin=1.5em]
    \item original selected variables that were already approved,
    \item expert feedback,
    \item variables by section: a list of variables grouped under category and subcategory.
\end{itemize}

The variables by section list is already filtered to only the sections selected in Step A.

\vspace{1em}

\textbf{Protocol:}

\begin{enumerate}[leftmargin=1.5em]

    \item \textbf{Map original variables to taxonomy, if needed.}

    \begin{itemize}[leftmargin=1.5em]
        \item For each item in original variables, try to find an exact match in variables by section.
        \item If an original variable does not appear exactly, choose the closest proxy from variables by section and record it in \texttt{original\_proxy\_mapping}.
        \item The mapped original set is called \texttt{original\_mapped}, using exact matches and/or proxies.
    \end{itemize}

    \item \textbf{Extract clinician requests.}

    \begin{itemize}[leftmargin=1.5em]
        \item From expert feedback, list the explicit variables or concepts clinicians are asking for using short clinical labels.
    \end{itemize}

    \item \textbf{Add needed variables from the provided section lists.}

    \begin{itemize}[leftmargin=1.5em]
        \item Choose variables only from variables by section and match names exactly.
        \item If a requested concept lacks an exact match, choose the closest proxy from variables by section and record it in \texttt{feedback\_proxy\_mapping}.
        \item Be conservative and clinically focused; avoid redundancy and near-duplicates.
    \end{itemize}

    \item \textbf{Build outputs.}

    \begin{itemize}[leftmargin=1.5em]
        \item \texttt{added\_from\_feedback}: items added for feedback that were not in \texttt{original\_mapped}.
        \item \texttt{refined\_variables}: \texttt{original\_mapped} $\cup$ \texttt{added\_from\_feedback}, deduplicated.
    \end{itemize}

    \item \textbf{Coverage check.}

    \begin{itemize}[leftmargin=1.5em]
        \item \texttt{all\_original\_mapped} is true only if every original variable is covered either by an exact match in \texttt{original\_mapped} or via \texttt{original\_proxy\_mapping}.
        \item \texttt{all\_requested\_included} is true only if every requested concept is covered either by an exact variable in \texttt{refined\_variables} or via \texttt{feedback\_proxy\_mapping}.
        \item \texttt{missing\_original}: original variables not covered, empty if none.
        \item \texttt{missing\_requested}: requested concepts not covered, empty if none.
    \end{itemize}

\end{enumerate}

\vspace{1em}

\textbf{Original variables selected, approved by at least one annotator:}

\vspace{0.5em}

\begin{lstlisting}
{variables_str}
\end{lstlisting}

\vspace{1em}

\textbf{Expert feedback:}

\vspace{0.5em}

\begin{lstlisting}
{feedback_text}
\end{lstlisting}

\vspace{1em}

\textbf{Variables by section. Select only from these and match names exactly:}

\vspace{0.5em}

\begin{lstlisting}
{variables_by_section}
\end{lstlisting}

\vspace{1em}

\textbf{Return only the following JSON, with no text outside JSON:}

\vspace{0.5em}

\begin{lstlisting}
{
  "audit": {
    "requested_from_feedback": ["..."],
    "original_selected": ["..."],
    "coverage_check": {
      "all_original_mapped": true or false,
      "missing_original": ["..."],
      "all_requested_included": true or false,
      "missing_requested": ["..."]
    },
    "original_proxy_mapping": [
      {"concept": "original variable label", "proxy": "ExactVariableName"}
    ],
    "feedback_proxy_mapping": [
      {"concept": "requested label", "proxy": "ExactVariableName"}
    ],
    "omitted_with_reason": [
      {"concept": "requested label", "reason": "brief clinical justification"}
    ]
  },
  "original_mapped": ["Var1", "Var2", "..."],
  "added_from_feedback": ["VarA", "VarB"],
  "refined_variables": ["Var1", "Var2", "..."]
}
\end{lstlisting}

\vspace{1em}

\textbf{Constraints:}

\begin{itemize}[leftmargin=1.5em]
    \item Every item in \texttt{original\_mapped}, \texttt{added\_from\_feedback}, and \texttt{refined\_variables} must be drawn exactly from variables by section or included via the corresponding proxy mappings.
    \item Do not invent new variables or rename variables.
    \item If nothing new is required from feedback, return \texttt{added\_from\_feedback}: \texttt{[]} and keep \texttt{refined\_variables} equal to \texttt{original\_mapped}.
\end{itemize}

\end{promptbox}

\begin{promptbox}{Prompt 7}
\footnotesize
\textbf{Prompt for generating refined reasoning from clinician-selected variables and feedback on ICU-REACT test set.}

\vspace{0.5em}

You are an ICU clinician. Refine the reasoning paragraph for the decision-making question using the provided variables grouped by category and sub-category. Do not change the variable list. Output only valid JSON.

\vspace{1em}

\textbf{Goal:}

\begin{itemize}[leftmargin=1.5em]
    \item Produce a single natural-language paragraph explaining how the provided variables support answering the question.
    \item Align the reasoning with the expert feedback.
\end{itemize}

\vspace{1em}

\textbf{Introduction:}

Start the paragraph with one introductory sentence that:

\begin{enumerate}[leftmargin=1.5em]
    \item briefly re-states the patient context and the decision at hand, and
    \item provides a brief overview of the key types of data that need to be reviewed, such as laboratory results, physiology and vital signs, diagnoses, imaging, medications, or fluid balance, without listing every variable.
\end{enumerate}

\vspace{1em}

\textbf{Length and style requirements:}

\begin{itemize}[leftmargin=1.5em]
    \item Write eight to twelve sentences total, including the introductory sentence.
    \item Target approximately 220--350 words.
    \item Keep the writing fluent and clinical.
    \item Avoid repetitive phrasing such as ``X matters because ...''.
    \item Do not sound like a taxonomy dump or checklist.
\end{itemize}

\vspace{1em}

\textbf{How to use the taxonomy grouping:}

\begin{itemize}[leftmargin=1.5em]

    \item Use categories and sub-categories as light structural cues.

    \item Do not write sentences of the form:

\end{itemize}

\vspace{0.5em}

\begin{lstlisting}
<Category> <Sub-category> helps with ...
<Sub-category> helps ...
\end{lstlisting}

\vspace{0.5em}

\textbf{Bad example to avoid:}

\vspace{0.5em}

\begin{lstlisting}
Laboratory Measurements Comprehensive Metabolic Panel helps with ...
\end{lstlisting}

\vspace{0.5em}

Instead, follow this pattern:

\begin{enumerate}[leftmargin=1.5em]

    \item When first introducing a category, write one sentence explaining why that category matters for the decision. This should be a general rationale and should not yet list variables.

    Examples:

\begin{lstlisting}
Laboratory Measurements are essential to quantify severity and
immediacy of physiologic derangements.

Physiology helps assess stability, trajectory,
and response to interventions.
\end{lstlisting}

    \item For each sub-category within that category, write one subsequent sentence that:

    \begin{itemize}[leftmargin=1.5em]
        \item starts with the sub-category name, or a natural transition referencing it,
        \item lists the variables as a compact cluster using comma-separated formatting,
        \item provides one concise rationale for the entire cluster.
    \end{itemize}

    Example:

\begin{lstlisting}
Comprehensive Metabolic Panel (BUN, creatinine,
potassium, bicarbonate) frames renal clearance
and immediate electrolyte and acid-base threats.
\end{lstlisting}

\end{enumerate}

\vspace{1em}

\textbf{Additional writing guidance:}

\begin{itemize}[leftmargin=1.5em]
    \item Vary transitions across categories.
    \item Avoid repeatedly starting sentences with the same structure.
    \item Examples include:
    \begin{itemize}
        \item Laboratory Measurements...
        \item Physiology...
        \item On the diagnosis side...
        \item Imaging...
        \item Medication context...
        \item Fluid balance...
    \end{itemize}
    \item Do not stack category and sub-category names back-to-back in the same clause unless grammatically necessary.
    \item When referencing categories or sub-categories, use their full names exactly as provided in VARIABLES.
    \item Match spelling and capitalization exactly.
    \item Avoid abbreviations unless the full name is written first.
    \item Only single out an individual variable with extra detail if it is central to the decision, and limit this to at most one or two variables.
\end{itemize}

\vspace{1em}

\textbf{Coverage requirement:}

Every variable must be addressed by appearing explicitly somewhere in the paragraph:

\begin{itemize}[leftmargin=1.5em]
    \item either as part of its sub-category cluster,
    \item or as an individually highlighted variable.
\end{itemize}

If the variable list is long, coverage should primarily occur through clustered mentions rather than repeated references.

\vspace{1em}

\textbf{Hard rules:}

\begin{itemize}[leftmargin=1.5em]
    \item Do not invent patient values.
    \item Do not make a final recommendation.
    \item Do not include meta-commentary about the task.
    \item If the current reasoning is incomplete or inconsistent, rewrite it into a complete and internally consistent paragraph suitable for fine-tuning.
\end{itemize}

\vspace{1em}

\textbf{Question:}

\begin{lstlisting}
{question}
\end{lstlisting}

\vspace{1em}

\textbf{Context:}

\begin{lstlisting}
{context}
\end{lstlisting}

\vspace{1em}

\textbf{Variables (grouped by Category/Sub-category; do not modify):}

\begin{lstlisting}
{refined_variables}
\end{lstlisting}

\vspace{1em}

\textbf{Current reasoning:}

\begin{lstlisting}
{reasoning_text}
\end{lstlisting}

\vspace{1em}

\textbf{Expert feedback:}

\begin{lstlisting}
{feedback_text}
\end{lstlisting}

\vspace{1em}

\textbf{Return only the following JSON, with no text outside JSON:}

\vspace{0.5em}

\begin{lstlisting}
{
  "audit": {
    "coverage_check": {
      "all_variables_addressed": true or false,
      "missing_in_reasoning": ["..."]
    }
  },
  "refined_reasoning":
    "One paragraph (8-12 sentences) starting with an
     introductory sentence that re-states the context
     and decision and summarizes what data must be
     reviewed, then explains how the grouped variables
     support the decision, aligned with expert feedback."
}
\end{lstlisting}

\vspace{1em}

\textbf{Constraints:}

\begin{itemize}[leftmargin=1.5em]
    \item Do not add, remove, rename, or reorder variables.
    \item Use exactly the variables provided in VARIABLES.
\end{itemize}

\end{promptbox}

\begin{promptbox}{Prompt 8}
\footnotesize

\textbf{Prompt for creating case-specific LLM-judge rubrics on ICU-REACT test set.}

\vspace{0.5em}

You are helping build an ICU decision-making benchmark called ICU-REACT.

For each single evaluation sample, you will design an explicit rubric that can be used by another LLM-judge to grade candidate responses.

\vspace{1em}

\textbf{Critical constraint:}

The LLM-judge will only see:

\begin{enumerate}[leftmargin=1.5em]
    \item the patient context,
    \item the decision-making question,
    \item the candidate model response,
    \item the rubric JSON you write.
\end{enumerate}

The judge will not have access to the reference answer. Therefore, every rubric criterion must be self-contained, observable from the candidate response, and grounded in the patient context and question. Do not write criteria that require access to hidden ground truth wording.

\vspace{1em}

\textbf{Each sample consists of:}

\begin{itemize}[leftmargin=1.5em]
    \item Patient context, representing an ICU scenario.
    \item A decision-making question.
    \item A reference answer written by an expert, visible to you only, which indicates:
    \begin{itemize}[leftmargin=1.5em]
        \item which categories, subcategories, and variables are relevant,
        \item why they are relevant,
        \item how they influence the decision,
        \item relevant safety considerations.
    \end{itemize}
\end{itemize}

Your task is to read the sample and create a case-specific rubric that evaluates the quality and correctness of the response's reasoning.

\vspace{1em}

\textbf{Important goal:}

The rubric should primarily evaluate whether the candidate response gives the correct clinical reason for including categories, subcategories, and variables.

This means:

\begin{itemize}[leftmargin=1.5em]
    \item Do not reward a response simply for mentioning the right variable.
    \item Instead, reward whether the response correctly explains:
    \begin{itemize}[leftmargin=1.5em]
        \item why that variable or category matters,
        \item what clinical role it plays,
        \item and how it influences the decision.
    \end{itemize}
    \item A response should score well only if it provides correct, decision-relevant, mechanistic reasoning.
\end{itemize}

\vspace{1em}

\textbf{Target behavior of the rubric:}

\begin{itemize}[leftmargin=1.5em]
    \item A strong response explains the correct reason that each important factor matters.
    \item A weak response may mention relevant factors but give vague, generic, incomplete, or incorrect reasons.
    \item A response should not receive high scores for broad variable listing alone.
    \item A concise response with correct and well-explained reasoning should outperform a longer but shallow response.
    \item The rubric should reward explanation quality, not mention count.
\end{itemize}

\vspace{1em}

\textbf{Task-faithfulness requirement:}

\begin{itemize}[leftmargin=1.5em]
    \item Candidate responses should not make a specific final recommendation or decision.
    \item They should focus on explaining the decision factors: categories, subcategories, variables, and how each would influence the decision.
    \item Therefore, the rubric must evaluate faithfulness to this format: explaining decision factors without recommending a final action.
\end{itemize}

\vspace{1em}

\textbf{Example ICU-REACT case and rubric, for guidance only.}

Do not reuse the same wording. Adapt the style to the new sample.

\vspace{0.5em}

\textbf{Example context:}

\begin{lstlisting}
60-year-old male with acute kidney injury and electrolyte disturbances in the ICU.
\end{lstlisting}

\vspace{0.5em}

\textbf{Example question:}

\begin{lstlisting}
Should we initiate dialysis for this patient with severe electrolyte imbalances?
\end{lstlisting}

\vspace{0.5em}

\textbf{Example reference answer, visible to you only and not to the judge:}

\begin{lstlisting}
In this 60-year-old man in the ICU with acute kidney injury and severe electrolyte disturbances, the decision to initiate dialysis depends on integrating laboratory derangements, physiologic stability and fluid status, cardiac/imaging evidence of electrolyte toxicity or overload, neurologic/physical exam findings, and current medication exposures that may worsen kidney function or electrolytes. Medications are critical to interpret because recent or ongoing therapies can precipitate acute kidney injury, aggravate electrolyte abnormalities, and determine whether medical management has been maximized or is limited by toxicity. Fluids (Albumin 25%, Albumin 5%, Lactated Ringers, Normal Saline) inform the resuscitation strategy and whether administered volume or oncotic support could be contributing to or mitigating overload while kidney function is impaired. Diuretics (Bumetanide, Furosemide, Metolazone, Spironolactone) clarify attempts to manage volume and potassium balance, and lack of response despite these agents supports concern for refractory overload or electrolyte control. Antibiotics (Cefepime, Gentamicin, Levofloxacin, Meropenem, Piperacillin-Tazobactam, Vancomycin) are particularly relevant as potential nephrotoxins or contributors to acute kidney injury and may necessitate dialysis for clearance or to safely continue therapy. Antihypertensives (Enalaprilat, Nitroprusside) provide context for renal perfusion and hemodynamic tolerance of ultrafiltration in a patient with labile pressures. Laboratory Measurements are essential to quantify the severity and immediacy of physiologic derangements that dialysis can rapidly correct when conservative therapy is insufficient. Comprehensive Metabolic Panel (BUN, Calcium (total), Chloride, Creatinine, Magnesium, Phosphorus, Potassium, Sodium) frames the degree of renal failure and the urgency of threats such as potassium elevation or progressive azotemia, while also characterizing concurrent abnormalities including hypomagnesemia and disordered calcium/phosphate handling. Physiology helps assess stability, trajectory, and end-organ impact of electrolyte and volume disturbances to judge whether the patient is tolerating ongoing medical management. Vitals (Blood pressure diastolic, Blood pressure systolic, Heart rate, Mean arterial pressure, Oxygen saturation, Respiratory rate) highlight hemodynamic adequacy and respiratory compromise, with declining Oxygen saturation or rising Respiratory rate raising concern for fluid-overload physiology that may push toward dialysis-based ultrafiltration. Fluidics (Fluid balance, Fluid intake, Insensible losses) and Excretion (Hourly urine output, Urine output) together determine whether the patient is accumulating volume despite therapy and whether oliguria suggests limited capacity to correct electrolytes without extracorporeal support. Imaging provides objective evidence of cardiopulmonary consequences and cardiac electrical instability that can make electrolyte derangements immediately dangerous. EKG (EKG Findings, EKG Rhythm) identifies arrhythmias or electrocardiographic manifestations consistent with severe potassium or magnesium disturbances, while Basic Radiography (Pulmonary edema (X-ray)) supports clinically significant overload. Scores and Assessments help determine systemic impact and urgency, as Neuro-Cognitive Assessments (Glasgow Coma Scale) can signal encephalopathy from uremia or electrolyte shifts and Physical Exam Findings (Peripheral edema) corroborate volume overload. Diagnosis anchors interpretation by confirming Renal (acute kidney injury) as the substrate for impaired clearance and highlighting Metabolic (Hypomagnesemia) as a documented abnormality that, if severe or refractory, may accompany other dialysis-relevant electrolyte threats.
\end{lstlisting}

\vspace{0.5em}

\textbf{Example rubric. JSON shape to imitate:}

\begin{lstlisting}
{
  "rubric": [
    {
      "id": "R1",
      "dimension": "task_faithfulness",
      "description": "Evaluates whether the response stays focused on explaining decision factors and their influence on the dialysis decision, without making a final recommendation to initiate or withhold dialysis.",
      "weight": 5,
      "type": "0_1_2_3_scale"
    },
    {
      "id": "R2",
      "dimension": "reasoning_correctness",
      "description": "Evaluates whether the response correctly explains why chemistry and renal laboratory abnormalities are relevant: they define the severity of kidney dysfunction and electrolyte derangement, and help determine whether abnormalities are severe, refractory, or dangerous enough that dialysis becomes more relevant.",
      "weight": 4,
      "type": "0_1_2_3_scale"
    },
    {
      "id": "R3",
      "dimension": "reasoning_correctness",
      "description": "Evaluates whether the response correctly explains why diuretics are relevant to the dialysis decision: they reflect attempts to manage volume status and potassium balance, and lack of response despite these agents increases concern for refractory volume overload or inability to control electrolytes with medical therapy alone.",
      "weight": 4,
      "type": "0_1_2_3_scale"
    },
    {
      "id": "R4",
      "dimension": "reasoning_correctness",
      "description": "Evaluates whether the response correctly explains why fluid administration is relevant: fluids such as albumin or crystalloids inform the resuscitation strategy and help determine whether administered volume may be contributing to, worsening, or attempting to correct hemodynamic or volume problems while renal clearance is impaired.",
      "weight": 3,
      "type": "0_1_2_3_scale"
    },
    {
      "id": "R5",
      "dimension": "reasoning_correctness",
      "description": "Evaluates whether the response correctly explains why nephrotoxic or renally relevant medications, especially antibiotics, matter: they may worsen acute kidney injury, aggravate electrolyte disturbances, or influence whether dialysis is needed to support clearance or permit ongoing treatment safely.",
      "weight": 4,
      "type": "0_1_2_3_scale"
    },
    {
      "id": "R6",
      "dimension": "reasoning_correctness",
      "description": "Evaluates whether the response correctly explains why urine output and fluid balance matter: reduced urine output or ongoing positive balance suggests limited ability to clear volume and electrolytes without extracorporeal support, increasing concern for refractory overload or failed medical management.",
      "weight": 4,
      "type": "0_1_2_3_scale"
    },
    {
      "id": "R7",
      "dimension": "reasoning_correctness",
      "description": "Evaluates whether the response correctly explains why vitals and physiologic status matter: blood pressure, MAP, oxygenation, respiratory rate, and related instability help assess both the consequences of overload/electrolyte derangement and the patient's likely tolerance of dialysis or ultrafiltration.",
      "weight": 3,
      "type": "0_1_2_3_scale"
    },
    {
      "id": "R8",
      "dimension": "reasoning_correctness",
      "description": "Evaluates whether the response correctly explains why EKG findings are relevant: they provide objective evidence of cardiac electrical instability from electrolyte abnormalities, especially dangerous potassium or magnesium disturbances, which increases concern for urgent dialysis-relevant complications.",
      "weight": 4,
      "type": "0_1_2_3_scale"
    },
    {
      "id": "R9",
      "dimension": "reasoning_correctness",
      "description": "Evaluates whether the response correctly explains why imaging findings such as pulmonary edema matter: they provide objective evidence of clinically significant volume overload, supporting concern that fluid cannot be adequately managed without dialysis-based ultrafiltration.",
      "weight": 3,
      "type": "0_1_2_3_scale"
    },
    {
      "id": "R10",
      "dimension": "reasoning_correctness",
      "description": "Evaluates whether the response correctly explains why neurologic or exam findings such as reduced GCS, encephalopathy, or peripheral edema matter: they reflect systemic consequences of uremia, electrolyte disturbance, or volume overload that increase concern for dialysis-relevant deterioration.",
      "weight": 3,
      "type": "0_1_2_3_scale"
    },
    {
      "id": "R11",
      "dimension": "reasoning_synthesis",
      "description": "Evaluates whether the response correctly synthesizes multiple categories into a coherent clinical explanation, showing how labs, medications, fluids, output, vitals, imaging, and exam findings jointly shape the dialysis decision factors rather than functioning as isolated observations.",
      "weight": 5,
      "type": "0_1_2_3_scale"
    },
    {
      "id": "R12",
      "dimension": "safety",
      "description": "Evaluates whether the response's explanatory reasoning is clinically safe: it should not ignore major dangerous factors, should not assign clearly incorrect meanings to variables, and should not rely heavily on weak or irrelevant rationale.",
      "weight": 5,
      "type": "0_1_2_3_scale"
    }
  ]
}
\end{lstlisting}

This example is only to show the level of specificity and style. For the new sample below, you must design a new rubric with different, case-appropriate criteria.

\vspace{1em}

\textbf{Now design a rubric for the following sample.}

\vspace{0.5em}

\textbf{Sample ID:}

\begin{lstlisting}
{sample_id}
\end{lstlisting}

\vspace{0.5em}

\textbf{Patient context:}

\begin{lstlisting}
{context}
\end{lstlisting}

\vspace{0.5em}

\textbf{Decision-making question:}

\begin{lstlisting}
{question}
\end{lstlisting}

\vspace{0.5em}

\textbf{Reference answer, expert ground truth, visible to you only and not to the judge:}

\begin{lstlisting}
{reference_answer}
\end{lstlisting}

\vspace{1em}

\textbf{General guidelines for your rubric:}

\begin{enumerate}[leftmargin=1.5em]

    \item \textbf{The rubric must be specific to this sample, not generic.}

    \item \textbf{Judge-visibility constraint.}

    The judge will not see the reference answer. Do not write criteria like ``mentions the variables in the reference answer.'' Each criterion must stand on its own and be checkable from the candidate response, using only what is clinically justified by the context and question.

    \item \textbf{Primary goal: reasoning correctness.}

    The rubric must mainly evaluate whether the response explains the correct reason that a category, subcategory, or variable is relevant. The main target is not coverage by itself, but correctness of explanation. Prefer criteria of the form:

\begin{lstlisting}
factor -> why it matters -> how it influences the decision
\end{lstlisting}

    Good criteria should evaluate whether the response captures the clinical role of the factor, the correct mechanism or interpretation, and the correct decision implication.

    \item \textbf{Do not over-reward mention count.}

    Do not create criteria that mainly reward listing many factors. Mentioning a factor should only help when the response correctly explains why it matters and how it affects the decision. A vague mention, boilerplate mention, or shallow mention should score low.

    \item \textbf{Use specific reasoning-correctness criteria.}

    Create criteria for the most important categories, subcategories, or variables from the reference answer. For each important factor, ask whether the response correctly explains why it is relevant, what it clarifies, what concern it raises or lowers, or how it changes interpretation of the decision.

    Example pattern:

\begin{lstlisting}
Evaluates whether the response correctly explains that [factor] matters because [clinical role], and that it influences the decision by [decision implication].
\end{lstlisting}

    \item \textbf{Use synthesis criteria in addition to local correctness criteria.}

    Include at least one criterion evaluating whether the response integrates multiple factors into a coherent clinical explanation rather than listing them as isolated facts. This criterion should reward cross-category reasoning and good synthesis.

    \item \textbf{Task faithfulness.}

    Include task-faithfulness criteria that evaluate whether the response remains in decision-factor analysis mode. The response should explain what factors matter and how they influence the decision. The response should not make a final recommendation such as initiating, withholding, or choosing a definitive action. Penalize recommendation-making or drifting into final action selection.

    \item \textbf{Safety.}

    Include safety-oriented evaluation of the explanatory reasoning. Safety here means the response should not:

    \begin{itemize}[leftmargin=1.5em]
        \item ignore major dangerous factors,
        \item assign clearly incorrect meanings to important variables,
        \item rely on obviously weak or irrelevant reasoning,
        \item or present clinically misleading interpretations.
    \end{itemize}

    Safety should be evaluated in terms of reasoning quality, not final management choice.

    \item \textbf{Criterion granularity.}

    Criteria should be atomic and checkable. Each criterion should evaluate one clear reasoning behavior. Avoid bundling multiple unrelated expectations into one criterion. However, it is acceptable for a criterion to cover a tightly connected cluster if those elements share the same clinical role and decision implication.

    \item \textbf{Number of criteria.}

    There is no fixed required number of criteria. Include as many criteria as needed to cover the important reasoning components of the sample well. Prefer completeness of important reasoning over forcing an arbitrary count. In most cases, the rubric will likely have roughly eight to fourteen criteria, but this is not a hard rule.

    \item \textbf{Weighting philosophy.}

    Reasoning correctness should dominate the rubric. High-value criteria should reward correct explanation of important factors. Synthesis, task faithfulness, and safety should also be represented. Do not overuse low-value binary criteria. Prefer giving more weight to clinically central reasoning elements.

    \item \textbf{Preferred criterion types.}

    Use \texttt{0\_1\_2\_3\_scale} for most reasoning-correctness criteria. This scale should reflect quality of explanation, for example:

    \begin{itemize}[leftmargin=1.5em]
        \item 0 = absent, incorrect, or clinically misleading.
        \item 1 = factor is mentioned but reasoning is vague, shallow, incomplete, or partly incorrect.
        \item 2 = reasoning is mostly correct but incomplete or insufficiently specific.
        \item 3 = reasoning is clearly correct, specific, and decision-relevant.
    \end{itemize}

    Use \texttt{0\_1\_2\_scale} when a simpler partial-credit scale is sufficient. Use \texttt{present\_absent} only for rare must-have anchors, not for most reasoning criteria.

    \item Do not invent new patient details.

    \item \textbf{Do not design the rubric around final recommendation correctness.}

    The response is not supposed to choose the final action. Therefore, do not reward or penalize based on whether the response recommends a specific final decision. Instead, evaluate whether the response correctly explains the decision factors.

\end{enumerate}

\vspace{1em}

\textbf{Allowed fields:}

For each criterion, output:

\begin{itemize}[leftmargin=1.5em]
    \item \texttt{id}: a short string like \texttt{R1}, \texttt{R2}, etc.
    \item \texttt{dimension}: one of:
    \begin{itemize}[leftmargin=1.5em]
        \item \texttt{reasoning\_correctness}
        \item \texttt{reasoning\_synthesis}
        \item \texttt{task\_faithfulness}
        \item \texttt{safety}
        \item \texttt{critical\_anchor}
    \end{itemize}
    \item \texttt{description}: concise and case-specific, written so a judge without the reference answer can apply it using only the context, question, and response.
    \item \texttt{weight}: integer points from 1 to 5.
    \item \texttt{type}: one of:
    \begin{itemize}[leftmargin=1.5em]
        \item \texttt{present\_absent}
        \item \texttt{0\_1\_2\_scale}
        \item \texttt{0\_1\_2\_3\_scale}
    \end{itemize}
\end{itemize}

\vspace{1em}

\textbf{Additional guidance on dimensions:}

\begin{itemize}[leftmargin=1.5em]
    \item \texttt{reasoning\_correctness}: use for criteria that evaluate whether the response correctly explains why a factor matters and how it influences the decision.
    \item \texttt{reasoning\_synthesis}: use for criteria that evaluate integration across multiple factors or domains.
    \item \texttt{task\_faithfulness}: use for criteria that evaluate staying in decision-factor explanation mode without making a final recommendation.
    \item \texttt{safety}: use for criteria that evaluate whether the explanatory reasoning is clinically safe.
    \item \texttt{critical\_anchor}: use sparingly for rare must-not-miss anchors.
\end{itemize}

\vspace{1em}

\textbf{Weighting guidance:}

\begin{itemize}[leftmargin=1.5em]
    \item \texttt{reasoning\_correctness}: typically 3 to 5.
    \item \texttt{reasoning\_synthesis}: typically 4 to 5.
    \item \texttt{task\_faithfulness}: typically 4 to 5.
    \item \texttt{safety}: typically 4 to 5.
    \item \texttt{critical\_anchor}: typically 1 to 3.
\end{itemize}

\vspace{1em}

\textbf{Output format:}

Return a single JSON object with the following structure:

\begin{lstlisting}
{
  "rubric": [
    {
      "id": "R1",
      "dimension": "reasoning_correctness",
      "description": "...",
      "weight": 4,
      "type": "0_1_2_3_scale"
    }
  ]
}
\end{lstlisting}

\vspace{1em}

\textbf{Rules:}

\begin{itemize}[leftmargin=1.5em]
    \item Do not include explanations outside of this JSON.
    \item Do not include comments, markdown, or trailing commas.
    \item Make sure the JSON is syntactically valid.
    \item Ensure every criterion is case-specific.
    \item Ensure the rubric emphasizes reasoning correctness over mere coverage.
\end{itemize}

\end{promptbox}

\subsection{Dataset augmentation prompts}\label{augment_prompts}

\begin{promptbox}{Prompt 9}
\footnotesize

\textbf{Prompt for generating ICU-REACT context--question pairs for dataset augmentation.}

\vspace{0.5em}

You are an ICU clinical authoring assistant. Your task is to generate \texttt{\{n\}} new context--question pairs to augment a dataset for instruction tuning.

\vspace{1em}

\textbf{General constraints, applied to every generated pair:}

\begin{itemize}[leftmargin=1.5em]
    \item Language: \texttt{\{language\}}.
    \item Do not include any specific numeric values for laboratory results, vital signs, ventilator settings, medication doses, or flows. Age is allowed.
    \item Keep each pair ICU-relevant, clinically plausible, and free of protected health information, including names, exact dates, addresses, and medical record numbers.
    \item Each pair must be distinct and should plausibly lead to different reasoning or variable selections during annotation.
    \item The context should be one to two sentences and include key clinical features, etiologies, or constraints.
    \item Age is allowed, but avoid numeric measurements.
    \item The context should follow this format:
\end{itemize}

\vspace{0.5em}

\begin{quote}
{\scriptsize
\texttt{[age][gender] patient with [past\_medical\_history] admitted to the ICU due to [surgery\_or\_current\_complication] is developing [critical\_condition]}
}
\end{quote}

\vspace{0.5em}

\begin{itemize}[leftmargin=1.5em]
    \item The question should be one sentence, action-oriented, and decision-focused.
    \item The question may focus on diagnostic, therapeutic, monitoring, escalation, or modality-choice decisions.
    \item Avoid wording that implies numeric levels. For example, say ``concern for oxygenation'' instead of ``high PEEP''.
    \item Each pair must be categorized under exactly one of the ICU topics listed below, based on the primary clinical issue and decision focus.
\end{itemize}

\vspace{1em}

When creating the context--question pairs, try to ground them in the ICU topics below. For each topic, ideas are suggested but may be adapted to the specific patient story.

\vspace{0.5em}

\begin{itemize}[leftmargin=1.5em]

    \item \textbf{Topic 1}

    \begin{itemize}[leftmargin=1.5em]
        \item \textbf{Example clinical scenarios:} \ldots
        \item \textbf{Typical ICU presentations to highlight:} \ldots
        \item \textbf{Decisions the question can target:} \ldots
        \item \textbf{Possible modifiers or constraints:} \ldots
    \end{itemize}

    \textbf{\ldots}

    \item \textbf{Topic n}

    \begin{itemize}[leftmargin=1.5em]
        \item \textbf{Example clinical scenarios:} \ldots
        \item \textbf{Typical ICU presentations to highlight:} \ldots
        \item \textbf{Decisions the question can target:} \ldots
        \item \textbf{Possible modifiers or constraints:} \ldots
    \end{itemize}

\end{itemize}

\vspace{1em}

\textbf{Few-shot examples for style and content guidance. Do not copy blindly:}

\vspace{0.5em}

\begin{lstlisting}
{few_shot_examples}
\end{lstlisting}

\vspace{1em}

\textbf{Output only a JSON object with this exact schema, with no extra keys and no prose outside JSON:}

\vspace{0.5em}

\begin{lstlisting}
{
  "output": [
    {
      "context": "Generated patient context here",
      "question": "Generated decision-making question here",
      "topic": "One of the ICU topics listed above"
    }
  ]
}
\end{lstlisting}

\end{promptbox}

\begin{promptbox}{Prompt 10}
\footnotesize

\textbf{Prompt for generating initial ICU-REACT reasoning from context--question pairs.}

\vspace{0.5em}

You are an ICU clinical authoring assistant. You will receive an ICU patient context and a decision-making question.

\vspace{1em}

\textbf{Your task:}

\begin{enumerate}[leftmargin=1.5em]
    \item Write one initial reasoning paragraph that explains what information would matter to answer the question.
    \item The reasoning should be clinically plausible, relevant, and coherent, but it does not need to be perfect.
    \item The reasoning should feel like an early draft that could later be improved by expert refinement.
\end{enumerate}

\vspace{1em}

\textbf{Initial reasoning guidelines:}

\begin{itemize}[leftmargin=1.5em]
    \item Language: English.
    \item Write exactly one paragraph.
    \item Do not output bullets, numbered lists, section headers, or JSON-like text inside the reasoning.
    \item Keep the reasoning decision-focused and clinically grounded.
    \item Emphasize what information, physiology, trajectories, risks, or alternative explanations matter.
    \item Avoid numeric specifics for laboratory results, vital signs, ventilator settings, medication doses, or flows.
    \item Age is allowed if already present in the context.
    \item Do not mention that the reasoning is incomplete or intentionally imperfect.
    \item Do not copy the question verbatim as the answer.
\end{itemize}

\vspace{1em}

\textbf{Few-shot examples for style guidance. Do not copy blindly:}

\vspace{0.5em}

\begin{lstlisting}
{few_shot_examples}
\end{lstlisting}

\vspace{1em}

\textbf{Input:}

\vspace{0.5em}

\textbf{Context:}

\begin{lstlisting}
{context}
\end{lstlisting}

\vspace{0.5em}

\textbf{Question:}

\begin{lstlisting}
{question}
\end{lstlisting}

\vspace{1em}

\textbf{Output only a JSON object with this exact schema, with no extra keys and no prose outside JSON:}

\vspace{0.5em}

\begin{lstlisting}
{
  "output": [
    {
      "question": "The input question here",
      "context": "The input context here",
      "initial_reasoning": "Initial clinical reasoning here"
    }
  ]
}
\end{lstlisting}

\end{promptbox}

\begin{promptbox}{Prompt 11}
\footnotesize

\textbf{Prompt for generating refined ICU-REACT reasoning from an initial reasoning draft.}

\vspace{0.5em}

You are an ICU clinical educator. You will receive:

\begin{itemize}[leftmargin=1.5em]
    \item an ICU patient context,
    \item a decision-making question,
    \item an initial reasoning draft.
\end{itemize}

\vspace{1em}

\textbf{Your task:}

\begin{enumerate}[leftmargin=1.5em]
    \item Analyze the initial reasoning and identify how it should be improved.
    \item Write a concise explanation of the main refinement opportunity.
    \item Write one improved reasoning paragraph.
\end{enumerate}

\vspace{1em}

\textbf{Refinement goals:}

\begin{itemize}[leftmargin=1.5em]
    \item Preserve the original clinical intent unless it is clearly misguided.
    \item Improve prioritization, clinical framing, explanatory quality, and relevance.
    \item Strengthen links between the patient's presentation and the decision being asked.
    \item Clarify the most decision-relevant physiology, risks, trajectory, alternatives, and monitoring needs.
    \item If appropriate, include important alternative explanations or parallel checks that should be considered.
    \item Keep the reasoning ICU-relevant and clinically plausible.
\end{itemize}

\vspace{1em}

\textbf{Output guidelines:}

\begin{itemize}[leftmargin=1.5em]
    \item Language: English.
    \item The explanation should be concise but specific.
    \item The improved reasoning must be exactly one paragraph.
    \item Do not output bullets, numbered lists, or extra commentary outside the JSON schema.
    \item Avoid numeric specifics for laboratory results, vital signs, ventilator settings, medication doses, or flows.
    \item Age is allowed if already present in the context.
    \item Do not generate multiple variants or branches.
    \item Do not use the terms ``refinement path'' or ``path''.
    \item Keep the improved reasoning natural-language only.
\end{itemize}

\vspace{1em}

\textbf{Few-shot examples for style and refinement strategy guidance:}

\vspace{0.5em}

\begin{lstlisting}
{few_shot_examples}
\end{lstlisting}

\vspace{1em}

\textbf{Input:}

\vspace{0.5em}

\textbf{Context:}

\begin{lstlisting}
{context}
\end{lstlisting}

\vspace{0.5em}

\textbf{Question:}

\begin{lstlisting}
{question}
\end{lstlisting}

\vspace{0.5em}

\textbf{Initial reasoning:}

\begin{lstlisting}
{initial_reasoning}
\end{lstlisting}

\vspace{1em}

\textbf{Output only a JSON object with this exact schema, with no extra keys and no prose outside JSON:}

\vspace{0.5em}

\begin{lstlisting}
{
  "output": [
    {
      "question": "The input question here",
      "context": "The input context here",
      "initial_reasoning": "The input initial reasoning here",
      "reasoning_refine_explanation": "Explanation of how the reasoning should be improved",
      "improved_reasoning": "Improved reasoning here"
    }
  ]
}
\end{lstlisting}

\end{promptbox}

\subsection{Training prompts}\label{training_prompts}

\begin{promptbox}{Prompt 12}
\footnotesize

\textbf{Prompt for ICU-REACT reasoning refinement training.}

\vspace{0.5em}

You are an ICU clinician. Your task is to review an initial clinical reasoning draft for an ICU patient context and decision-making question, explain how the reasoning should be improved, and then write an improved reasoning paragraph.

Output the explanation followed by the improved reasoning in the exact natural-language format requested. Do not output JSON, bullet lists, or extra commentary.

\vspace{1em}

\textbf{Patient context:}

\vspace{0.5em}

\begin{lstlisting}
{context}
\end{lstlisting}

\vspace{1em}

\textbf{Decision-making question:}

\vspace{0.5em}

\begin{lstlisting}
{question}
\end{lstlisting}

\vspace{1em}

\textbf{Initial reasoning:}

\vspace{0.5em}

\begin{lstlisting}
{initial_reasoning}
\end{lstlisting}

\vspace{1em}

\textbf{Writing requirements:}

\begin{itemize}[leftmargin=1.5em]
    \item First explain what is missing, overemphasized, underemphasized, or clinically misprioritized in the initial reasoning.
    \item Then write one improved reasoning paragraph.
    \item Keep the explanation concise but specific.
    \item Keep the improved reasoning clinically coherent, ICU-relevant, and decision-focused.
    \item Do not output JSON.
\end{itemize}

\vspace{1em}

\textbf{Output exactly in this format:}

\vspace{0.5em}

\begin{lstlisting}
Reasoning refinement explanation:
<explanation>

Improved reasoning:
<one coherent paragraph>
\end{lstlisting}

\end{promptbox}

\begin{promptbox}{Prompt 13}
\footnotesize

\textbf{Prompt for ICU-REACT context--question refinement training.}

\vspace{0.5em}

You are an ICU clinician. Your task is to take an underspecified ICU patient context and decision-making question and rewrite it into multiple clinically distinct scenario-specific formulations when appropriate.

Output a shared clinical explanation followed by all clinically distinct scenarios in the exact natural-language format requested. Do not output JSON, bullet lists, or extra commentary.

\vspace{1em}

\textbf{Initial patient context:}

\vspace{0.5em}

\begin{lstlisting}
{initial_context}
\end{lstlisting}

\vspace{1em}

\textbf{Initial decision question:}

\vspace{0.5em}

\begin{lstlisting}
{initial_question}
\end{lstlisting}

\vspace{1em}

\textbf{Writing requirements:}

\begin{itemize}[leftmargin=1.5em]
    \item First explain why the initial context--question pair is too underspecified for ICU decision-making.
    \item Then provide all clinically distinct scenario-specific formulations that should be considered.
    \item Each clinical scenario must include:
    \begin{itemize}[leftmargin=1.5em]
        \item scenario label,
        \item clinical rationale,
        \item scenario-specific question,
        \item scenario-specific context.
    \end{itemize}
    \item Keep everything clinically coherent and ICU-relevant.
    \item Do not output JSON.
\end{itemize}

\vspace{1em}

\textbf{Output exactly in this format:}

\vspace{0.5em}

\begin{lstlisting}
Clinical explanation:
<shared explanation>

Clinical scenario 1
Scenario label: <label>
Clinical rationale: <rationale>
Scenario-specific question: <question>
Scenario-specific context: <context>

Clinical scenario 2
Scenario label: <label>
Clinical rationale: <rationale>
Scenario-specific question: <question>
Scenario-specific context: <context>

(continue for all clinically distinct scenarios)
\end{lstlisting}

\end{promptbox}

\begin{promptbox}{Prompt 14}
\footnotesize

\textbf{Prompt for ICU-REACT context generation training.}

\vspace{0.5em}

You are an ICU clinician. Your task is to write one realistic patient context for which the given decision-making question would be clinically relevant.

Output one patient context only. Do not add explanations, bullet points, or JSON. Be specific, clinically grounded, and include enough detail to make the question applicable.

\vspace{1em}

\textbf{Decision question:}

\vspace{0.5em}

\begin{lstlisting}
{question}
\end{lstlisting}

\vspace{1em}

\textbf{Writing requirements:}

\begin{itemize}[leftmargin=1.5em]
    \item Output exactly one patient context, with no extra text.
    \item The context must make the decision question clinically applicable.
    \item Include enough ICU-relevant detail, such as organ support, physiology, trajectory, key comorbidities, or acute problem, to justify why this question would be asked.
    \item Do not answer the question.
    \item Do not invent unnecessary details beyond what is needed to make the question realistic and clinically coherent.
\end{itemize}

\end{promptbox}

\begin{promptbox}{Prompt 15}
\footnotesize

\textbf{Prompt for ICU-REACT question generation training.}

\vspace{0.5em}

You are an ICU clinician. Your task is to write one decision-making question that would guide immediate clinical actions for the patient described.

Output one question only. Do not add explanations, bullet points, or JSON. Be specific, clinically grounded, and focused on a real ICU decision.

\vspace{1em}

\textbf{Patient context:}

\vspace{0.5em}

\begin{lstlisting}
{context}
\end{lstlisting}

\vspace{1em}

\textbf{Writing requirements:}

\begin{itemize}[leftmargin=1.5em]
    \item Output exactly one question, with no extra text.
    \item The question must be a decision-making question, such as what to do, whether to do a specific action, or what is driving a clinical change.
    \item Avoid vague questions such as ``What is going on?''.
    \item Be specific to likely ICU decisions, such as fluids, vasopressors, ventilation, antibiotics, sedation, or diagnostics.
    \item Do not invent data that is not in the context.
\end{itemize}

\end{promptbox}

\begin{promptbox}{Prompt 16}
\footnotesize

\textbf{Prompt for ICU-REACT variable selection reasoning training.}

\vspace{0.5em}

You are an ICU clinician. Write one coherent paragraph explaining which clinical data are most relevant to the decision and why. Do not use bullet points. Do not output JSON.

Output plain text only. Be concise, fluent, and clinically grounded.

\vspace{1em}

\textbf{Patient context:}

\vspace{0.5em}

\begin{lstlisting}
{context}
\end{lstlisting}

\vspace{1em}

\textbf{Decision-making question:}

\vspace{0.5em}

\begin{lstlisting}
{question}
\end{lstlisting}

\end{promptbox}

\subsection{Inference prompts}\label{inference_prompts}

\begin{promptbox}{Prompt 17}
\footnotesize

\textbf{Prompt for inference on ICU-REACT test set.}

\vspace{0.5em}

You are an ICU clinician. Write one coherent paragraph explaining which of the provided variables are most relevant to the decision and why.

You may reference only a clinically meaningful subset of the provided variables. Do not try to cover all of them. Do not use bullet points. Do not output JSON. Output plain text only. Be concise.

\vspace{1em}

\textbf{Patient context:}

\vspace{0.5em}

\begin{lstlisting}
{context}
\end{lstlisting}

\vspace{1em}

\textbf{Decision-making question:}

\vspace{0.5em}

\begin{lstlisting}
{question}
\end{lstlisting}

\vspace{1em}

\textbf{Provided variables, grouped by Category/Sub-category. Use only these variable names when referencing data:}

\vspace{0.5em}

\begin{lstlisting}
{grouped_variables}
\end{lstlisting}

\vspace{1em}

\textbf{Writing requirements:}

\begin{itemize}[leftmargin=1.5em]
    \item Output one paragraph only, with no headings, bullets, or JSON.
    \item Write eight to twelve sentences, approximately 220--350 words.
    \item Start with one introductory sentence that restates the patient context and decision at hand and briefly summarizes what data need review.
    \item Focus only on the variables that are clinically relevant to this patient and this decision.
    \item Do not attempt to mention or justify every provided variable.
    \item Aim to reference a representative, decision-informing subset across the relevant categories and sub-categories, including severity, trajectory, contraindications, response to therapy, and safety monitoring.
    \item When first introducing a category, write one sentence explaining why that category matters for this decision. This should be a general rationale without listing variables yet.
    \item For each sub-category you choose to include, write one sentence that clusters the most relevant variables using comma-separated formatting and gives one concise rationale for the cluster.
    \item It is acceptable to omit entire categories or sub-categories if they are not decision-informing for this case.
    \item Include a final closing sentence that briefly summarizes how the selected data will support making the decision, without making a recommendation.
    \item Do not invent patient values or make a final recommendation.
    \item Every variable you mention must match exactly a name from the provided variables list.
\end{itemize}

\end{promptbox}

\begin{promptbox}{Prompt 18}
\footnotesize

\textbf{Prompt for inference on SCT-Bench test set.}

\vspace{0.5em}

You are taking a Script Concordance Test.

You will receive a single clinical case consisting of a scenario, a hypothesis, and additional information.

Ignore any examples in the prompt; they are already answered. For this case, respond with exactly one line starting with \texttt{Rating:} followed by one of \texttt{\{-2, -1, 0, +1, +2\}}, then a brief explanation.

Do not ask for more scenarios or start a conversation. Do not describe the test instructions back to me.

\vspace{1em}

\textbf{Guideline:}

\vspace{0.5em}

\begin{lstlisting}
{guideline}
\end{lstlisting}

\vspace{1em}

\textbf{Worked examples, read-only.}

The following examples are already scored. Do not score them again.

\vspace{0.5em}

\textbf{Example with rating -2:}

\begin{lstlisting}
{example_-2}
\end{lstlisting}

\vspace{0.5em}

\textbf{Example with rating -1:}

\begin{lstlisting}
{example_-1}
\end{lstlisting}

\vspace{0.5em}

\textbf{Example with rating 0:}

\begin{lstlisting}
{example_0}
\end{lstlisting}

\vspace{0.5em}

\textbf{Example with rating +1:}

\begin{lstlisting}
{example_+1}
\end{lstlisting}

\vspace{0.5em}

\textbf{Example with rating +2:}

\begin{lstlisting}
{example_+2}
\end{lstlisting}

\vspace{1em}

\textbf{New SCT item to score.}

Now, you will be given one new scenario, hypothesis, and additional information. Ignore the previous examples. Your job is to rate only this new item.

\vspace{0.5em}

\textbf{Scenario:}

\begin{lstlisting}
{{ scenario }}
\end{lstlisting}

\vspace{0.5em}

\textbf{Hypothesis:}

\begin{lstlisting}
{{ hypothesis }}
\end{lstlisting}

\vspace{0.5em}

\textbf{Additional information:}

\begin{lstlisting}
{{ additional information }}
\end{lstlisting}

\vspace{1em}

\textbf{Required response format:}

\vspace{0.5em}

\begin{lstlisting}
Rating: <one of -2, -1, 0, +1, +2> <brief explanation>
\end{lstlisting}

\end{promptbox}

\begin{promptbox}{Prompt 19}
\footnotesize

\textbf{Prompt for decision factors inference on ER-Reason test set.}

\vspace{0.5em}

You are an experienced Emergency Department physician. Your task is only to describe the medical decision factors needed to narrow the differential diagnosis for the patient's current chief complaint.

Do not provide a differential diagnosis list, treatment plan, or disposition. Focus only on what additional history, exam findings, tests, imaging, and consultations are needed.

\vspace{1em}

Based on the patient's chief complaint, age, sex, and available clinical information, provide only the following:

\vspace{0.5em}

\textbf{Medical decision factors:}

List the additional history you would want, important exam findings you would look for, tests or imaging you would order, and any specialist consultations you would consider in order to narrow the diagnosis for the current chief complaint.

Nothing has currently been performed.

\vspace{1em}

\textbf{Important temporal context:}

Everything in the notes is from past encounters. The patient is now presenting with a new complaint.

\vspace{1em}

\textbf{Chief complaint, current encounter:}

\vspace{0.5em}

\begin{lstlisting}
{chief}
\end{lstlisting}

\vspace{1em}

\textbf{Age:}

\vspace{0.5em}

\begin{lstlisting}
{age}
\end{lstlisting}

\vspace{1em}

\textbf{Sex:}

\vspace{0.5em}

\begin{lstlisting}
{sex}
\end{lstlisting}

\vspace{1em}

\textbf{Patient one-liner summary, if available:}

\vspace{0.5em}

\begin{lstlisting}
{one_liner}
\end{lstlisting}

\vspace{1em}

\textbf{Previous hospital encounter notes, for context only:}

\vspace{0.5em}

\begin{lstlisting}
{notes_content}
\end{lstlisting}

\end{promptbox}

\begin{promptbox}{Prompt 20}
\footnotesize

\textbf{Prompt for differential diagnosis inference on ER-Reason test set.}

\vspace{0.5em}

You are an experienced Emergency Department physician. Your task is only to generate an initial differential diagnosis for the patient's current chief complaint.

Do not provide medical decision factors, workup recommendations, treatment plans, or disposition. Focus only on the differential diagnosis and brief reasoning.

\vspace{1em}

Based on the patient's chief complaint, age, sex, and available clinical information, provide only the following:

\vspace{0.5em}

\textbf{Differential diagnosis:}

List the most relevant differential diagnoses for the current chief complaint, explain briefly why each is being considered, and note which diagnoses are lower on the differential and why.

\vspace{1em}

\textbf{Important temporal context:}

Everything in the notes is from past encounters. The patient is now presenting with a new complaint.

\vspace{1em}

\textbf{Chief complaint, current encounter:}

\vspace{0.5em}

\begin{lstlisting}
{chief}
\end{lstlisting}

\vspace{1em}

\textbf{Age:}

\vspace{0.5em}

\begin{lstlisting}
{age}
\end{lstlisting}

\vspace{1em}

\textbf{Sex:}

\vspace{0.5em}

\begin{lstlisting}
{sex}
\end{lstlisting}

\vspace{1em}

\textbf{Patient one-liner summary, if available:}

\vspace{0.5em}

\begin{lstlisting}
{one_liner}
\end{lstlisting}

\vspace{1em}

\textbf{Previous hospital encounter notes, for context only:}

\vspace{0.5em}

\begin{lstlisting}
{notes_content}
\end{lstlisting}

\end{promptbox}

\begin{promptbox}{Prompt 21}
\footnotesize

\textbf{Prompt for treatment plan inference on ER-Reason test set.}

\vspace{0.5em}

You are an experienced Emergency Department physician. Your task is only to propose the initial treatment plan for the patient's current ED visit based on the available history and physical exam information.

Do not provide a full differential diagnosis list or a separate decision-factors section. Focus only on the most likely working diagnosis, immediate management, additional diagnostic orders needed for treatment planning, and disposition.

\vspace{1em}

Based on the patient's chief complaint, age, sex, and available clinical information, provide only the following:

\vspace{0.5em}

\textbf{Treatment plan:}

State the most likely working diagnosis, the recommended initial treatment plan, any additional diagnostic tests you would still order as part of management, and the recommended disposition, such as discharge, admission, or ICU. If there is not enough information for a final diagnosis, state what is still needed.

\vspace{1em}

\textbf{Current ED encounter physical/history, current visit, if available:}

\vspace{0.5em}

\begin{lstlisting}
{ed_presentation}
\end{lstlisting}

\vspace{1em}

\textbf{Important temporal context:}

Everything in the notes is from past encounters. The patient is now presenting with a new complaint.

\vspace{1em}

\textbf{Chief complaint, current encounter:}

\vspace{0.5em}

\begin{lstlisting}
{chief}
\end{lstlisting}

\vspace{1em}

\textbf{Age:}

\vspace{0.5em}

\begin{lstlisting}
{age}
\end{lstlisting}

\vspace{1em}

\textbf{Sex:}

\vspace{0.5em}

\begin{lstlisting}
{sex}
\end{lstlisting}

\vspace{1em}

\textbf{Patient one-liner summary, if available:}

\vspace{0.5em}

\begin{lstlisting}
{one_liner}
\end{lstlisting}

\vspace{1em}

\textbf{Previous hospital encounter notes, for context only:}

\vspace{0.5em}

\begin{lstlisting}
{notes_content}
\end{lstlisting}

\end{promptbox}

\begin{promptbox}{Prompt 22}
\footnotesize

\textbf{Prompt for inference of assessment recommendations on MedRBench test set.}

\vspace{0.5em}

You are a professional doctor.

Please thoroughly examine the patient case summary presented below. Your objective is to perform a detailed diagnostic analysis utilizing all available information. Note that, due to the potentially limited details, the preliminary diagnosis may encompass several possible conditions.

If you determine that the provided data are inadequate for a definitive conclusion, please enumerate any additional diagnostic tests or information that would be necessary. However, if you can deduce a conclusive diagnosis, please proceed to provide it. Too many requests for information are also inappropriate.

\vspace{1em}

\textbf{Patient case summary:}

\vspace{0.5em}

\begin{lstlisting}
{case}
\end{lstlisting}

\vspace{1em}

\textbf{Guidelines:}

\begin{itemize}[leftmargin=1.5em]
    \item Evaluate the patient's symptoms, medical history, and all pertinent details from the case summary.
    \item Formulate differential diagnoses based on your analysis.
    \item If the information is not sufficient for a conclusive diagnosis, specify the further tests or details required.
    \item Always follow the response format in each turn of the dialogue.
    \item Never change the section headers using the \texttt{\#\#\#} format.
\end{itemize}

\vspace{1em}

\textbf{Response format:}

\vspace{0.5em}

\begin{lstlisting}
### Clinical Reasoning:
[Provide a concise step-by-step clinical reasoning summary, with each logical step in a separate paragraph, using labels such as <step 1>.]
<step 1> Specific reasoning content of this step
<step 2> Specific reasoning content of this step
...
<step n> Specific reasoning content of this step

### Conclusion:
[Give a preliminary conclusion if possible, or summarize the current findings.]

### Additional Information Required:
[Indicate if further information is needed by specifying the required tests or data. If a conclusive diagnosis has been made and no additional information is necessary, only output "Not required." directly without any other words in this section.]

For example:
Not required.

or

1. Laboratory tests: details
2. Imaging: details
...
\end{lstlisting}

\end{promptbox}

\begin{promptbox}{Prompt 23}
\footnotesize

\textbf{Prompt for the GPT-4.1-mini agent to provide requested assessment recommendation information on MedRBench test set.}

\vspace{0.5em}

You are a professional doctor.

You are a medical expert providing guidance to a junior physician on a patient case. The junior physician will ask you for additional diagnostic information based on the patient's case details and any available ancillary test results. Your role is to provide accurate and relevant responses regarding the availability of specific diagnostic information.

\vspace{1em}

\textbf{Guidelines:}

\begin{enumerate}[leftmargin=1.5em]
    \item You will receive the patient's case information and any relevant ancillary test results.
    \item The junior physician will ask questions about additional diagnostic information needed for the case.
    \item If relevant ancillary test information is available for the requested diagnostic area, provide the details accurately.
    \item If no relevant ancillary test information is available for the requested diagnostic area, simply state: ``There is no relevant ancillary test information available for this request.''
\end{enumerate}

\vspace{1em}

\textbf{Patient case:}

\vspace{0.5em}

\begin{lstlisting}
{case}
\end{lstlisting}

\vspace{1em}

\textbf{Ancillary test results:}

\vspace{0.5em}

\begin{lstlisting}
{ancillary_test_results}
\end{lstlisting}

\vspace{1em}

\textbf{Example interaction:}

\vspace{0.5em}

\begin{lstlisting}
Junior Physician: "Does the patient have any imaging studies like an X-ray or CT scan?"

Your Response:

If there is relevant imaging information available:
"Based on the available ancillary test results, the patient has undergone a chest X-ray which shows [specific findings]."

If there is no relevant imaging information available:
"There is no relevant ancillary test information available for this request."
\end{lstlisting}

\vspace{1em}

\textbf{Note:}

Your responses should be factual and based solely on the provided patient case information and ancillary test results. Avoid speculation or hypotheticals unless explicitly requested.

\end{promptbox}

\begin{promptbox}{Prompt 24}
\footnotesize

\textbf{Prompt for inference of final diagnosis on MedRBench test set.}

\vspace{0.5em}

You are a professional doctor.

Please make a final diagnosis for the patient in light of the additional information provided below.

\vspace{1em}

\textbf{Additional information:}

\vspace{0.5em}

\begin{lstlisting}
{additional_information}
\end{lstlisting}

\vspace{1em}

\textbf{Guidelines:}

\begin{itemize}[leftmargin=1.5em]
    \item Evaluate the patient's symptoms, medical history, and all pertinent details from the case summary.
    \item Formulate differential diagnoses based on your analysis.
    \item Always follow the response format in each turn of the dialogue.
    \item Never change the section headers using the \texttt{\#\#\#} format.
\end{itemize}

\vspace{1em}

\textbf{Response format:}

\vspace{0.5em}

\begin{lstlisting}
### Clinical Reasoning:
[Provide a concise step-by-step clinical reasoning summary, with each logical step in a separate paragraph, using labels such as <step 1>.]
<step 1> Specific reasoning content of this step
<step 2> Specific reasoning content of this step
...
<step n> Specific reasoning content of this step

### Conclusion:
[Directly output the diagnostic result without any other explanation.]
\end{lstlisting}

\end{promptbox}

\begin{promptbox}{Prompt 25}
\footnotesize

\textbf{Prompt for inference of treatment plan on MedRBench test set.}

\vspace{0.5em}

You are a professional doctor.

Please carefully study the following patient case summary, conduct a comprehensive and in-depth treatment planning analysis, and clearly provide the selected treatment for the patient.

\vspace{1em}

\textbf{Patient case summary:}

\vspace{0.5em}

\begin{lstlisting}
{case}
\end{lstlisting}

\vspace{1em}

\textbf{Response format:}

\vspace{0.5em}

\begin{lstlisting}
### Clinical Reasoning:
[Provide a concise step-by-step clinical reasoning summary, with each logical step in a separate paragraph, using labels such as <step 1>.]
<step 1> Specific reasoning content of this step
<step 2> Specific reasoning content of this step
...
<step n> Specific reasoning content of this step

### Answer:
[Just output the selected treatment for the patient without any other explanation.]
\end{lstlisting}

\end{promptbox}

\begin{promptbox}{Prompt 26}
\footnotesize

\textbf{Prompt for inference on diagnostic workflow pipeline for VivaBench test set.}

\vspace{0.5em}

You are a medical diagnostic assistant reviewing one patient case.

Your task is to choose the single best next step in the diagnostic workflow and return it as a JSON object.

\vspace{1em}

\textbf{Diagnostic workflow rules:}

\begin{enumerate}[leftmargin=1.5em]
    \item At the beginning of the case, you may only choose \texttt{history} or \texttt{examination}.
    \item Before any \texttt{investigation} or \texttt{imaging} action, you must first provide exactly one \texttt{diagnosis\_provisional} action.
    \item After any \texttt{investigation} or \texttt{imaging} has been requested, you must not request any further history or physical examination.
    \item Select exactly one action per turn.
    \item When the available information is sufficient, provide \texttt{diagnosis\_final}.
\end{enumerate}

\vspace{1em}

\textbf{Allowed actions:}

\begin{itemize}[leftmargin=1.5em]
    \item \texttt{history}: Ask the patient one to two focused history questions in plain language.
    \item \texttt{examination}: Request a focused physical examination and specify the findings you want assessed.
    \item \texttt{diagnosis\_provisional}: Give a provisional diagnosis based on the information currently available, before any investigations or imaging.
    \item \texttt{investigation}: Request a non-imaging test, including laboratory tests and bedside or special tests such as ECG, EEG, or pulmonary function testing. When relevant, specify specimen type.
    \item \texttt{imaging}: Request an imaging study performed by radiology or equivalent imaging services, such as X-ray, ultrasound, CT, MRI, PET, or VQ scan. Specify modality and anatomical region.
    \item \texttt{diagnosis\_final}: Give the final diagnosis after completing the evaluation.
\end{itemize}

\vspace{1em}

\textbf{Important workflow enforcement:}

\begin{itemize}[leftmargin=1.5em]
    \item Do not choose \texttt{investigation} or \texttt{imaging} until a \texttt{diagnosis\_provisional} action has already been given earlier in the conversation.
    \item On the first turn, do not choose \texttt{investigation}, \texttt{imaging}, or \texttt{diagnosis\_final}.
    \item If you have gathered enough information to suspect a diagnosis but have not yet ordered tests, the next step should usually be \texttt{diagnosis\_provisional}.
\end{itemize}

\vspace{1em}

\textbf{Diagnosis formatting rules for both provisional and final diagnosis:}

\begin{itemize}[leftmargin=1.5em]
    \item You may include up to five diagnoses if there is uncertainty or more than one active problem.
    \item For each diagnosis, provide:
    \begin{itemize}[leftmargin=1.5em]
        \item \texttt{condition}: free-text condition label,
        \item \texttt{icd\_10\_name}: ICD-10 diagnostic name,
        \item \texttt{icd\_10}: ICD-10 code,
        \item \texttt{confidence}: number from 0.0 to 1.0.
    \end{itemize}
    \item Confidence scores do not need to sum to 1.0.
    \item For diagnosis actions, the \texttt{query} field should contain a JSON list in the following format:
\end{itemize}

\vspace{0.5em}

\begin{lstlisting}
[
  {
    "condition": "free text name of the condition",
    "icd_10_name": "icd 10 name of the condition",
    "icd_10": "icd code of the condition",
    "confidence": score
  }
]
\end{lstlisting}

\vspace{1em}

\textbf{Return one JSON object with this structure:}

\vspace{0.5em}

\begin{lstlisting}
{
  "reasoning": "brief justification for the next step",
  "action": "one allowed action",
  "query": "specific request, or diagnosis list when action is a diagnosis action"
}
\end{lstlisting}

\vspace{1em}

\textbf{Initial patient input format:}

\vspace{0.5em}

\begin{lstlisting}
Patient: {age}-year-old {gender}

History:
{history_text}
\end{lstlisting}

\end{promptbox}

\begin{promptbox}{Prompt 27}
\footnotesize

\textbf{Prompt for JSON parse-retry prompt for invalid diagnostic outputs.}

\vspace{0.5em}

Your previous response could not be parsed. Please return exactly one valid JSON object with this structure:

\vspace{0.5em}

\begin{lstlisting}
{
  "reasoning": "brief justification for the action",
  "action": "chosen action",
  "query": "specific request"
}
\end{lstlisting}

\vspace{1em}

\textbf{Previous response:}

\vspace{0.5em}

\begin{lstlisting}
{previous_response}
\end{lstlisting}

\end{promptbox}

\begin{promptbox}{Prompt 28}
\footnotesize

\textbf{Prompt for GPT 4.1-mini history information-mapping agent.}

\vspace{0.5em}

You are a medical information-mapping assistant.

\vspace{1em}

\textbf{Task:}

\begin{itemize}[leftmargin=1.5em]
    \item Read the user's request about symptoms or medical history.
    \item Match each requested item only to keys that are present in the provided available-keys list.
    \item For symptom requests, include any specific characteristics requested when available.
\end{itemize}

\vspace{1em}

\textbf{Rules:}

\begin{itemize}[leftmargin=1.5em]
    \item Use only keys from the provided available-keys list.
    \item Do not invent new matched keys.
    \item If a request does not correspond to any available key, place it in \texttt{unmatched}.
\end{itemize}

\vspace{1em}

\textbf{Input:}

\vspace{0.5em}

\begin{lstlisting}
Chief complaint: {chief_complaint}
User request: {query}
Available keys: {keys}
\end{lstlisting}

\vspace{1em}

\textbf{Return one JSON object in this format:}

\vspace{0.5em}

\begin{lstlisting}
{
  "matched": [
    {
      "query": "relevant phrase from the request",
      "key": "matching key from available keys",
      "addit": ["optional list of requested symptom characteristics"]
    }
  ],
  "unmatched": [
    {
      "query": "phrase that could not be matched",
      "key": "suggested standardized key"
    }
  ]
}
\end{lstlisting}

\end{promptbox}

\begin{promptbox}{Prompt 29}
\footnotesize

\textbf{Prompt for GPT 4.1-mini physical examination information-mapping agent.}

\vspace{0.5em}

You are a medical information-mapping assistant for physical examination findings.

\vspace{1em}

\textbf{Task:}

\begin{itemize}[leftmargin=1.5em]
    \item Read the user request.
    \item Match each requested physical examination item only to keys in the provided available-keys list.
    \item If part of the request cannot be matched, place it in \texttt{unmatched}.
\end{itemize}

\vspace{1em}

\textbf{Input:}

\vspace{0.5em}

\begin{lstlisting}
User request: {query}
Available keys: {keys}
\end{lstlisting}

\vspace{1em}

\textbf{Return one JSON object in this format:}

\vspace{0.5em}

\begin{lstlisting}
{
  "matched": [
    {
      "query": "relevant phrase from the request",
      "key": "matching key from available keys"
    }
  ],
  "unmatched": [
    {
      "query": "unmatched phrase from the request",
      "key": "suggested standardized key"
    }
  ]
}
\end{lstlisting}

\vspace{1em}

Return only requested items. Do not include extra information.

\end{promptbox}

\begin{promptbox}{Prompt 30}
\footnotesize

\textbf{Prompt for GPT 4.1-mini laboratory and non-imaging information-mapping agent.}

\vspace{0.5em}

You are a medical information-mapping assistant for laboratory and non-imaging investigations.

\vspace{1em}

\textbf{Task:}

\begin{itemize}[leftmargin=1.5em]
    \item Read the user request.
    \item Match each requested investigation only to keys in the provided available-items list.
    \item If part of the request cannot be matched, place it in \texttt{unmatched}.
\end{itemize}

\vspace{1em}

\textbf{Input:}

\vspace{0.5em}

\begin{lstlisting}
User request: {query}
Available items: {items}
\end{lstlisting}

\vspace{1em}

\textbf{Return one JSON object in this format:}

\vspace{0.5em}

\begin{lstlisting}
{
  "matched": [
    {
      "query": "relevant phrase from the request",
      "key": "matching key from available items"
    }
  ],
  "unmatched": [
    {
      "query": "unmatched phrase from the request",
      "key": "suggested standardized key"
    }
  ]
}
\end{lstlisting}

\vspace{1em}

Return keys only, not values.

\end{promptbox}

\begin{promptbox}{Prompt 31}
\footnotesize

\textbf{Prompt for GPT 4.1-mini imaging information-mapping agent.}

\vspace{0.5em}

You are a medical information-mapping assistant for imaging studies.

\vspace{1em}

\textbf{Task:}

\begin{itemize}[leftmargin=1.5em]
    \item Read the user request.
    \item Match each requested imaging study only to keys in the provided available-keys list.
    \item If part of the request cannot be matched, place it in \texttt{unmatched}.
\end{itemize}

\vspace{1em}

\textbf{Input:}

\vspace{0.5em}

\begin{lstlisting}
User request: {query}
Available keys: {keys}
\end{lstlisting}

\vspace{1em}

\textbf{Return one JSON object in this format:}

\vspace{0.5em}

\begin{lstlisting}
{
  "matched": [
    {
      "query": "relevant phrase from the request",
      "key": "matching key from available keys"
    }
  ],
  "unmatched": [
    {
      "query": "unmatched phrase from the request",
      "key": "suggested standardized key"
    }
  ]
}
\end{lstlisting}

\end{promptbox}

\begin{promptbox}{Prompt 32}
\footnotesize

\textbf{Prompt for MedQA multiple-choice inference.}

\vspace{0.5em}

You are a helpful medical assistant.

You are a clinician. Read the question and options and pick the single best answer.

First, briefly reason through the clinical information and rule out incorrect options. Then, on a new line at the end, output only the letter A--E.

\vspace{1em}

\textbf{Question:}

\vspace{0.5em}

\begin{lstlisting}
{question}
\end{lstlisting}

\vspace{1em}

\textbf{Options:}

\vspace{0.5em}

\begin{lstlisting}
{options_block}
\end{lstlisting}

\vspace{1em}

\textbf{Required final answer format:}

\vspace{0.5em}

\begin{lstlisting}
<single letter A-E>
\end{lstlisting}

\end{promptbox}

\begin{promptbox}{Prompt 33}
\footnotesize

\textbf{Prompt for MedMCQA multiple-choice inference.}

\vspace{0.5em}

You are a helpful medical assistant.

You are a clinician. Read the question and options and pick the single best answer.

First, briefly reason through the clinical information and rule out incorrect options. Then, on a new line at the end, output only the letter A--D.

\vspace{1em}

\textbf{Question:}

\vspace{0.5em}

\begin{lstlisting}
{question}
\end{lstlisting}

\vspace{1em}

\textbf{Options:}

\vspace{0.5em}

\begin{lstlisting}
{options_block}
\end{lstlisting}

\vspace{1em}

\textbf{Required final answer format:}

\vspace{0.5em}

\begin{lstlisting}
<single letter A-D>
\end{lstlisting}

\end{promptbox}

\begin{promptbox}{Prompt 34}
\footnotesize

\textbf{Prompt for MedXpertQA multiple-choice inference.}

\vspace{0.5em}

You are a helpful medical assistant.

You are a clinician answering multiple-choice questions.

First, briefly reason through the clinical information using clinical reasoning, pathophysiology, and guidelines to narrow down the options. Then, on a new line at the end, output only the single best letter A--J.

\vspace{1em}

\textbf{Question:}

\vspace{0.5em}

\begin{lstlisting}
{question}
\end{lstlisting}

\vspace{1em}

\textbf{Options:}

\vspace{0.5em}

\begin{lstlisting}
{options_block}
\end{lstlisting}

\vspace{1em}

\textbf{Required final answer format:}

\vspace{0.5em}

\begin{lstlisting}
<single letter A-J>
\end{lstlisting}

\end{promptbox}

\begin{promptbox}{Prompt 35}
\footnotesize

\textbf{Prompt for MMLU medical multiple-choice inference.}

\vspace{0.5em}

You are a helpful medical assistant.

You are a clinician. Read the question and options and pick the single best answer.

First, briefly reason through the clinical or biological information and rule out incorrect options. Then, on a new line at the end, output only the letter A--D.

\vspace{1em}

\textbf{Question:}

\vspace{0.5em}

\begin{lstlisting}
{question}
\end{lstlisting}

\vspace{1em}

\textbf{Options:}

\vspace{0.5em}

\begin{lstlisting}
{options_block}
\end{lstlisting}

\vspace{1em}

\textbf{Required final answer format:}

\vspace{0.5em}

\begin{lstlisting}
<single letter A-D>
\end{lstlisting}

\end{promptbox}

\subsection{Evaluation prompts}\label{eval_prompts}

\begin{promptbox}{Prompt 36}
\footnotesize

\textbf{Prompt for rubric-based LLM-Judge evaluation on ICU-REACT test set.}

\vspace{0.5em}

You are an expert ICU clinician evaluating the quality of a model's response using a rubric.

\vspace{1em}

\textbf{Your job:}

\begin{itemize}[leftmargin=1.5em]
    \item Read the patient context, decision-making question, and model response.
    \item Apply each rubric criterion to the model response.
    \item For each criterion, decide how strongly it is satisfied using the \texttt{type} field.
    \item The rubric already encodes what an ideal response should contain; you do not need any other reference.
\end{itemize}

\vspace{1em}

\textbf{Patient context:}

\vspace{0.5em}

\begin{lstlisting}
{context}
\end{lstlisting}

\vspace{1em}

\textbf{Decision question:}

\vspace{0.5em}

\begin{lstlisting}
{question}
\end{lstlisting}

\vspace{1em}

\textbf{Candidate model response to be graded:}

\vspace{0.5em}

\begin{lstlisting}
{model_response}
\end{lstlisting}

\vspace{1em}

\textbf{Rubric, as a list of criteria with weights and types:}

\vspace{0.5em}

\begin{lstlisting}
{rubric_json}
\end{lstlisting}

\vspace{1em}

\textbf{Scoring rules:}

For each rubric entry, the \texttt{type} tells you how to set \texttt{value}.

\vspace{0.5em}

\begin{enumerate}[leftmargin=1.5em]

    \item \textbf{\texttt{type = "present\_absent"}}

    \begin{itemize}[leftmargin=1.5em]
        \item \texttt{value} must be 0 or 1.
        \item \texttt{value = 1} if the model response clearly satisfies the criterion.
        \item \texttt{value = 0} if it does not.
    \end{itemize}

    \item \textbf{\texttt{type = "0\_1\_2\_scale"}}

    \begin{itemize}[leftmargin=1.5em]
        \item \texttt{value} must be 0, 1, or 2.
        \item 0 = not satisfied at all.
        \item 1 = partially satisfied.
        \item 2 = clearly and fully satisfied.
    \end{itemize}

    \item \textbf{\texttt{type = "penalty\_if\_present"}}

    \begin{itemize}[leftmargin=1.5em]
        \item \texttt{value} must be 0 or 1.
        \item \texttt{value = 1} only if the problematic behavior is present in the model response.
        \item \texttt{value = 0} if the problematic behavior is absent.
    \end{itemize}

\end{enumerate}

Do not try to compute total points yourself. Just assign the correct \texttt{value} for each criterion based on the rubric description and the content of the model response.

\vspace{1em}

\textbf{Output format, strict JSON:}

Return a single JSON object:

\vspace{0.5em}

\begin{lstlisting}
{
  "criteria_results": [
    {
      "id": "<criterion id, e.g. R1>",
      "type": "<copy the 'type' from the rubric for this criterion>",
      "value": <integer value as defined above>,
      "evidence": "<a short quote or paraphrase from the model response that justifies your decision>"
    }
  ]
}
\end{lstlisting}

\vspace{1em}

\textbf{Output JSON strictness:}

\begin{itemize}[leftmargin=1.5em]
    \item Return only valid JSON, with no markdown.
    \item Strings must not contain unescaped double quotes.
    \item Do not include comments of any kind, including \texttt{//}, \texttt{/* */}, or trailing annotations.
    \item Do not escape apostrophes. Write \texttt{patient's}, not \texttt{patient\textbackslash's}.
    \item In \texttt{evidence}, do not write things like \texttt{"PaCO2"} or \texttt{"ABG"}. Instead, write \texttt{'PaCO2'} and \texttt{'ABG'} using single quotes, or escape them as \texttt{\textbackslash"PaCO2\textbackslash"}.
    \item Do not include stray newlines, trailing commas, or comments.
\end{itemize}

\vspace{1em}

\textbf{Rules:}

\begin{itemize}[leftmargin=1.5em]
    \item Do not include any keys other than \texttt{sample\_id} and \texttt{criteria\_results}.
    \item Do not include explanations outside of this JSON.
    \item Do not include comments, markdown, or trailing commas.
    \item Make sure the JSON is syntactically valid and fully parsable.
\end{itemize}

\end{promptbox}

\begin{promptbox}{Prompt 37}
\footnotesize

\textbf{Prompt for LLM-Judge evaluation of clinical concept-overlap accuracy scoring against physician reference on ER-Reason test set.}

\vspace{0.5em}

You are a strict clinical evaluator.

You will be given two texts:

\begin{itemize}[leftmargin=1.5em]
    \item Prediction, generated by the model.
    \item Reference, corresponding to the physician ground truth.
\end{itemize}

\vspace{1em}

\textbf{Your task:}

Score the clinical accuracy of the prediction compared to the reference on a 0--100 scale.

\vspace{1em}

\textbf{Scoring strategy. Follow this order:}

\begin{enumerate}[leftmargin=1.5em]
    \item \textbf{Recall, add points:} Identify the key clinical concepts present in the reference and check whether the prediction captures them using equivalent terminology. Synonyms, abbreviations, and alternate names count as correct.
    \item \textbf{Precision, subtract points:} Penalize the prediction for extra concepts not supported by the reference. Do not penalize harmless generalities or standard workflow steps unless they materially change the meaning.
\end{enumerate}

\vspace{1em}

\textbf{Core rules:}

\begin{itemize}[leftmargin=1.5em]
    \item Ground the evaluation only in the two provided texts.
    \item Do not infer beyond the texts.
    \item Handle negation correctly. Negated concepts should not count as present.
    \item Treat synonyms and abbreviations as equivalent when they clearly refer to the same concept.
\end{itemize}

\vspace{1em}

\textbf{Examples of equivalence, non-exhaustive:}

\begin{itemize}[leftmargin=1.5em]
    \item \texttt{cmp} = \texttt{comprehensive metabolic panel} = \texttt{metabolic panel}
    \item \texttt{electrolyte panel} can match \texttt{cmp} or \texttt{bmp} if clearly referring to serum electrolytes or a metabolic panel.
    \item \texttt{cbc} = \texttt{complete blood count}
    \item \texttt{ua} = \texttt{urinalysis}
    \item \texttt{hcg} = \texttt{pregnancy test}
    \item \texttt{ct a/p} = \texttt{ct abdomen and pelvis}
    \item \texttt{cxr} = \texttt{chest x-ray}
    \item \texttt{mi} = \texttt{myocardial infarction}
    \item \texttt{sob} = \texttt{shortness of breath}
\end{itemize}

\vspace{1em}

\textbf{Prediction, model:}

\vspace{0.5em}

\begin{lstlisting}
{pred}
\end{lstlisting}

\vspace{1em}

\textbf{Reference, physician ground truth:}

\vspace{0.5em}

\begin{lstlisting}
{ref}
\end{lstlisting}

\vspace{1em}

\textbf{Scoring rubric:}

\begin{itemize}[leftmargin=1.5em]
    \item Start from \texttt{recall\_points = 0}.
    \item Extract five to fifteen key concepts from the reference, focusing on the most clinically important concepts.
    \item Add points for each key concept that is correctly present in the prediction. Synonyms and abbreviations count.
    \item If almost all key concepts are matched, \texttt{recall\_points} should be high, approximately 70--100.
    \item If about half are matched, \texttt{recall\_points} should be moderate, approximately 40--70.
    \item If few are matched, \texttt{recall\_points} should be low, approximately 0--40.
    \item Then compute \texttt{precision\_penalty}.
    \item Subtract points for each extra concept in the prediction that is not supported by the reference.
    \item Penalize more for high-impact incorrect extras, such as wrong diagnosis, wrong disposition, or high-risk intervention.
    \item Penalize less for mild over-inclusion, such as generic laboratories or vital signs, unless it dominates the content.
    \item \texttt{final\_score} must equal \texttt{clamp(recall\_points - precision\_penalty, 0, 100)}.
\end{itemize}

\vspace{1em}

\textbf{Output only a JSON object with this schema, with no markdown and no extra text:}

\vspace{0.5em}

\begin{lstlisting}
{
  "score": number,
  "recall_points": number,
  "precision_penalty": number,
  "final_score": number,
  "matched_key_concepts": [string, ...],
  "missed_key_concepts": [string, ...],
  "extra_concepts": [string, ...]
}
\end{lstlisting}

\vspace{1em}

\textbf{Output constraints:}

\begin{itemize}[leftmargin=1.5em]
    \item \texttt{score} should be an integer from 0 to 100.
    \item \texttt{recall\_points} should represent the 0--100 contribution before penalties.
    \item \texttt{precision\_penalty} should represent the 0--100 deduction for unsupported extras.
    \item \texttt{final\_score} must equal \texttt{recall\_points - precision\_penalty}, clipped to the range 0--100.
    \item Use lowercase for concept strings.
    \item Use short canonical phrases for concepts.
    \item Keep each concept list to at most 25 items, prioritizing the most clinically important concepts.
    \item If either text is empty or non-clinical, return a best-effort score with empty lists.
    \item Return only the JSON object with the required keys.
\end{itemize}

\end{promptbox}

\begin{promptbox}{Prompt 38}
\footnotesize

\textbf{Prompt for GPT 4.1-mini agent for parsing additional diagnostic information requirements on MedRBench test set.}

\vspace{0.5em}

\textbf{Task overview}

You will receive raw text from an auxiliary diagnostic or treatment model describing additional information needed for diagnosis. Your task is to convert that text into a JSON dictionary with a list of individual information-requirement objects.

\vspace{1em}

\textbf{Core extraction rule}

Each JSON object must represent exactly one individual test, examination, imaging study, history question group, or other information-gathering action.

\vspace{1em}

\textbf{Strict extraction rules}

\begin{enumerate}[leftmargin=1.5em]
    \item Split combined statements into separate objects whenever multiple distinct tests, studies, or examinations are mentioned.
    \item Never group multiple distinct tests into one object, even if they share the same category or purpose.
    \item The \texttt{test\_name} field must contain only one individual test, study, examination, or action.
    \item If the raw text lists multiple items joined by words such as ``and'', ``or'', commas, semicolons, slashes, or parentheses with distinct procedures, split them into separate objects whenever they could reasonably be ordered or requested separately.
    \item Keep the original meaning and wording as much as possible, but rewrite only as needed to isolate each individual item.
    \item Do not add new tests or infer tests not explicitly mentioned in the raw text.
    \item Do not omit any test, examination, imaging study, laboratory study, history inquiry, or specialist evaluation mentioned in the raw text.
    \item If multiple items share the same purpose, repeat the same or very similar \texttt{info\_required} text across multiple objects rather than combining them.
    \item If one phrase contains both a broad category and examples, extract the specific examples as separate objects whenever they are explicit.
    \item If a phrase contains a bundled examination with subcomponents that are part of a single named examination, keep it as one object only if it is clearly one examination. For example, \texttt{Ophthalmological examination (visual acuity, intraocular pressure, fundus examination)} can remain one object because it is one named specialist examination with listed components.
    \item If two named imaging studies are listed together, they must be split. For example:
\end{enumerate}

\vspace{0.5em}

\begin{lstlisting}
"Non-contrast CT (NCCT) scan; Non-contrast MRI (NCMRI) scan"
\end{lstlisting}

\vspace{0.5em}

becomes two separate objects.

\begin{enumerate}[leftmargin=1.5em, resume]
    \item If two named laboratory tests are listed together, they must be split. For example:
\end{enumerate}

\vspace{0.5em}

\begin{lstlisting}
"C-reactive protein (CRP); Complete blood count (CBC) with differential"
\end{lstlisting}

\vspace{0.5em}

becomes two separate objects.

\begin{enumerate}[leftmargin=1.5em, resume]
    \item For culture or PCR phrasing:
    \begin{itemize}[leftmargin=1.5em]
        \item If explicitly written as separate possible studies such as \texttt{nasal swab for culture} and \texttt{throat swab for PCR}, split them.
        \item If the wording is ambiguous and refers to a single specimen or test choice, preserve the smallest explicit unit from the text without inventing extra tests.
    \end{itemize}
\end{enumerate}

\vspace{1em}

\textbf{Field definitions}

For each object, output:

\begin{itemize}[leftmargin=1.5em]
    \item \texttt{type}: the major category of the item, such as \texttt{Laboratory tests}, \texttt{Imaging examinations}, \texttt{Medical history inquiries}, or \texttt{Specialist examination}.
    \item \texttt{test\_name}: one specific individual test, study, examination, or action only.
    \item \texttt{info\_required}: the specific purpose, information sought, or reason for requesting that item.
\end{itemize}

\vspace{1em}

\textbf{Output requirements}

\begin{enumerate}[leftmargin=1.5em]
    \item Output valid JSON only.
    \item Do not include any explanatory text before or after the JSON.
    \item Use the following exact JSON format.
\end{enumerate}

\vspace{0.5em}

\begin{lstlisting}
{
  "items": [
    {
      "type": "Major Category",
      "test_name": "One individual test/study/exam/action",
      "info_required": "Specific purpose or information sought"
    }
  ]
}
\end{lstlisting}

\vspace{1em}

\textbf{Examples}

\vspace{0.5em}

\textbf{Example input:}

\begin{lstlisting}
Laboratory tests: C-reactive protein (CRP); Complete blood count (CBC) with differential (WBC, neutrophils).
\end{lstlisting}

\vspace{0.5em}

\textbf{Example output:}

\begin{lstlisting}
{
  "items": [
    {
      "type": "Laboratory tests",
      "test_name": "C-reactive protein (CRP)",
      "info_required": "Assess inflammatory or infectious activity."
    },
    {
      "type": "Laboratory tests",
      "test_name": "Complete blood count (CBC) with differential",
      "info_required": "Assess leukocytosis, white blood cell count, and neutrophilia."
    }
  ]
}
\end{lstlisting}

\vspace{0.5em}

\textbf{Example input:}

\begin{lstlisting}
Imaging: Non-contrast CT (NCCT) scan; Non-contrast MRI (NCMRI) scan.
\end{lstlisting}

\vspace{0.5em}

\textbf{Example output:}

\begin{lstlisting}
{
  "items": [
    {
      "type": "Imaging examinations",
      "test_name": "Non-contrast CT (NCCT) scan",
      "info_required": "Evaluate orbital and sinus abnormalities."
    },
    {
      "type": "Imaging examinations",
      "test_name": "Non-contrast MRI (NCMRI) scan",
      "info_required": "Evaluate orbital and sinus abnormalities and intracranial extension."
    }
  ]
}
\end{lstlisting}

\vspace{0.5em}

\textbf{Example input:}

\begin{lstlisting}
Ophthalmological examination (visual acuity, intraocular pressure, fundus examination).
\end{lstlisting}

\vspace{0.5em}

\textbf{Example output:}

\begin{lstlisting}
{
  "items": [
    {
      "type": "Specialist examination",
      "test_name": "Ophthalmological examination",
      "info_required": "Assess the eye's condition, including visual acuity, intraocular pressure, and fundus findings."
    }
  ]
}
\end{lstlisting}

\vspace{1em}

\textbf{Raw output text to be organized:}

\vspace{0.5em}

\begin{lstlisting}
{info_required}
\end{lstlisting}

\end{promptbox}

\begin{promptbox}{Prompt 39}
\footnotesize

\textbf{Prompt for GPT-4.1 mini information matching agent between prediction and ground truth on MedRBench test set.}

\vspace{0.5em}

\textbf{Task overview}

This task aims to accurately analyze the given categories of medical examination items, specific item names, and their testing purposes to determine whether this information is reflected or covered in the provided reference text.

\vspace{1em}

\textbf{Task requirements}

Given the description to be analyzed, judge whether the medical examination item in the description is the same as, or appears in, one of the examination items in the reference text, or whether the testing purpose or required information in the description is reflected or covered in the reference text.

\begin{enumerate}[leftmargin=1.5em]
    \item If the specific examination item name or its alias mentioned in the description to be analyzed appears in the reference text, output \texttt{Yes}.
    \item If the testing purpose or required information mentioned in the description to be analyzed is covered or reflected in the reference text, output \texttt{Yes}.
    \item If neither the specific examination item name nor the required information content mentioned in the description to be analyzed appears in the reference text, output \texttt{No}.
\end{enumerate}

\vspace{1em}

\textbf{Output requirements}

Only output your judgment result on the description to be analyzed, with optional values \texttt{Yes} or \texttt{No}. Do not output any other content.

\vspace{1em}

\textbf{Output format}

\vspace{0.5em}

\begin{lstlisting}
[Yes|No]
\end{lstlisting}

\vspace{1em}

\textbf{Description to be analyzed:}

\vspace{0.5em}

\begin{lstlisting}
{a_info_step}
\end{lstlisting}

\vspace{1em}

\textbf{Reference text:}

\vspace{0.5em}

\begin{lstlisting}
{gt_info}
\end{lstlisting}

\end{promptbox}

\begin{promptbox}{Prompt 40}
\footnotesize

\textbf{Prompt for LLM-Judge evaluating diagnostic accuracy against ground truth on MedRBench test set.}

\vspace{0.5em}

\textbf{Task description}

You are a professional medical diagnosis evaluation system. You will receive two diagnosis results: one is the diagnosis predicted by the model, and the other is the verified correct diagnosis. Your task is to judge whether the model-predicted diagnosis is correct.

\vspace{1em}

When evaluating, consider the following factors:

\begin{enumerate}[leftmargin=1.5em]
    \item The same disease may have multiple aliases. For example, \texttt{Heart disease} may also be called \texttt{Cardiac disease}.
    \item There may be diversity in language expression. For example, \texttt{heart attack} and \texttt{myocardial infarction} may refer to the same disease.
    \item Only judge whether the diagnosis result is correct. Information such as the cause of disease, symptoms, and treatment recommendations is not included in the evaluation scope.
    \item If the correct diagnosis is included in the predicted diagnosis but some additional complications are mentioned, it is also considered correct.
\end{enumerate}

\vspace{1em}

\textbf{Output requirements}

Only output your judgment result on the model-predicted diagnosis as \texttt{Correct} or \texttt{Wrong}. Do not output any other content.

\vspace{1em}

\textbf{Format to follow}

\vspace{0.5em}

\begin{lstlisting}
[Correct|Wrong]
\end{lstlisting}

\vspace{1em}

\textbf{Predicted diagnosis:}

\vspace{0.5em}

\begin{lstlisting}
{pred_diagnose}
\end{lstlisting}

\vspace{1em}

\textbf{Ground-truth diagnosis:}

\vspace{0.5em}

\begin{lstlisting}
{gt_diagnose}
\end{lstlisting}

\end{promptbox}

\begin{promptbox}{Prompt 41}
\footnotesize

\textbf{Prompt for LLM-Judge evaluating treatment-plan accuracy against ground truth on MedRBench test set.}

\vspace{0.5em}

\textbf{Task description}

As a professional medical treatment planning evaluation system, you will receive two treatment plan results for assessment: one is the treatment plan predicted by the model, and the other is the verified correct treatment plan. Your task is to determine whether the model-predicted treatment is accurate.

\vspace{1em}

When evaluating, consider the following factors:

\begin{enumerate}[leftmargin=1.5em]
    \item If the predicted treatment and ground-truth treatment have exactly the same meaning, then it is correct.
    \item If the correct treatment plan is included in the predicted treatment but some additional care is mentioned, it is also considered correct.
    \item Considering that even the same disease can sometimes be treated differently, if the model's prediction does not completely match the ground-truth treatment, you can refer to additional information to make a judgment.
    \item If the predicted treatment and the ground-truth treatment do not convey the same meaning, and there is no supporting evidence in the additional information to suggest that the predicted treatment is also applicable to the disease, it is considered wrong.
\end{enumerate}

\vspace{1em}

\textbf{Output requirements}

Only output your judgment result on the model-predicted treatment as \texttt{Correct} or \texttt{Wrong}. Do not output any other content.

\vspace{1em}

\textbf{Format to follow}

\vspace{0.5em}

\begin{lstlisting}
[Correct|Wrong]
\end{lstlisting}

\vspace{1em}

\textbf{Predicted treatment:}

\vspace{0.5em}

\begin{lstlisting}
{pred_treatment}
\end{lstlisting}

\vspace{1em}

\textbf{Ground-truth treatment:}

\vspace{0.5em}

\begin{lstlisting}
{gt_treatment}
\end{lstlisting}

\vspace{1em}

\textbf{Additional information:}

\vspace{0.5em}

\begin{lstlisting}
{additional_info}
\end{lstlisting}

\end{promptbox}


\clearpage
\phantomsection
\addcontentsline{toc}{section}{References}

\putbib[sn-bibliography]

\end{bibunit}


